\documentclass[preprint,12pt]{elsarticle}
\usepackage{amssymb}
\usepackage{amsmath}
\usepackage{times}
\usepackage{soul}
\usepackage{url}
\usepackage[hidelinks]{hyperref}
\usepackage[utf8]{inputenc}
\usepackage[small]{caption}
\usepackage{graphicx}
\usepackage{amsthm}
\usepackage{booktabs}
\usepackage{algorithm}
\usepackage[switch]{lineno}

\usepackage{balance} % for balancing columns on the final page
\usepackage[T1]{fontenc}    % use 8-bit T1 fonts
\usepackage{amsfonts}       % blackboard math symbols
\usepackage{nicefrac}       % compact symbols for 1/2, etc.
\usepackage{microtype}      % microtypography
\usepackage{xcolor}         % colors

\usepackage{algorithmicx}
\usepackage{algpseudocode}

\usepackage{verbatim}
\usepackage{multirow}
\usepackage{capt-of}
\usepackage{subcaption}
\usepackage{caption}

\usepackage{fancyhdr}
\usepackage{diagbox}
\usepackage{makecell}

\usepackage{marvosym}

\definecolor{purple}{HTML}{7D54B2}
\definecolor{green}{HTML}{229487}
\definecolor{red}{HTML}{DA4C4C}
\definecolor{blue}{HTML}{5387DD}
\definecolor{b}{HTML}{5387DD}
\definecolor{o}{HTML}{E57439}
\definecolor{y}{HTML}{EDB732}
\definecolor{r}{HTML}{DA4C4C}
\definecolor{g}{HTML}{479A5F}

\newcommand{\answerTODO}[1][]{\textcolor{red}{\bf [TODO]}}
\newcommand{\justificationTODO}[1][]{\textcolor{red}{\bf [TODO]}}

\journal{Arxiv}

\begin{document}

\begin{frontmatter}

%% Title, authors and addresses

%% use the tnoteref command within \title for footnotes;
%% use the tnotetext command for theassociated footnote;
%% use the fnref command within \author or \affiliation for footnotes;
%% use the fntext command for theassociated footnote;
%% use the corref command within \author for corresponding author footnotes;
%% use the cortext command for theassociated footnote;
%% use the ead command for the email address,
%% and the form \ead[url] for the home page:
%% \title{Title\tnoteref{label1}}
%% \tnotetext[label1]{}
%% \author{Name\corref{cor1}\fnref{label2}}
%% \ead{email address}
%% \ead[url]{home page}
%% \fntext[label2]{}
%% \cortext[cor1]{}
%% \affiliation{organization={},
%%             addressline={},
%%             city={},
%%             postcode={},
%%             state={},
%%             country={}}
%% \fntext[label3]{}

\title{D3HRL: A Distributed Hierarchical Reinforcement Learning Approach Based on Causal Discovery and Spurious Correlation Detection} %% Article title

%% use optional labels to link authors explicitly to addresses:
%% \author[label1,label2]{}
%% \affiliation[label1]{organization={},
	%%             addressline={},
	%%             city={},
	%%             postcode={},
	%%             state={},
	%%             country={}}
%%
%% \affiliation[label2]{organization={},
	%%             addressline={},
	%%             city={},
	%%             postcode={},
	%%             state={},
	%%             country={}}

\author[1]{Chenran Zhao}%\ead{judylalala2021@163.com} %% Author name
\author[1,2]{Dianxi Shi\Letter}%\ead{dxshi@nudt.edu.cn}	% \corref{dxshi@nudt.edu.cn}
\author[4]{Mengzhu Wang}
\author[5]{Jianqiang Xia}%\ead{jianqiang.xia@foxmail.com}
\author[1]{Huanhuan Yang}%\ead{yanghh94@126.com}
\author[2]{Songchang Jin}%\ead{jsc04@tsinghua.org.cn}
\author[1]{Shaowu Yang}%\ead{shaowu.yang@nudt.edu.cn}
\author[2]{Chunping Qiu}%\ead{chunping.qiu@aliyun.com}

%% Author affiliation
\affiliation[1]{organization={College of Computer Science and Technology, National University of Defense Technology},%Department and Organization
	%addressline={}, 
	city={China},
	%postcode={}, 
	%state={},
	country={Changsha}}

\affiliation[2]{organization={Intelligent Game and Decision Lab (IGDL)},%Department and Organization
	%addressline={}, 
	city={China},
	%postcode={}, 
	%state={},
	country={Beijing}}

\begin{comment}
\affiliation[3]{organization={National Key Laboratory of Parall and Distributed Computing, College of Computer Science and Technology, National University of Defense Technology},%Department and Organization
	%addressline={}, 
	city={China},
	%postcode={}, 
	%state={},
	country={Changsha}}
\end{comment}

\affiliation[4]{organization={College of Artificial Intelligence, Hebei University of Technology},%Department and Organization
	%addressline={}, 
	city={China},
	%postcode={}, 
	%state={},
	country={Tianjin}}

\affiliation[5]{organization={School of Electronic information and Electrical Engineering, Shanghai Jiao Tong University},%Department and Organization
	%addressline={}, 
	city={China},
	%postcode={}, 
	%state={},
	country={Shanghai}}

%% Abstract
\begin{abstract}
	%% Text of abstract
	Current Hierarchical Reinforcement Learning (HRL) algorithms excel in long-horizon sequential decision-making tasks but still face two challenges: delay effects and spurious correlations. To address them, we propose a causal HRL approach called D3HRL. First, D3HRL models delayed effects as causal relationships across different time spans and employs distributed causal discovery to learn these relationships. Second, it employs conditional independence testing to eliminate spurious correlations. Finally, D3HRL constructs and trains hierarchical policies based on the identified true causal relationships. These three steps are iteratively executed, gradually exploring the complete causal chain of the task. Experiments conducted in 2D-MineCraft and MiniGrid show that D3HRL demonstrates superior sensitivity to delay effects and accurately identifies causal relationships, leading to reliable decision-making in complex environments.
\end{abstract}

\begin{comment}
%%Graphical abstract
\begin{graphicalabstract}
	%\includegraphics{grabs}
\end{graphicalabstract}

%%Research highlights
\begin{highlights}
	\item Research highlight 1
	\item Research highlight 2
\end{highlights}
\end{comment}

%% Keywords
\begin{keyword}
	%% keywords here, in the form: keyword \sep keyword
	Causality \sep Hierarchical Reinforcement Learning \sep Long-horizon Sequential Decision-making Tasks
	
	%% PACS codes here, in the form: \PACS code \sep code
	
	%% MSC codes here, in the form: \MSC code \sep code
	%% or \MSC[2008] code \sep code (2000 is the default)
	
\end{keyword}
\end{frontmatter}

%\linenumbers

\section{Introduction}\label{sec:introduction}

%Long-horizon sequential decision-making tasks require a series of decisions to achieve a complex goal, typically requiring the agent to accomplish multiple intermediate objectives step by step.

%Long-horizon sequential decision-making tasks (hereafter abbreviated as long-horizon tasks) require accomplishing multiple intermediate objectives step by step to achieve the final goal.

%Hierarchical Reinforcement Learning (HRL)

Long-horizon sequential decision-making tasks, which we will refer to as long-horizon tasks, involve achieving a series of intermediate objectives in sequence to accomplish the ultimate goal. HRL algorithms excel in these tasks due to their inherent framework advantages, yet still face two challenges.

%In the real world, long-horizon tasks often involve "delay effects," which can be seen as variable-length state transitions or temporally extended actions. 
% commonly found in long-horizon tasks in the real world
%Semi-Markov Decision Processes (SMDPs)
%significant 
%, referred to as variable-length state transitions.
% (hereafter referred to as variable-length state transitions)
%The first challenge is the "delay effect". This can be viewed as temporally extended actions that trigger state transitions spanning multiple time steps, referred to as variable-length state transitions. 
The first challenge is the "delay effect," where temporally extended actions trigger state transitions over multiple time steps, referred to as variable-length state transitions. Submitting papers to a conference exemplifies a temporally extended action with a delay effect, where authors must wait for several months before receiving notification. SMDPs are typically used to model variable-length state transitions. \cite{sutton1999between} introduced options to simplify SMDPs, proving that SMDPs can be represented as MDPs augmented with options. This abstraction allows agents to make decisions at a more macroscopic level. \cite{schaul2015universal} proposed goals as another form of abstraction, which are typically intermediate states needed to complete the final task. However, both options and goals primarily focus on abstract skills, such as "press the button"~\cite{roder2020curious} or "explore the first room of the maze"~\cite{campos2020explore}. These are combinations of multiple primitive actions rather than the temporal extension of a single action. We aim to incorporate "temporally extended primitive actions" to enable decision-making in more complex tasks.
%In other words, a temporally extended action triggers a state transition with delay effect. 
%a temporally extended action triggers a state transition with delay effect.
%For example, submitting a paper to a conference and awaiting the decision is a long-duration state transition spanning several months.
%Submitting a research paper to an academic conference is a classic example of a temporally extended action, as it involves a waiting period of several months before authors receive the results of their submission.

%, which can be viewed as MDPs

%Although any primitive action can be further decomposed, we do not pursue further decomposition and consider the current environment's action space as the primitive action space.\cite{cui2018review,liu2019efficient}

%However, even if we have addressed delayed effects, agents' decision-making will still be limited if they cannot understand and apply causal relationships. 
%there is a significant gap between being "usable" and "reliably usable"
%they fall short of being "reliably usable"~\cite{deng2023causal}.
The second challenge is the spurious correlations. In long-horizon tasks, the complex interactions can lead to widespread data correlations, many of which are spurious. In fact, most agents rely on data correlations rather than causal relationships. \cite{willig2023causal} show that even powerful ChatGPT is nothing more than an echoing parrot, lacking causal understanding. While these correlation-seeking agents perform well, there is a significant gap between being "usable" and "reliably usable"~\cite{deng2023causal}. For example, a fire breaks out in a residential building, leading to the dispatch of fire trucks and casualties. From a causal perspective, assuming the data is comprehensive and evenly distributed, there are causal relationships as follows: \textit{the dispatch of fire trucks}\(\leftarrow\)\textit{fire}\(\rightarrow\)\textit{casualties}. This creates an open path between the dispatch of fire trucks and casualties, leading to a correlation between them. In fact, this correlation is spurious and not causal~\cite{runge2019detecting}. If an agent learns from these correlations, it might incorrectly infer that dispatching fire trucks will result in casualties, leading to the wrong decision to avoid dispatching them.

%From a causal perspective, assuming the data is comprehensive and evenly distributed, there is a causal relationship between a fire and both the dispatch of fire trucks and casualties.

\begin{comment}
\begin{figure}
	\centering
	\includegraphics[width=0.9\linewidth]{figures/causal or not}
	\caption{The true causal relationship and spurious correlation.}\label{fig:causal_or_not}
\end{figure}
\end{comment}

%the aforementioned
%A distributed causal discovery mechanism enables parallel learning across various time spans.
%we introduce a distributed causal discovery mechanism to perform parallel learning of causal relationships across various time horizons.
%By assessing the likelihood of effects given different causes, we identify causal relationships. 
%and systematically achieving sub-goals 
To address these challenges, we propose \textbf{D3HRL}, a \textbf{D}istributed \textbf{HRL} approach based on causal \textbf{D}iscovery and spurious correlation \textbf{D}etection. It iteratively execute the following three modules, progressively uncovering causal chain until the task is completed. \textbf{(1) Modeling Delay Effects}: We model delay effects as causal relationships and introduce a distributed causal discovery mechanism to perform parallel learning of causal relationships across various time spans. \textbf{(2) Detecting Spurious Correlations}: We conduct conditional independence testing to screen for genuine causal relationships and determine their true time span. \textbf{(3) Building Hierarchical Policies}: We construct hierarchical policy networks and train sub-goals based on the identified causal relationships.

\begin{comment}
\begin{itemize}
	\item \textbf{Distributed Causal Discovery}: We formalize "delay effects" as causal relationships. By employing distributed causal discovery, D3HRL learns causal relationships at different time spans
	%, thereby enhancing the agent's ability to handle complex temporal dependencies.
	
	\item \textbf{Spurious Correlation Detection}: We employ conditional independence testing to filter out spurious correlations and determine the time spans of true causal relationships, ensuring the agent learns reliable causality.
	
	\item \textbf{Hierarchical Policy Network Construction}: We use the identified true causal relationships and their corresponding time spans to construct a hierarchical policy network and learn the associated sub-goals. 
\end{itemize}
\end{comment}

We tested our method in two environments with long-horizon tasks: 2D-MineCraft~\cite{sohn2018hierarchical} and MiniGrid~\cite{MinigridMiniworld23}. We modified them to enable variable-length state transitions. Results show that D3HRL can identify variable-length state transitions and construct the causal chains accurately, significantly improving the learning efficiency of the hierarchical framework.

\section{Related works}
\subsection{Hierarchical Reinforcement Learning}
%,achiam2018variational
%frans2017meta,
%kim2021landmark,
% without managing every detail
%,cui2018review
%,liu2019efficient
HRL falls into two categories: option-based (O-HRL) and goal-based (G-HRL). \cite{sutton1999between} introduced options to model variable-length state transitions in SMDPs, enabling agents to focus on high-level decisions. \cite{schaul2015universal} later introduced goals, breaking tasks into sub-goals to progressively reach the final goal. O-HRL involves a lower-level network learning skills and an upper-level network using these skills, either asynchronously~\cite{eysenbach2018diversity,campos2020explore,achiam2018variational} or synchronously~\cite{frans2017meta,song2019diversity,song2019playing}. G-HRL involves an upper-level network generating sub-goals and a lower-level network achieving them, typically done synchronously~\cite{park2024hiql,zou2023sample,li2025highly,ou2024modular}. However, these methods primarily focus on abstract skills rather than temporally extended actions. We propose a distributed causal discovery module and a hierarchical architecture to effectively model them both.

\subsection{Causal Reinforcement Learning}
Causality integrates with RL in three main ways. First, counterfactual reasoning is used: \cite{foerster2018counterfactual} use counterfactual differential advantage functions to evaluate actions retrospectively, \cite{buesing2018woulda} employ counterfactual inference for decisions, and \cite{madumal2020explainable} extend Structural Causal Mechanism (SCM) to explain counterfactual behaviors. Second, intrinsic rewards are used: \cite{herlau2022reinforcement} generate rewards by analyzing the causal effect of intermediate variables, guiding policy learning. Third, causal graphs are employed: \cite{ding2022generalizing} transform RL into a likelihood maximization problem, learning causal graphs through classification and using them as latent variables for variational inference. However, none of them consider spurious correlations. D3HRL includes a spurious correlation detection module to ensure reliable causal relationships.
%However, none of these approaches achieve a complete exploration of causal chains in long-horizon sequential decision-making tasks. D3HRL integrates distributed causal discovery methods, spurious correlation detection techniques, and distributed hierarchical policy learning in an organic manner, ensuring the accuracy of learned causal relationships while progressively explore the causal chain.

\subsection{Causal Hierarchical Reinforcement Learning}
%SCALE~\cite{lee2023scale} generates intervention data in a simulated environment, identifying causal features and building a diverse, compact skill library. However, SCALE's data is generated in a parameter-controlled simulation, which limits its autonomous exploration in uncontrolled tasks.
%SCALE \cite{lee2023scale} uses simulated data to identify causal features and build a skill library, but its reliance on controlled simulations limits autonomous exploration in uncontrolled tasks.
%SCALE~\cite{lee2023scale} identifies causal features and builds a skill library using simulated intervention data. However, its reliance on parameter-controlled simulations limits autonomous exploration in uncontrolled tasks.
CHRL is an emerging area. CEHRL~\cite{corcoll2022disentangling} uses meta-supervised learning with counterfactual reasoning and advantage functions to learn the hierarchical structure of control effects. CDHRL~\cite{hu2022causality} gradually learns causal chain through intervention data and build hierarchical policies based on it. Both CEHRL and CDHRL can learn causal chains in long-horizon tasks but fail to model "delayed effects". SCALE \cite{lee2023scale} uses simulated data to identify causal features and build a skill library, but its reliance on controlled simulations limits autonomous exploration in uncontrolled tasks. COInS~\cite{chuckgranger} identifies key interaction states and constructs skill chains, controlling increasingly difficult factors. However, its skill exploration is manually predefined, not automatic. HCPI-HRL~\cite{chen2025hcpi} leverages human-guided causal discovery to automatically infer effective and diverse subgoal structures from dynamic environments. The method's dependence on human causal insights curtails its capacity for fully autonomous learning of subgoals and causal structures. VACERL~\cite{nguyen2024variable} eliminates hand-crafted causal variables by using transformer attention to isolate the observation-action steps most predictive of future returns, then builds a causal graph over those key steps. That graph is converted into intrinsic rewards or sub-goals, sharply boosting exploration in sparse-reward and Noisy-TV tasks. Different from them, our D3HRL employs the automatic gradual exploration of causal chains to complex environments with "delayed effects".

%However, its reliance on human-provided causal knowledge limits its autonomy and adaptability, as it cannot fully learn causal structures and subgoals automatically from raw data without expert guidance. 
% and enhances the efficiency of hierarchical policies by introducing a spurious correlation detection module.
%It associates multiple sub-goals with a single skill. 
%COInS~\cite{chuckgranger} identifies key interaction states and constructs skill chains, controlling increasingly difficult factors. However, COInS cannot automatically explore causal chains; its skill exploration is manually pre-defined.
%In summary, D3HRL is the first CHRL algorithm that simultaneously addresses variable-length state transitions and spurious correlation detection in long-horizon tasks. It can sensitively perceive variable-length state transitions while ensuring the accuracy of the learned causal relationships, thereby enabling more reliable decision-making in complex and dynamic real-world environments.

\section{Preliminary}
%Causality~\cite{pearl2000models,spirtes2001causation} refers to a relationship where an intervention on one variable (cause \(C\)) directly leads to a change in another variable (effect \(E\)), denoted as \(C \rightarrow E\). 
%Generally, the cause precedes the effect with variable-length time spans. 
%Granger causality~\cite{granger1969investigating} is defined as: if one variable (cause \(C\)) significantly improves the prediction of another variable (effect \(E\)), then there is a Granger causality between \(C\) and \(E\), denoted as \(C \rightarrow E\). 
Causality~\cite{pearl2000models} refers to a relationship where an intervention on one variable (cause \(C\)) directly leads to a change in another variable (effect \(E\)), denoted as \(C \rightarrow E\). Time series causal discovery involves inferring and quantifying causality from time series data~\cite{runge2019inferring}, helping to understand the influence mechanisms between variables. Structural causal mechansim (Section~\ref{sec:scm}) is used to represent causal relationships. Time series graphical models (Section~\ref{sec:tsgm}) model causal relationships among multiple variables across multiple time steps, and conditional independence testing (Section~\ref{sec:cit}) is used to detect spurious correlations. 

\subsection{Structural Causal Mechanism}\label{sec:scm}
%Structural Causal Mechanism (SCM) consists of two parts. First is the causal graph: if \(C\) has a causal effect on \(E\) , there is a directed edge \(C \rightarrow E\). Second is the generating function: if \(C\) is the cause of \(E\), there exists a generating function such that we can predict \(E\) based on \(C\). See \ref{app:granger} for details.
\paragraph{Structural Causal Mechanism (SCM)} SCM~\cite{peters2017elements} consists of two parts. First is the causal graph, where \(C \rightarrow E\) indicates a causal relationship. Second is the generating function, which predicts \(E\) based on \(C\). 
%the value of \(E\) can be derived from the value of \(C\). See \ref{app:granger} for details.
%SCM~\cite{peters2017elements} consists of two parts: the causal graph, where \(C \rightarrow E\) indicates a causal relationship; and the generating function, which predicts \(E\) based on \(C\). Our SCM's generating functions are implemented by predicting the effect given different causes, similar to the concept of Granger causality~\cite{granger1969investigating}. Therefore, the causal relationships identified in our method refer to Granger causality (see \ref{app:granger}).
%Our SCM's generating functions are implemented by predicting the effect given different causes, . Therefore, our SCM is implemented based on GC to some degree, and all causal relationships in our method refer to Granger causality. Details are described in \ref{app:granger}.
%We leverage the principles of the Granger causality test to implement our SCM. 

\paragraph{Granger causality} According to Granger causality~\cite{granger1969investigating}, if there is a Granger causal relationship between cause \(C\) and effect \(E\) (\(C \rightarrow E\)), then \(C\) contains unique information related to \(E\) that is not contained in all past information (excluding \(C\)). The Granger causality test can determine whether a Granger causal relationship exists between two variables. Leveraging Granger-causality principles, we sample candidate causes \(C\) and forecast the effect \(E\); gains in predictive accuracy reveal the true causal edges. This procedure simultaneously reconstructs the causal graph and learns the generative mapping \(E = f(C)\), yielding an instantiation of the SCM. %See \ref{app:granger} for more details. 

\paragraph{Intervention} In causal inference, deliberately setting a variable to a specific value and observing the resulting changes is called an \emph{intervention}, denoted by Pearl’s do-operator \(\operatorname{do}(C{=}c)\). In our method, the random sampling of different candidate causes and evaluation of their impact on the prediction of \(E\) is tantamount to performing a collection of interventions \(\{\operatorname{do}(C{=}c_i)\}\).

\subsection{Time Series Graphical Models}\label{sec:tsgm}
\begin{figure*}[htbp]
	\centering
	\includegraphics[width=0.45\linewidth]{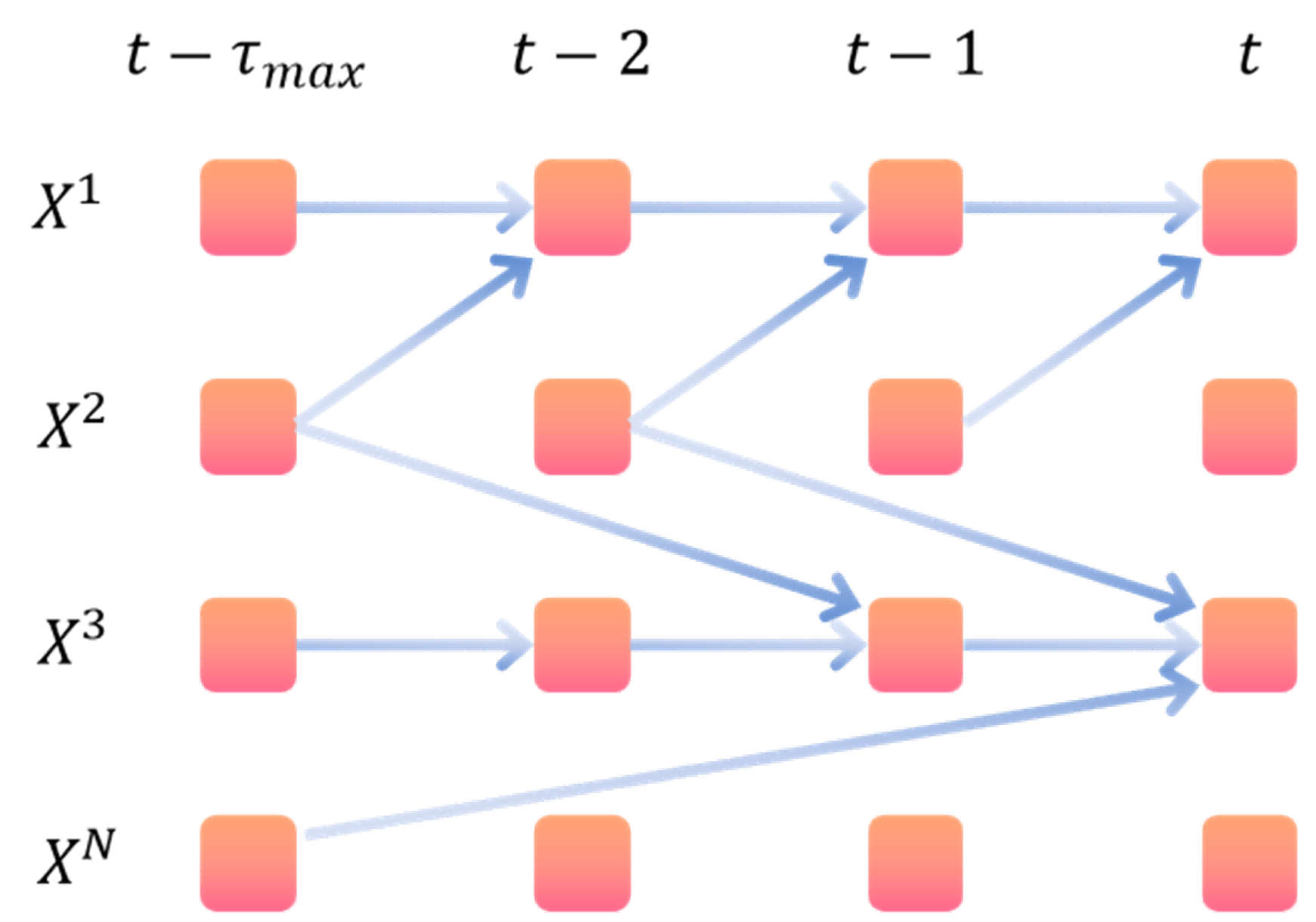}
	\caption{An example of Time Series Graphical Models.}\label{fig:tsgm}
\end{figure*}

%Time Series Graphical Models (TSGM)~\cite{eichler2012graphical} can model variable-length causal relationships among multiple variables and their time-lagged variables.
Time Series Graphical Models (TSGM)~\cite{eichler2012graphical} can model causal relationships across multiple time steps. Figure~\ref{fig:tsgm} shows a schematic diagram of a TSGM. Consider a multivariate process \(X\) consisting of \(N\) variables. Its TSGM can be defined as \(G = (\mathbf{V} \times \mathbf{T}, \mathbf{E})\):
\begin{itemize}
\item Vertex set \(\mathbf{V=\{X_t\}_{t=1}^T}\): a set of temporal nodes for \(N\) variables at each time step \(t \in \mathbf{T}\), where \(\mathbf{X_t}=\{X_t^1, X_t^2, ...,X_t^N\}\).

\item Edge set \(\mathbf{E}\): a set of causal relationships between nodes in \(\mathbf{V}\). If \(X_{t-\tau}^j\) causally influences \(X_t^i\), there is a directed edge \(X_{t-\tau}^j \rightarrow X_t^i\) (\(\tau>0\)).
%If there exists a causal relationship from \(X_{t-\tau}^j\) to \(X_t^i\), then there is a directed edge \(X_{t-\tau}^j \rightarrow X_t^i\) (\(\tau>0\)). 
\end{itemize}
%are the set of nodes that causally influence \(X^i_t\): \begin{equation}
%are the set of nodes involved in causal relationships where \(X^i_t\) is the effect: \begin{equation}
%The parent nodes of node \(X^i_t\), denoted as \(\mathcal{PA}(X_t^i)\), are the set of nodes that causally influence \(X^i_t\): \begin{equation}
The parent nodes of \(X^i_t\), denoted as \(\mathcal{PA}(X_t^i)\), are the set of nodes involved in causal relationships where \(X^i_t\) is the effect: \begin{equation}
	\mathcal{PA}(X_{t}^{i})=\left\{X_{t-\tau}^{k} | 
	\begin{array}{l}
		X_{t-\tau}^{k}\rightarrow X_{t}^{i} \in \mathbf{E}, \tau>0,\\
		X^{k}_{t-\tau}\in\mathbf{V}, X^{i}_t\in\mathbf{V}
	\end{array}
	\right\} 
	%X_{t-\tau}^{k}\rightarrow X_{t}^{j}, X^{k}_{t-\tau}\in\mathbf{V}, X^{j}_t\in\mathbf{V}, \tau>0\}.
\end{equation}
\(\tau_{max}\) is the maximum time span with significant correlations in complex systems~\cite{runge2019detecting}. In other words, \(\tau_{max}\) is an empirical value that incorporates prior knowledge. 

% We propose \(h\)-window based on \(\tau_{max}\). It refers to the time segment from time step \(t - h\) (\(h \in \{1, 2, \dots, \tau_{max}\)\}) to \(t\).

\subsection{Conditional Independence Testing}\label{sec:cit}
%it is hard to determine the information flow along the path \(X^j \rightarrow X^i\), making it challenging to identify the causal relationship between them
%%If the two variables are conditionally dependent, it indicates a causal relationship; otherwise, it suggests a spurious correlation:\begin{equation}
%Conditional dependence indicates a causal relationship; otherwise, the correlation is spurious
%By assessing the information flow along this path, we can determine if the correlation is spurious. Thus, conducting Conditional Independence Testing (CIT)~\cite{runge2019detecting} on the set of \(X\)'s parent nodes can identify spurious correlations. 
%it suggests 
Due to the environmental complexity, determining the information flow along the path \(X^j \rightarrow X^i\) is challenging, making it difficult to identify their causal relationship. However, causal theory~\cite{pearl2000models} states that \(X\)'s parent nodes \(\mathcal{PA}(X)\) form a sufficient conditioning set to block the backdoor paths, ensuring a unique open path. By evaluating the information flow along this path, we can determine if the variables are conditionally dependent. If they are, it indicates a causal relationship; otherwise, it suggests a spurious correlation. This is known as Conditional Independence Testing (CIT)~\cite{runge2019detecting}:\begin{equation}
	\ X_{t-\tau}^j \perp\!\!\!\perp X_t^i\mid
	\begin{array}{l}
		\mathcal{PA}(X_t^i)\setminus\{X_{t-\tau}^j\}\\
		\mathcal{PA}(X_{t-\tau}^j) 
	\end{array}
	%\mathcal{PA}(X_t^j)\setminus\{X_{t-\tau}^i\},\mathcal{PA}(X_{t-\tau}^i) 
	\Rightarrow X_{t-\tau}^{j} \nrightarrow X_{t}^{i}
\end{equation}

\section{Formalization}\label{cfsmdp}
Real-world variable-length state transitions can overlap in time but remain independent of each other. For example, after submitting a paper to a conference, one might wait for the notification while working on another project. This paper's result and the project's progress are separate sub-states that do not influence each other. We integrate the concepts of SMDPs to model variable-length state transitions and those of Factored-MDPs~\cite{boutilier1999decision} to handle overlapping sub-state transitions. Besides, state transitions follow objective causal laws. Thus, it is natural to represent variable-length state transitions as causal relationships spanning multiple time steps. Based on these ideas, we propose \textbf{Causal Factored-SMDPs}. It relies on several key assumptions:
%we propose a formalization called Causal Factored-SMDPs
%However, during the delay of one transition, another can be triggered. 
%, even though they occur concurrently.
%We discretize continuous time and model these relationships using TSGM, while using the generating function in SCM to model their causal effects. 

\textit{(1) Independent Sub-State Transition}: The state can be decomposed into disjoint sub-states~\cite{boutilier2000stochastic}, and the transitions of each sub-state are independent. 

%The cause \(X^j\) precedes the effect \(X^i\) by a temporal interval \(\tau > 0\). In other words, 
%In our setting, we assume \(1 \leq \tau \leq \tau_{max}\).
\textit{(2) Cause precedes Effect}~\cite{pearl2000models}: For all \(i \neq j\), it holds that \(X_{t}^{j} \nrightarrow X_{t}^{i}\) and \(X_{t}^{j} \nrightarrow X_{t-\tau}^{i}\) (\(1 \leq \tau \leq \tau_{max}\)).

\textit{(3) Causal Sufficiency Assumption}~\cite{spirtes2001causation}: All relevant causal factors are observed. In our setting, an effect corresponds to multiple causes, and the effect occurs only if all causes meet the conditions, spanning the same time span.

\textit{(4) Causal Markov Condition}~\cite{spirtes2001causation}: The true causes are sufficient to predict the effect, while other past variables are irrelevant. In our setting, all variables except actions exhibit autocorrelation. %See \ref{app:causal_markov_condition} for its connection with the Markov Property~\cite{sutton1999reinforcement}.

%We discretize continuous time and model causal relationships with TSGM and SCM.
%We replace the state and action spaces with cause and effect spaces, and substitute the state transition functions with the generating functions of SCM. We also introduce causal relationship matrices and the time span matrix to describe TSGM. We discretize continuous time and the policy is closed-loop.
% See \ref{app:cfsmdp} for details. 
%We discretize continuous time and the policy is closed-loop. 
We replace the state and action spaces with cause and effect spaces, and replace the state transition functions with the generating functions of SCM. We introduce causal relationship matrices and the time span matrix to describe TSGM. Now we present Causal Factored-SMDPs consisting of 5 components \(<\mathbf{C,E,P,T,F}>\):
%See \ref{app:cfsmdp}, \ref{app:connection_cfsmdp_smdp_mdp} for details. 
%Causal Factored-SMDPs consists of five components \(<\mathbf{C,E,P,T,F}>\):
%We discretize continuous time and model these relationships using TSGM, while using the generating function in SCM to model their causal effects. 
%, where decisions are based on the current state at each time step

\begin{itemize}
	\item \textbf{Cause Space \(\mathbf{C}\):} Comprises a decomposed state space \(\{S^i\}_{i=1}^M \in \mathbf{S} = \{S^1 \times S^2 \times \dots \times S^M\}\) and a decomposed action space \(\{A^j\}_{j=1}^N \in \mathbf{A} = \{A^1 \times A^2 \times \dots \times A^N\}\). %State sub-variables and action sub-variables can both serve as cause variables.
	%State sub-variables and action sub-variables serve as cause variables \(C\).
	\item \textbf{Effect Space \(\mathbf{E}\):} Comprises a decomposed state space \(\{S^i\}_{i=1}^M \in \mathbf{S} = \{S^1 \times S^2 \times \dots \times S^M\}\). The agent's decision is made to induce state transitions, so the action space cannot be considered part of the effect space.
	%The agent's decision is made to induce state transitions, so the action space cannot be considered part of the effect space.
	% and a reward \(R\). Reward \(R\) is only given upon the completion of the final task.
	%State sub-variables and reward serve as effect variables \(E\). Reward \(R\) is only given upon the completion of the final task.
	\item \textbf{Causal Relationship Matrices \(\mathbf{P}= \{P_h\}_{h=1}^{\tau_{max}}\):} \(P_h \in \{0,1\}^{(M) \times (M+N)}\) discribes the existence of the causal relationships under time span \(h\). \(P_h[i][j]=1\) indicates that \(C^j_{t-h}\) is a direct cause of \(E^i_t\), and \(0\) otherwise.
	%\(P_h[i][j]=0\)
	%Rows represent effects and columns represent causes. 
	%Comprises the matrices \(P_h \in \{0,1\}^{(M) \times (M+N)}\) for each time span \(h\).
	%while \(P^{ij}_h=0\) indicates otherwise.
	%Describes the time span of each causal relationships. 
	\item \textbf{Causal Relationship Time Span Matrix \(\mathbf{T \in \{1,2,...,\tau_{max}\}^{(M) \times (M+N)}}\):} The true time span of the causal relationship (\(C^j \rightarrow E^i\)) is \(T_{ij}\). % initiated by cause \(C^i\) leading to effect \(E^j\)
	\item \textbf{Causal Effect Generating Functions \(\mathbf{F}=\{F_h\}_{h=1}^{\tau_{max}}\):} It consists of generating functions for each time span \(h\) and each effect variable \(E^i\):\begin{equation}
		%=\left\{ f^i_h\left(E^i \mid \mathcal{PA}(E^i) \right) \mid \mathcal{PA}(E^i) \subseteq \mathbf{C}, E^i \in \mathbf{E}\right\}_{i=1}^{M}
		F_h=\left\{ f^i_h\left(E^i \mid \mathcal{PA}(E^i) \right) | 
		\begin{array}{l}
			\mathcal{PA}(E^i) \subseteq \mathbf{C} \\
			E^i \in \mathbf{E}
		\end{array}
		\right\}_{i=1}^{M}
	\end{equation}
	\(f^i_h\left(E^i|\mathcal{PA}(E^i)\right)\) predicts \(E^i\)'s value \(h\) steps later given its direct cause \(\mathcal{PA}(E^i)\). During these \(h\) steps, the value of \(E^i\) doesn't change. In short, \(\mathbf{F}\) describe the sub-state transitions of each effect variable. Thus, the complete state $S_t=\{E^1_t, E^2_t, ..., E^M_t\}$ is determined by \(\mathbf{F}\).
	%describes the causal effect of \(\mathcal{PA}(E^i) \xrightarrow{h} E^i\), 
	%all effect variables' generating functions.
\end{itemize}

\section{Approach}
\begin{figure*}[htbp]
	\centering
	\includegraphics[width=1.0\linewidth]{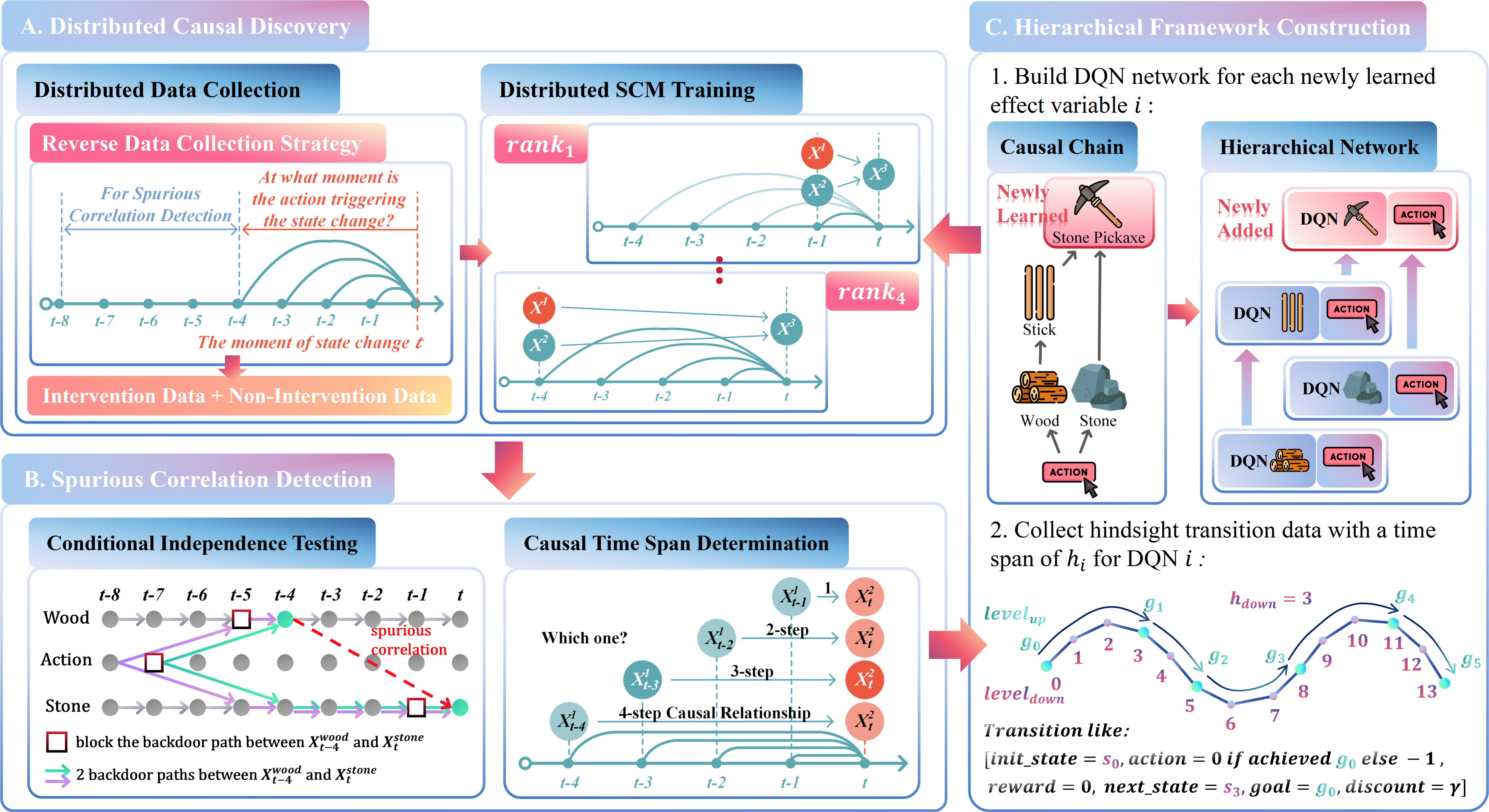}
	\caption{Overview of D3HRL framework.}\label{fig:overall} 
\end{figure*}

In this chapter, we introduce D3HRL, a distributed approach that accurately captures variable-length state transitions and constructs a high-quality hierarchical framework. The overall framework is shown in Figure~\ref{fig:overall}. In Module A, data is collected using a reverse data collection strategy, followed by distributed SCM training to preliminarily identify causal relationships at different time spans (Section~\ref{sec:causal_discovery}). In Module B, the identified "causal relationships" undergo spurious correlation detection, including conditional independence testing and determination of causal relationship time spans (Section~\ref{sec:global_causal_determination}). In Module C, based on the identified causal relationships and their time spans, we construct the hierarchical network and train the subgoals for effect variables (Section~\ref{sec:hierarchical_network}). These three modules iteratively execute until the causal chain is explored completely. %More details are provided in \ref{app:d3hrl}.

\subsection{Distributed Causal Discovery}\label{sec:causal_discovery}
To capture the variable-length state transitions, we propose a distributed causal discovery method consisting of distributed data collection and distributed SCM training. Based on the maximum time span \(\tau_{max}\), multiple processes \(\{rank_h\}_{h=1}^{\tau_{max}}\) are set up for distributed learning. 
%Details are shown in \ref{app:distributed_causal_discovery}.

\subsubsection{\textbf{Distributed Data Collection.}} 
Each process \(rank_h\) interacts with the environment using the current hierarchical policy in a offline manner. Complete trajectories are stored in the replay buffer \(\mathbf{R}\) and extracted for data collection. We employ a reverse data collection strategy: after an intervention variable $X_{do} \in List_{do}$ (initially $List_{do}=\{\mathbf{A}\}$) is triggered, if the state changes at $t$, given $\tau_{\max}$, the causes must lie within $[t-\tau_{max}, t)$. Therefore, starting from $t$, we collect $\tau_{max}+1$ steps of data in reverse:
\begin{equation}
	d^{scm}_t = \{s_{t-\tau_{max}}, a_{t-\tau_{max}}, \ldots, s_{t-1}, a_{t-1}, s_t, a_t\}
\end{equation}
Additionally, to facilitate the detection of spurious correlations, we also collect an additional $\tau_{max}$ steps of data in reverse: 
\begin{equation}
	%d^{for}_t = \{s_{t+1}, a_{t+1}, s_{t+2}, a_{t+2}, \ldots, s_{t+\tau_{max}}, a_{t+\tau_{max}}\}
	d^{cmi}_t = \{s_{t-2\tau_{max}}, a_{t-2\tau_{max}},\ldots, a_{t-\tau_{max}-1}\}
\end{equation}
For each iteration, we collect $batch\_size$ such segments of intervention data for each intervention variable parallelly:
\begin{equation}
	D_{do}=\left\{D_{do}^h=\left\{\text{\textit{cat}}\left(d^{cmi}_{t_k}, d^{scm}_{t_k}\right)\right\}_{k=1}^{batch\_size/\tau_{max}}\right\}_{h=1}^{\tau_{max}}
\end{equation}
%Furthermore, to contrast with cases where \(X_{do}\) is not triggered, we also collect an equal amount of non-intervention data \(D_{undo}\), which is relatively straightforward. As long as \(X_{do}\) is not triggered, we can slide and collect data segments from the start of each episode until the required amount is reached. After that, \(D=\{D_{do}, D_{undo}\}\) gathered by each process \(rank_h\) is aggregated and then redistributed to all processes. Reverse data collection better traces the causes of state changes and captures causal relationships more effectively than the forward one.
%The data \(\mathbf{D}=\{D_{do}, D_{undo}\}\) collected by each process is aggregated and then redistributed to all processes.
%The data gathered in parallel by all processes is subsequently redistributed to each process.
Furthermore, to contrast with cases where \(X_{do}\) is not triggered, we collect an equal amount of non-intervention data \(D_{undo}\). This is done by sliding and collecting data segments from the start of each episode until the required amount is reached, as long as \(X_{do}\) is not triggered. The parallelly collected data \(\mathbf{D}=\{D_{do}, D_{undo}\}\) is then redistributed to all processes. Reverse strategy more effectively traces the multiple causes of state changes compared to the forward strategy.	% (see \ref{app:data_collection_phase})
%, which triggers one intervention variable and then identifies the resulting state changes.
% and captures causal relationships

\subsubsection{\textbf{Distributed SCM Training.}} 
% setting a CMI threshold of 0.05 effectively distinguishes spurious correlations from true causal relationships. 
%process \(rank_1\) is responsible for learning \(1\)-step causal relationships from \(t-1\) to \(t\) and process \(rank_4\) handles \(4\)-step long causal relationships from \(t-4\) to \(t\).
%Each process trims the collected data based on its assigned time span \(h\) and then proceeds with the training of \(SCM_h\). 
%is responsible for learning
Each process \(rank_h\) trains SCM under the time span \(h\) based on the collected data \(\mathbf{D}\). As shown in Figure~\ref{fig:overall}.A, given \(\tau_{max}=4\), process \(rank_1\) handles \(1\)-step causal relationships \(X^j_{t-1} \rightarrow X^i_t\) and process \(rank_4\) handles \(4\)-step causal relationships \(X^j_{t-4} \rightarrow X^i_t\). We adapted the SCM training methods from CDHRL and SDI~\cite{ke2019learning}. For \(SCM_h\) with \(M+N\) variables \(\{X_k\}_{k=1}^{M+N}\), we construct two parts of networks: generating functions \(F_h=\{f_h^i(\theta)\}_{i=1}^{M}\) and a causal graph's parameter matrix \(\eta_h \in \mathbb{R}^{M \times (M+N)}\).
% (see \ref{app:distributed_scm_training})
%its corresponding generating function network 
% (see Figure~\ref{fig:s_params} in \ref{app:distributed_scm_training} for the sampling process)
Generating functions: \(f^i_h(\theta)\) predicts the value of \(X^i\) given its direct cause set. During training, based on the sampled causal relationship matrix \(P_h\) and collected data \(\mathbf{D}\), we input the causes $\mathcal{PA}(X^i_t)=\{X^j_{t-h}|P_h[i][j]=1\}$ of each effect \(X^i_t\) into \(f^i_h(\theta)\) to evaluate the likelihood \(\mathcal{L}\). The true value \(X^i_t\) serves as the label to maximize \(\mathcal{L}\), thereby updating \(f^i_h(\theta)\):
%Generating functions: \(f^i_h(\theta)\) predicts the value of \(X^i\) given its direct cause set. During training, for each causal relationship \(X^j_{t-h} \rightarrow X^i_t\) in each sampled causal relationship matrix \(P_h\) (the sampling process is shown in Figure~\ref{fig:s_params} in \ref{app:distributed_scm_training}), the cause \(X^j_{t-h}\) is fed into \(f^i_h(\theta)\) to predict the logits of \(\widehat{X}^i_t\). To optimize the paramters of \(f^i_h(\theta)\), the true value \(X^i_t\) is used as the label to maximize the likelihood: 
\begin{equation}
	%Maximize \quad likelihood(C)=log\_softmax \big(p_{\theta_j} (X_t^j|X_{t-h}^i;C) \big)
	%\sum_{(X^i_{t-h} \xrightarrow{h} X^j_t) \in C} 
	%Max \ \ \mathcal{L}(P_h[i]) = \text{\textit{log\_softmax}} \big(f_h^i (X_t^i|X_{t-h}^j;\theta) \big)
	max \ \ \mathcal{L}(\theta|X_t^i) = \text{\textit{log\_softmax}} \big(f_h^i (X_t^i|\mathcal{PA}(X^i_t);\theta) \big)
\end{equation}
%Causal graph's parameter matrix: \(\sigma(\eta^{ij}_h)=1/{(1 + e^{-\eta^{ij}_h})}\) represents the probability that \(X^j\) is a direct cause of \(X^i\) with a time span \(h\). During training, we sample \(K\) batches of data \(\{D^k\}_{k=1}^K\) from \(\mathbf{D}\) and \(N_p\) causal relationship matrices \(\{P^n_h\}_{n=1}^{N_p}\), calculate the likelihood \(\mathcal{L}(P^n_h, D^k)\) of the effect \(X^i\) for each causal relationship \(X^j \xrightarrow{h} X^i\) in \(P^n_h\) and compute the gradient \(\nabla_{\eta_h}\) as follows:
Causal graph's parameter matrix: \(\sigma(\eta^{ij}_h)=1/{(1 + e^{-\eta^{ij}_h})}\) represents the probability that \(X^j\) is a direct cause of \(X^i\) with a time span \(h\). During training, we sample \(K\) batches of data \(\{D^k\}_{k=1}^K\) from \(\mathbf{D}\) and \(N_p\) causal relationship matrices \(\{P^n_h\}_{n=1}^{N_p}\), calculate the likelihood \(\mathcal{L}(P^n_h, D^k)\) for each effect \(X^i\) and compute the gradient \(\nabla_{\eta_h}\) as follows:\begin{equation}
	%\nabla_{\eta}=\sum_{k}\left\{\left[\sum_{n}\left(\sigma(\eta)-P^n_h\right)\right] \left( \frac{e^{\sum_{n}\left(\mathcal{L}(P^n_h)^{(k')}\right)}}{\sum_{k}e^{\Sigma_{n}\left(\mathcal{L}(P^n_h)^{(k)}\right)}} \right) \right\}
	\nabla_{\eta_h}=\sum_{k}\left[\sum_{n}\left(\sigma(\eta)-P^n_h\right)\right] \left( \frac{e^{\sum_{n}\mathcal{L}(P^n_h, D^k)}}{\sum_{k'}e^{\Sigma_{n}\mathcal{L}(P^n_h, D^{k'})}} \right) 
\end{equation}
In each iteration, matrix \(\eta_h\) is trained after functions \(F_h\) have converged. Then we use \(\sigma(\eta_h)\) to preliminarily identify causal relationships where causes are in \(List_{do}\) and effects are not, ensuring the learned causal graph remains a DAG. Typically, if \(\sigma(\eta^{ij}_h) \geq 0.8\), then \(X^j \xrightarrow{h} X^i\) exists~\cite{hu2022causality}.% We conduct sensitivity tests to different thresholds in \ref{app:sensitivity_to_causal_threshold}. 
%intervention variables $X_{do} \in List_{do}$
%This ensures that no cycles are introduced, maintaining the learned causal graph as a DAG
%Then we can use \(\sigma(\eta_h)\) to preliminarily identify causal relationships.

\subsection{Spurious Correlation Detection}\label{sec:global_causal_determination}
%In causal theory, common causes can induce non-zero information flow between variables that originally have no causal relationship, resulting in the misidentification, known as spurious correlations. 
%In causal theory, 
Common causes can create non-zero information flow between variables that originally have no causal relationship, leading to spurious correlations. Besides, autocorrelation may result in the identification of causal relationships with incorrect time spans, another form of spurious correlation. Therefore, after training \(SCM_h\), it is essential to filter out spurious correlations to ensure the validity of the learned causal graph. 
%See \ref{app:spurious} for details.
%The pseudocode is shown in Algorithm~\ref{alg:global_causal_deter} in \ref{app:spurious}.
%resulting in the misidentification when using SCM to infer causal links,

\subsubsection{\textbf{Conditional Independence Testing.}}
%As discussed in Section~\ref{sec:cit}, CIT helps identify spurious correlations. In fact, CIT can be transformed into the calculation of whether the corresponding Conditional Mutual Information (CMI) is greater than zero~\cite{wang2022causal,runge2018causal}: \(X \perp\!\!\!\perp Y|Z\Leftrightarrow I(X;Y|Z)=0\). Therefore, we use CMI to perform CIT for a given correlation \(X^j_{t-h} \rightarrow X^i_t\):
%By assessing the information flow along this path, we can determine if the correlation is spurious. 
%As discussed in Section~\ref{sec:cit}, CIT helps identify spurious correlations.
As shown in Figure~\ref{fig:overall}.B, the causality \textit{stone}$\xleftarrow{3}$\textbf{A}$\xrightarrow{3}$\textit{wood} and \textit{wood}'s autocorrelation create an information flow between $X^{wood}_{t-4}$ and $X^{stone}_{t}$, leading the SCM to misidentify a spurious correlation as a causal link. As discussed in Section~\ref{sec:cit}, conditioning on the parents (denoted by \includegraphics[width=0.02\textwidth]{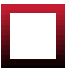}) blocks backdoor paths, ensuring $X^{wood}_{t-4} \rightarrow X^{stone}_{t}$ is the only open path between them. Then we can use CIT to determine whether this correlation is spurious. In fact, CIT can be transformed into the evaluation of whether the corresponding Conditional Mutual Information (CMI) is greater than zero~\cite{wang2022causal,runge2018causal}:$X \perp\!\!\!\perp Y \mid Z \Leftrightarrow I(X;Y \mid Z) = 0$. Thus we use CMI to perform CIT for any given correlation $X^j_{t-h} \rightarrow X^i_t$:\begin{equation}\label{eq:cit}
	\begin{aligned}
		&I\Big(X_{t-h}^{j};X_{t}^{i}|\mathcal{PA}(X_{t}^{i})\setminus X_{t-h}^{j},\mathcal{PA}(X_{t-h}^{j})\Big) \\
		=& H\Big[X_{t}^{i}|\mathcal{PA}(X_{t}^{i})\setminus X_{t-h}^{j},\mathcal{PA}(X_{t-h}^{j})\Big] \\
		&\qquad \qquad - H\Big[X_{t}^{i}|\mathcal{PA}(X_{t}^{i}),\mathcal{PA}(X_{t-h}^{j})\Big] \\
		=& E\Big[-\log p\left(X_{t}^{i}|\mathcal{PA}(X_{t}^{i})\setminus X_{t-h}^{j},\mathcal{PA}(X_{t-h}^{j})\right) \\
		&\qquad \qquad + \log p\left(X_{t}^{i}|\mathcal{PA}(X_{t}^{i}),\mathcal{PA}(X_{t-h}^{j})\right)\Big]
	\end{aligned}
\end{equation}
%After training, according to Equation~\ref{eq:cit}, we used the data \(\mathbf{D}\) collected before to evaluate the expected CMI of the correlation
%we randomly sampled a sequence data of \(2\tau_{max} + 1\) consecutive steps from replay buffer \(\mathbf{R}\).
For each effect variable \(X^i\), we construct a 3-layer fully connected neural network \(g(\phi_i)\) to compute the likelihood conditioned on the parents: \(\widehat{p}_{\phi_i} (X_{t}^{i}|\mathcal{PA}(X_{t}^{i})\setminus X_{t-h}^{j},\mathcal{PA}(X_{t-h}^{j}) )\) and \(\widehat{p}_{\phi_i} (X_{t}^{i}|\mathcal{PA}(X_{t}^{i}),\mathcal{PA}(X_{t-h}^{j}))\). During training, we randomly sample sequences of \(2\tau_{max} + 1\) consecutive data points from replay buffer \(\mathbf{R}\). Input the values of \(\mathcal{PA}(X_{t}^{i})\) and \(\mathcal{PA}(X_{t-h}^{j})\) into \(g(\phi_i)\), mask either \(\{\)\textit{None}, \(X_{t-h}^j\}\) randomly and update \(\phi_i\) by maximizing the likelihood until convergence. After training, we use the collected data \(\mathbf{D}\) to evaluate the expected CMI of the correlation according to Equation~\ref{eq:cit}:\begin{equation}\label{eq:cmi}
	\begin{aligned}
		&\widehat{I}_{\phi_i} \big(X_{t-h}^{j} \rightarrow X_{t}^{i}\big)\\
		&=E\Big[-\log \widehat{p}_{\phi_i}\left(X_{t}^{i}|\mathcal{PA}(X_{t}^{i})\setminus X_{t-h}^{j},\mathcal{PA}(X_{t-h}^{j})\right) \\
		&\qquad +\log \widehat{p}_{\phi_i} \left(X_{t}^{i}|\mathcal{PA}(X_{t}^{i}),\mathcal{PA}(X_{t-h}^{j})\right)\Big]
	\end{aligned}
\end{equation}
%In theory, \(\widehat{I}_{\phi_i} \big(X_{t-h}^{j} \rightarrow X_{t}^{i}\big)=0\) implies a spurious correlation, while a value greater than 0 indicates a causal relationship. 
In practice, due to data errors, we consider correlations with CMI above threshold \(\epsilon_{cmi}\) as genuine causal relationships:\begin{equation}
	\widehat{I}_{\phi_i} \big(X_{t-h}^{j} \rightarrow X_{t}^{i}\big) > \epsilon_{cmi} \Rightarrow X^j_{t-h} \rightarrow X^i_t
\end{equation}

\subsubsection{\textbf{True Causal Relationships and Time Span Determination.}}\label{sec:time_span}
A causal relationship between two variables might be judged to be valid across multiple different time spans. TSGM reveals that, apart from the  true time span \(h_i\), other cases may arise from the combination of a direct causal relationship (\(X^j_{t-h} \xrightarrow{h_i} X^i_t\)) and a variable's autocorrelation (\(X^i_t \rightarrow X^i_{t+1}\)), leading to indirect causation (\(X^j_{t-h} \xrightarrow{h_i+1} X^i_{t+1}\)). Typically, causal paths with collider structures lead to independence, making it unlikely for a causal relationship to appear with a time span shorter than the true time span. Therefore, if \(X^j \rightarrow X^i\) is judged to be valid across multiple different time spans, the shortest one is identified as true.

Throughout training, any correlation identified by SCM is tested for spuriousness. Once confirmed as a causal relationship and successfully trained as a subgoal, it is not retested. Furthermore, it is important to note that early intervention data may not be sufficient to uncover deeper causality. Spurious correlations identified in earlier iterations might become genuine later. Thus, each iteration's identification of spurious correlations serves as a reference only for that iteration.
% and not as a basis for subsequent iterations

\subsection{Hierarchical Policy Training}\label{sec:hierarchical_network}
%~\cite{hu2022causality}
D3HRL adopts the progressive causal chain learning idea from CDHRL~\cite{hu2022causality}, constructing and training hierarchical policies based on the learned causal chain. %The similarities and differences between CDHRL and D3HRL are provided in \ref{app:difference}.

\subsubsection{The Construction of the Hierarchical Policy Network.}
We implement D3HRL based on multi-level DQN~\cite{mnih2015human} with HER~\cite{andrychowicz2017hindsight}\footnote{We implement D3HRL based on the code of CDHRL}. In each iteration, after determining the true causality, a sub-goal network is established for each newly-learned effect variable with its action space set to include the causes of the effect and the primitive action space \(\mathbf{A}\). For example, as shown in Figure~\ref{fig:overall}.C, we add a new DQN sub-goal network for the newly-learned effect variable \(\text{\textit{stonepickaxe}}\) (denoted as \(X^{sp}\)), given the newly-learned causality (\textit{stone}\(\xrightarrow{h_{sp}}\)\textit{stonepickaxe}\(\xleftarrow{h_{sp}}\)\textit{stick}). It's action space is set to \(A_{sp}=\{g_{stone},\ g_{stick},\ \mathbf{A}\}\). In addition, there is always a top-level network with action space \(A_{top}=\{\)\textit{trained sub-goals, currently training sub-goals}, \(\mathbf{A}\}\). When the causal chain in the environment is fully explored, the resulting hierarchical policy network closely mirrors the causal chain. At this point, we simply input the target into the top-level network, which can then recursively trace the causes and sequentially achieve them.
\subsubsection{The Training of the Hierarchical Policy Network.}
%During the interaction between the \(X^{sp}\)'s sub-goal network and the environment, we collect transition data. 
During the training of the sub-goal network for \(X^{sp}\), it learns to invoke the sub-goal networks for \(stone\) (\textit{action}=\(g_{stone}\)) and \(stick\) (\textit{action}=\(g_{stick}\)) to increase or decrease their quantities, and to generate \(\text{\textit{stonepickaxe}}\) (sub-goal \(g_{sp}\)) after obtaining the necessary materials (\textit{action} \(\in \mathbf{A}\)). To handle variable-length state transitions, we adapt the hindsight transition collection mechanisms of CDHRL and HER. Specifically, at each time step \(t\) during interaction with the environment, we log the 1-step transition and extract the continuous \(h_{sp} + 1\) steps of data: \(\{s_{t-h_{sp}}, a_{t-h_{sp}},\ldots, s_t, a_t\}\). Based on whether $g_{sp}$ is achieved within this period, we generate hindsight goal transition: \([\)\textit{initial state=}\(s_{t-h_{sp}}\), \textit{action=}\(a_{t-h_{sp}}\), \textit{reward=}\(0\) \textit{if achieved} \(g_{sp}\) \textit{else} \(-1\), \textit{next state=}\(s_t\), \textit{goal=}\(g_{sp}\), \textit{discount=}\(\gamma]\). We then obtain the sub-goals achieved during this period, denoted as \(G_{ach}\). For each sub-goal \(g_i \in G_{ach}\), we generate hindsight action transition: \([\)\textit{initial state=}\(s_{t-h_{sp}}\), \textit{action=}\(g_i\), \textit{reward=}\(0\) \textit{if achieved} \(g_{sp}\) \textit{else} \(-1\), \textit{next state=}\(s_t\), \textit{goal=}\(g_{sp}\), \textit{discount=}\(\gamma]\). Each sub-goal network is trained on the collected hindsight transition data and, after successfully trained, added to \(List_{do}=\{\mathbf{A}, X^{sp}\}\) for the next iteration to begin.
%and the next iteration begins.
%the effect variable \(X^{sp}\)

\section{Experiment}

\subsection{Experiment Setup}
%\textbf{Environment.} 
\paragraph{Environment}
%2D-MineCraft~\cite{sohn2018hierarchical} and MiniGrid~\cite{MinigridMiniworld23}
We introduce two environments: 2D-MineCraft~\cite{sohn2018hierarchical} and MiniGrid~\cite{MinigridMiniworld23}. Both feature long-horizon sequential decision-making tasks with rewards provided only upon task completion. We designed two tasks for each: GetIron and GetSilverore for 2D-MineCraft, and Fire2Burn and Wood2Wet for MiniGrid. Take the GetIron task as an example, where the agent must obtain iron within a limited time. This involves sequentially gathering wood and stone, crafting sticks, creating a stone axe, collecting iron ore and coal, and finally smelting the iron. All tasks in our setting are sequential decision-making tasks like this. We modified these tasks to support variable-length state transitions. "T0\(\sim\)3" suffixes denote various time span configurations, where "T0" indicates that all state transitions in the environment are single-step ($\tau_{max}=1$), and "T1", "T2", and "T3" represent three distinct configurations of state transition spans under $\tau_{max}=4$. "R0\(\sim\)1" suffixes denote resource levels: "R1" for abundant, "R0" for scarce. %See \ref{app:environment} for details. 

%Further details are provided in \ref{app:environment}.
%Suffixes starting with "T" denote different time span configurations. Suffixes starting with "R" denote different resource configurations: "R1" indicates abundant resources, while "R0" indicates scarce resources. Further details are provided in \ref{app:environment}.
%In MineCraft, the agent must gather materials and craft tools within a limited number of steps to obtain iron or silver ore; the causal chains are deep (depth 5 for GetSilverOre and 7 for GetIron) but the environment is resource-rich. For MiniGrid, in Fire2Burn, the agent must pick up a flame and ignite wood, while in Wood2Wet, it must find water to extinguish burning wood; these tasks have shorter causal chains (both depth 2) but face challenges due to scarce resources and irrelevant causal chains. 

\paragraph{Evaluation Metrics}
We use Average Success Ratio (\textbf{ASR}) and Average Distance to Complete task (\textbf{ADC}) over the most recent 100 episodes during training as metrics. Higher ASR indicates better task completion, and lower ADC indicates deeper causal chain exploration. We also use Structural Hamming Distance (\textbf{SHD}) to measure the accuracy of the learned causal relationship graphs. Lower SHD indicates a more accurate causal relationship graph. All results are averaged over 5 random seeds.

%\textbf{Baselines.} 
\paragraph{Baselines}
We compare D3HRL with CDHRL~\cite{hu2022causality}, HAC~\cite{levy2017learning}, Option-Critic~\cite{bacon2017option} and LESSON~\cite{kim2023lesson}. \textbf{CDHRL} and D3HRL are both CHRL methods that learn goal-based hierarchical policies using sequential causal chain learning. \textbf{HAC} is a powerful goal-based HRL method that utilizes three types of transitions to steadily train hierarchical policies. \textbf{Option-Critic} is the first framework to use options for modeling variable-length state transitions. \textbf{LESSON} is a unified exploration framework based on Option-Critic. 
\textbf{We enhanced the baselines to identify causal relationships and adapt to variable-length state transitions.} For algorithms that cannot identify causal relationships, we designed a curriculum to help them learn sub-goals progressively. For those that cannot perceive variable-length state transitions, we set the transition data collection span to \(\tau_{max}\), enabling them to perceive transitions within \(\tau_{max}\) steps. %More details are in \ref{app:baselines}.
%\textbf{We enhanced the baselines to adapt to long-horizon tasks with variable-length state transitions.} For algorithms that cannot identify causal relationships, we designed a curriculum to help them learn sub-goals progressively. For those that cannot perceive variable-length state transitions, we set the transition data collection span to \(\tau_{max}\), enabling them to perceive transitions within \(\tau_{max}\) steps. More details are in \ref{app:baselines}.
%Its options integrate a range of exploration strategies, enabling the agent to adaptively choose the appropriate strategy based on the task.
%It utilizes three types of transition data collection methods to steadily train policies at different levels of the hierarchical network.

\subsection{Experimental Results}\label{sec:result}
We design experiments to answer the following questions:
\subsubsection{How does D3HRL perform on long-horizon sequential decision-making tasks compared to the baselines?}
\begin{figure*}[htbp]
	\centering
	\setlength{\tabcolsep}{-4pt} % 设置列间距为 2pt
	\begin{tabular}{cc}
		\subcaptionbox{GetIron-R0-T1}{
			\includegraphics[width=0.25\textwidth]{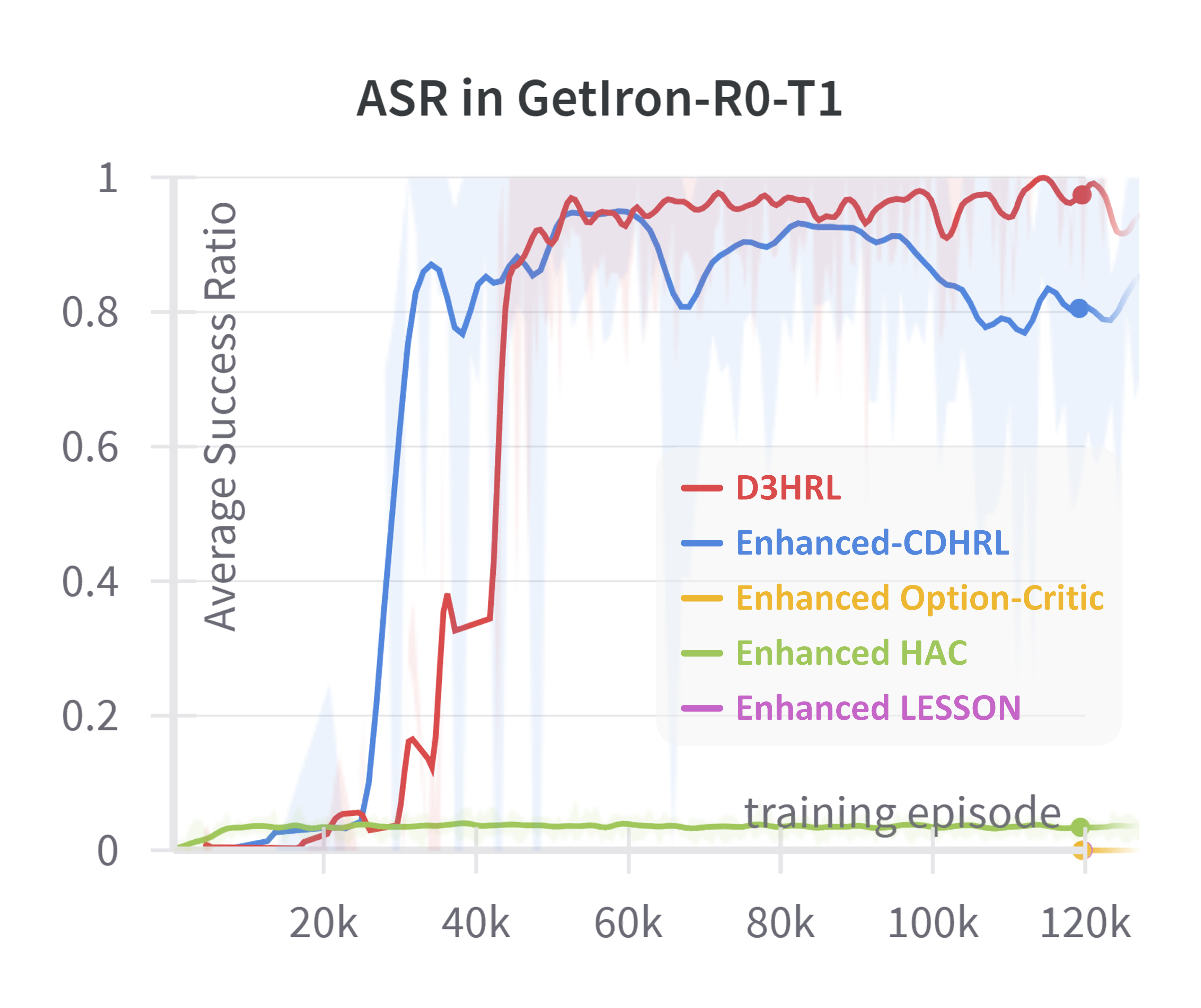}
			\hspace{-0.25cm}% 调整这里来控制图片间的间距
			\includegraphics[width=0.25\textwidth]{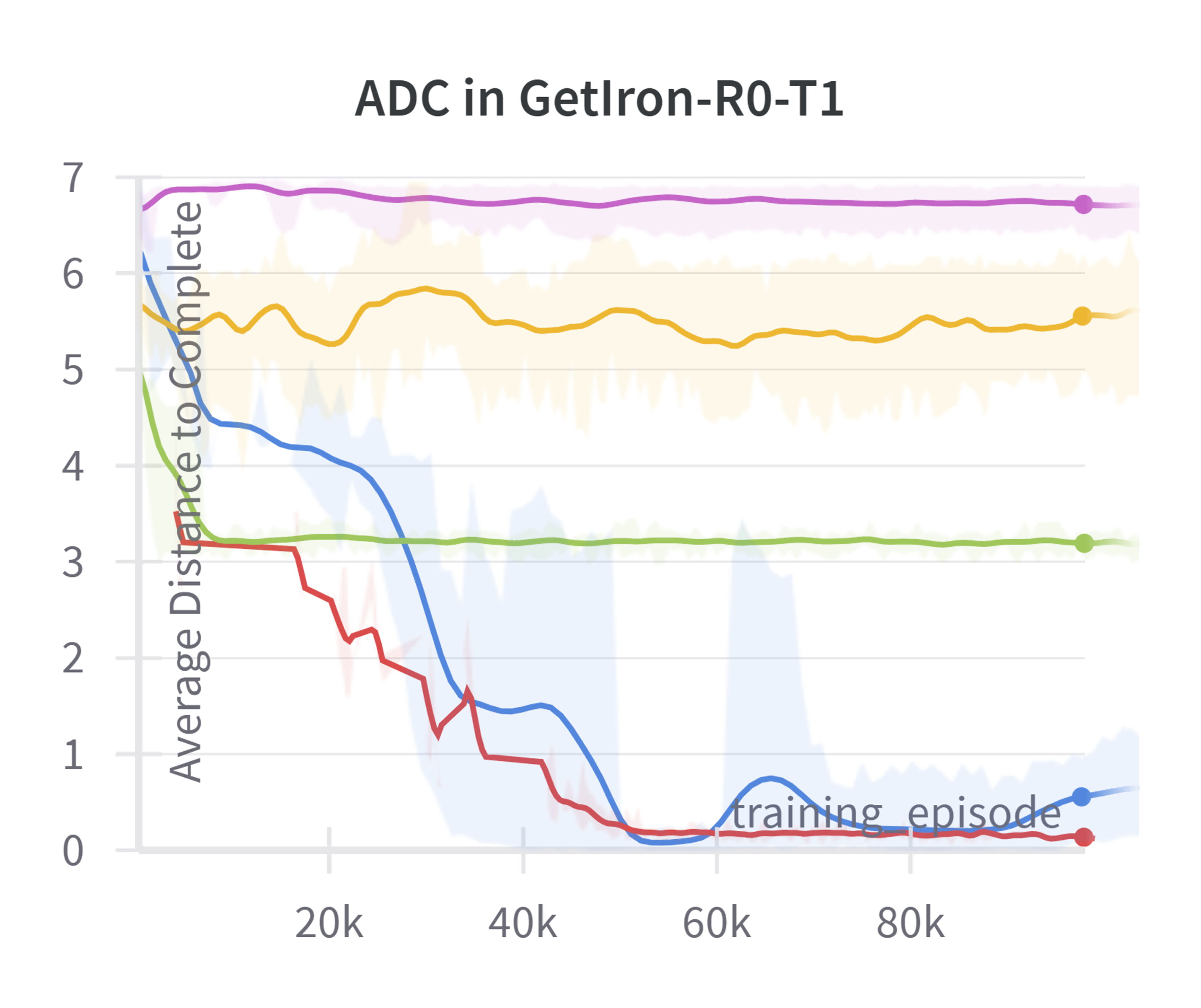}} &
		\subcaptionbox{GetIron-R1-T1}{
			\includegraphics[width=0.25\textwidth]{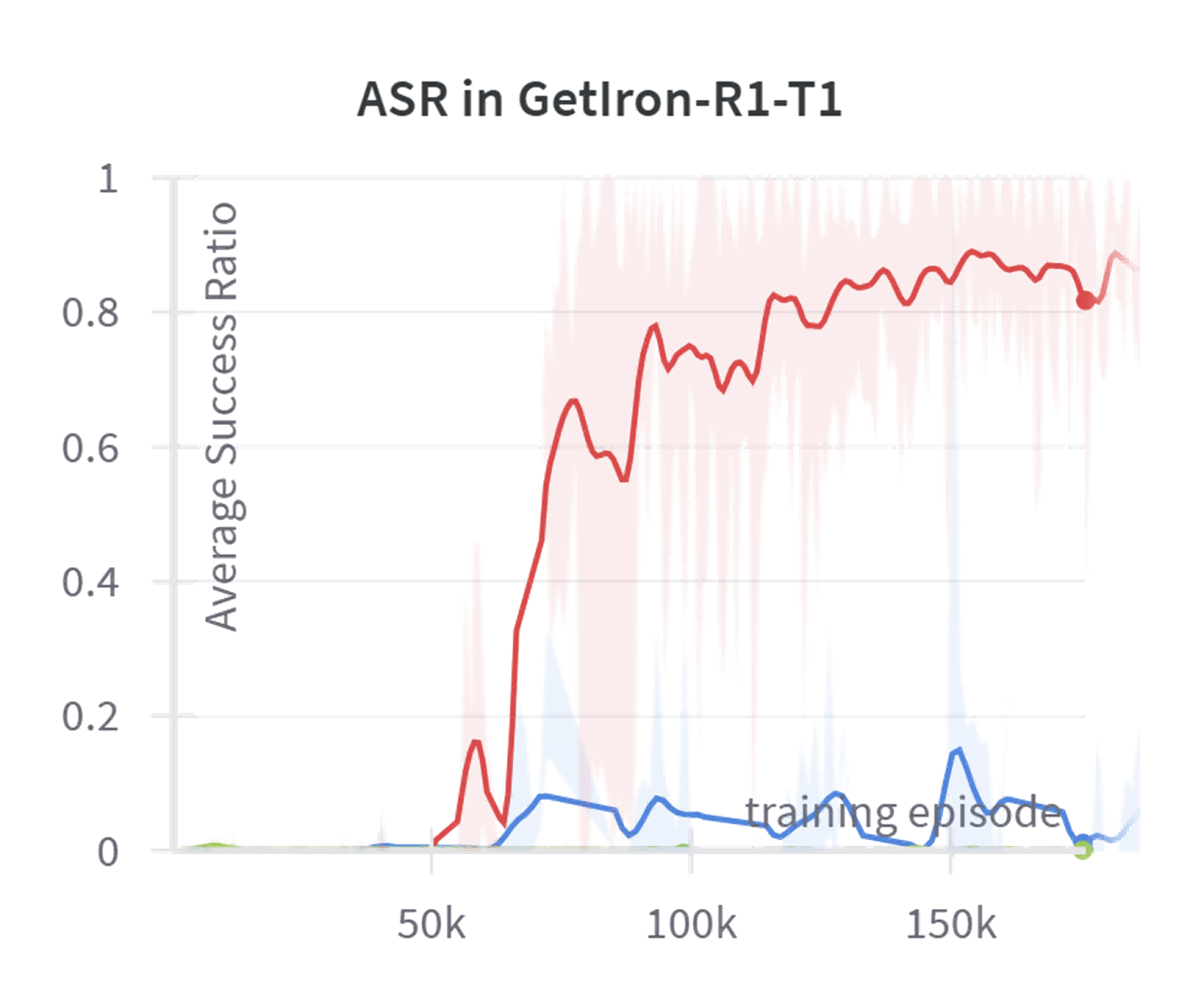}
			\hspace{-0.25cm}% 调整这里来控制图片间的间距
			\includegraphics[width=0.25\textwidth]{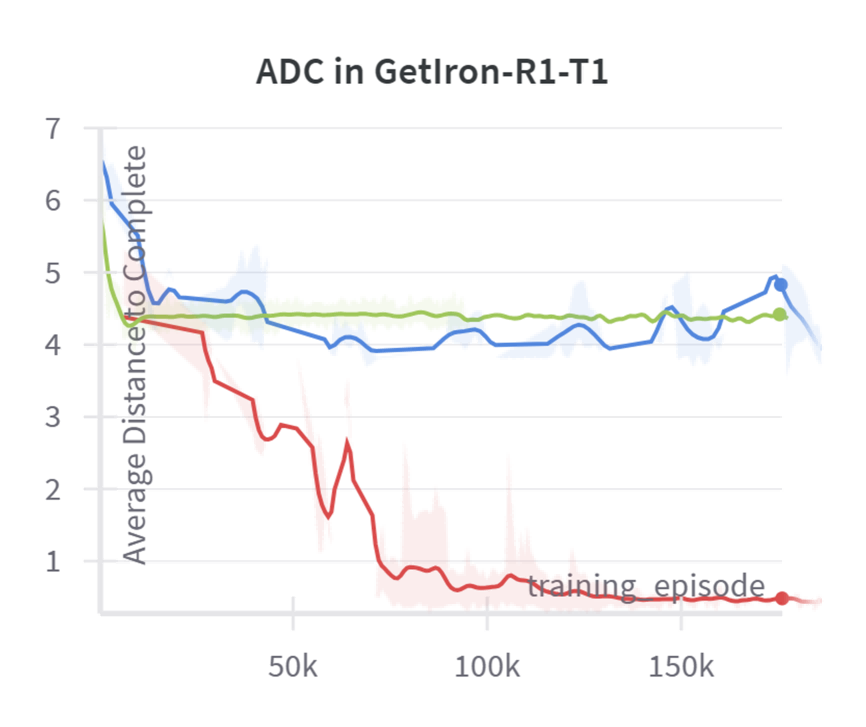}} \\
		\subcaptionbox{GetSilverore-R0}{
			\includegraphics[width=0.25\textwidth]{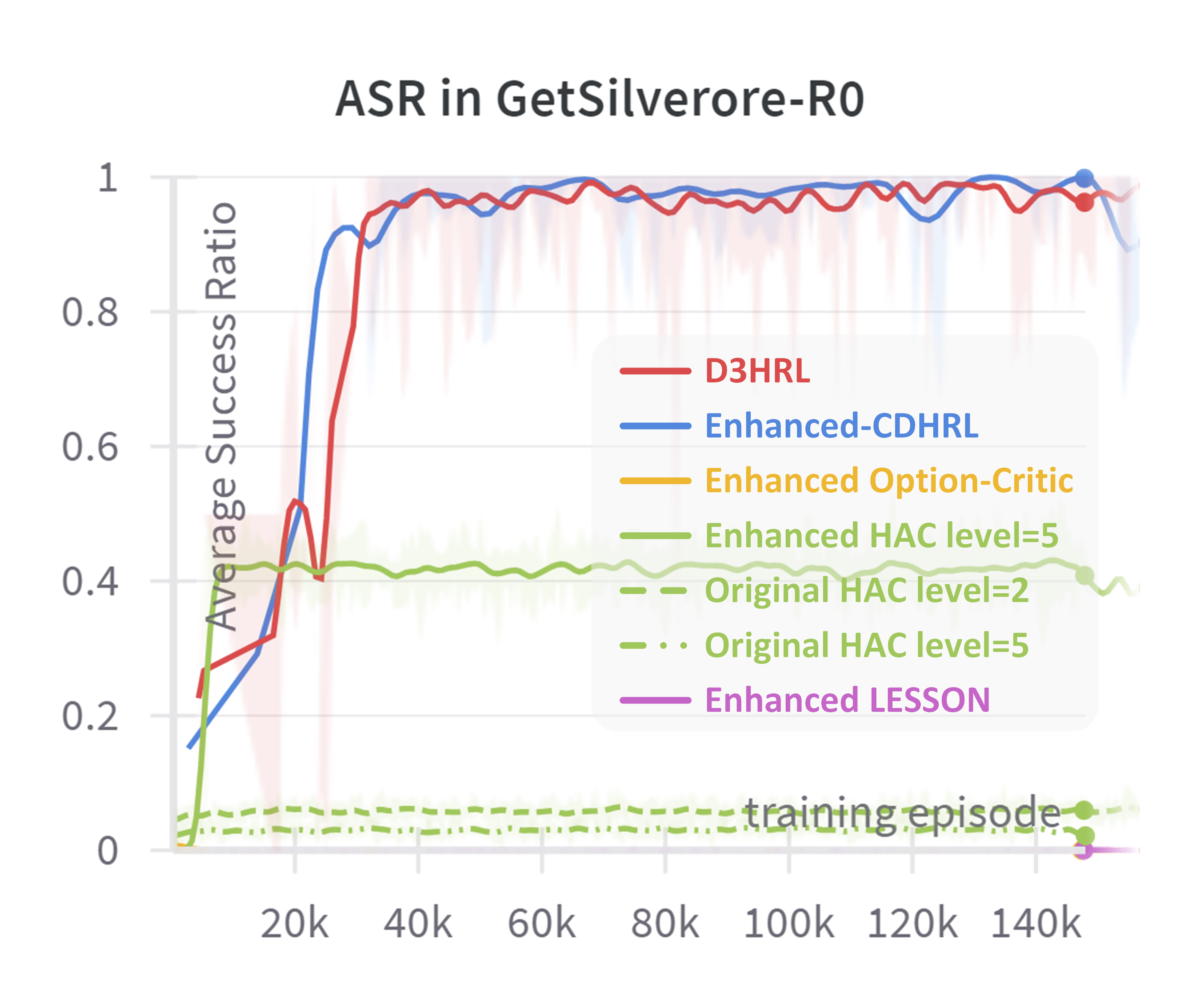}
			\hspace{-0.25cm}% 调整这里来控制图片间的间距
			\includegraphics[width=0.25\textwidth]{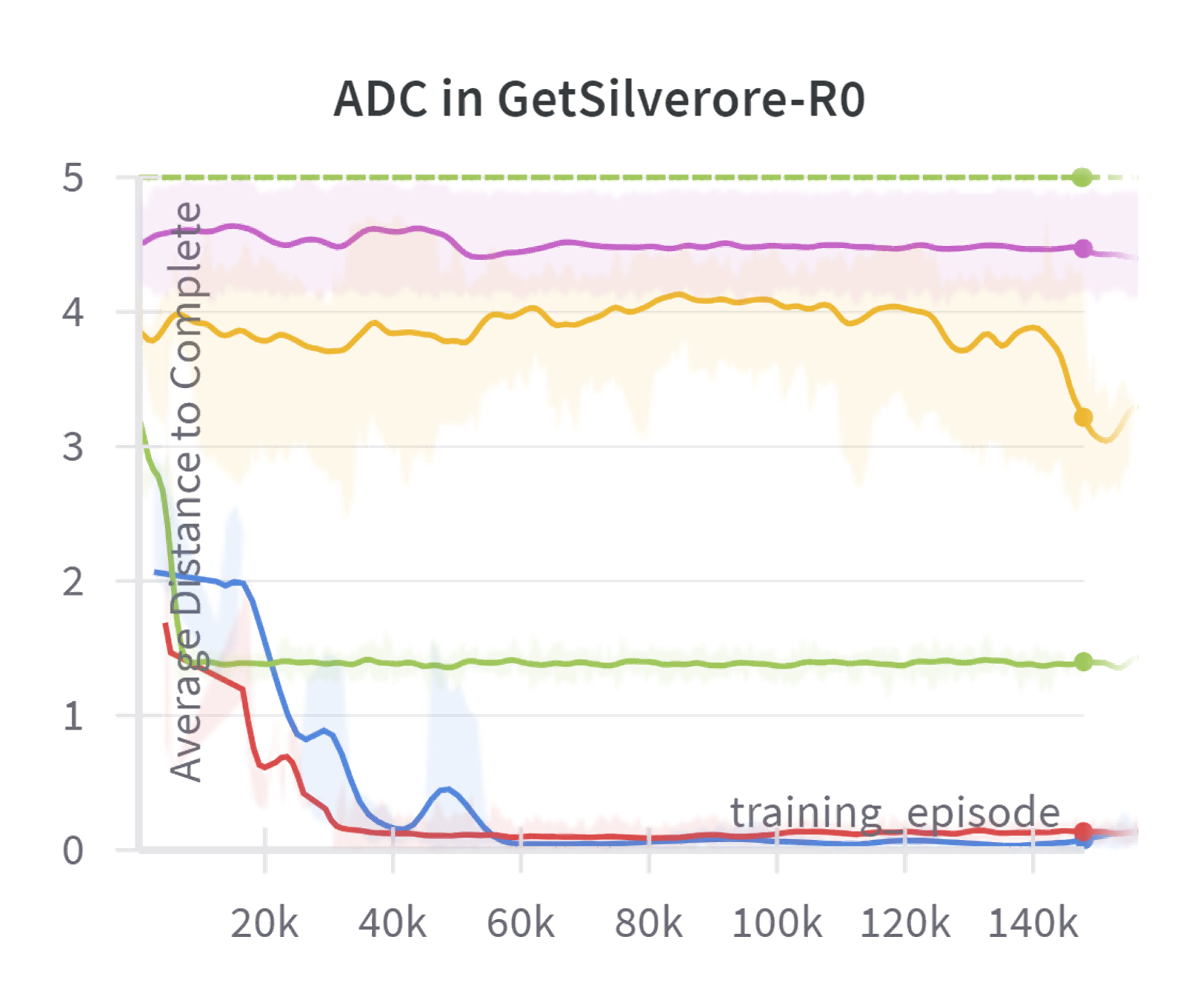}} &
		\subcaptionbox{GetSilverore-R1}{
			\includegraphics[width=0.25\textwidth]{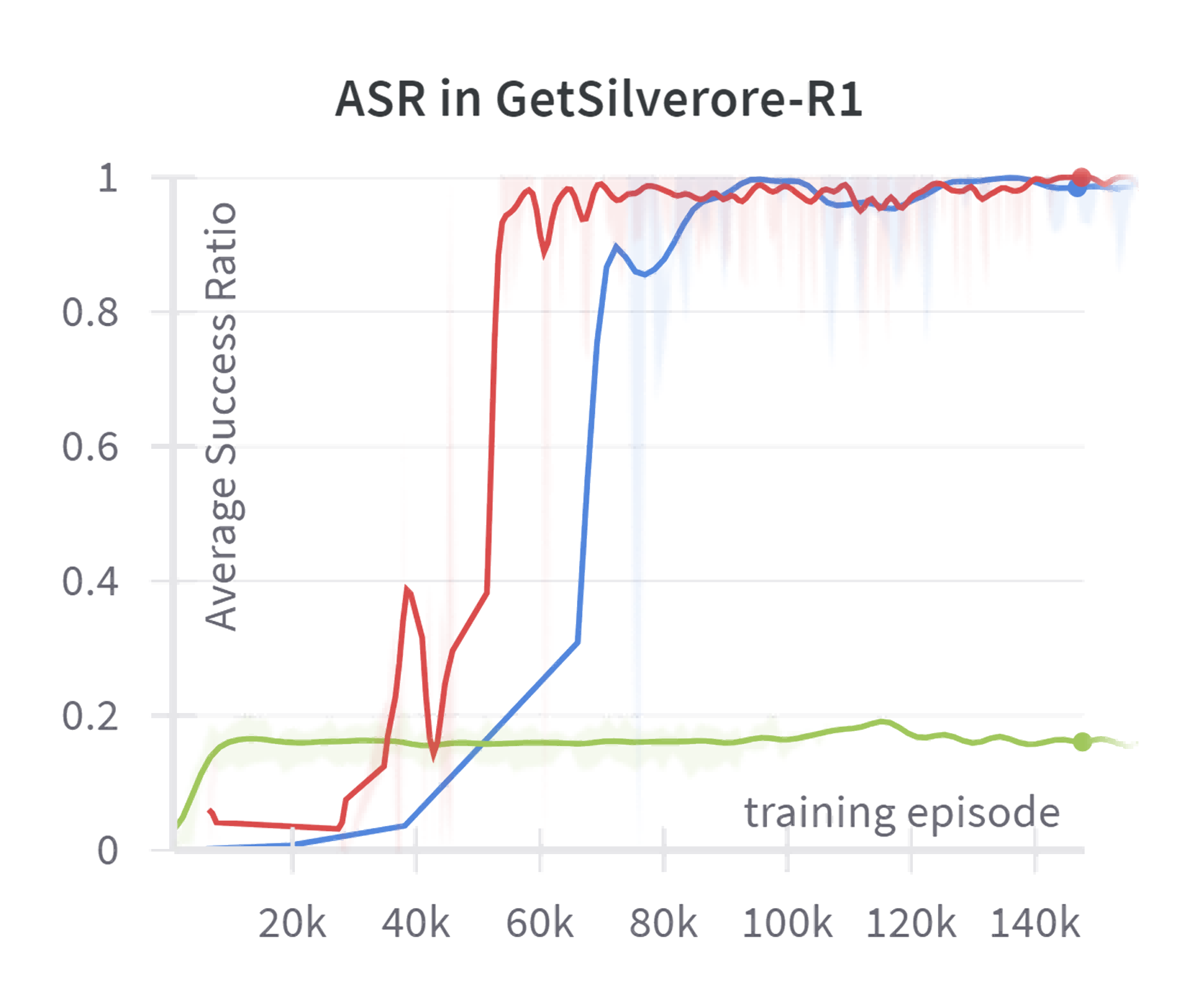}
			\hspace{-0.25cm}% 调整这里来控制图片间的间距
			\includegraphics[width=0.25\textwidth]{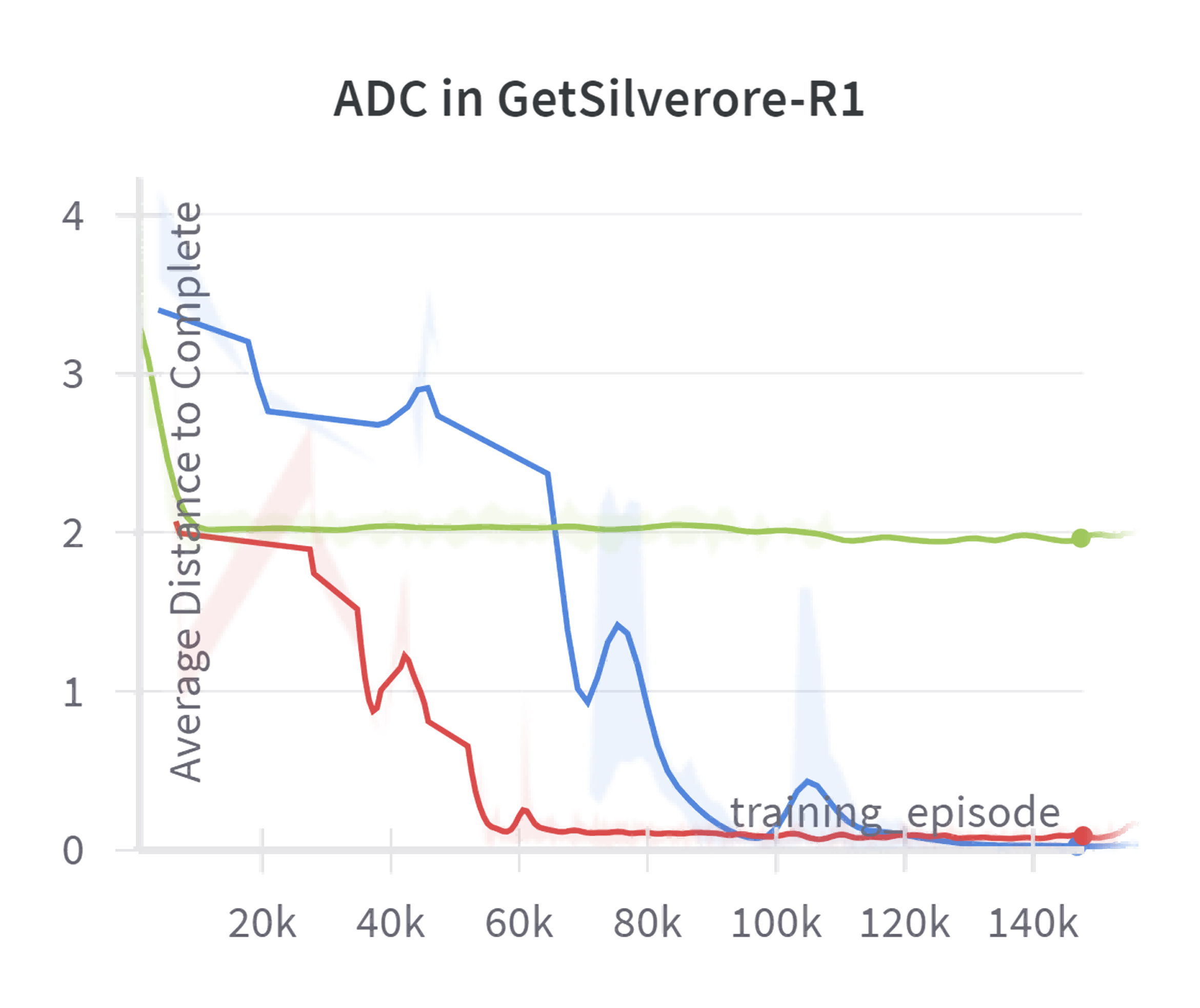}}\\
		\subcaptionbox{Wood2Wet}{
			\includegraphics[width=0.25\textwidth]{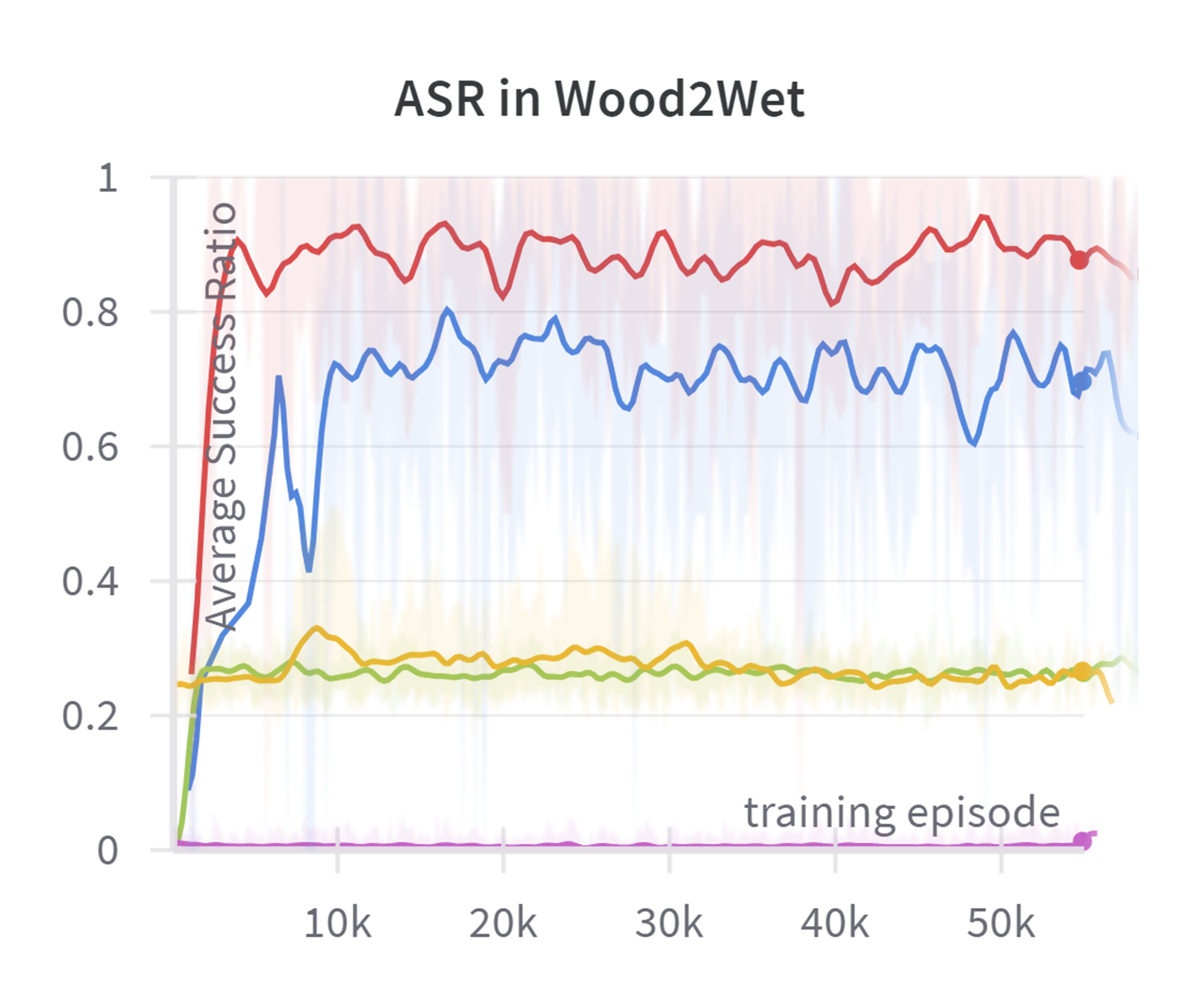}
			\hspace{-0.25cm}% 调整这里来控制图片间的间距
			\includegraphics[width=0.25\textwidth]{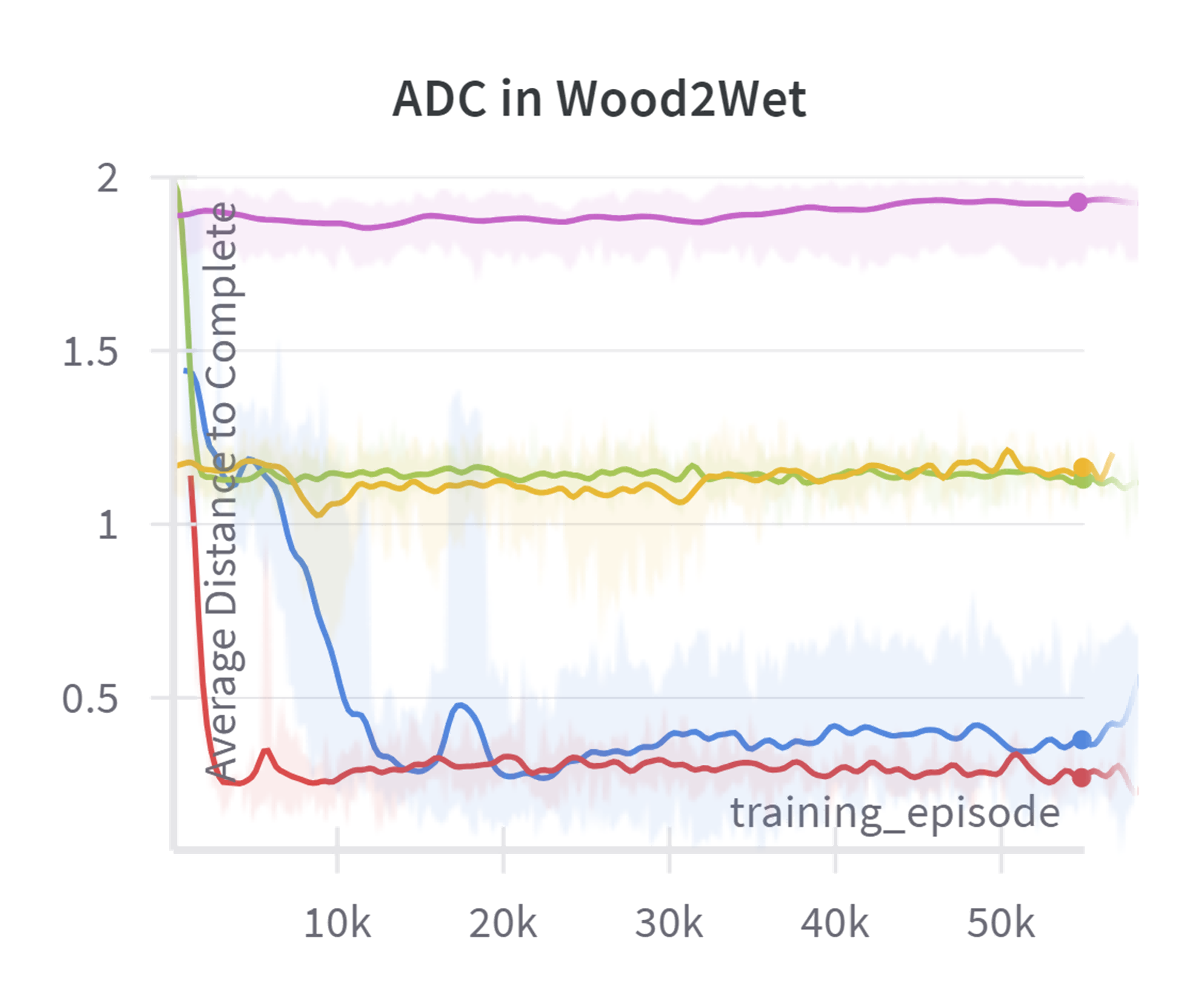}} &
		\subcaptionbox{Fire2Burn}{
			\includegraphics[width=0.25\textwidth]{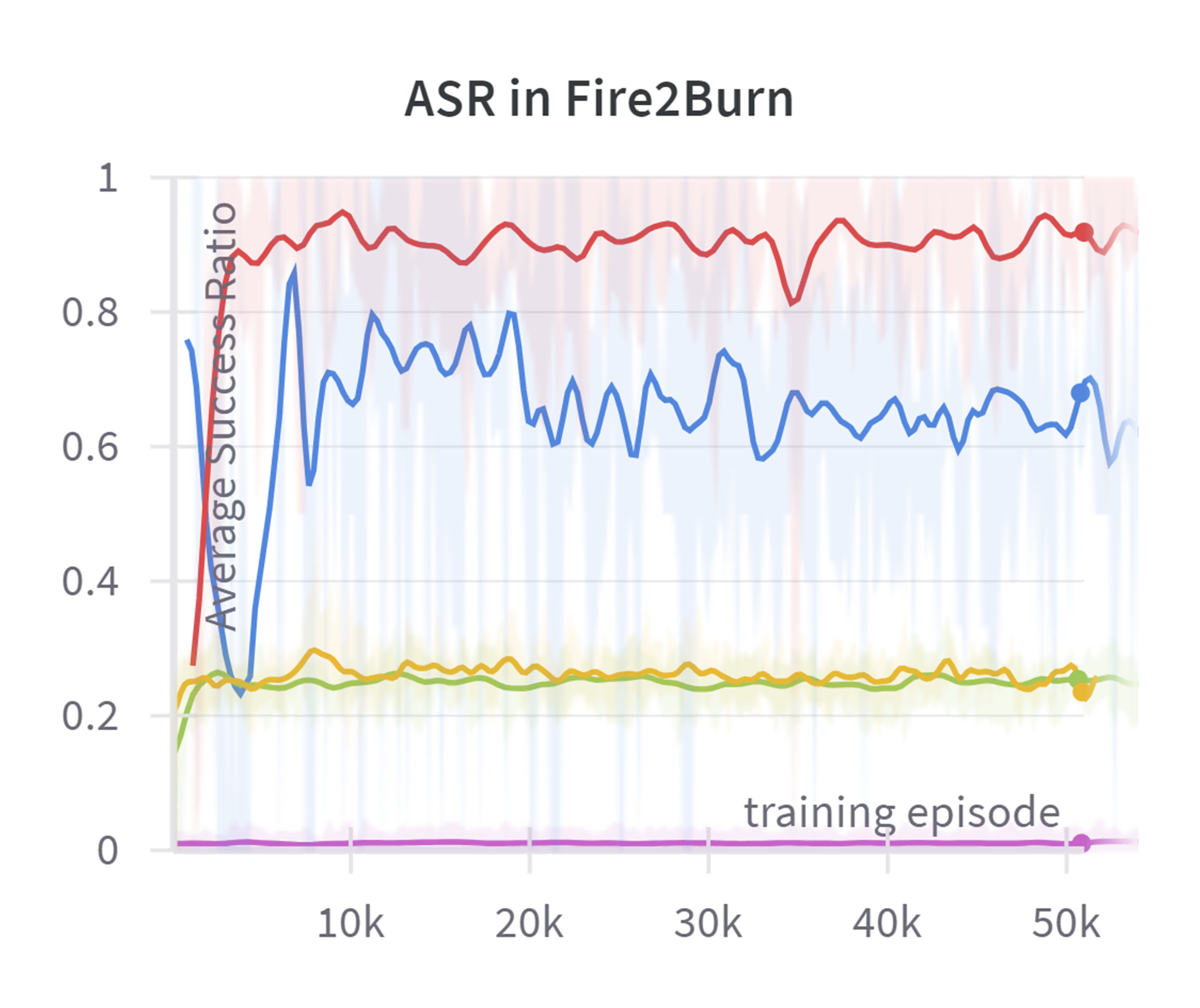}
			\hspace{-0.25cm}% 调整这里来控制图片间的间距
			\includegraphics[width=0.25\textwidth]{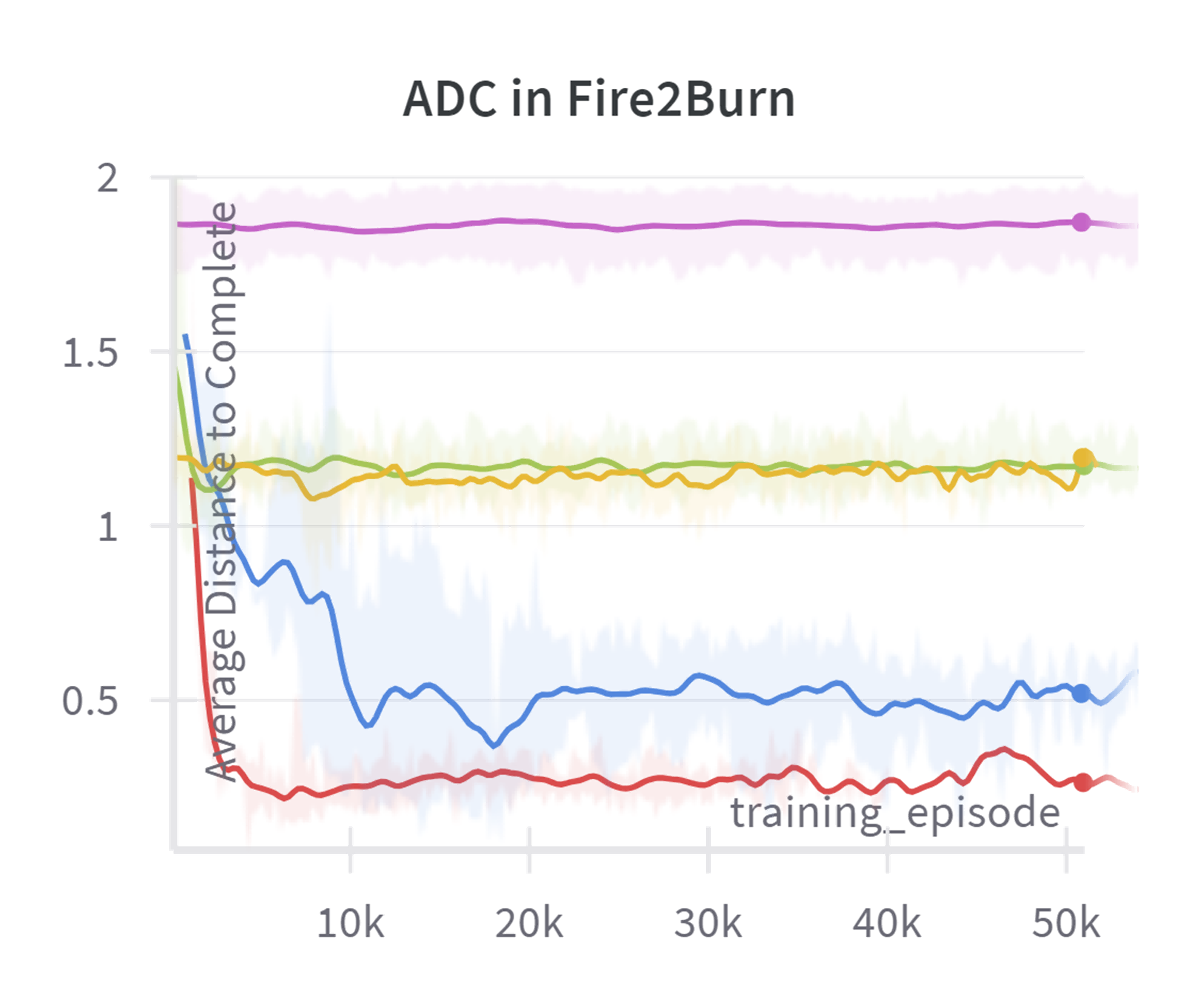}}
	\end{tabular}
	\caption{ASR and ADC under MiniGrid and MineCraft tasks.}
	% with different resource configurations
	\label{fig:ave_success_ratio}
\end{figure*}

%\noindent\textbf{1. How does D3HRL perform on long-horizon sequential decision-making tasks compared to the baselines?}
%\noindent\textit{\textbf{Experimental Design.}} 
\paragraph{Experimental Design} We compared D3HRL with the enhanced baselines in terms of ASR and ADC on four tasks with different configurations. The results are shown in Figure~\ref{fig:ave_success_ratio}. 

\paragraph{Experimental Results} \textbf{D3HRL} effectively identifies variable-length state transitions and filters out spurious correlations, performing best even in tasks with scarce resources. \textbf{CDHRL} initially outperforms D3HRL by learning indirect causality, enabling early exploration of deeper causality. However, its inability to accurately identify causality leads to greater complexity in sub-goal training, ultimately resulting in suboptimal performance. \textbf{HAC} can only explore up to 3.8 layers of the causal chain. Unlike CDHRL and D3HRL, which learn causality autonomously, HAC uses pre-defined curriculum to directly acquire causality, leading to earlier success. However, it struggles with variable-length state transitions. Multiple sub-goals in the same layer, with varying time spans, complicate training and hinder progress on deeper goals. Scarce resources make it hard for HAC and CDHRL to gather materials, resulting in low success. \textbf{Option-Critic} performs well in MiniGrid tasks with shallow causal chains but struggles in MineCraft tasks with deeper causal chains. While it can handle variable-length state transitions, its lack of a multi-layer policy structure prevents it from recalling previously learned sub-goals, despite having a curriculum for sequential sub-goal learning. This leads to decreasing efficiency in sub-goal learning and failure to complete the subsequent curriculum, especially in tasks with deep causal chains. \textbf{LESSON} consistently fails across all tasks since it lacks a multi-layer policy structure and cannot remember previously learned sub-goals, and it treats each option as an exploration strategy rather than a skill, making it ineffective in training sub-goals. %The time complexity of each algorithm is analyzed in \ref{app:each_algo_time_complexity}.
%consistently  as training progresses, 
%Training becomes complex when multiple sub-goals within the same layer have different time spans. This increases training difficulty and hinders the progress of training deeper sub-goals.
%Training is complicated by multiple sub-goals in the same layer but with different time spans, leading to great complexity for training them and hindering deeper sub-goals training.

\subsubsection{Experiment Validation: How does the reverse data collection strategy compare to the forward data collection strategy?}
\begin{figure*}[htbp]
	\centering
	\setlength{\tabcolsep}{1pt} % 设置列间距为 2pt
	\begin{tabular}{ccccc}
		\includegraphics[width=0.2\textwidth]{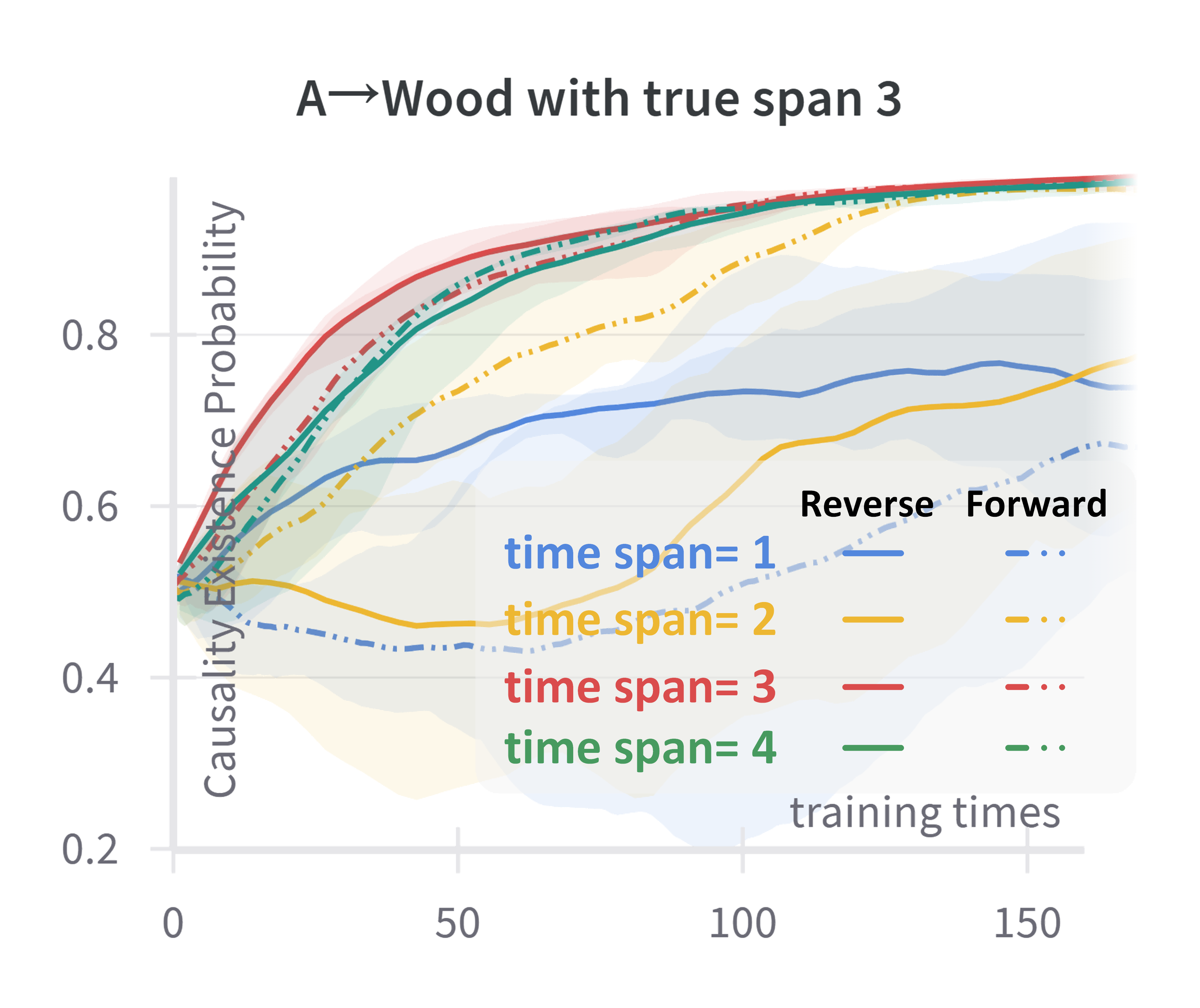} &
		\includegraphics[width=0.2\textwidth]{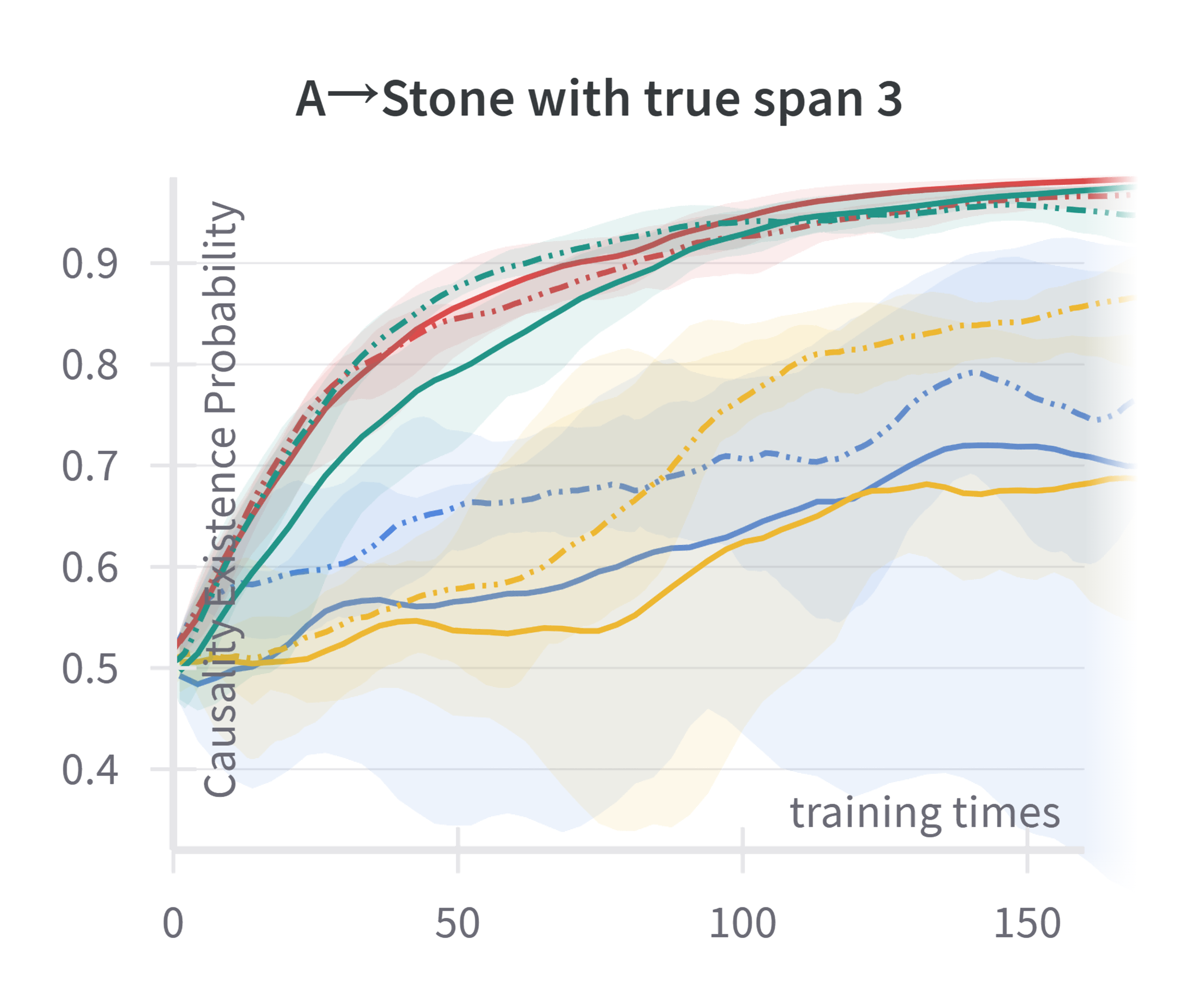} &
		\includegraphics[width=0.2\textwidth]{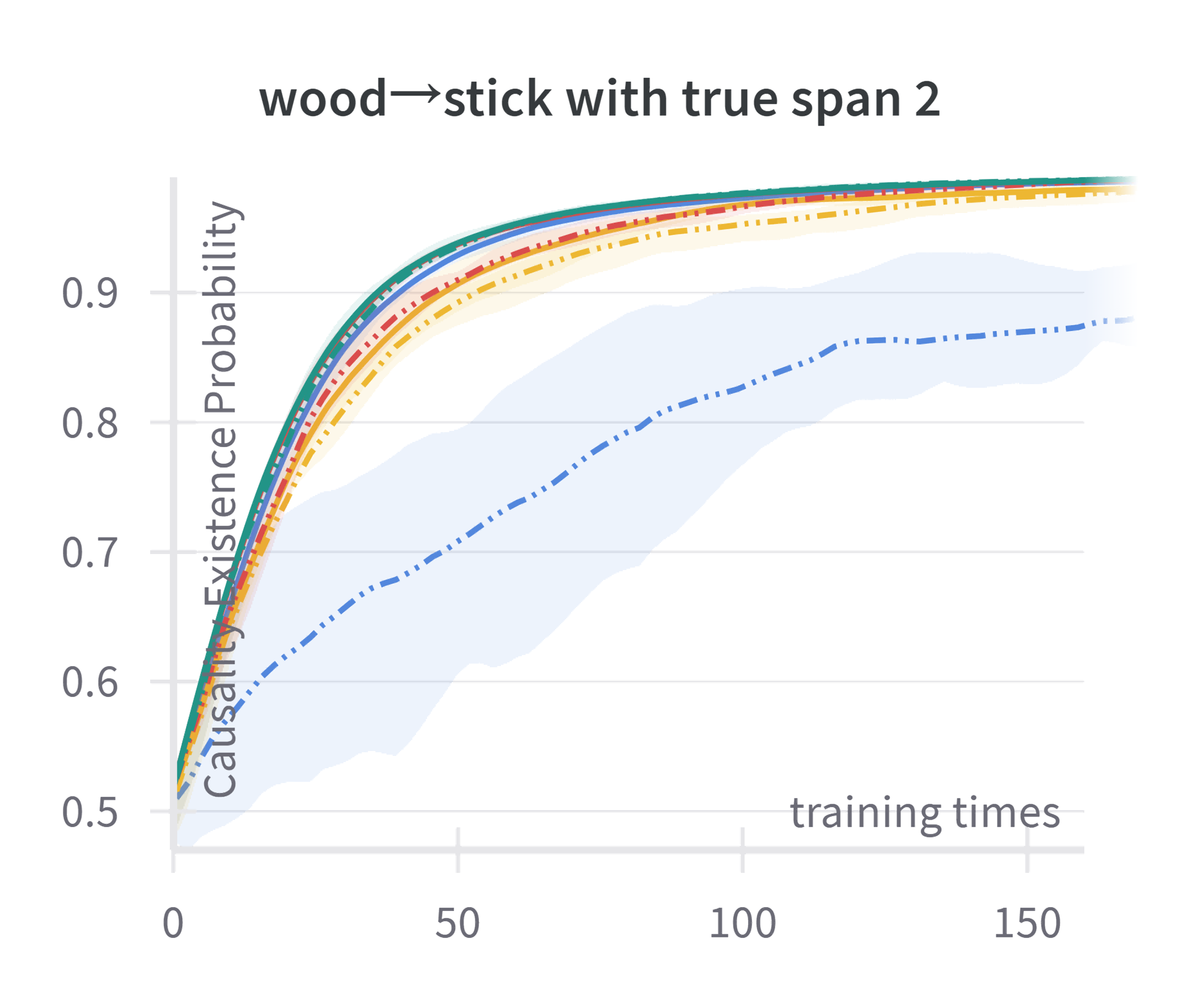} &
		\includegraphics[width=0.2\textwidth]{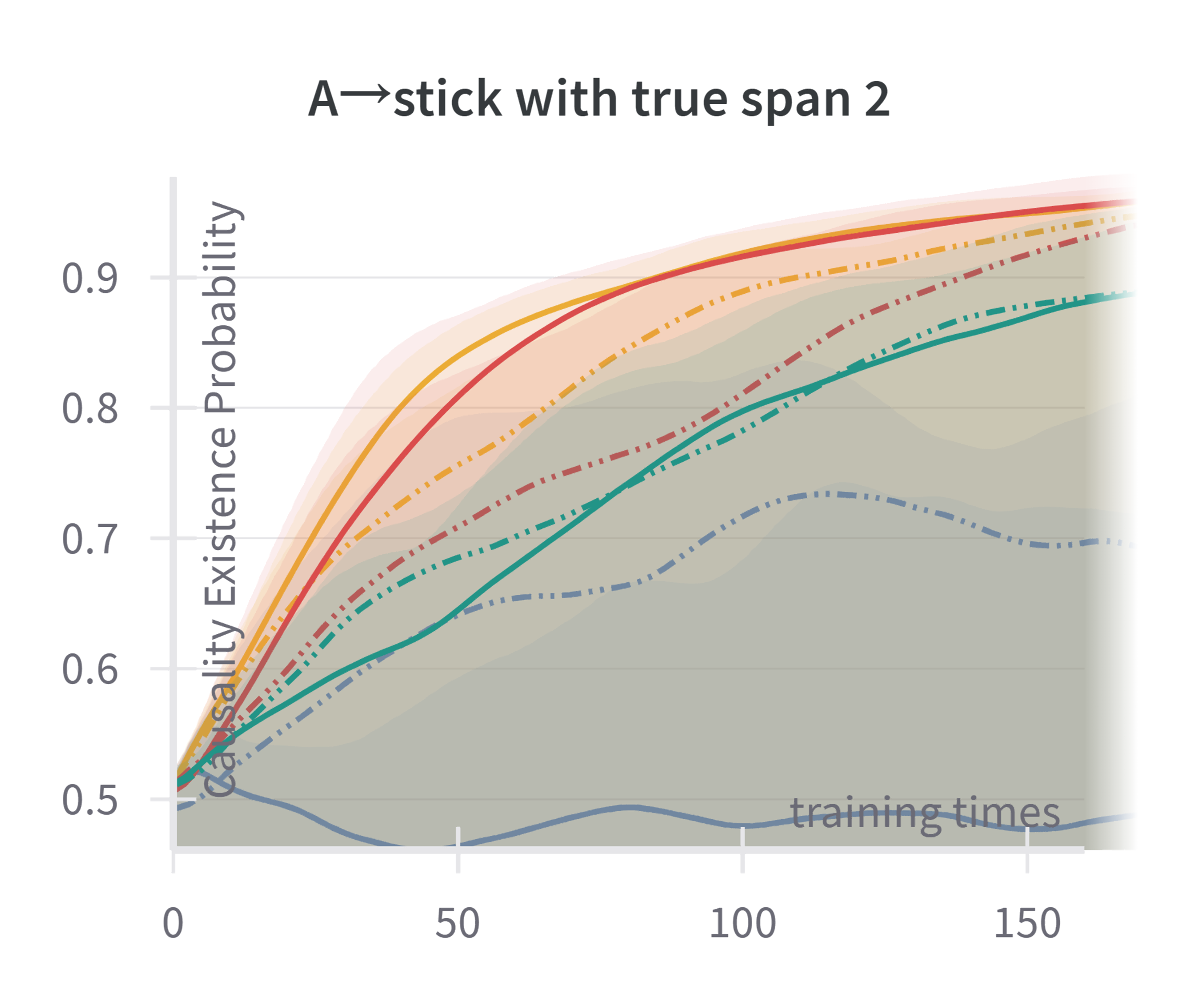} &
		\includegraphics[width=0.2\textwidth]{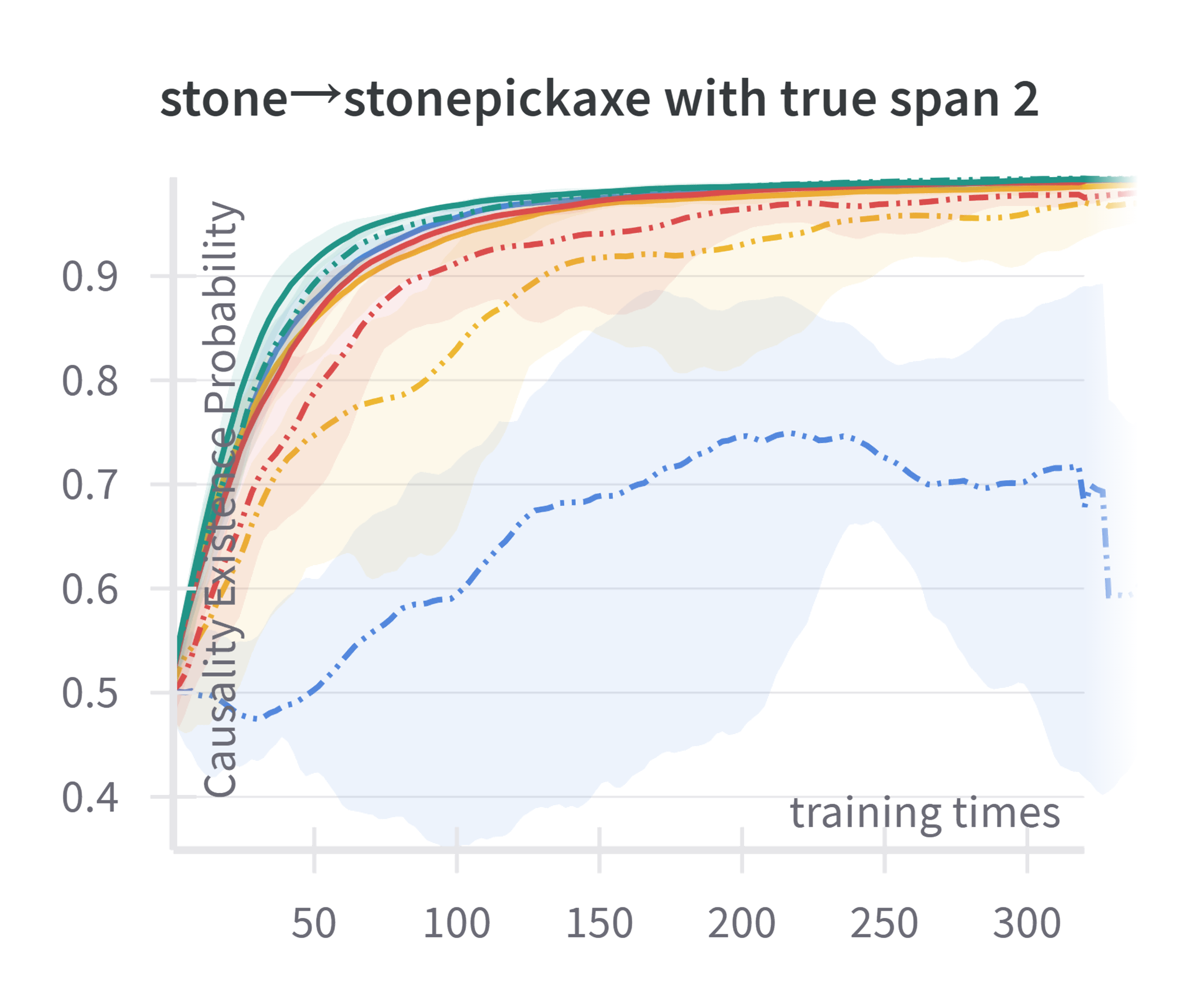} \\
		\includegraphics[width=0.2\textwidth]{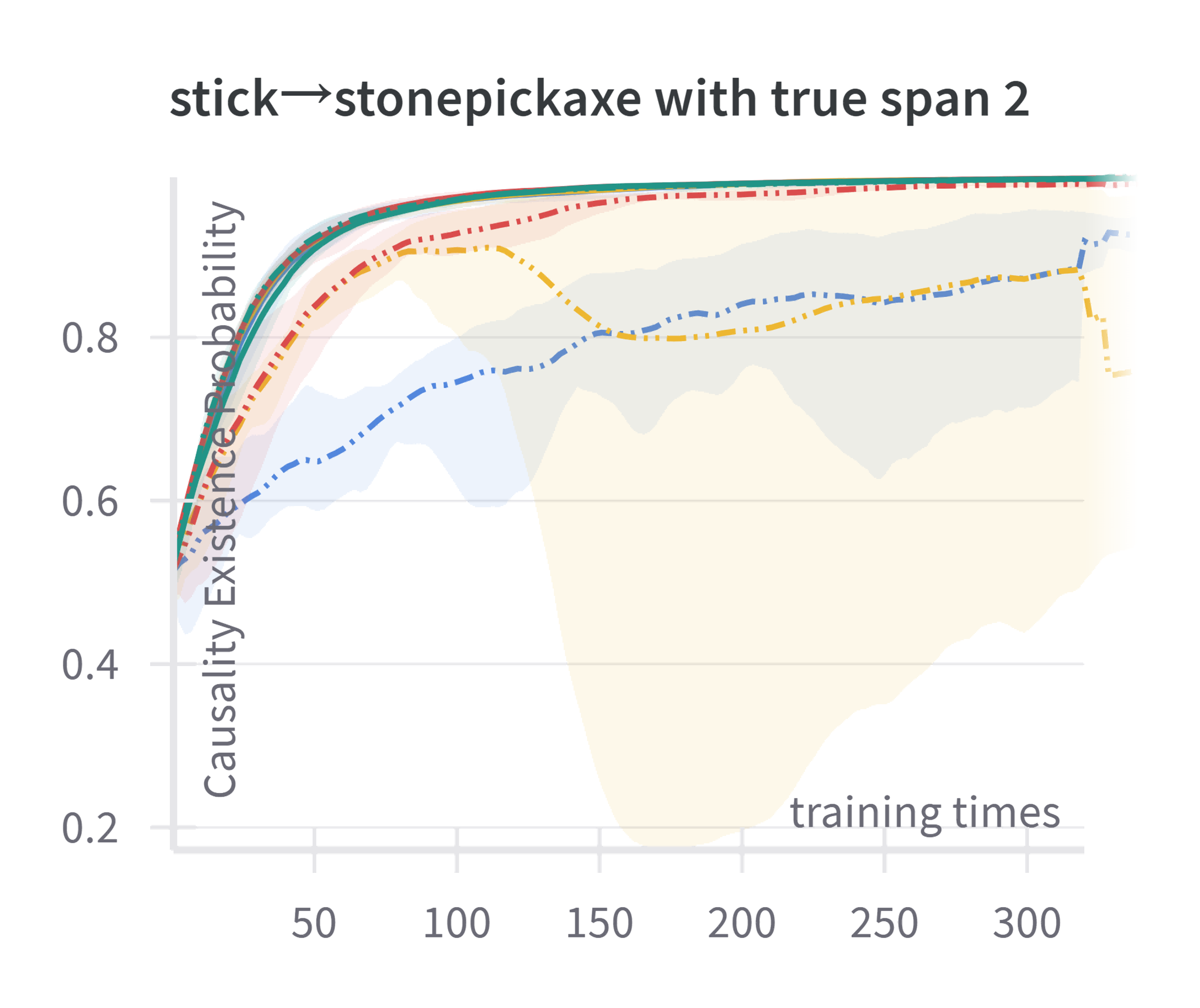} &
		\includegraphics[width=0.2\textwidth]{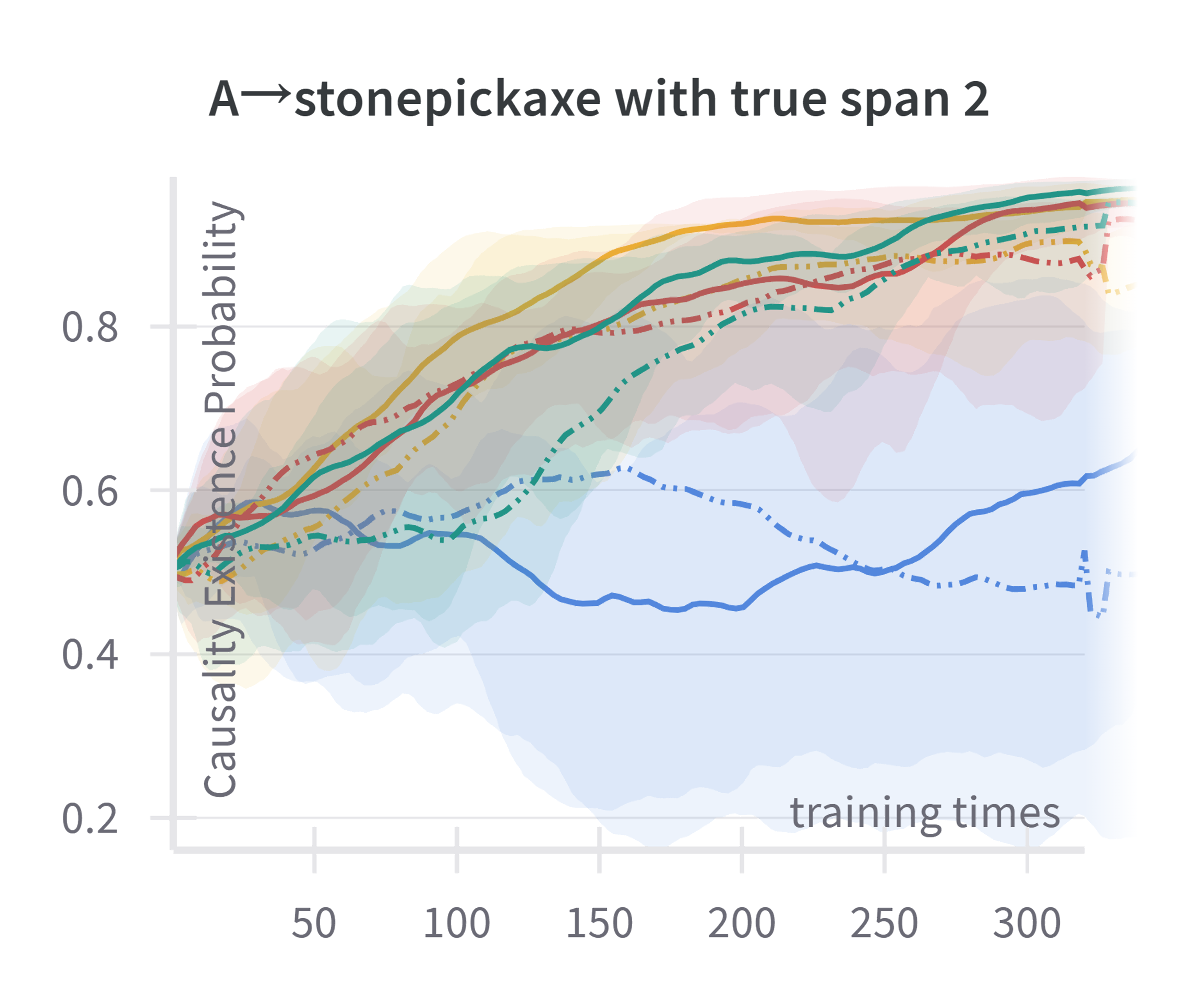} &
		
		\includegraphics[width=0.2\textwidth]{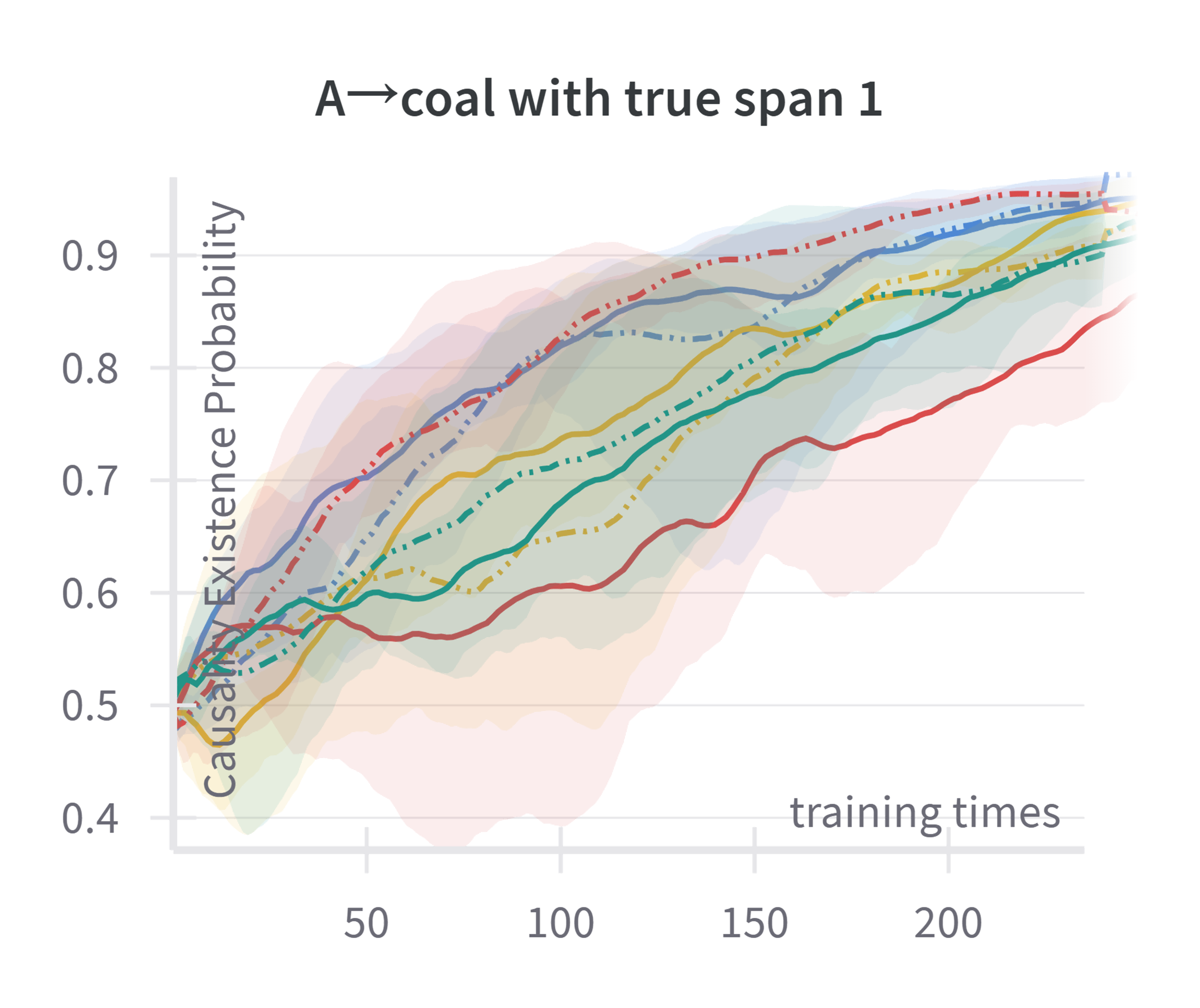} &
		\includegraphics[width=0.2\textwidth]{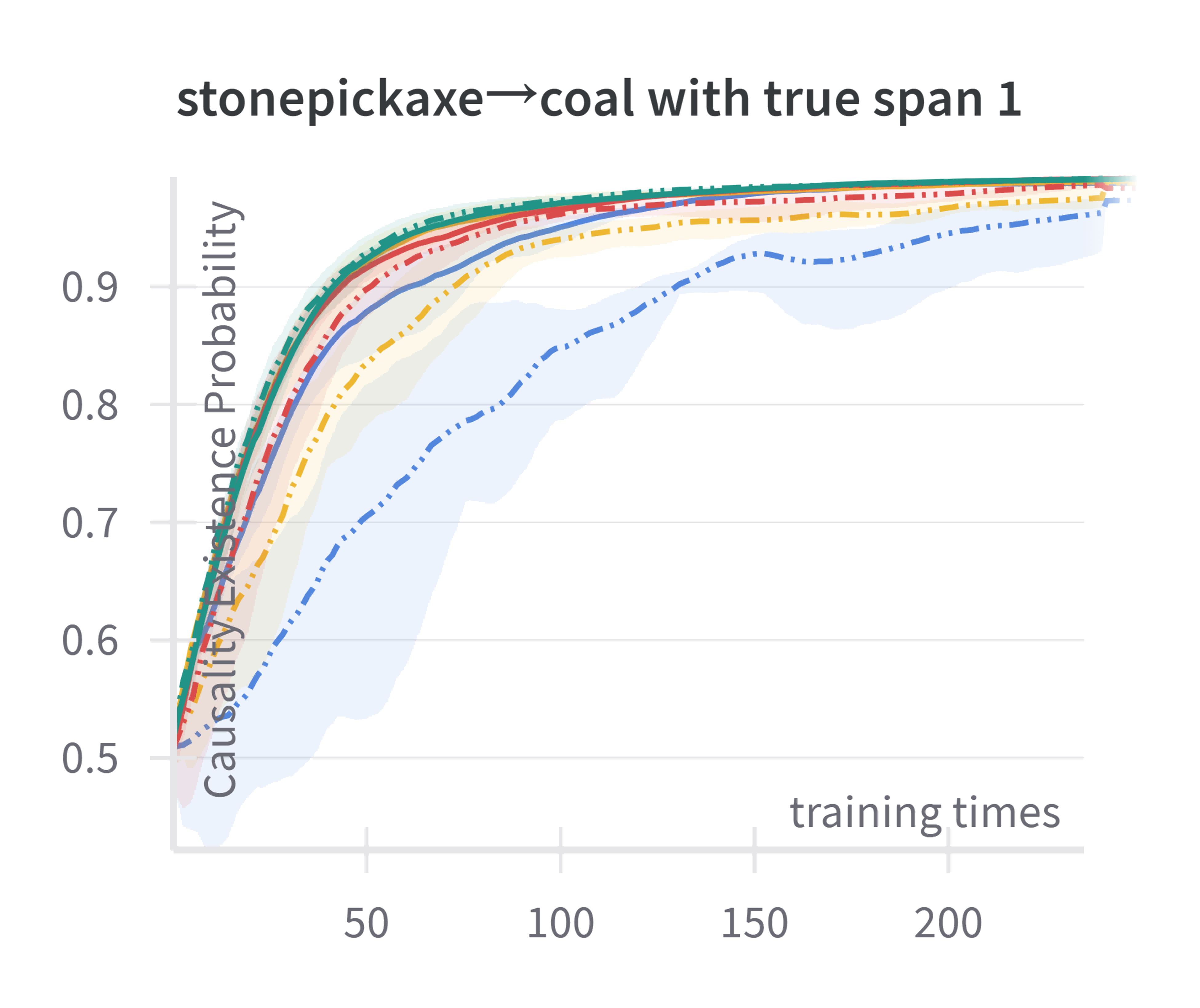} &
		\includegraphics[width=0.2\textwidth]{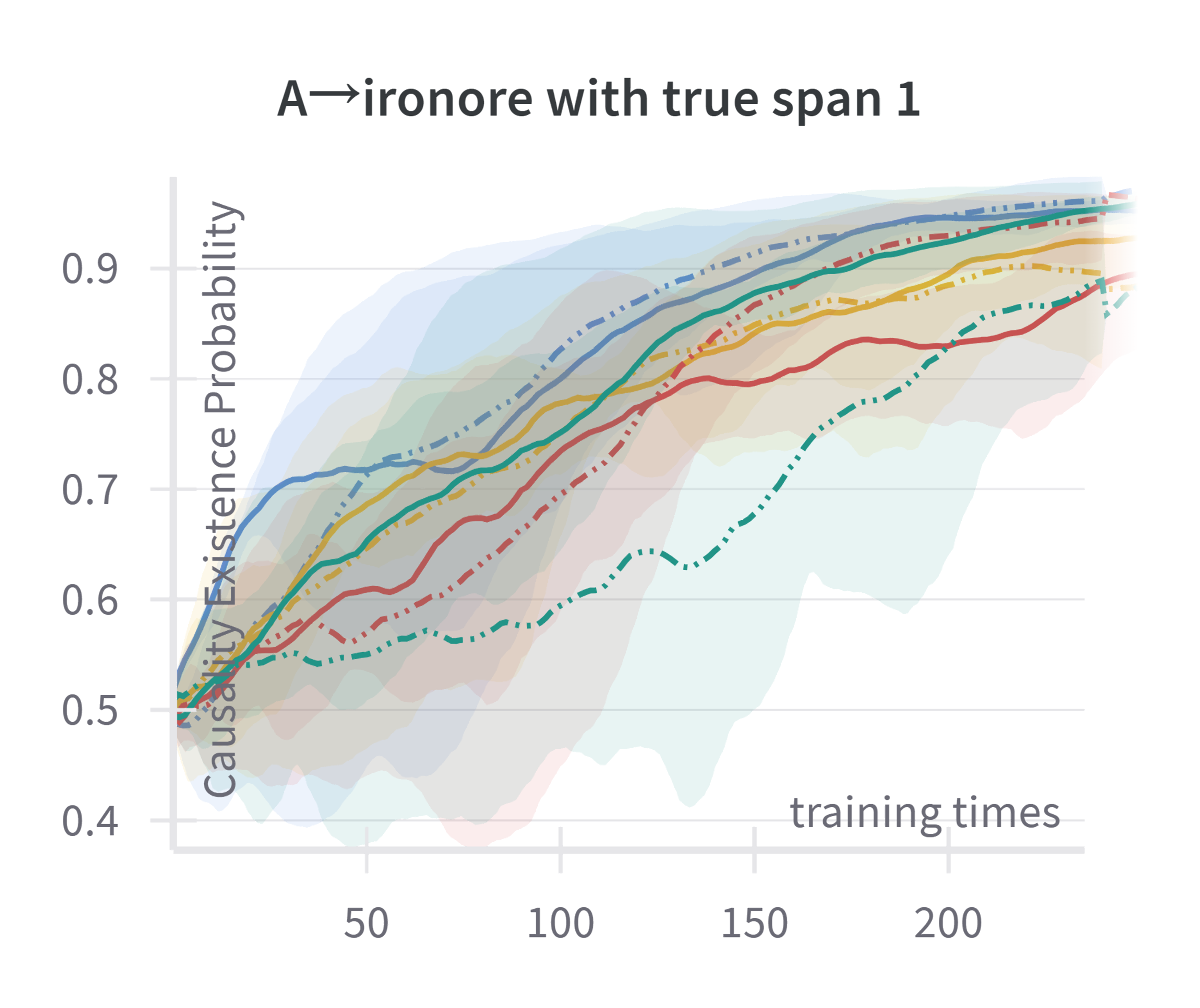} \\
		\includegraphics[width=0.2\textwidth]{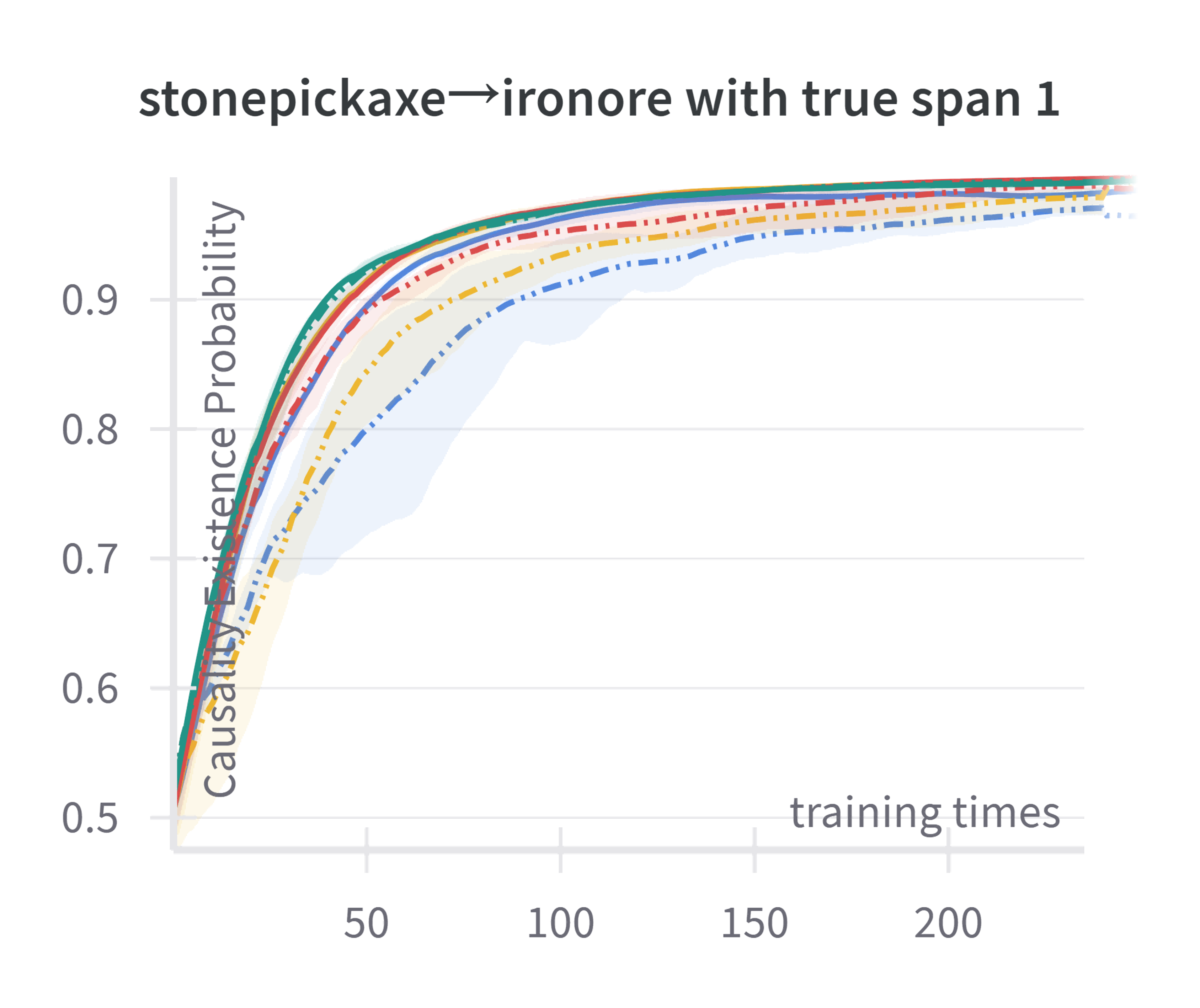} &
		\includegraphics[width=0.2\textwidth]{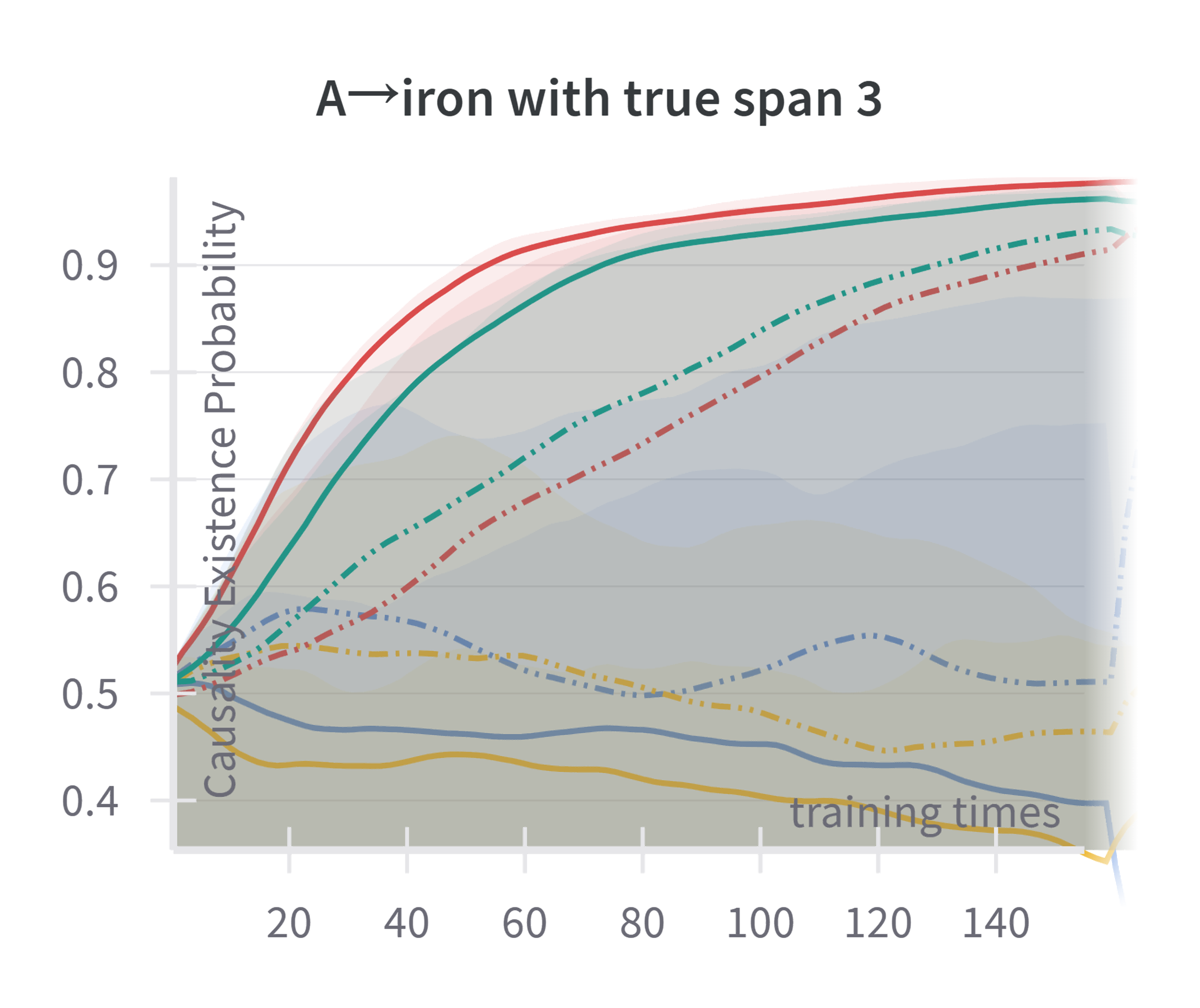} &
		\includegraphics[width=0.2\textwidth]{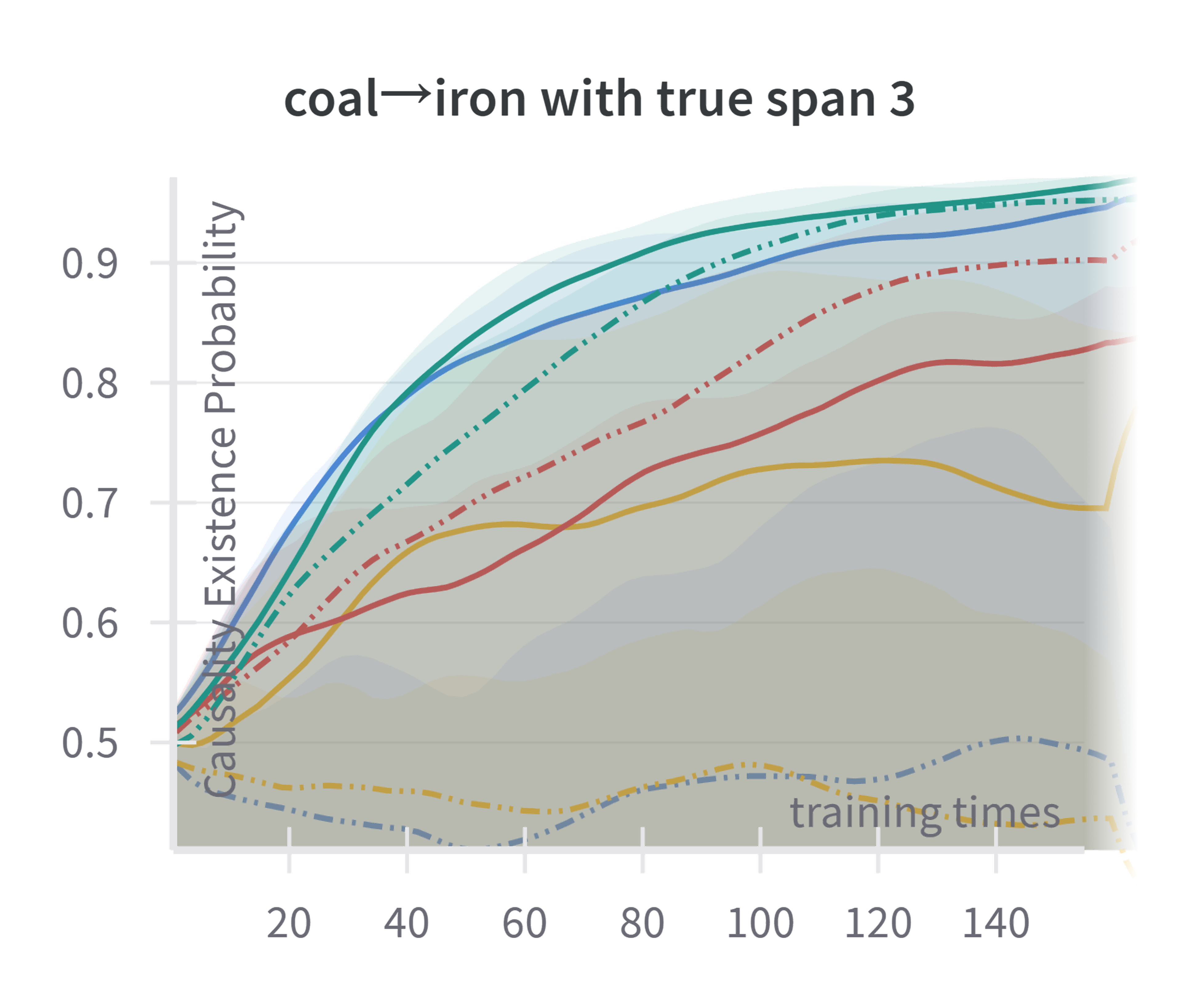} &
		\includegraphics[width=0.2\textwidth]{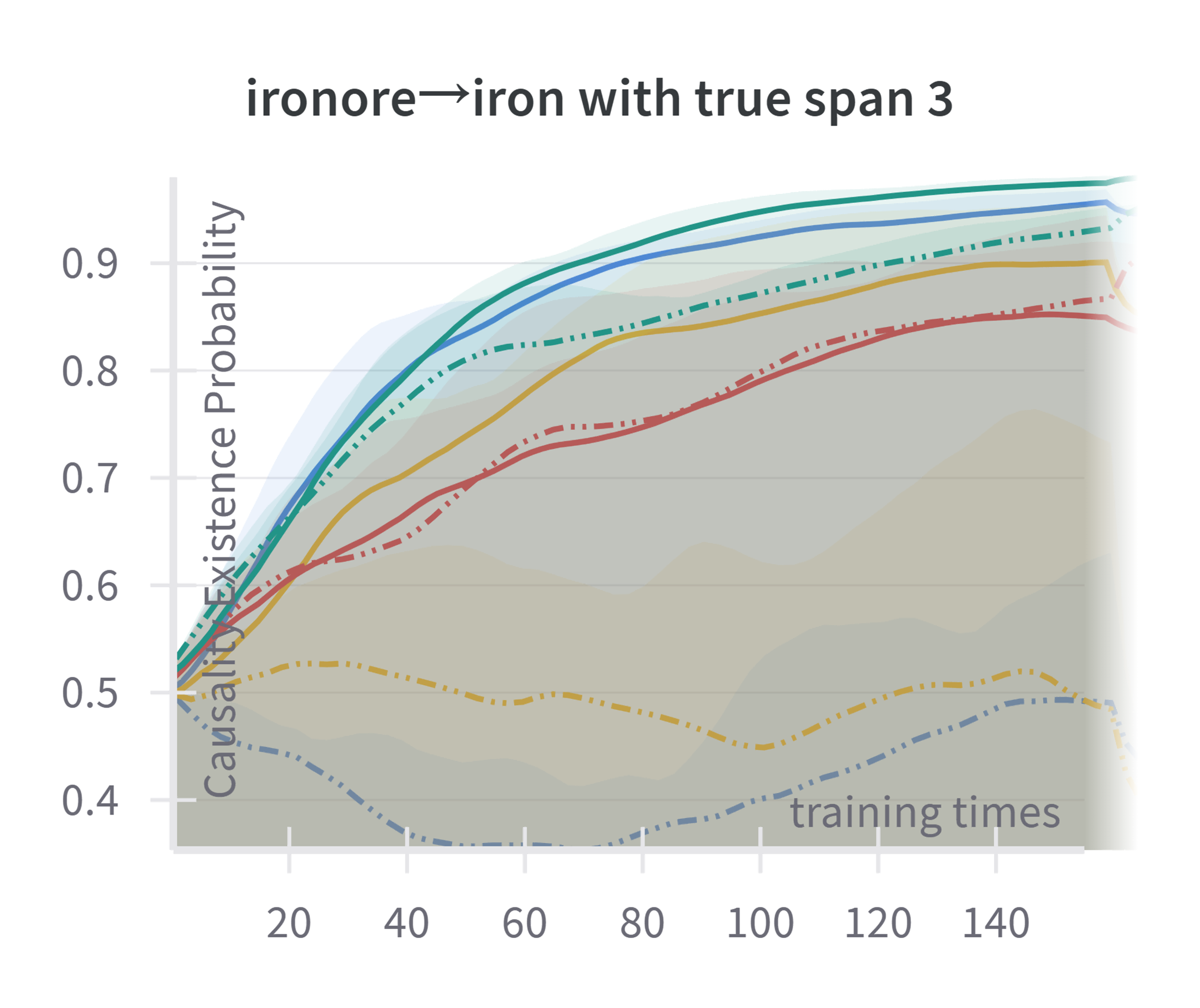}
	\end{tabular}
	\caption{The comparison between reverse and forward data collection strategy in GetIron-R0-T1 task with \(\tau_{max}=4\).}
	\label{fig:positive_reverse_}
\end{figure*}

%\noindent\textbf{5. How does the reverse data collection strategy compare to the forward data collection strategy?}
\paragraph{Experimental Design}
We evaluate two strategies on causal relationship learning in GetIron-R0 with \(\tau_{max}=4\). For a fair comparison, each causal relationship was learned assuming the preceding ones were already known. Additionally, the causal probability for each new segment was initialized to 0.5. Results are shown in Figure~\ref{fig:positive_reverse_}. Each figure contains up to 8 curves, representing the causality existence probability of the causal relationships as indicated in the figure titles at different time spans \(h \in \{\textcolor{b}{1},\textcolor{y}{2},\textcolor{r}{3},\textcolor{g}{4}\}\), four are solid lines for reverse strategy, four are dashed lines for forward strategy.

%To ensure a fair comparison, each causal relationship was learned under the assumption that the preceding causal relationships had already been learned. Additionally, each new segment of causal relationship learning began with the corresponding causal probability initialized to 0.5. 
%\noindent\textit{\textbf{Experimental Results.}}
\paragraph{Experimental Results}
%It can be observed that the convergence rate of the causal relationship learning curve with the reverse data collection strategy is higher than with the forward strategy for most causal relationships.	% slightly 
It can be observed that the convergence rate of the causal relationship learning curve is higher with the reverse data collection strategy compared to the forward strategy for most causal relationships, thereby validating the effectiveness of our reverse data collection strategy.

\subsubsection{How does D3HRL compare to CDHRL in terms of the accuracy of causal graph identification?}
\begin{figure*}[htbp]
	\centering
	\setlength{\tabcolsep}{-2pt} % 设置列间距为 2pt
	\begin{tabular}{ccccc}
		\includegraphics[width=0.2\textwidth]{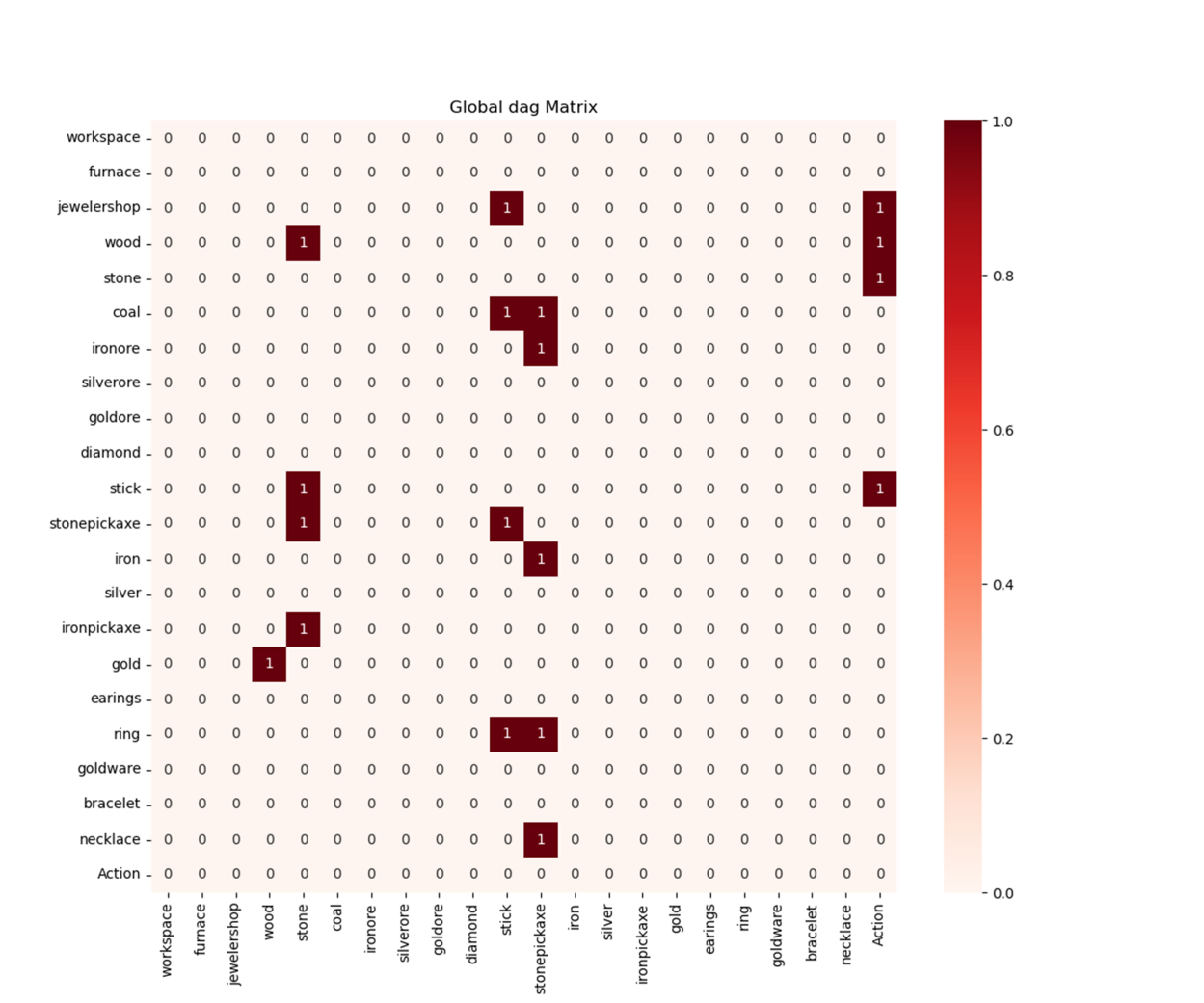} &
		\includegraphics[width=0.2\textwidth]{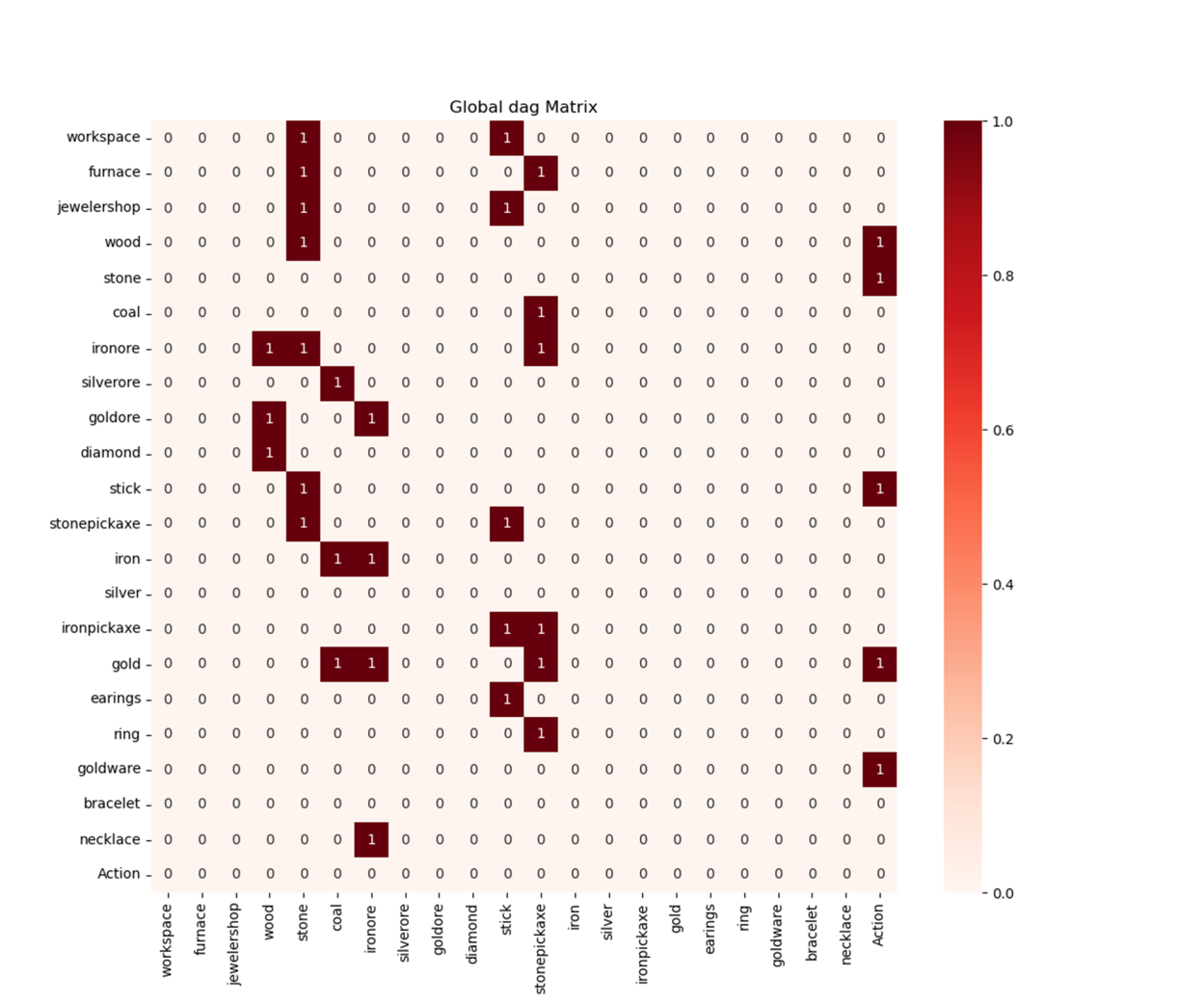} &
		\includegraphics[width=0.2\textwidth]{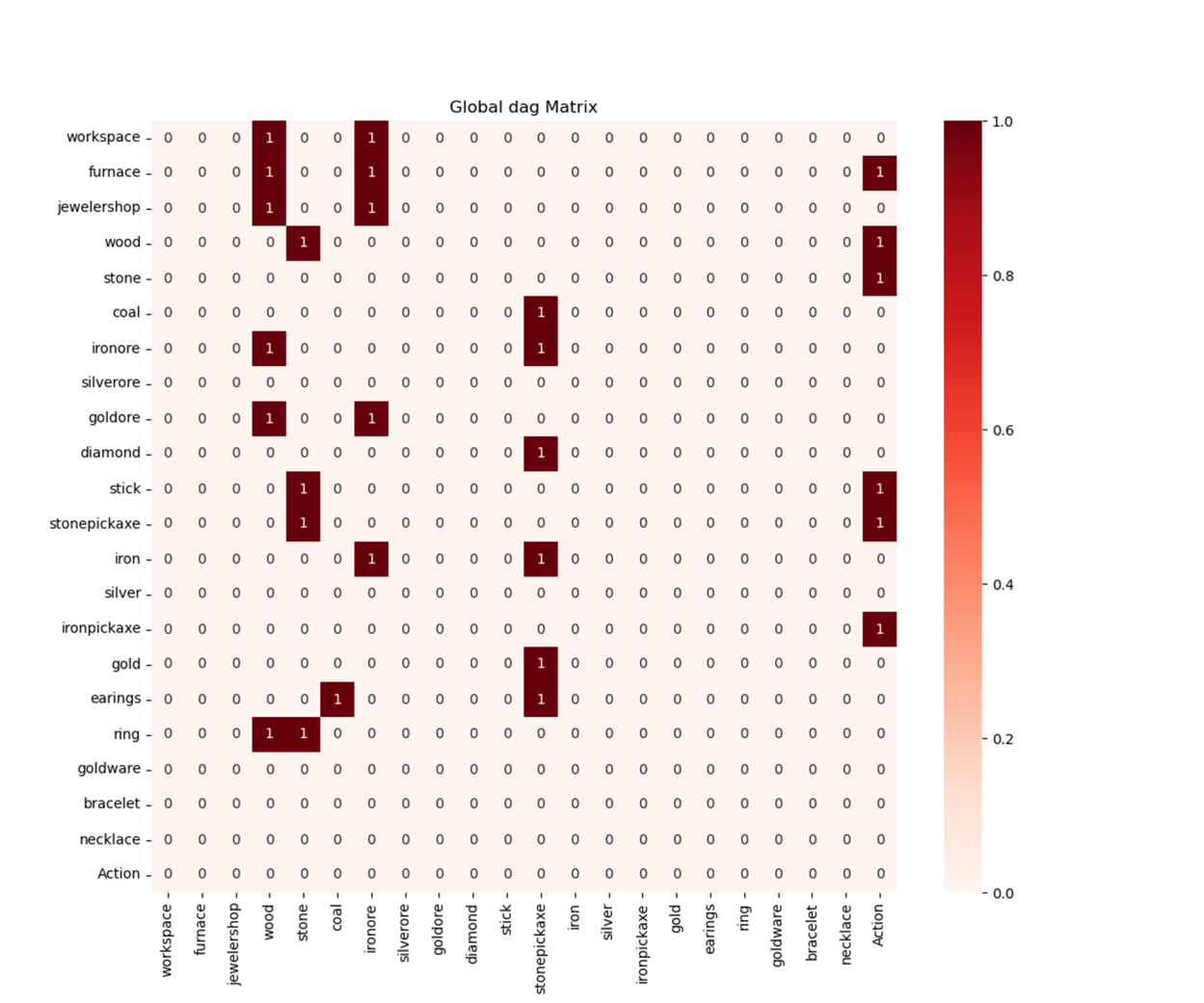} &
		\includegraphics[width=0.2\textwidth]{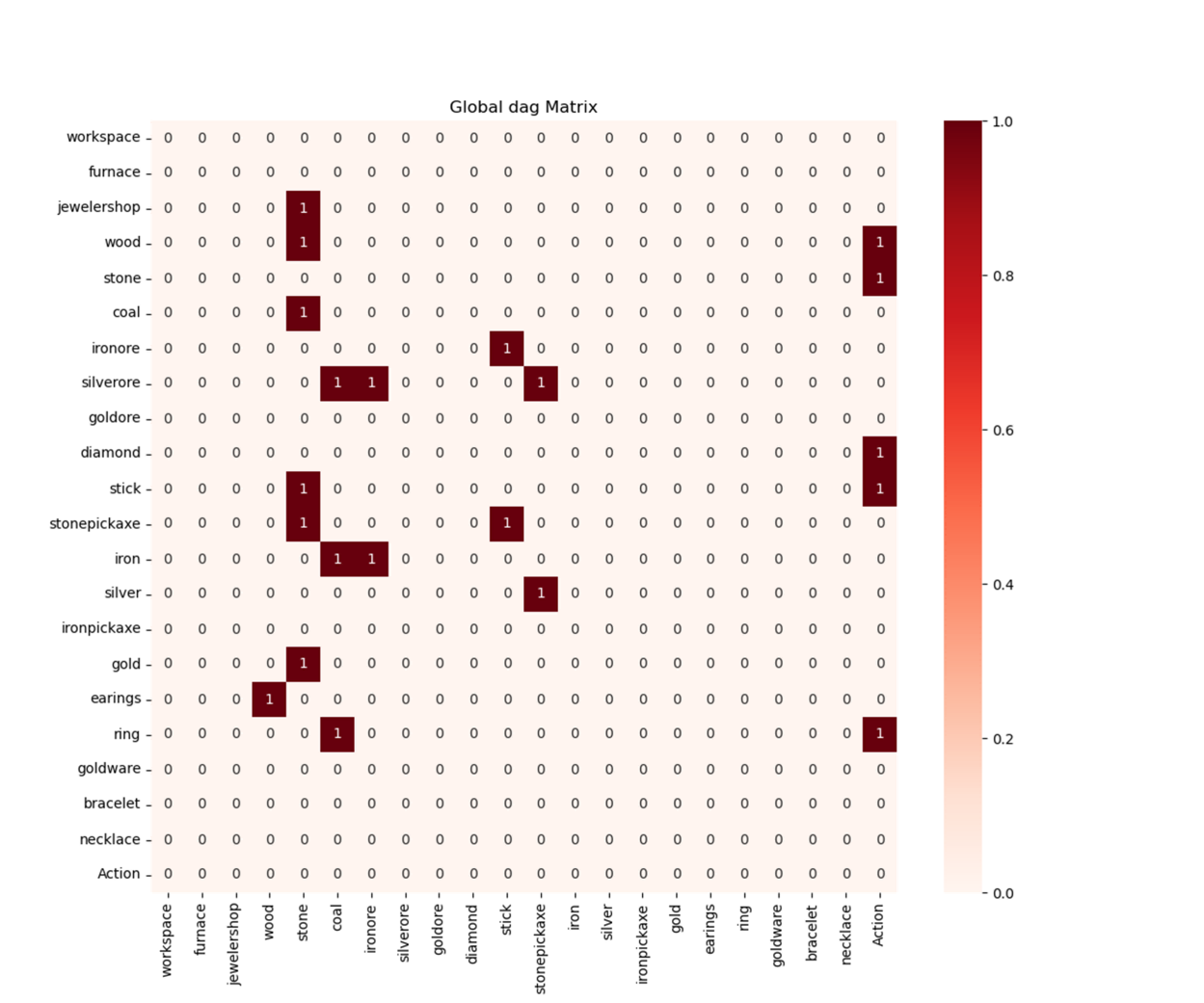} &
		\includegraphics[width=0.2\textwidth]{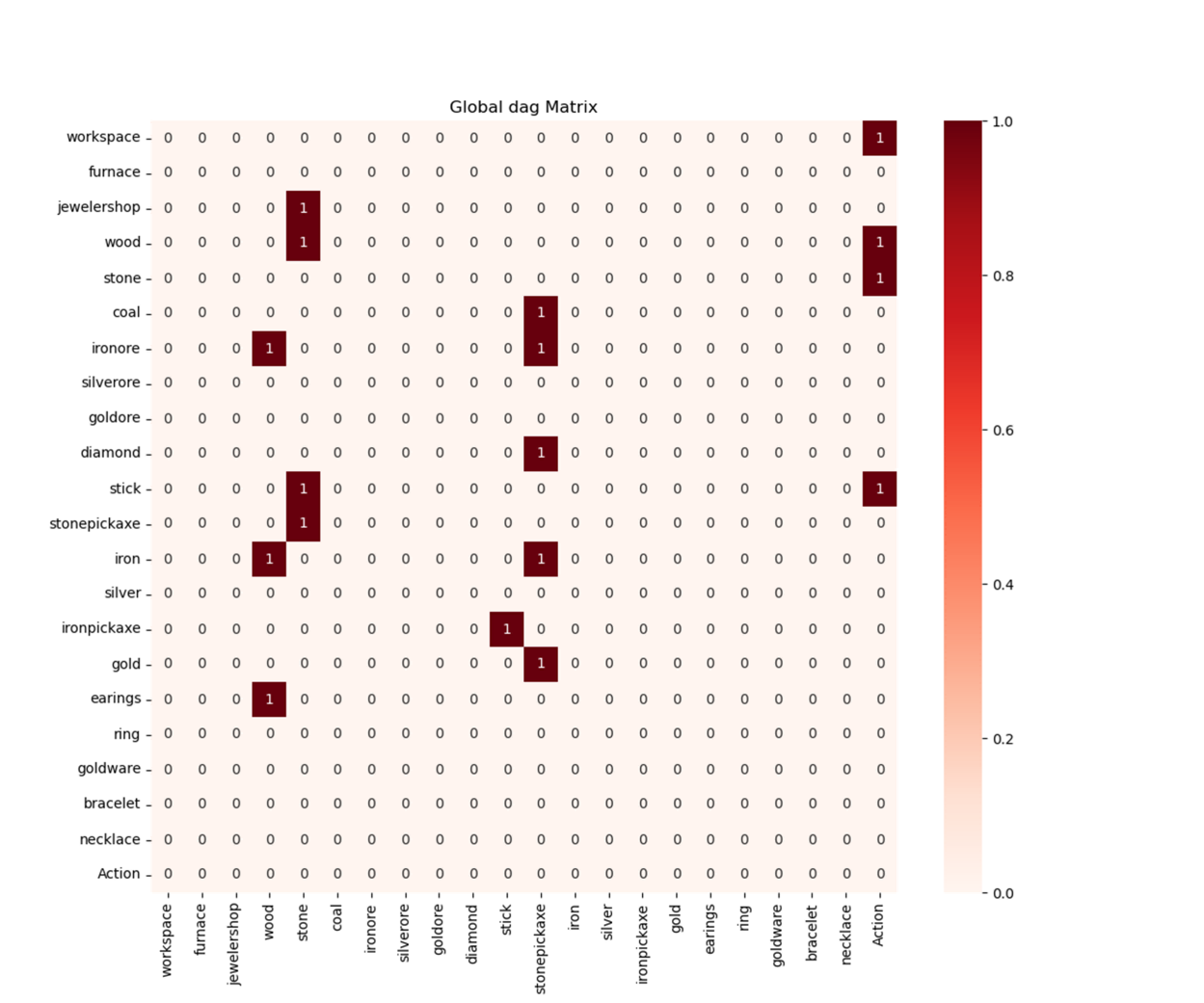} \\
		\includegraphics[width=0.2\textwidth]{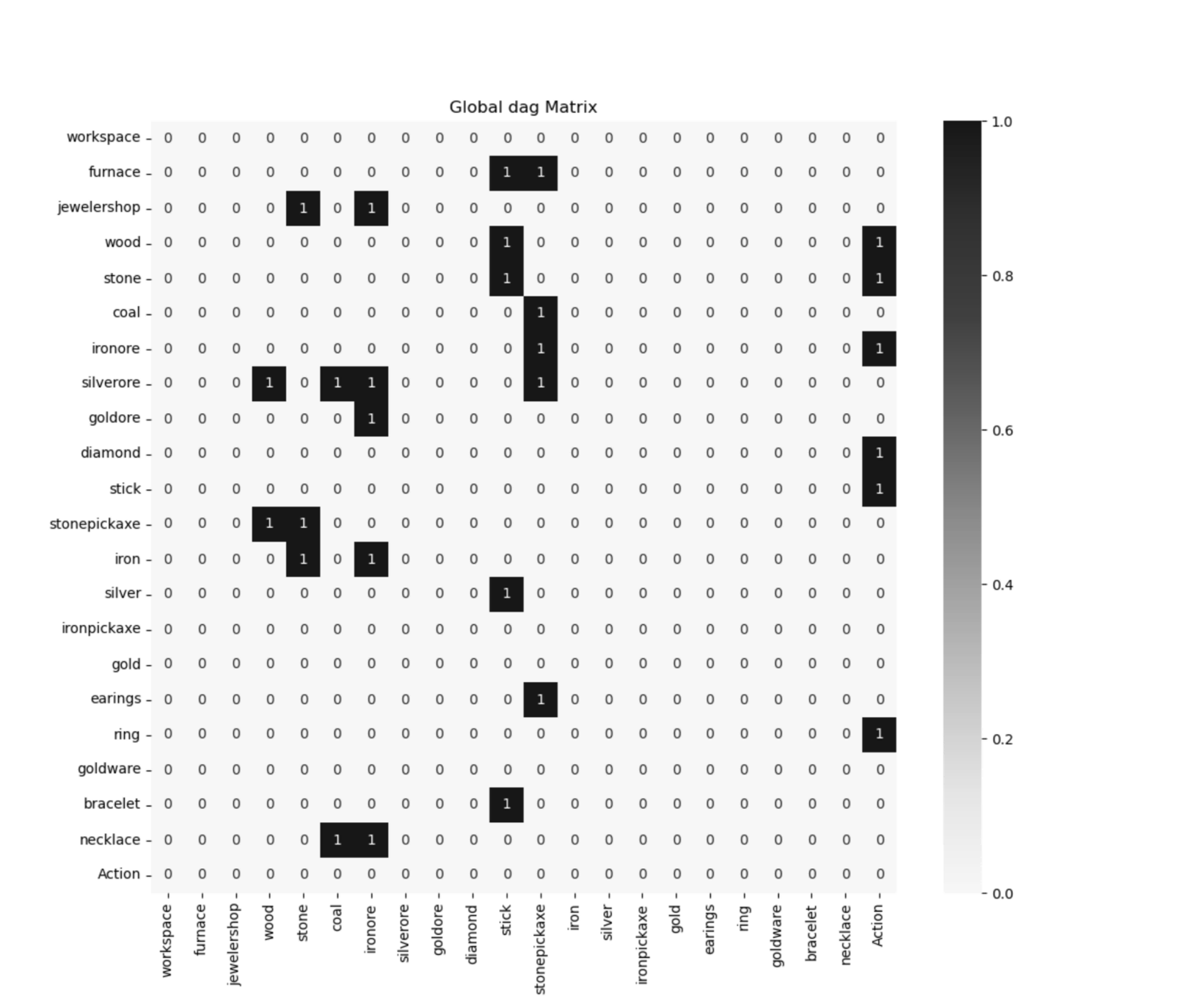} &
		\includegraphics[width=0.2\textwidth]{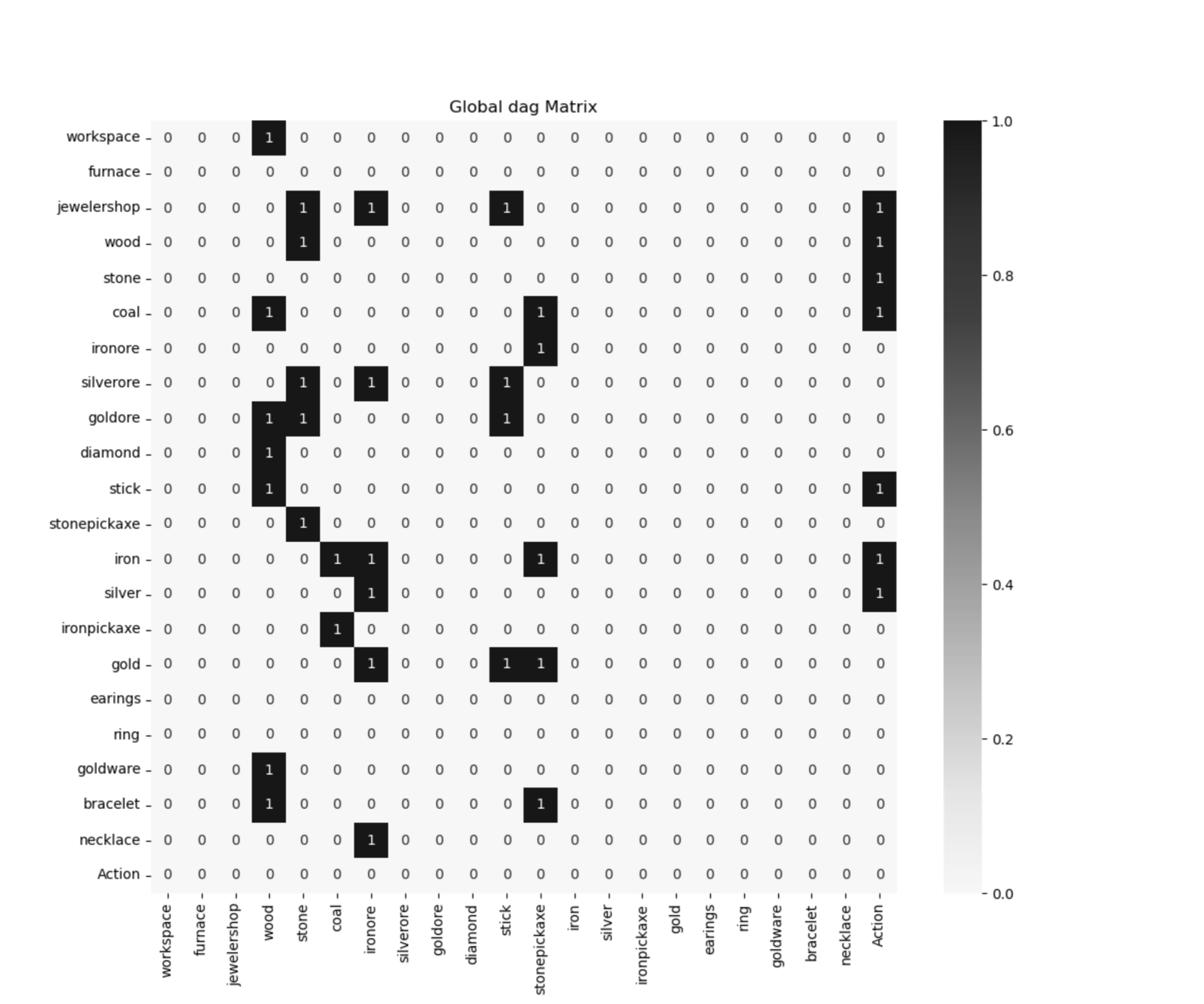} &
		\includegraphics[width=0.2\textwidth]{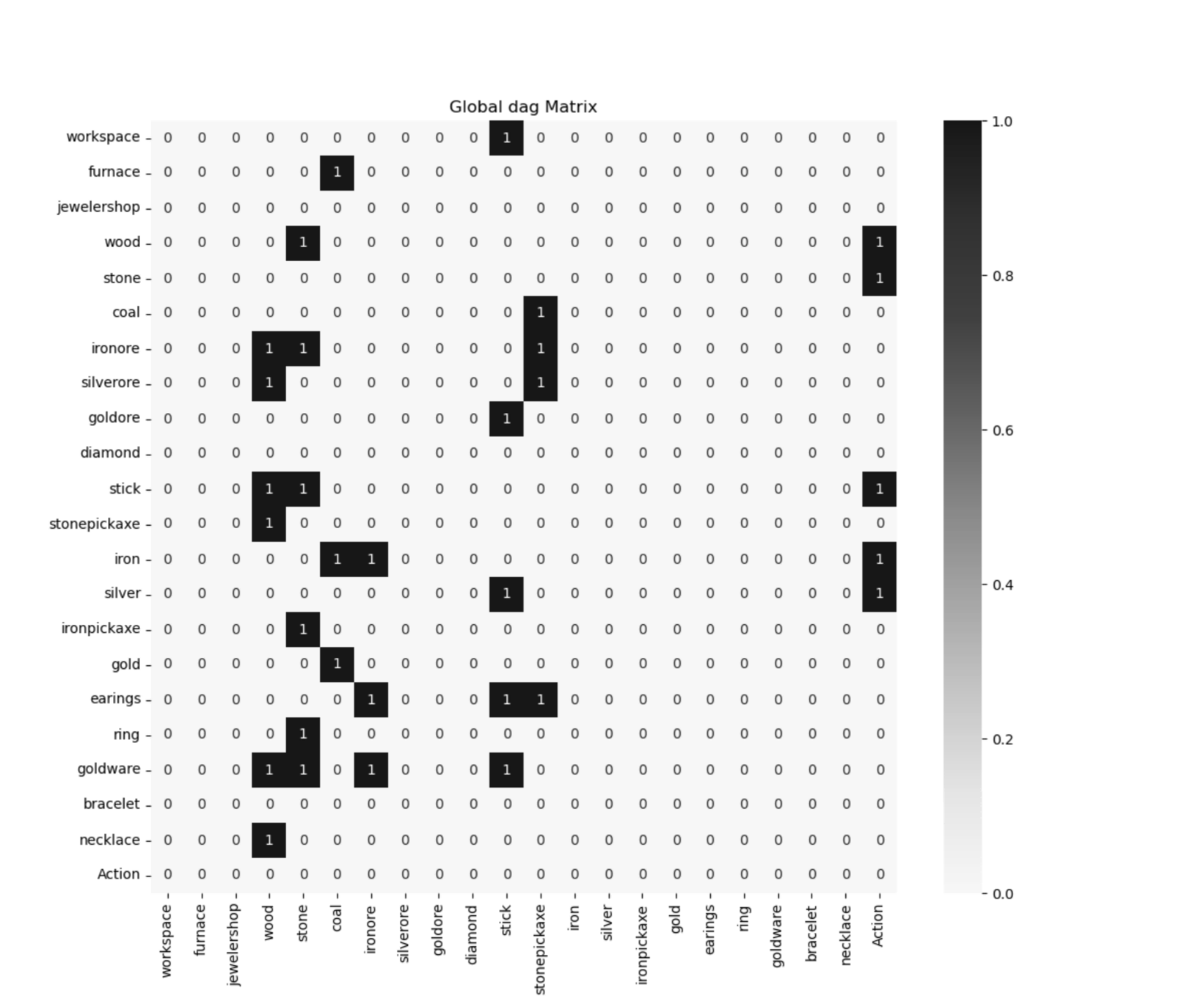} & 
		\includegraphics[width=0.2\textwidth]{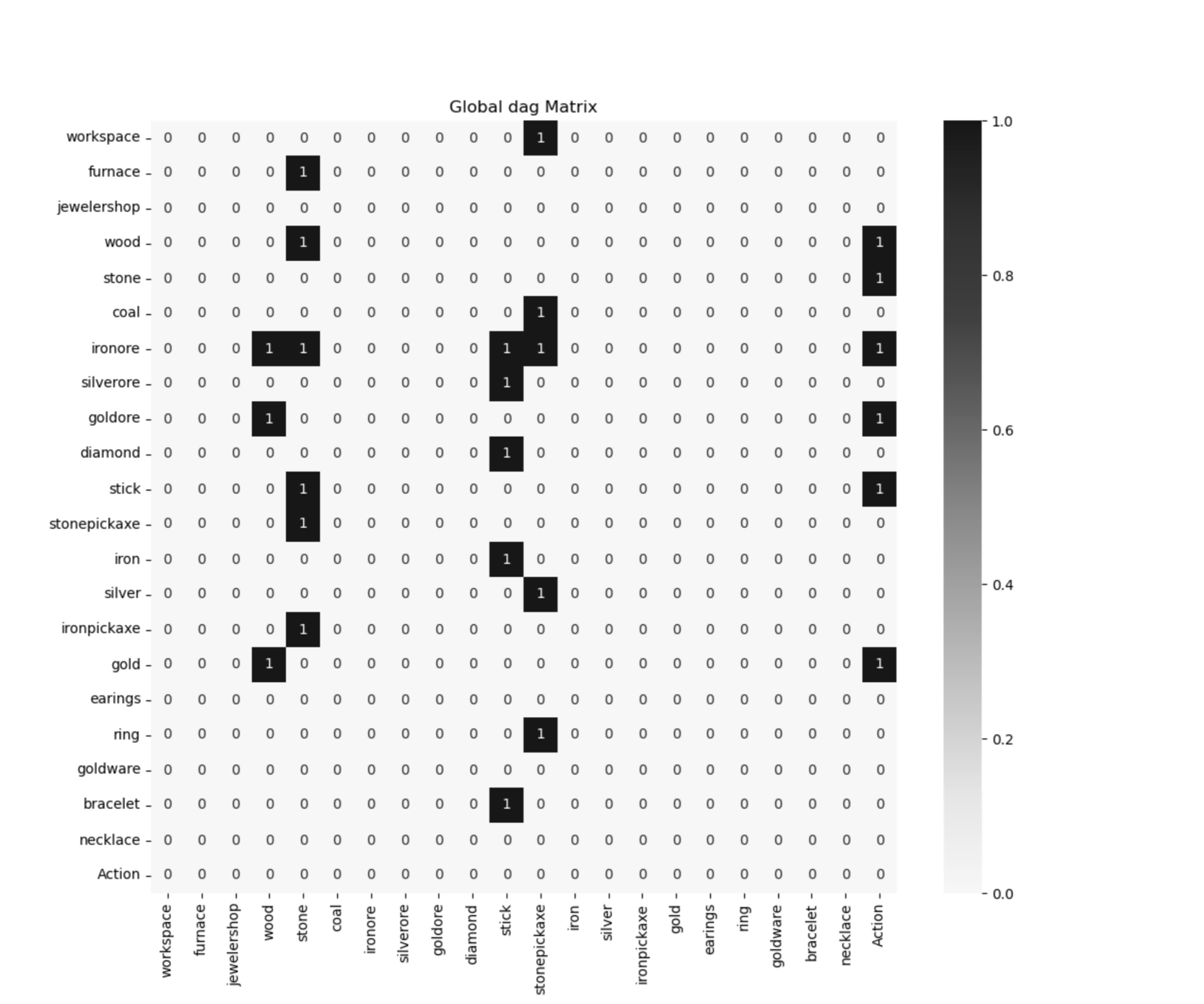} &
		\includegraphics[width=0.2\textwidth]{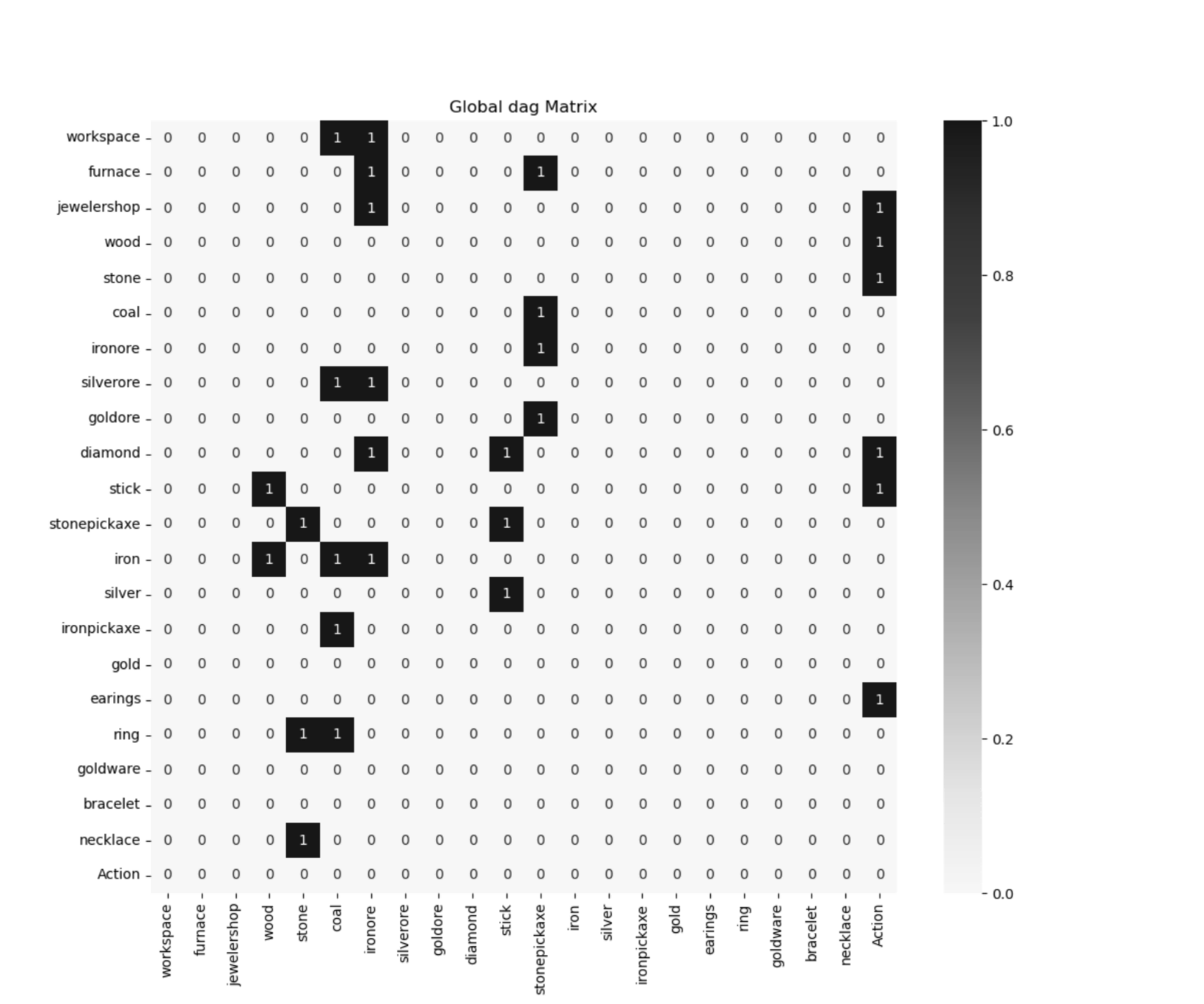}
	\end{tabular}
	\caption{The causal graph matrices of CDHRL in GetIron-R0-T0 task (up) and GetIron-R0-T1 task (down).}
	% with \(\tau_{max}=1\) (up) and \(\tau_{max}=4\) (down)
	\label{fig:cdhrl_multi_dag}
\end{figure*}

\begin{figure*}[htbp]
	\centering
	\setlength{\tabcolsep}{-2pt} % 设置列间距为 2pt
	\begin{tabular}{ccccc}
		\includegraphics[width=0.20\textwidth]{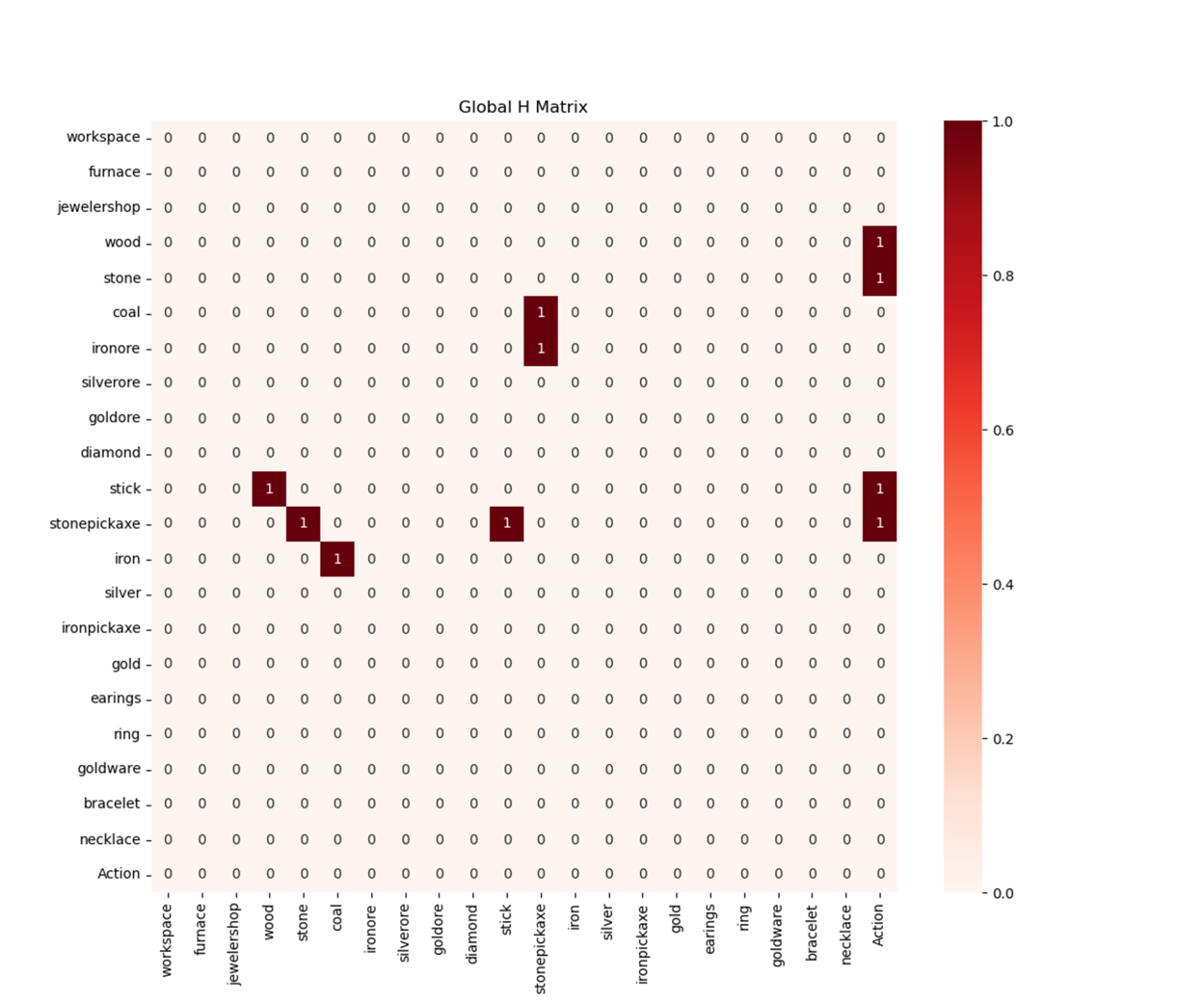} &
		\includegraphics[width=0.20\textwidth]{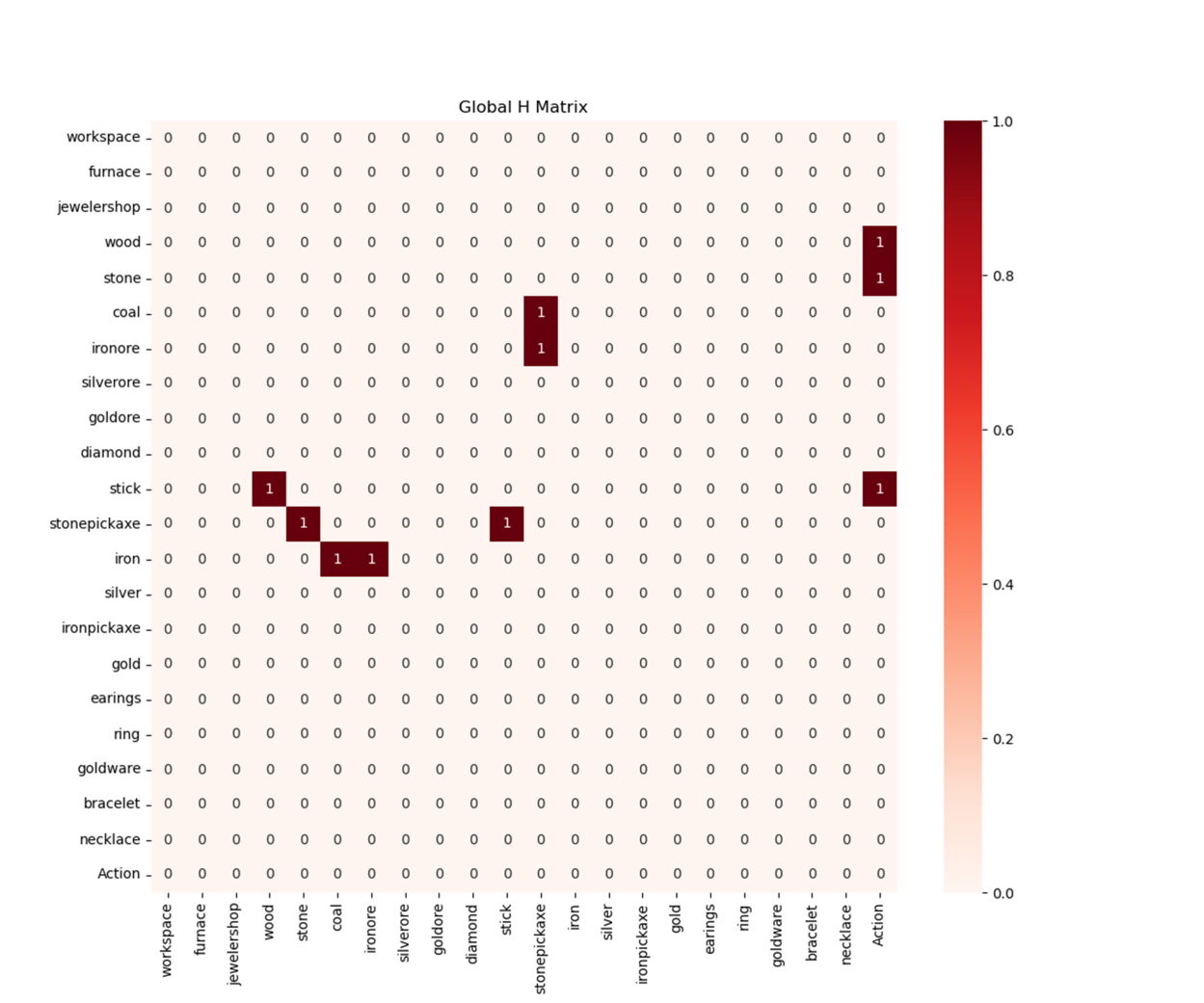} &
		\includegraphics[width=0.20\textwidth]{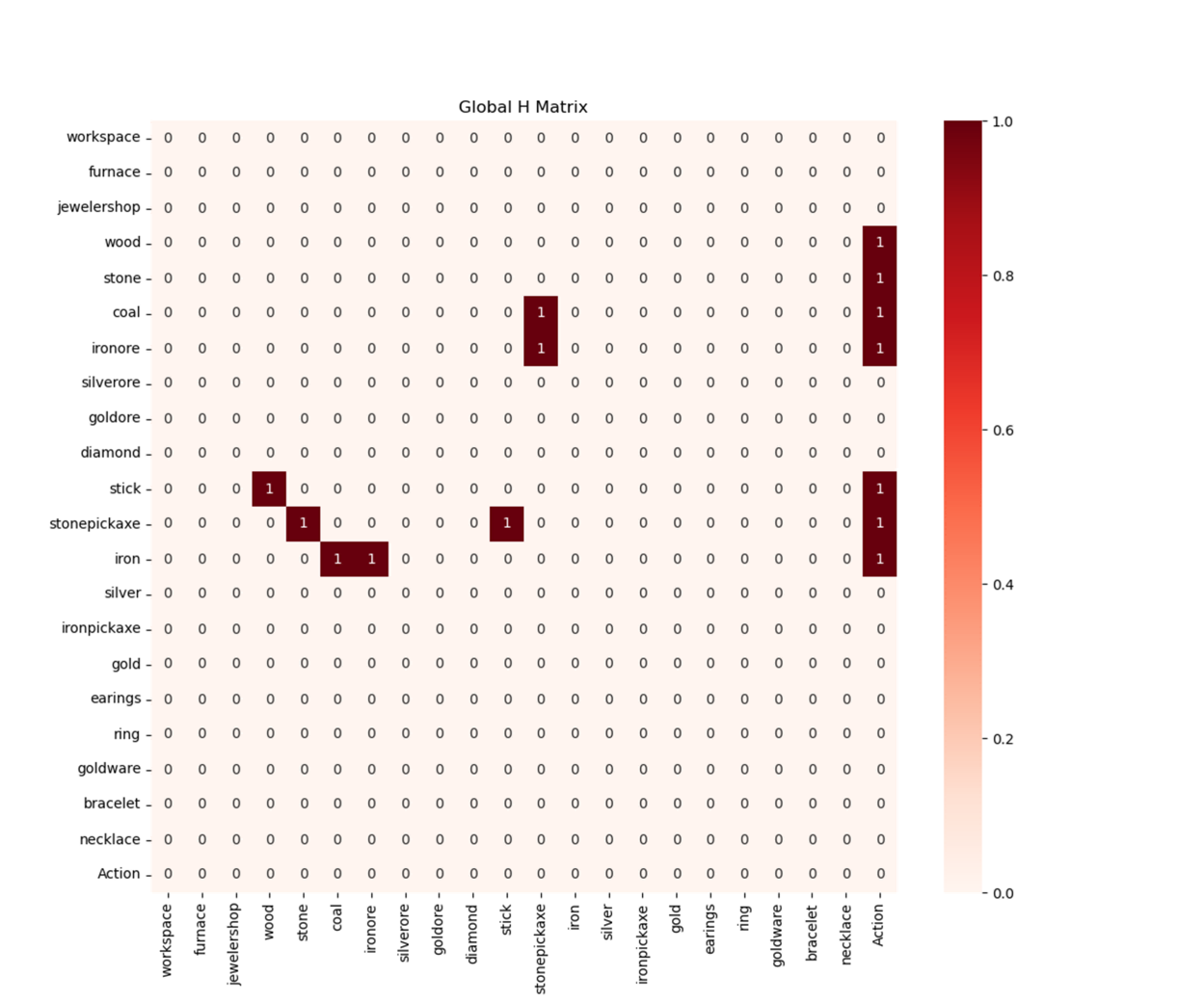} &
		\includegraphics[width=0.20\textwidth]{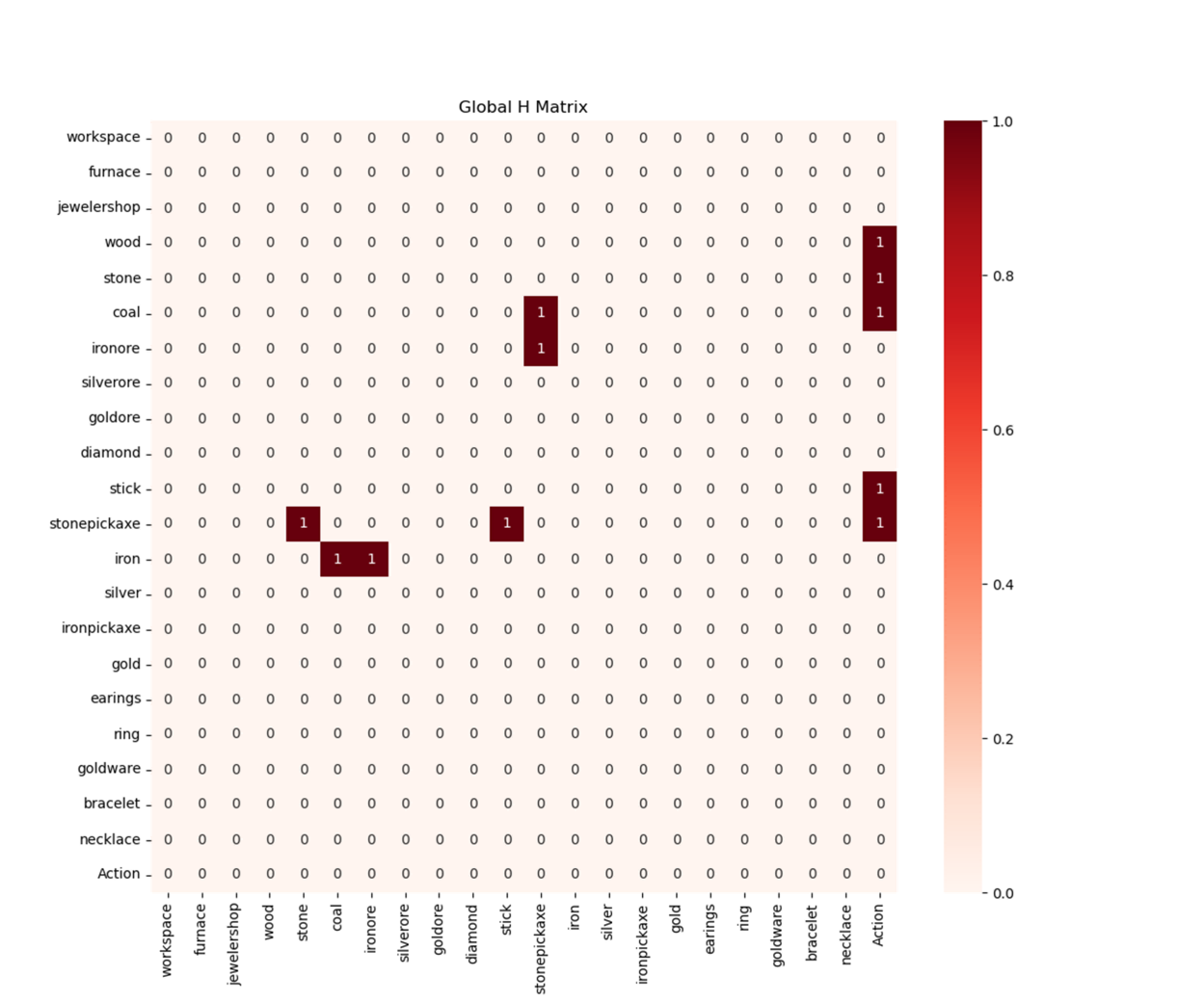} &
		\includegraphics[width=0.20\textwidth]{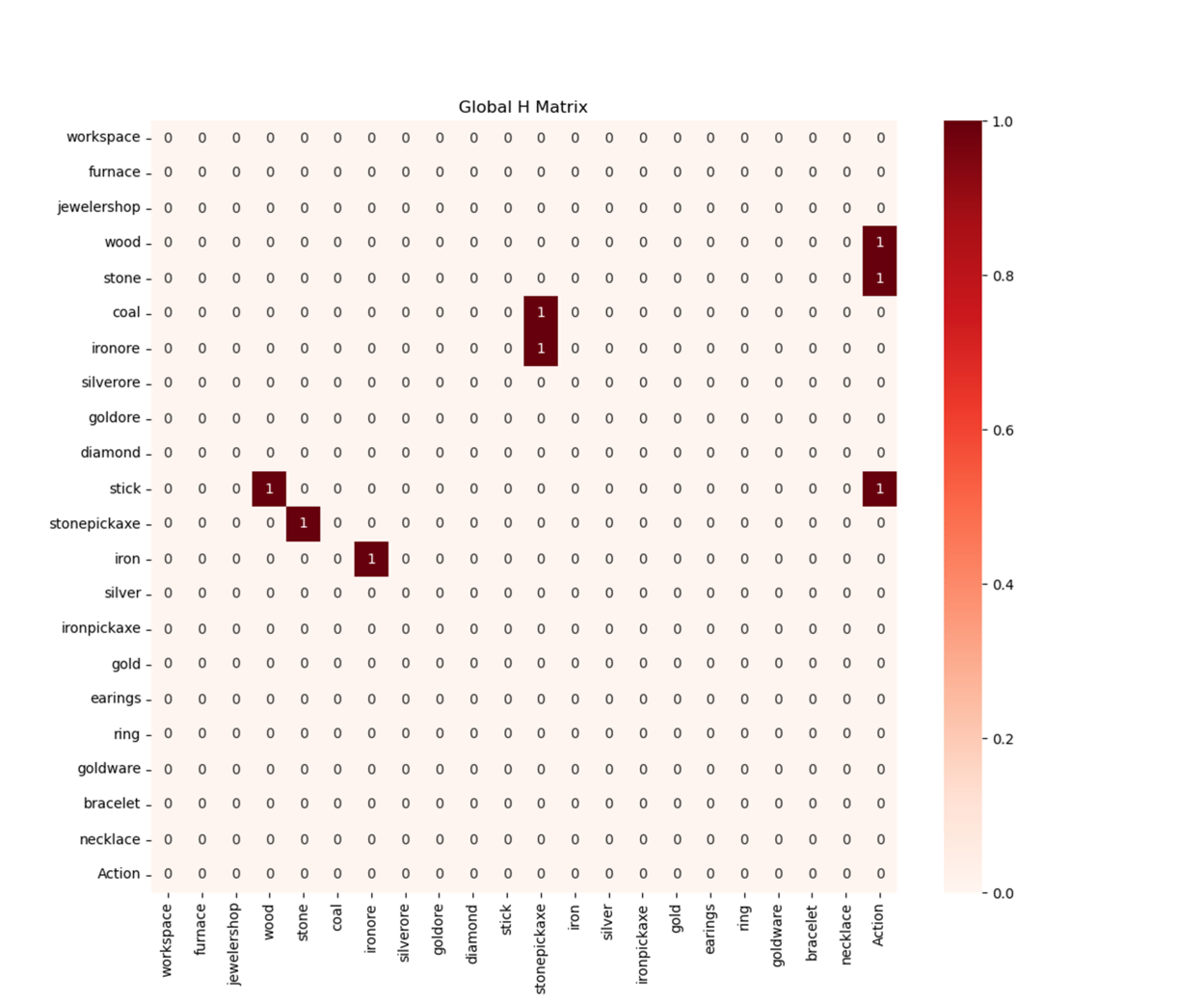} \\
		\includegraphics[width=0.20\textwidth]{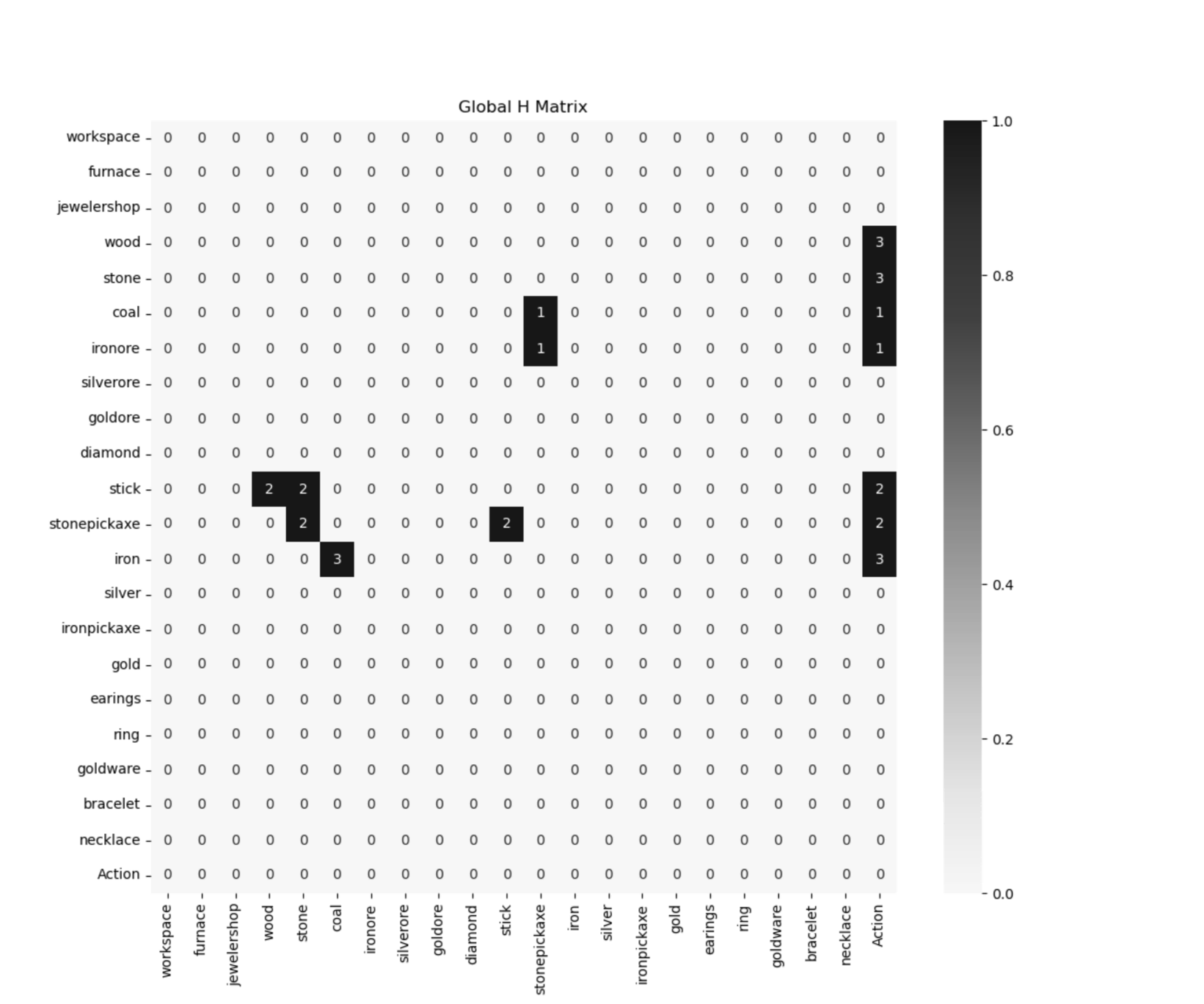} &
		\includegraphics[width=0.20\textwidth]{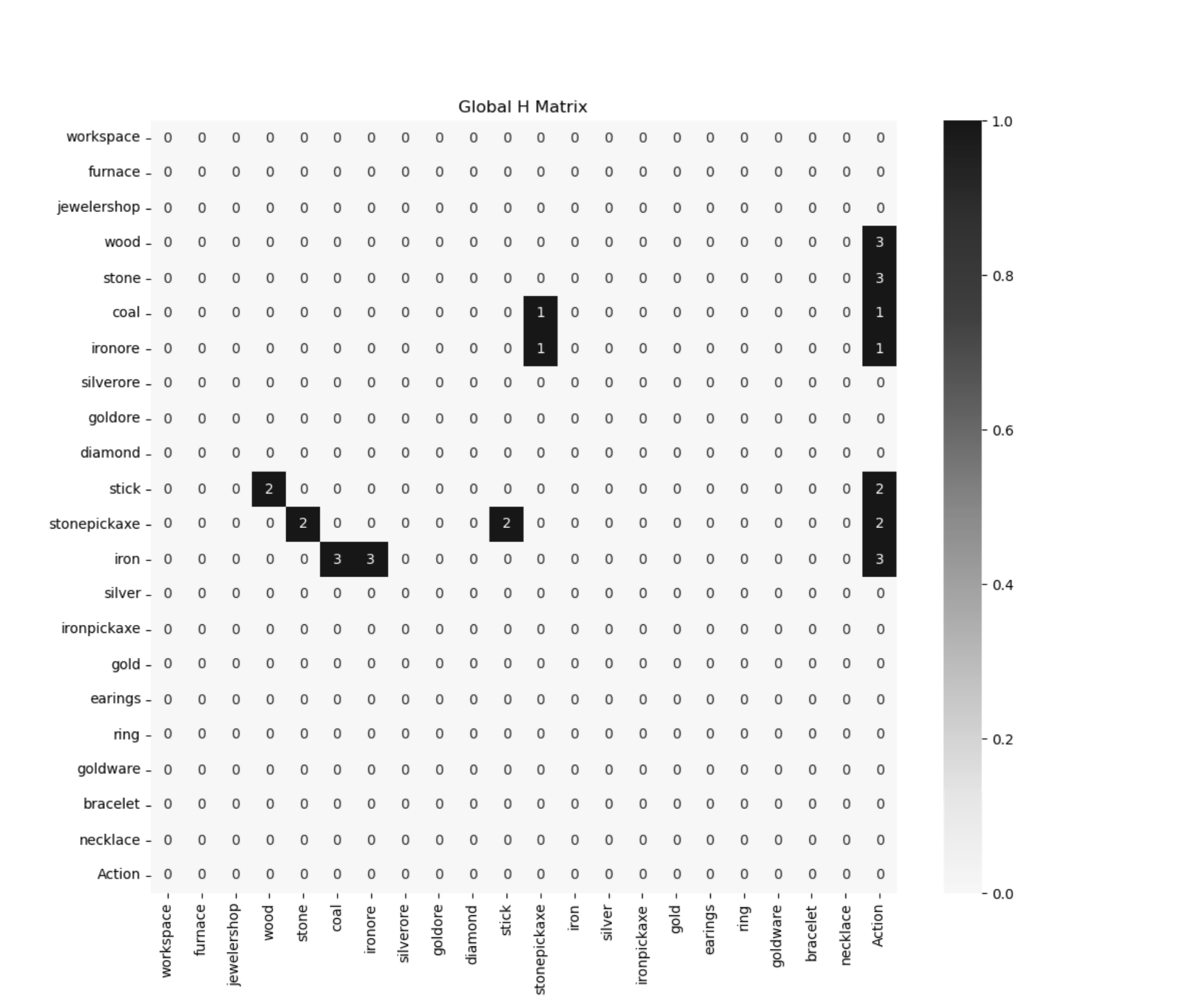} &
		\includegraphics[width=0.20\textwidth]{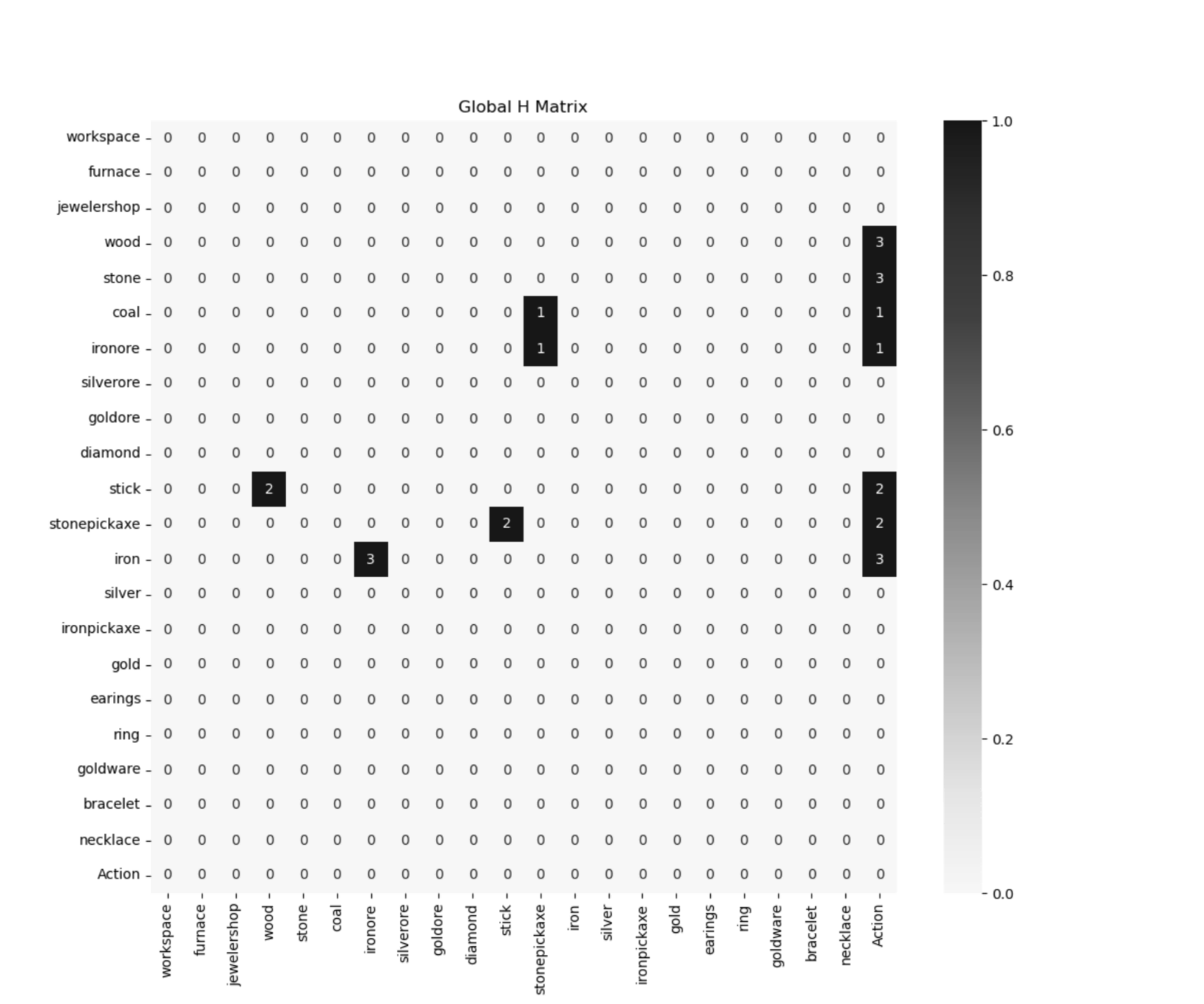} &
		\includegraphics[width=0.20\textwidth]{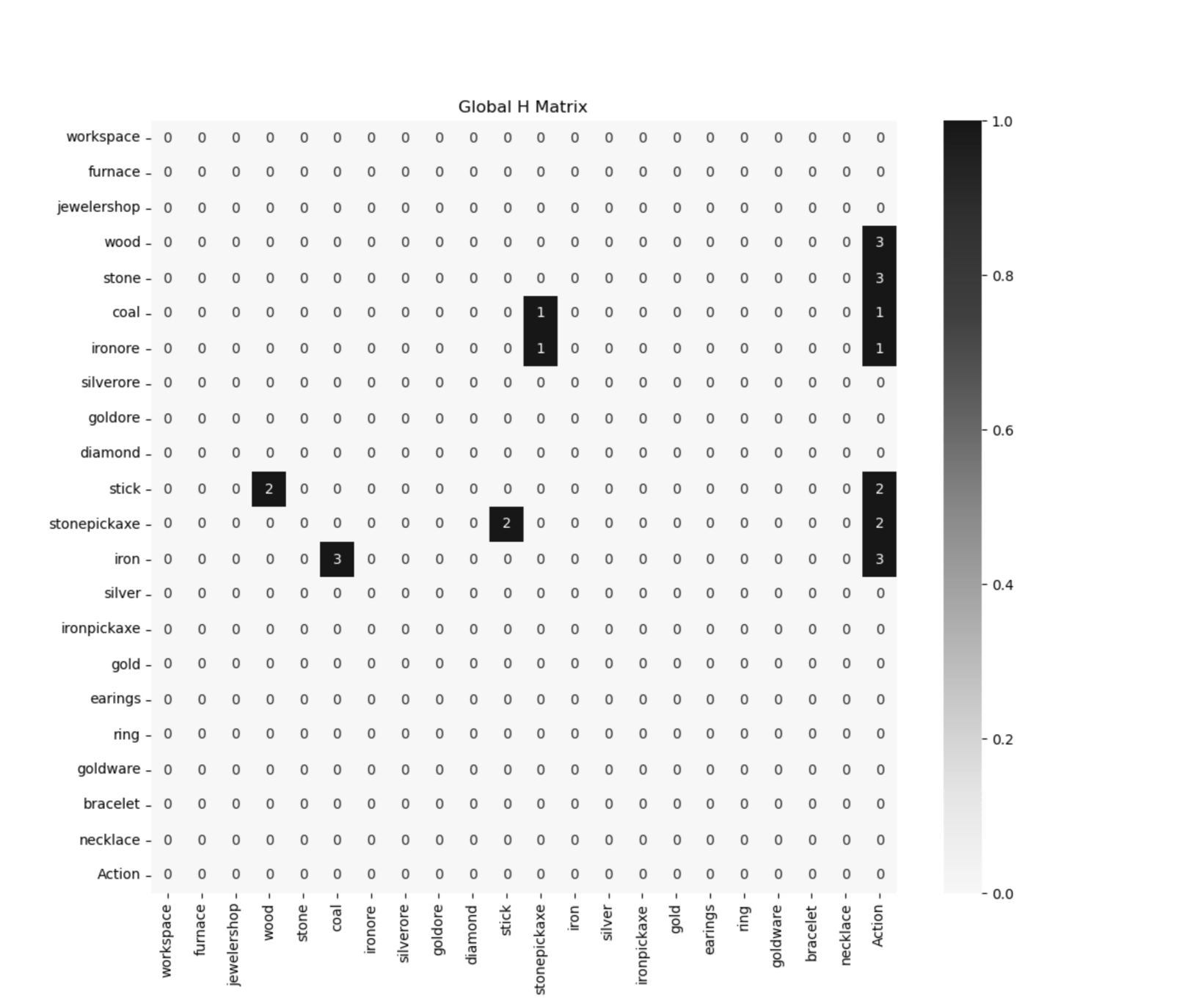} &
		\includegraphics[width=0.20\textwidth]{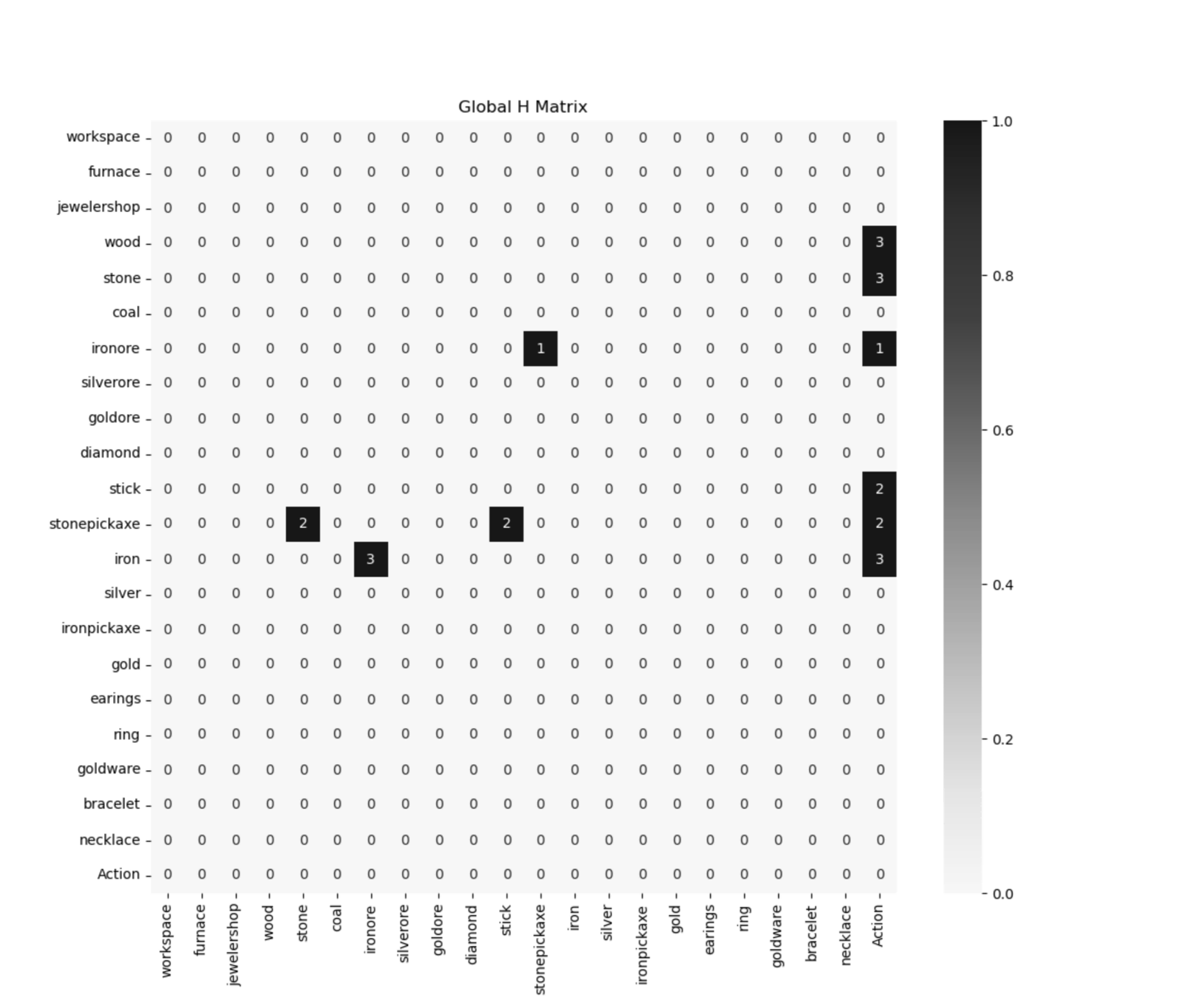} 
	\end{tabular}
	\caption{The causal graph matrices of D3HRL in GetIron-R0-T0 task (up) and GetIron-R0-T1 task (down).}
	% task with \(\tau_{max}=1\) (up) \(\tau_{max}=4\) (down)
	\label{fig:d3hrl_multi_dag}
\end{figure*}

%\noindent\textbf{2. How does D3HRL compare to CDHRL in terms of the accuracy of causal graph identification?}
\paragraph{Experimental Design}
To ensure a fair comparison, we evaluate the SHD of causal graphs identified by CDHRL and D3HRL in GetIron-R0 task and Wood2Wet task with \(\tau_{max}=1\) and \(\tau_{max}=4\), the result is shown in Table~\ref{tab:shd}. Figures~\ref{fig:cdhrl_multi_dag} presents the causal relationship matrices learned by CDHRL in GetIron-R0-T0 (\(\tau_{max}=1\)) task and GetIron-R0-T1 (\(\tau_{max}=4\)) respectively. Figures~\ref{fig:d3hrl_multi_dag} presents the causal relationship matrices learned by D3HRL in GetIron-R0-T0 (\(\tau_{max}=1\)) task and GetIron-R0-T1 (\(\tau_{max}=4\)) respectively. In these causal matrices, the rows represent effect variables \(X^j\), and the columns represent cause variables \(X^i\). If D3HRL identifies a causal relationship \(X^i \rightarrow X^j\), the corresponding element in the matrix is 1; otherwise, it is 0. Notably, in the causal relationship matrices learned by D3HRL, the numbers indicate the time spans of the causal relationships.

\begin{table*}[htbp]
	\centering
	\begin{tabular}{c|c|c|c|c|c|c}
		\toprule
		\multirow{2}{*}{\diagbox[width=4cm]{Method}{Task ($\tau_{max}$)}} & \multicolumn{2}{c|}{GetIron-R0} & \multicolumn{4}{c}{Wood2Wet} \\
		\cmidrule(r){2-7} & 1 & 4 & 1 & 4 & 8 & 16\\
		\midrule
		CDHRL & 21.4 & 24 & 1 & 1 & \diagbox{}{} & \diagbox{}{}\\
		D3HRL (ours) & \textbf{3.4} & \textbf{2} & \textbf{0.2} & \textbf{0.4} & 0.4 & 0.2\\
		\bottomrule
	\end{tabular}
	\caption{The SHD of CDHRL and D3HRL under various tasks and \(\tau_{max}\).}\label{tab:shd}
\end{table*}

%We use SHD as the evaluation metric. 
\paragraph{Experimental Results}
%significantly 
%As shown in Table~\ref{tab:shd}, D3HRL outperforms CDHRL in causal graph accuracy in both \(\tau_{max}=1\) and \(\tau_{max}=4\) tasks. Moreover, D3HRL accurately identifies each causal relationship's time span. The causal graph matrices learned by CDHRL and D3HRL are shown in Figures \ref{fig:cdhrl_multi_dag} to \ref{fig:d3hrl_multi_dag} in \ref{app:cdhrl_d3hrl_spurious_correlation_detection}, along with detailed analysis and validation of D3HRL's generalization in identifying causal graphs.
As shown in Table~\ref{tab:shd}, D3HRL consistently achieves lower SHD across all tasks, indicating that it outperforms CDHRL in terms of causal graph accuracy in both \(\tau_{max}=1\) and \(\tau_{max}=4\) tasks.
%Figures~\ref{fig:cdhrl_multi_dag} presents the causal graph matrices learned by CDHRL in GetIron-R0-T0 (\(\tau_{max}=1\)) task and GetIron-R0-T1 (\(\tau_{max}=4\)) respectively. Figures~\ref{fig:d3hrl_multi_dag} presents the causal graph matrices learned by D3HRL in GetIron-R0-T0 (\(\tau_{max}=1\)) task and GetIron-R0-T1 (\(\tau_{max}=4\)) respectively. In these causal matrices, the rows represent effect variables \(X^j\), and the columns represent cause variables \(X^i\). If the algorithm identifies a causal relationship \(X^i \rightarrow X^j\), the corresponding element in the matrix is 1; otherwise, it is 0. Notably, in the causal matrix learned by D3HRL, the numbers indicate the time spans of the causal relationships. 
By comparing these causal relationship matrices learned by CDHRL and D3HRL in Figure~\ref{fig:cdhrl_multi_dag} and Figure~\ref{fig:d3hrl_multi_dag}, it becomes evident that the causal relationship matrices learned by CDHRL includes numerous spurious correlations. In contrast, D3HRL accurately identifies the causal relationship matrices and the time spans of the corresponding causal relationships. 

\subsubsection{How does DEHRL perform in the generalization of identifying causal graphs?}
\begin{figure*}[htbp]
	\centering
	\setlength{\tabcolsep}{-2pt} % 设置列间距为 2pt
	\begin{tabular}{ccccc}
		\includegraphics[width=0.20\textwidth]{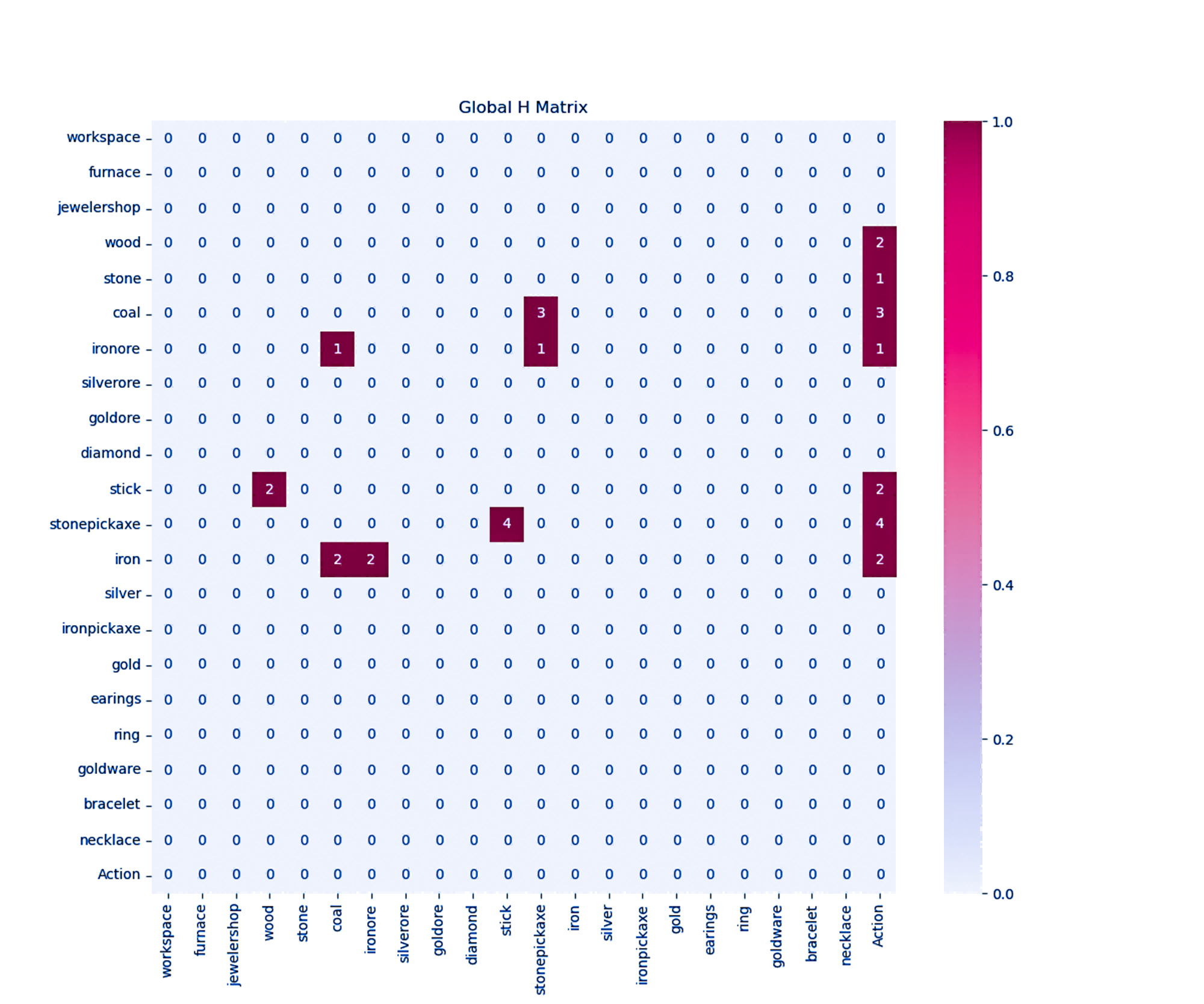} &
		\includegraphics[width=0.20\textwidth]{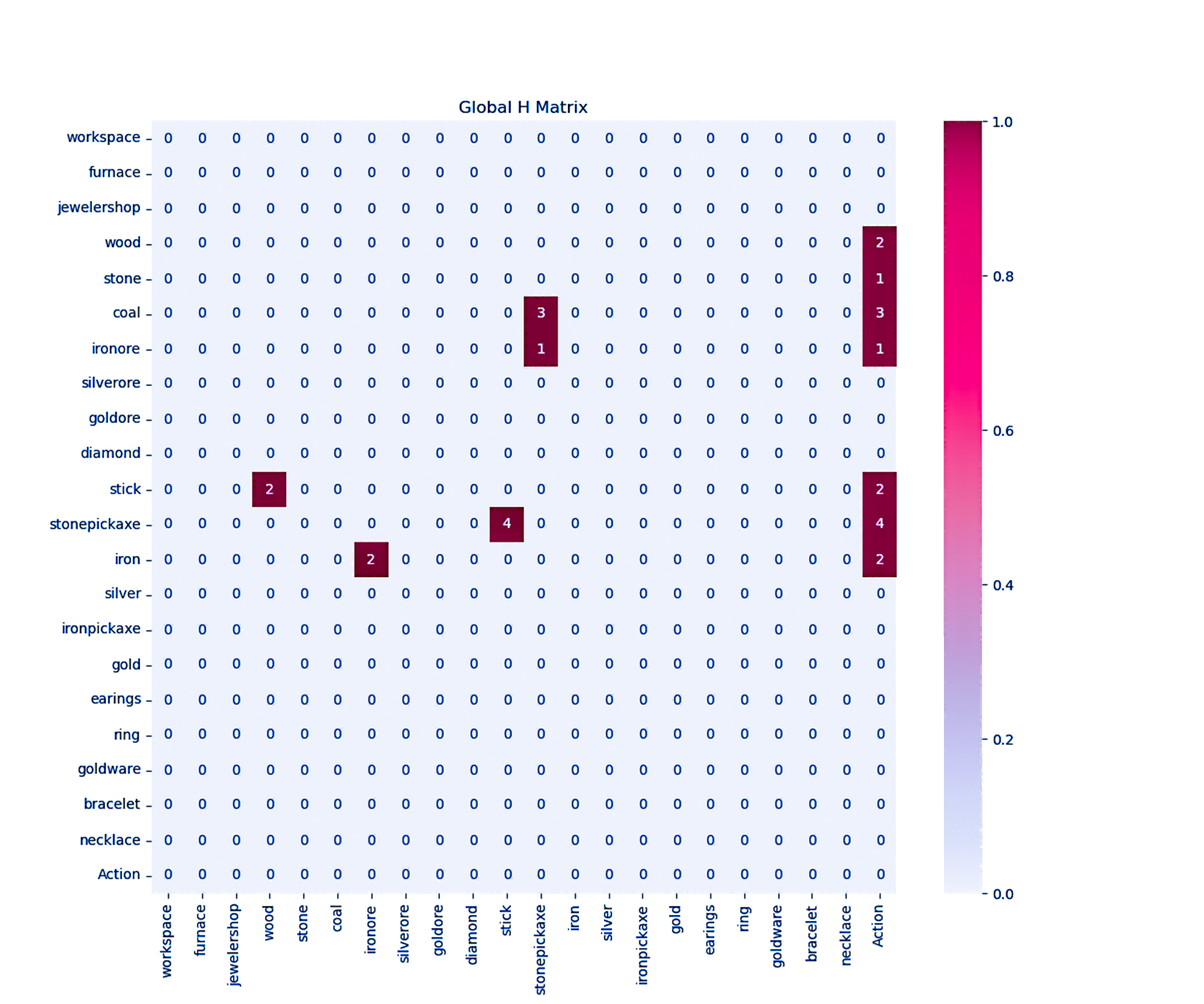} &
		\includegraphics[width=0.20\textwidth]{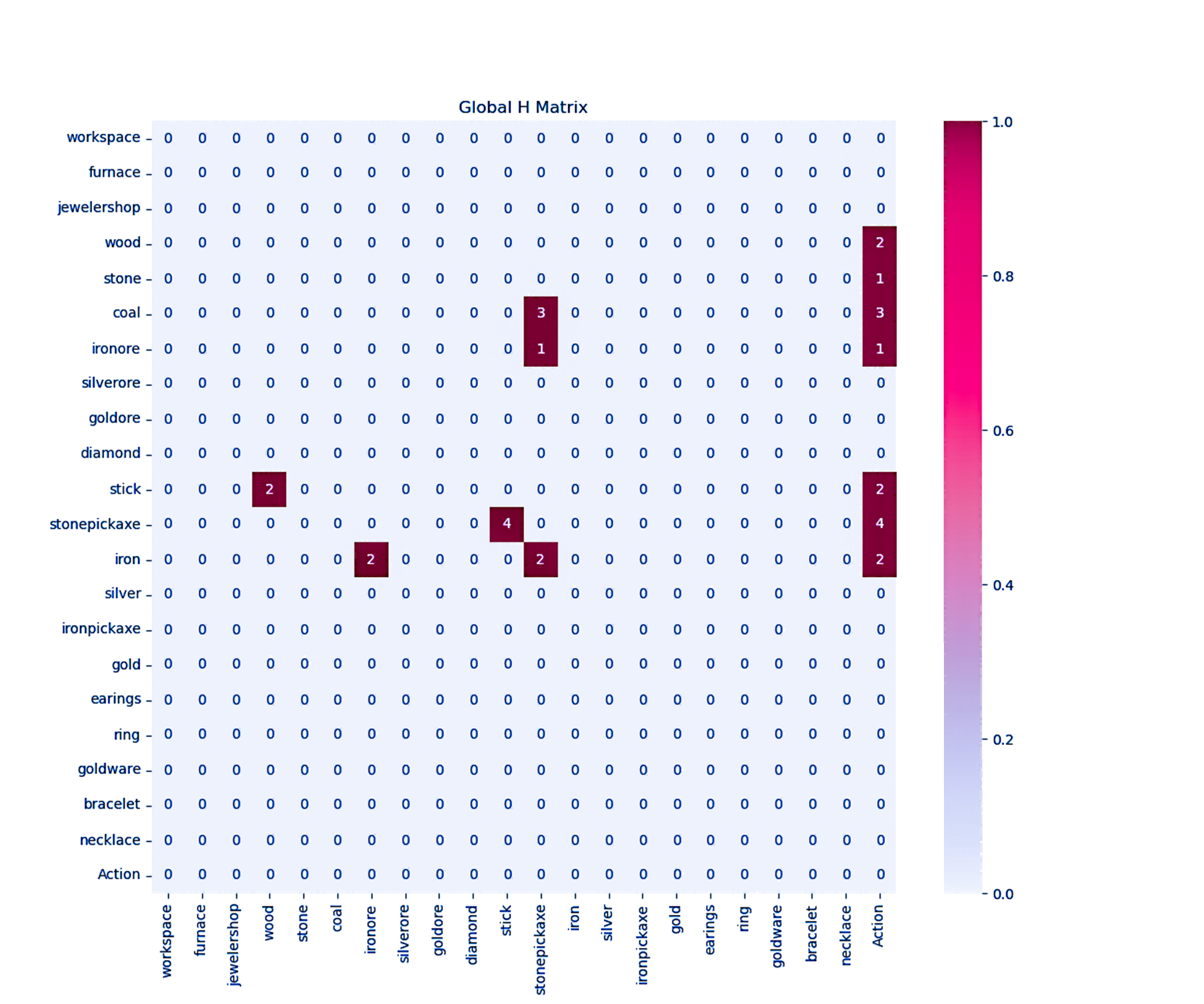} &
		\includegraphics[width=0.20\textwidth]{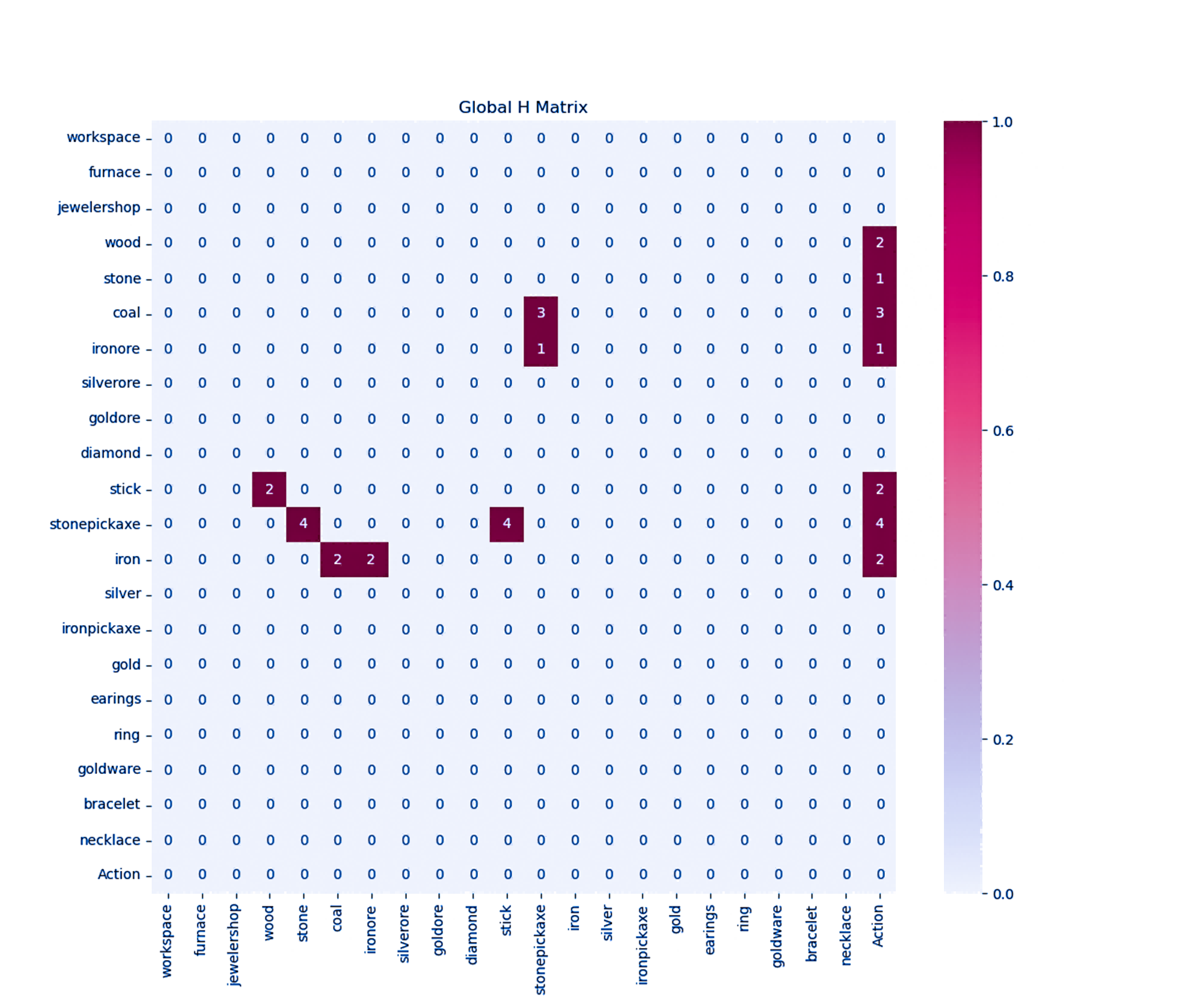} &
		\includegraphics[width=0.20\textwidth]{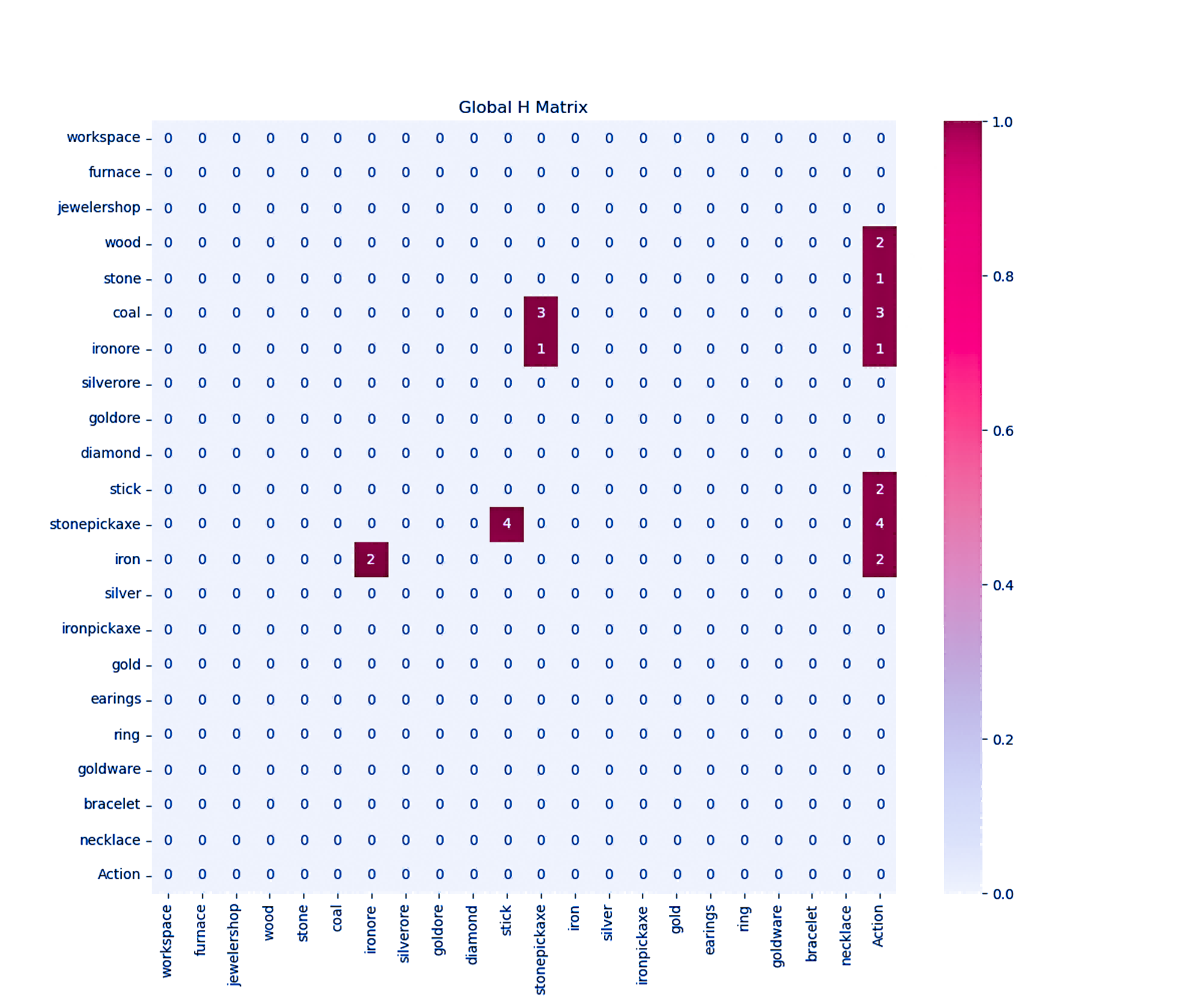} \\
		\includegraphics[width=0.20\textwidth]{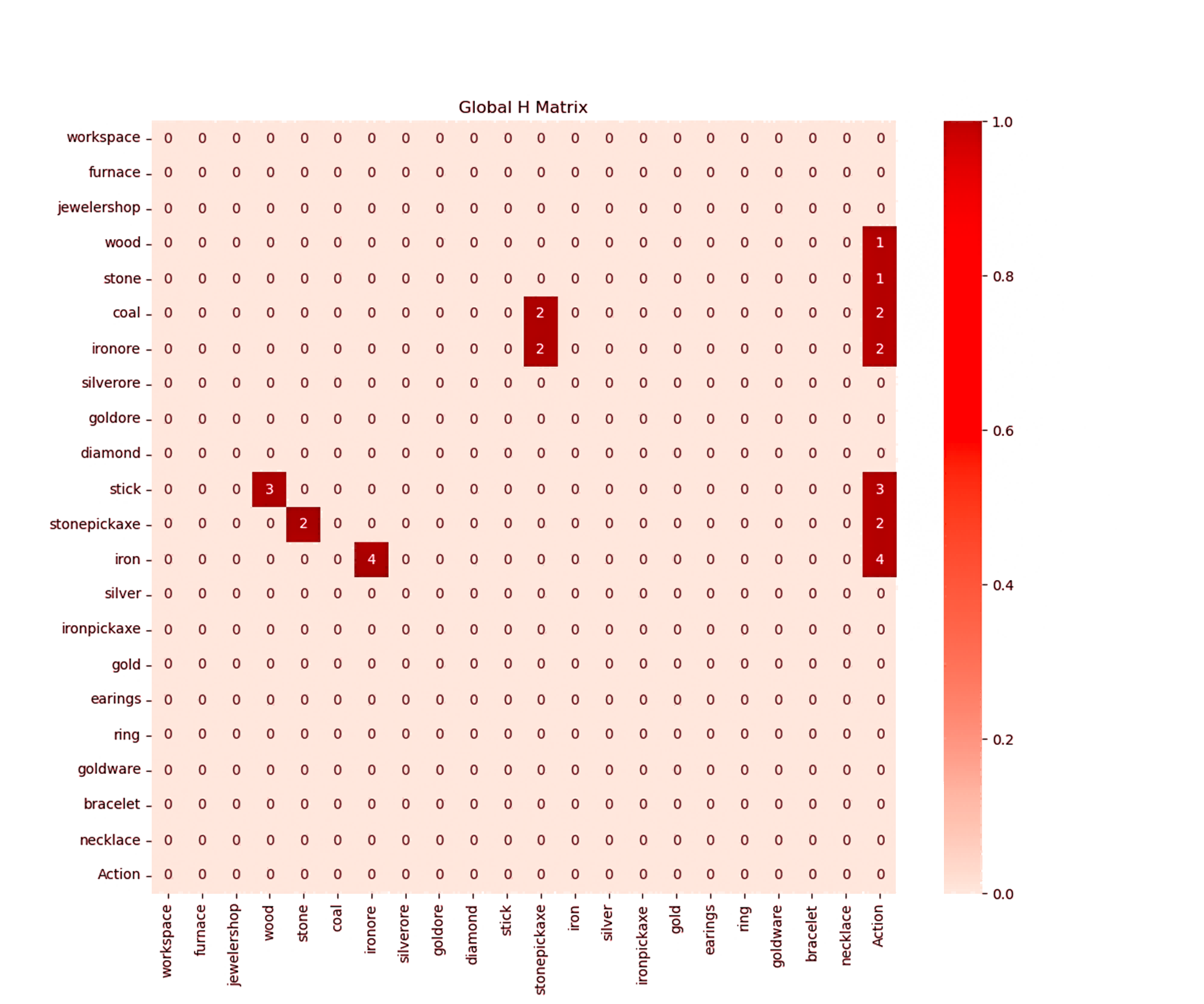} &
		\includegraphics[width=0.20\textwidth]{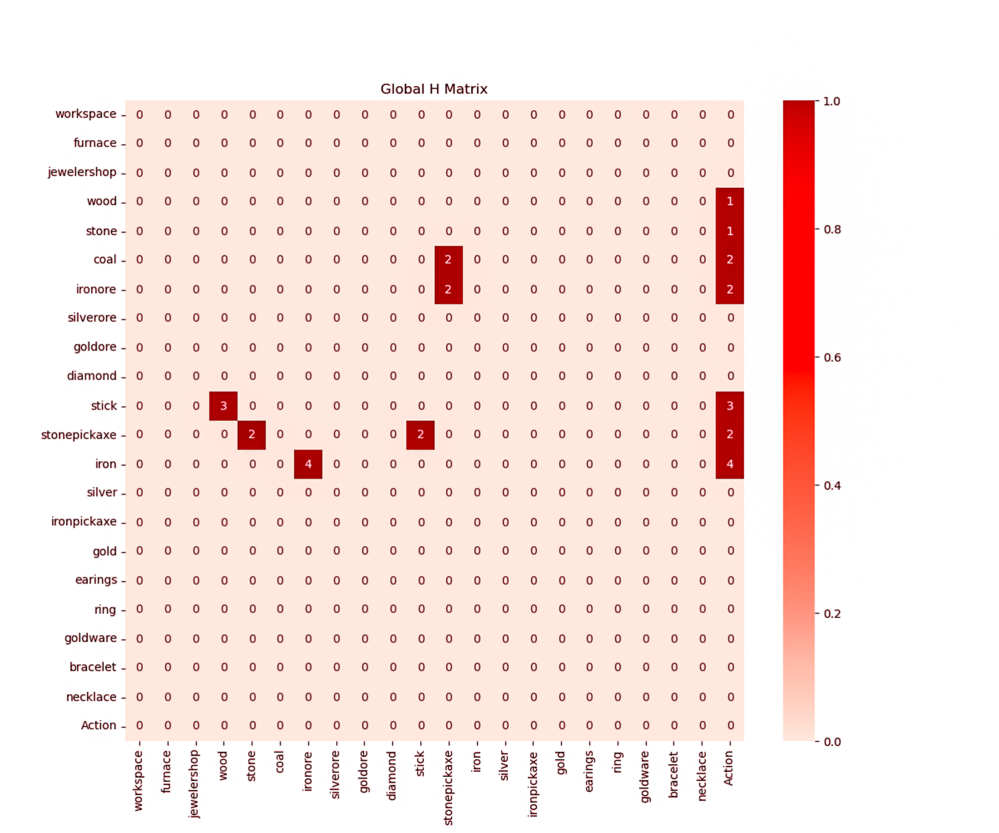} &
		\includegraphics[width=0.20\textwidth]{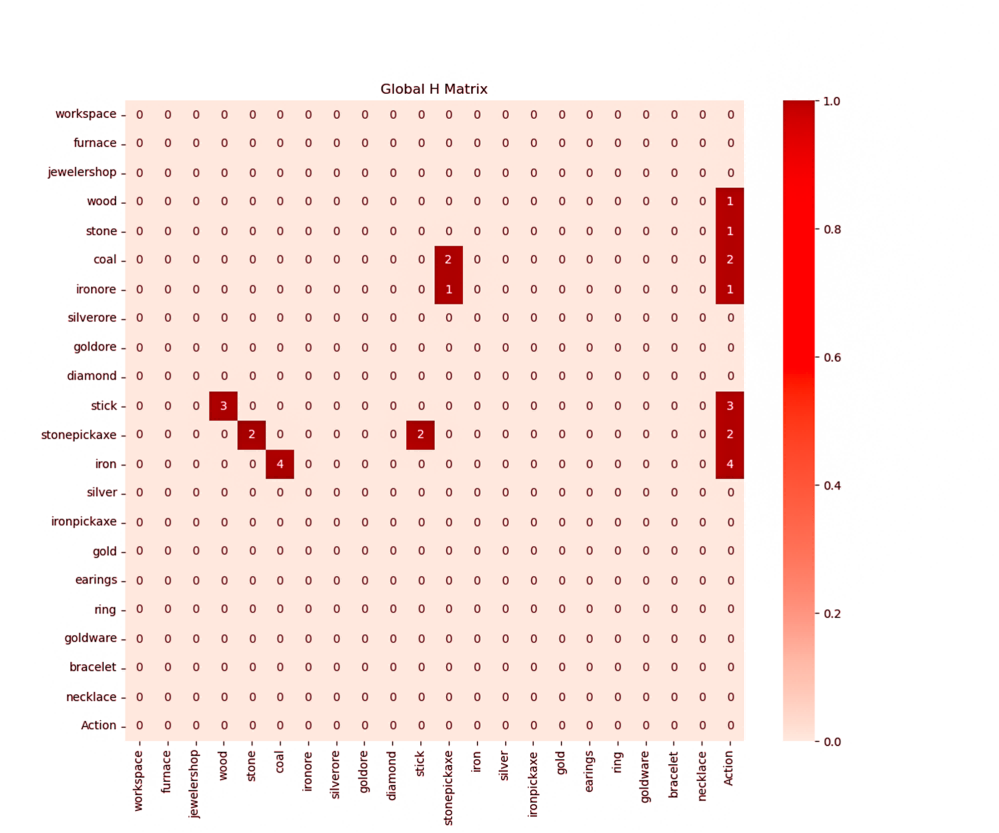} &
		\includegraphics[width=0.20\textwidth]{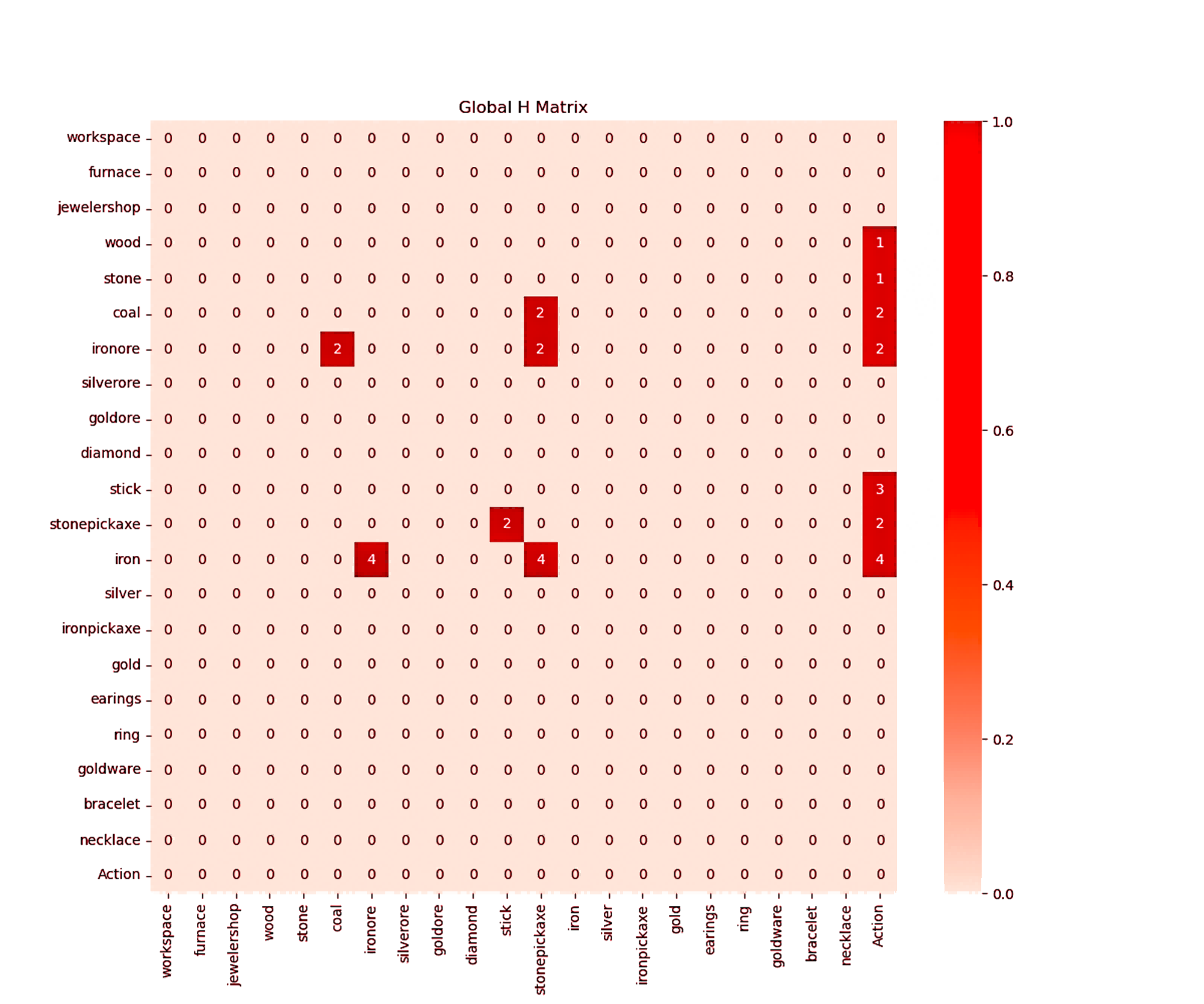} &
		\includegraphics[width=0.20\textwidth]{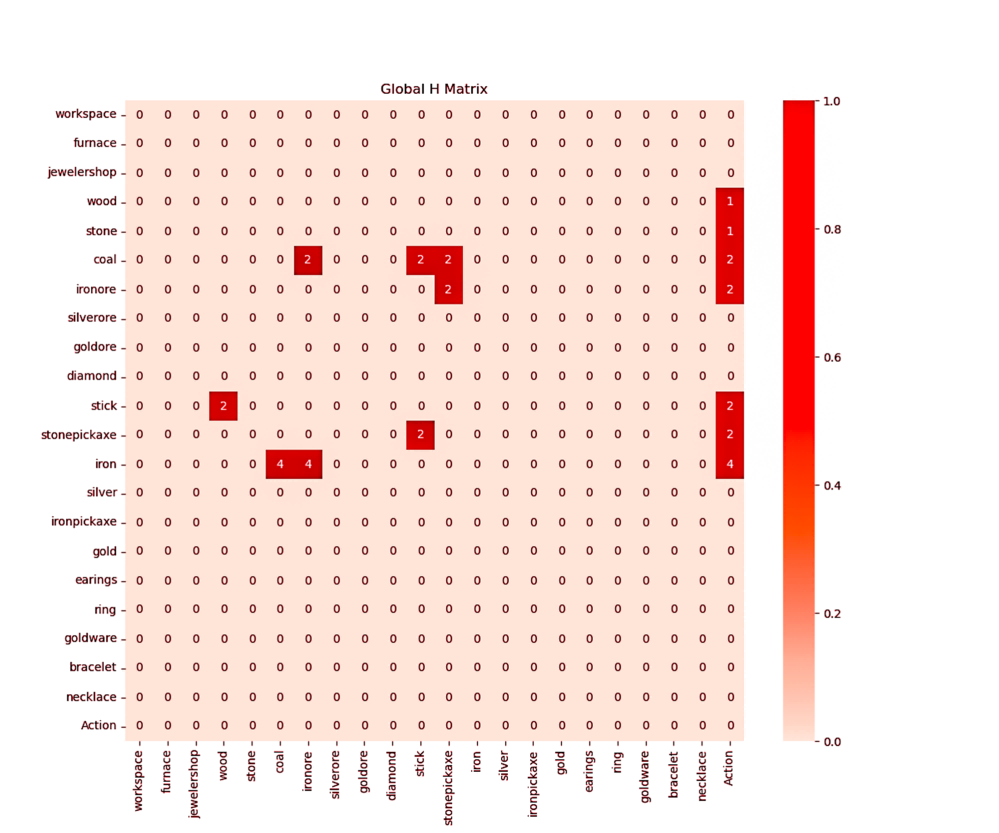} 
	\end{tabular}
	\caption{The learned causal graph matrices of D3HRL in GetIron-R0-T2 task (up) and GetIron-R0-T3 task (down).}
	% with \(\tau_{max}=4\) under two additional differnent causal relationship time span configuration
	\label{fig:d3hrl_v2v3}
\end{figure*}

\paragraph{Experimental Design} To validate the generalization of D3HRL in identifying causal graphs, we recorded the causal graphs identified by D3HRL on the GetIron task under various causal time span configurations, the results are shown in Figure~\ref{fig:d3hrl_v2v3}. %True State transitions and the true causal graph matrix for the GetIron task under two additional different causal time span configurations (GetIron-R0-T2 and GetIron-R0-T3) are shown in Figure~\ref{fig:get_iron}. 

\paragraph{Experimental Results} The results indicate that D3HRL consistently produces accurate causal graphs across different configurations. This clearly demonstrates D3HRL's robust generalization in causal graph identification.

%Moreover, D3HRL accurately identifies each causal relationship's time span. The causal graph matrices learned by CDHRL and D3HRL are shown in Figures \ref{fig:cdhrl_multi_dag} to \ref{fig:d3hrl_multi_dag}, along with detailed analysis and validation of D3HRL's generalization in identifying causal graphs.
% -----------------------------------------------------------

%Furthermore, we validate the generalization ability of D3HRL in identifying causal graphs, see \ref{app:cdhrl_d3hrl_spurious_correlation_detection} for details.
%As shown in Table~\ref{tab:shd}, D3HRL significantly outperforms CDHRL in terms of causal identification accuracy, in both \(\tau_{max}=1\) and \(\tau_{max}=4\) tasks. Moreover, D3HRL accurately identifies the time span of each causal relationship. We present the causal graph matrices learned by CDHRL and D3HRL in Figure \ref{fig:cdhrl_multi_dag} to \ref{fig:d3hrl_multi_dag} in Appendix \ref{app:cdhrl_d3hrl_spurious_correlation_detection}, accompanied by detailed analysis.
%Furthermore, to validate the generalization ability of D3HRL in identifying causal graphs, we evaluated and recorded the causal graphs identified by D3HRL using the GetIron task with varying causal time span configurations. The detailed results are provided in Appendix C.

\begin{comment}
	\begin{table}[h]
		\centering
		\caption{The SHD of CDHRL and D3HRL.}\label{tab:shd}
		\begin{tabular}{c|c|c|c|c}
			\toprule
			\multirow{2}{*}{Method} & \multicolumn{2}{c|}{MineCraft} & \multicolumn{2}{c}{MiniGrid} \\
			\cmidrule(r){2-5}
			& Single & Multi & Single & Multi \\
			\midrule
			CDHRL  & 14  &  23  &  1    & 1     \\
			D3HRL  & 5   &  1.8 &  0.2  & 0.4     \\
			\bottomrule
		\end{tabular}
	\end{table}
\end{comment}

\subsubsection{Does D3HRL filter out spurious correlations?}
\begin{figure*}[htbp]
	\centering
	\setlength{\tabcolsep}{1pt} % 设置列间距为 2pt
	\begin{tabular}{ccccc}
		\includegraphics[width=0.2\textwidth]{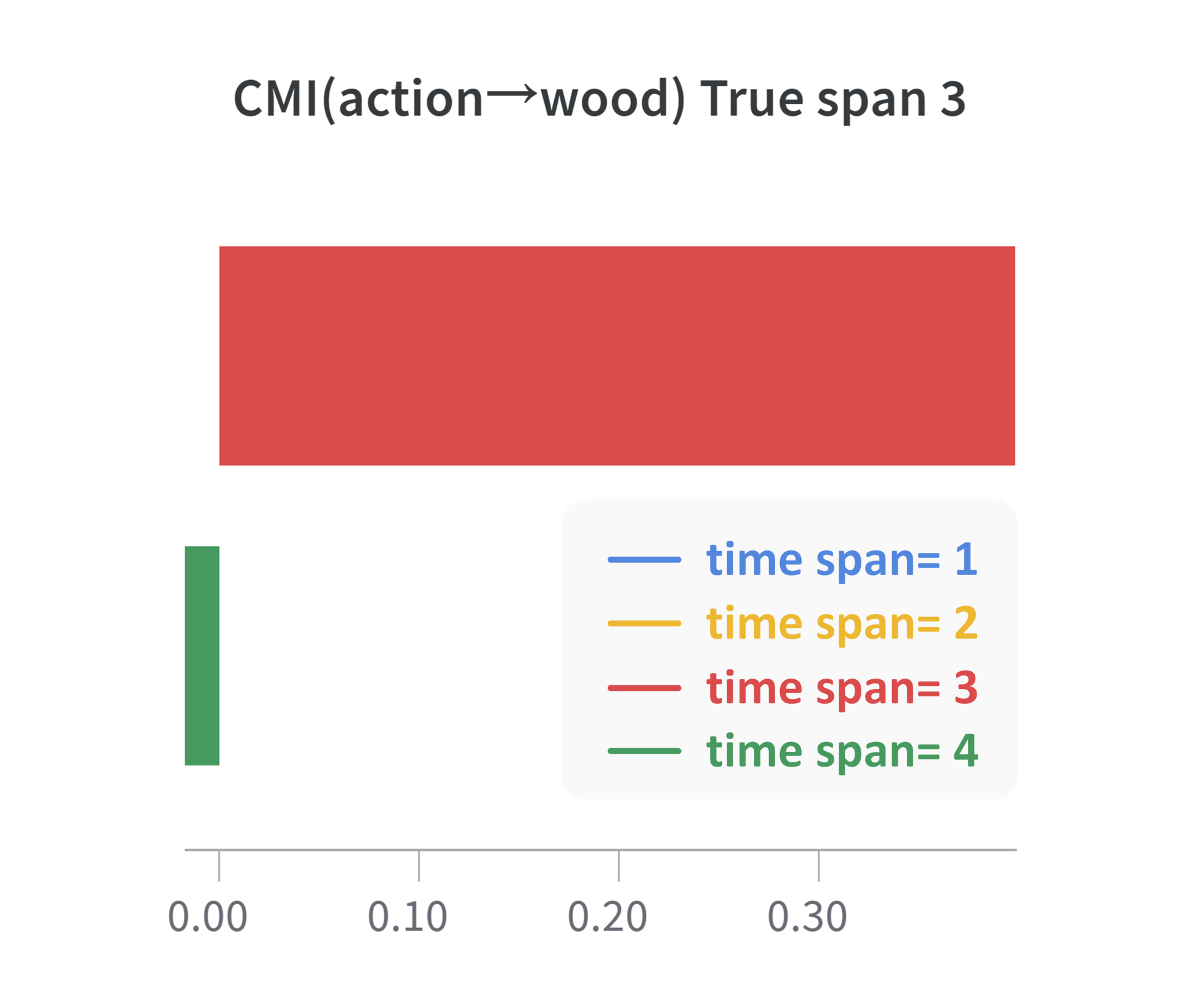} &
		\includegraphics[width=0.2\textwidth]{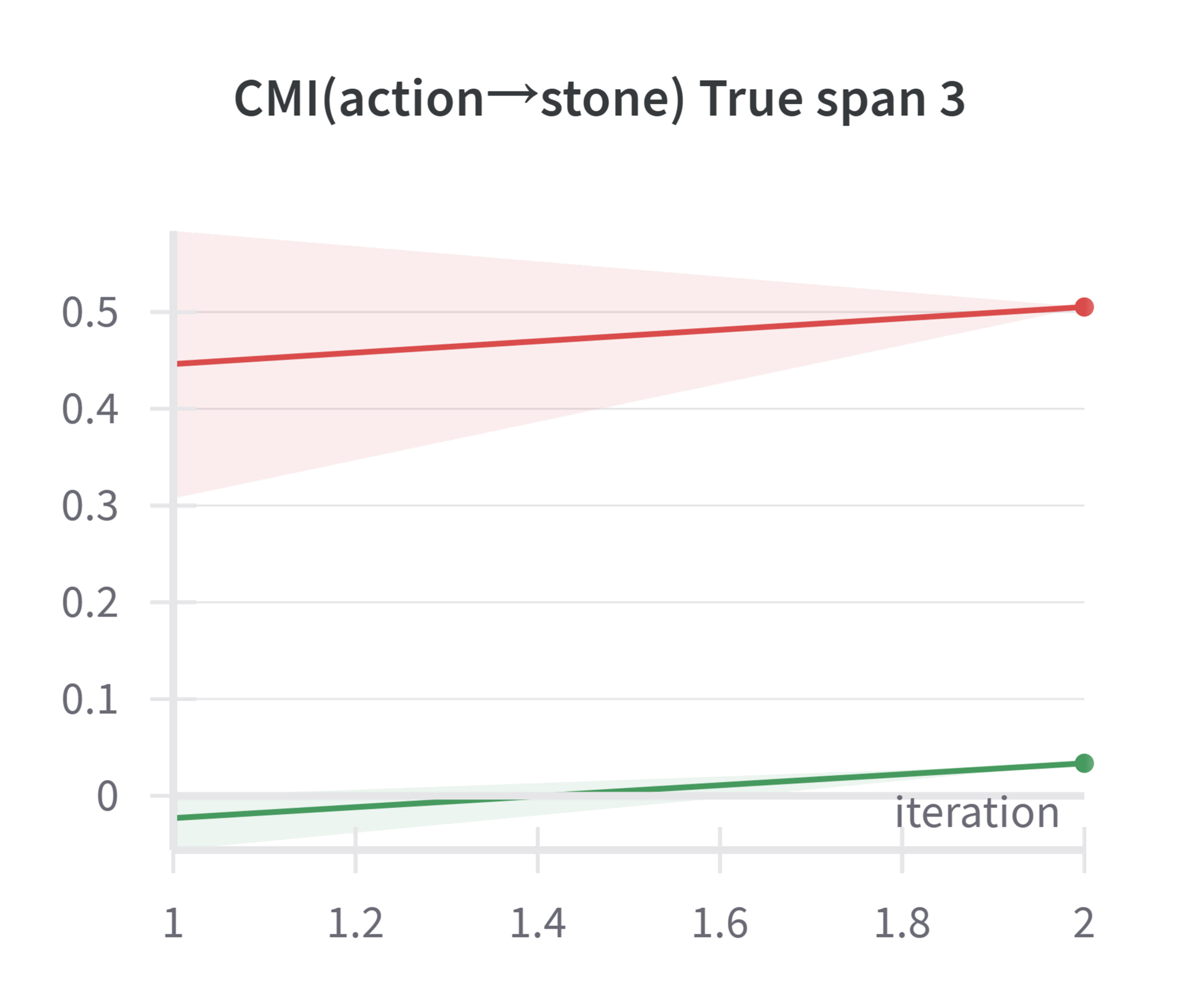} &
		\includegraphics[width=0.2\textwidth]{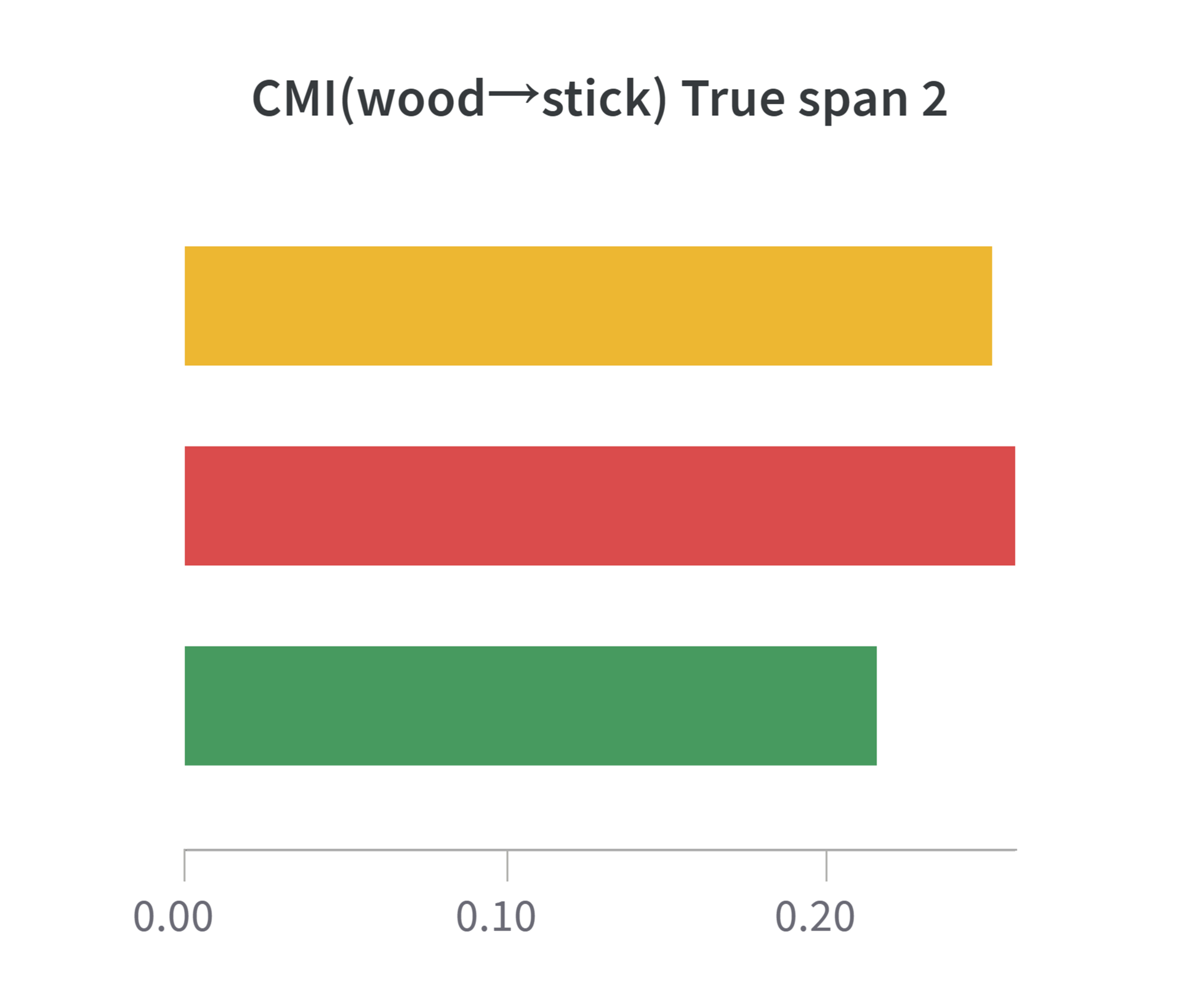} &
		\includegraphics[width=0.2\textwidth]{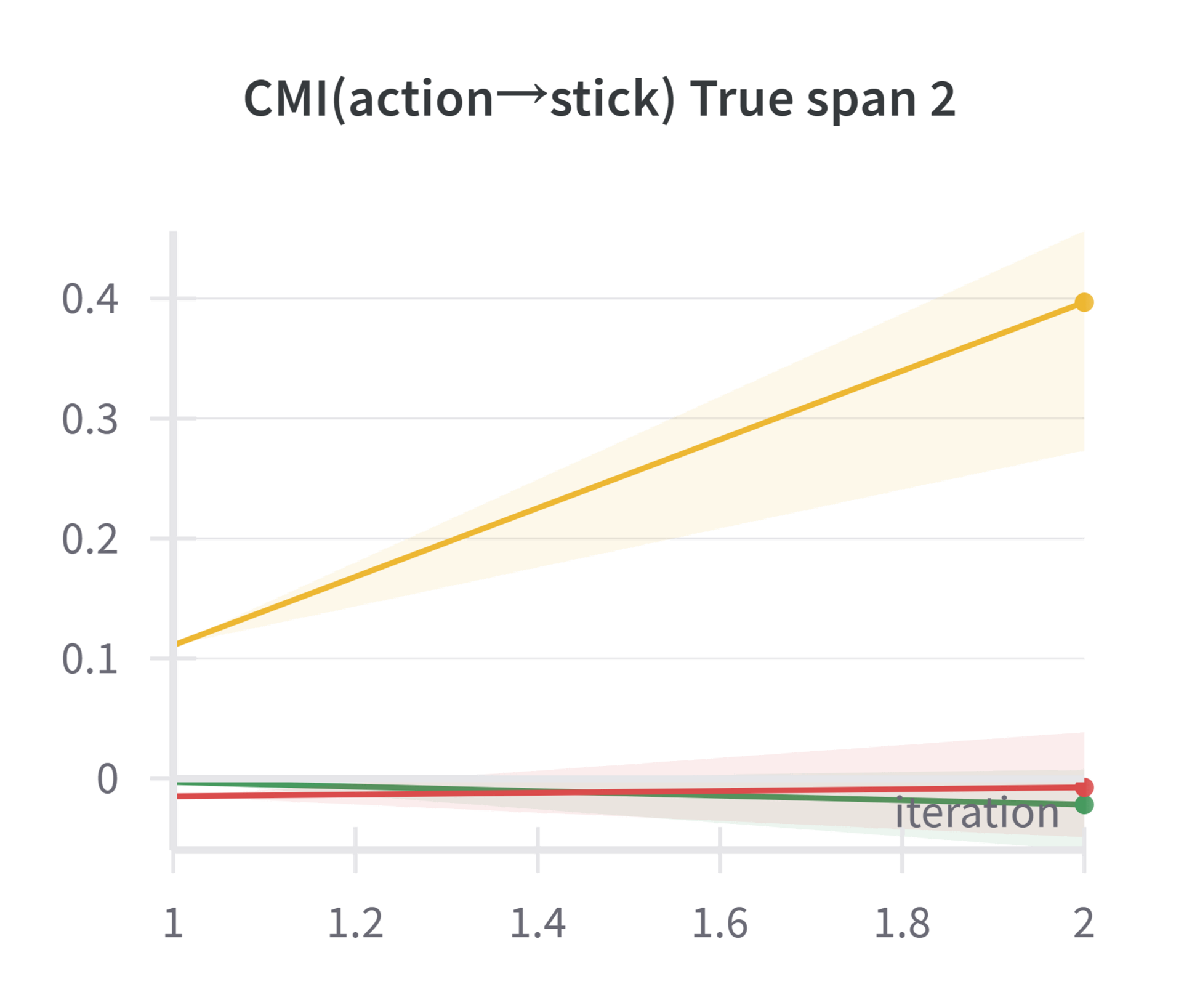} &
		\includegraphics[width=0.2\textwidth]{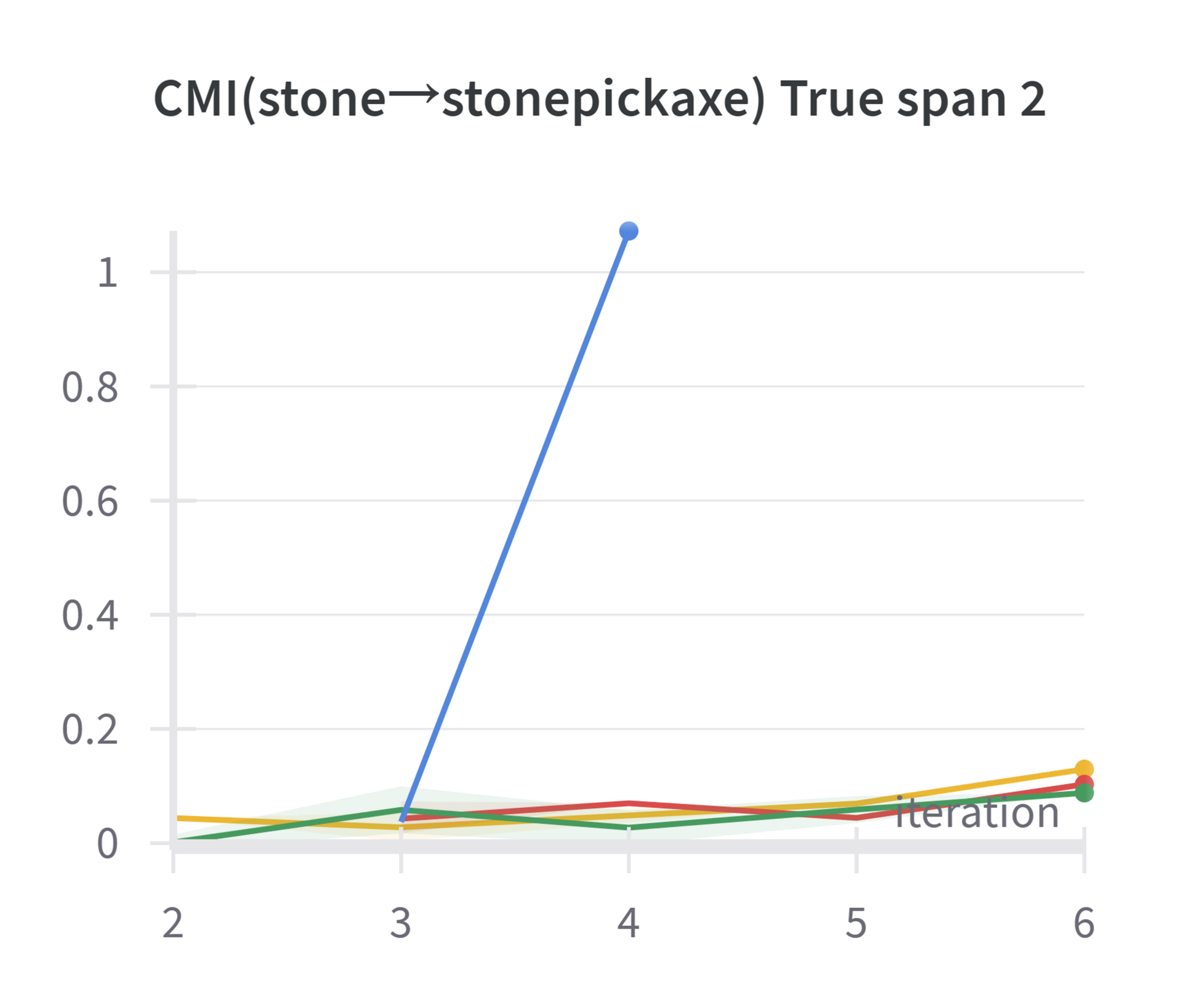} \\
		\includegraphics[width=0.2\textwidth]{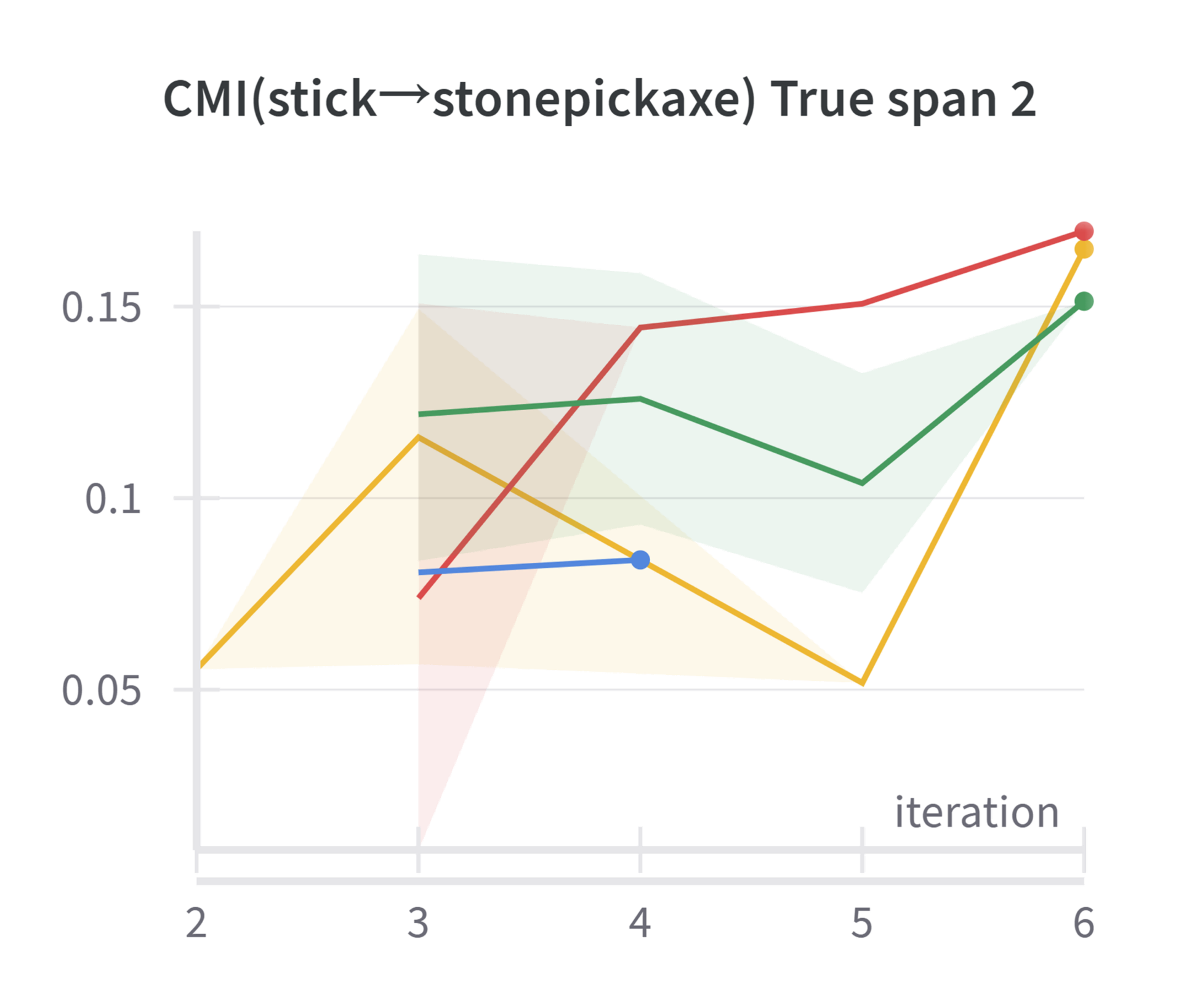} &
		\includegraphics[width=0.2\textwidth]{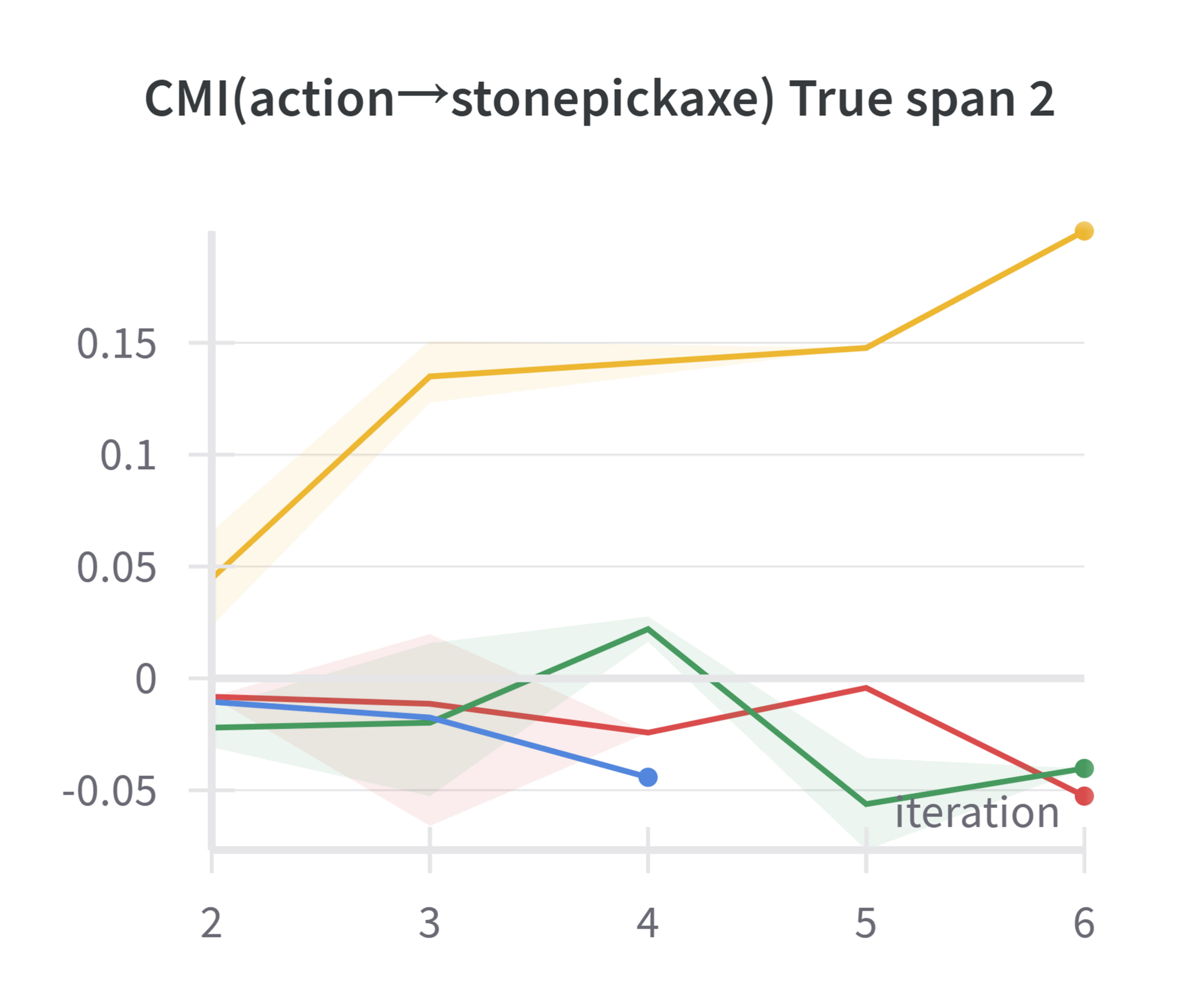} &
		\includegraphics[width=0.2\textwidth]{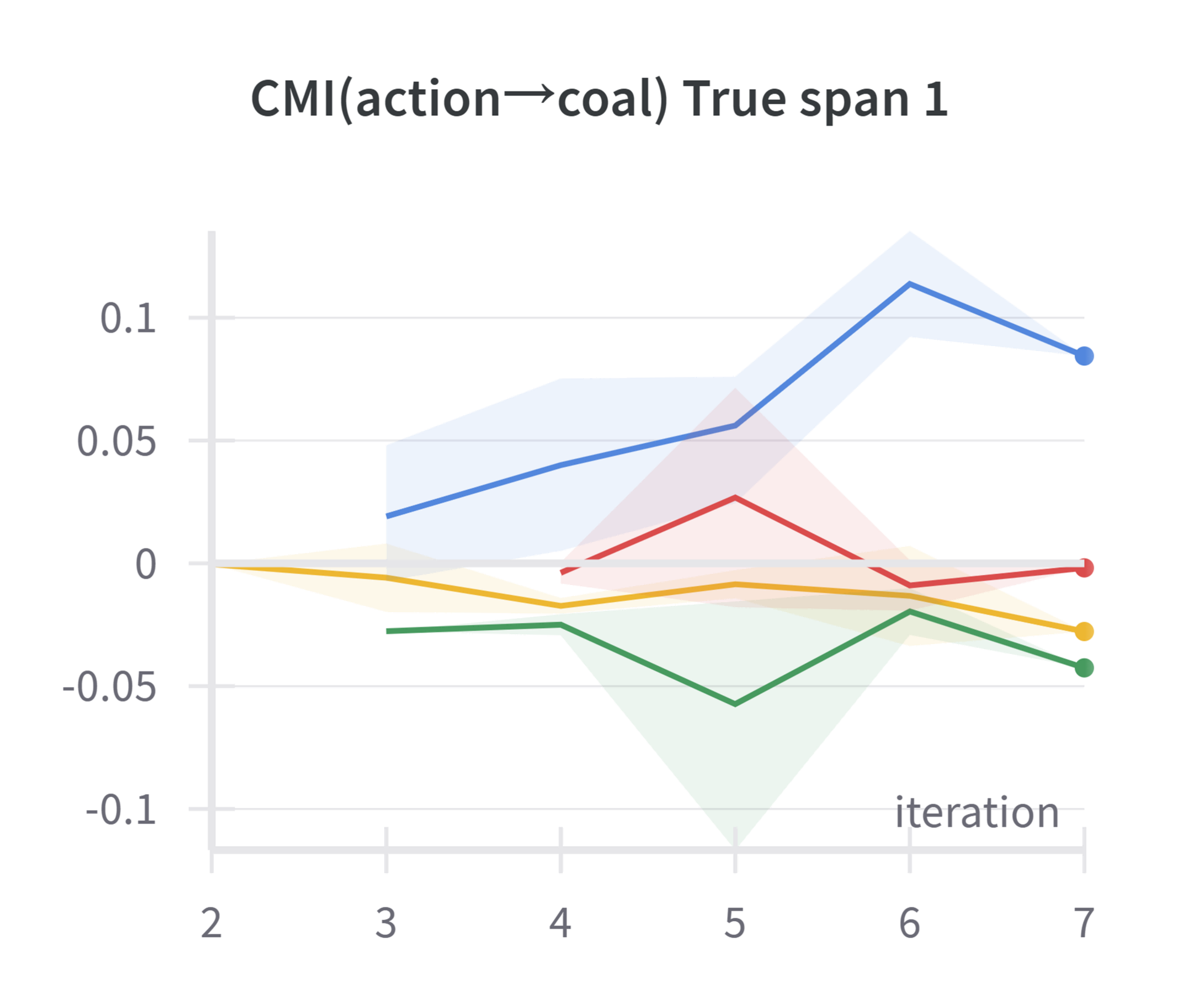} &
		\includegraphics[width=0.2\textwidth]{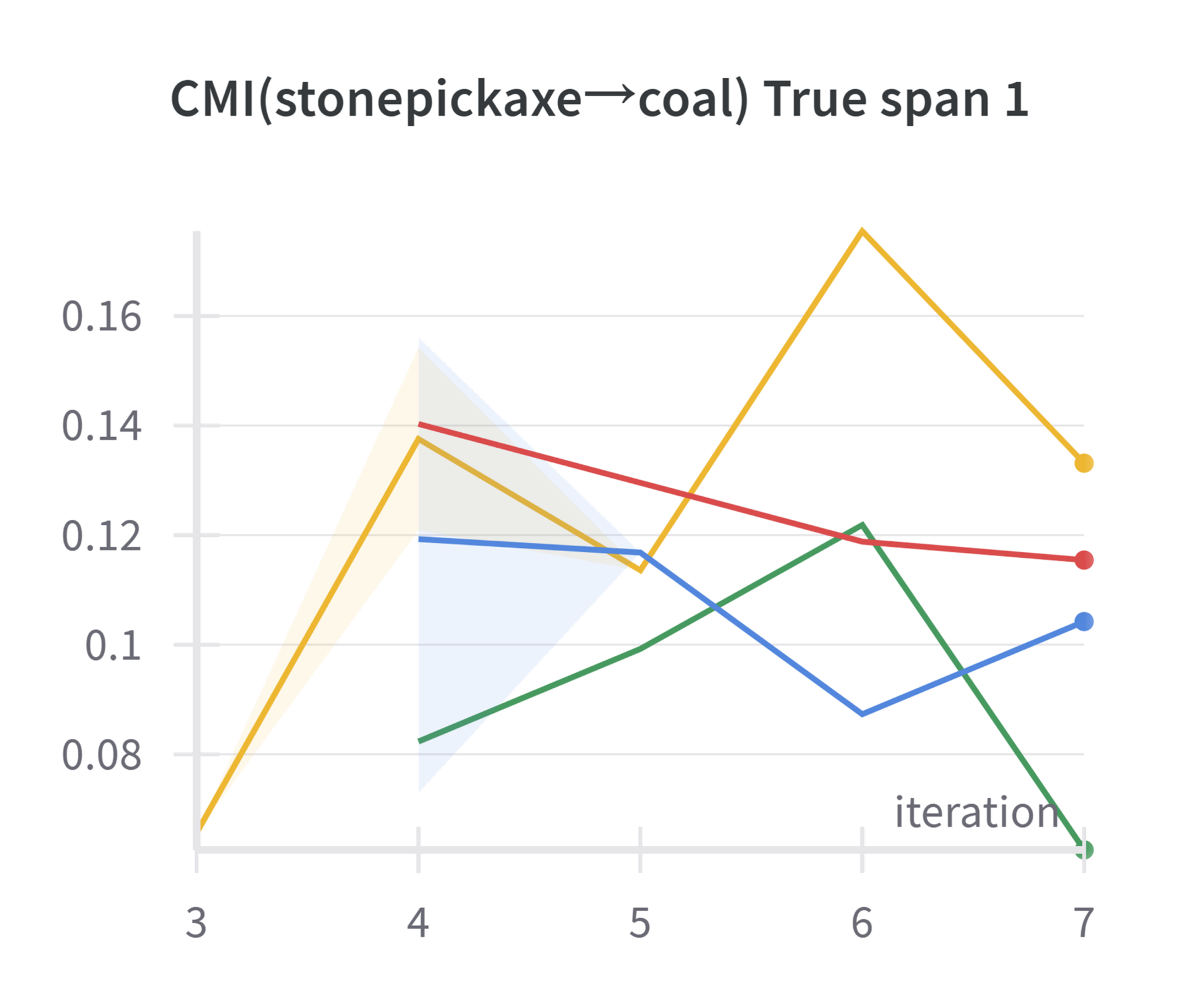} &
		\includegraphics[width=0.2\textwidth]{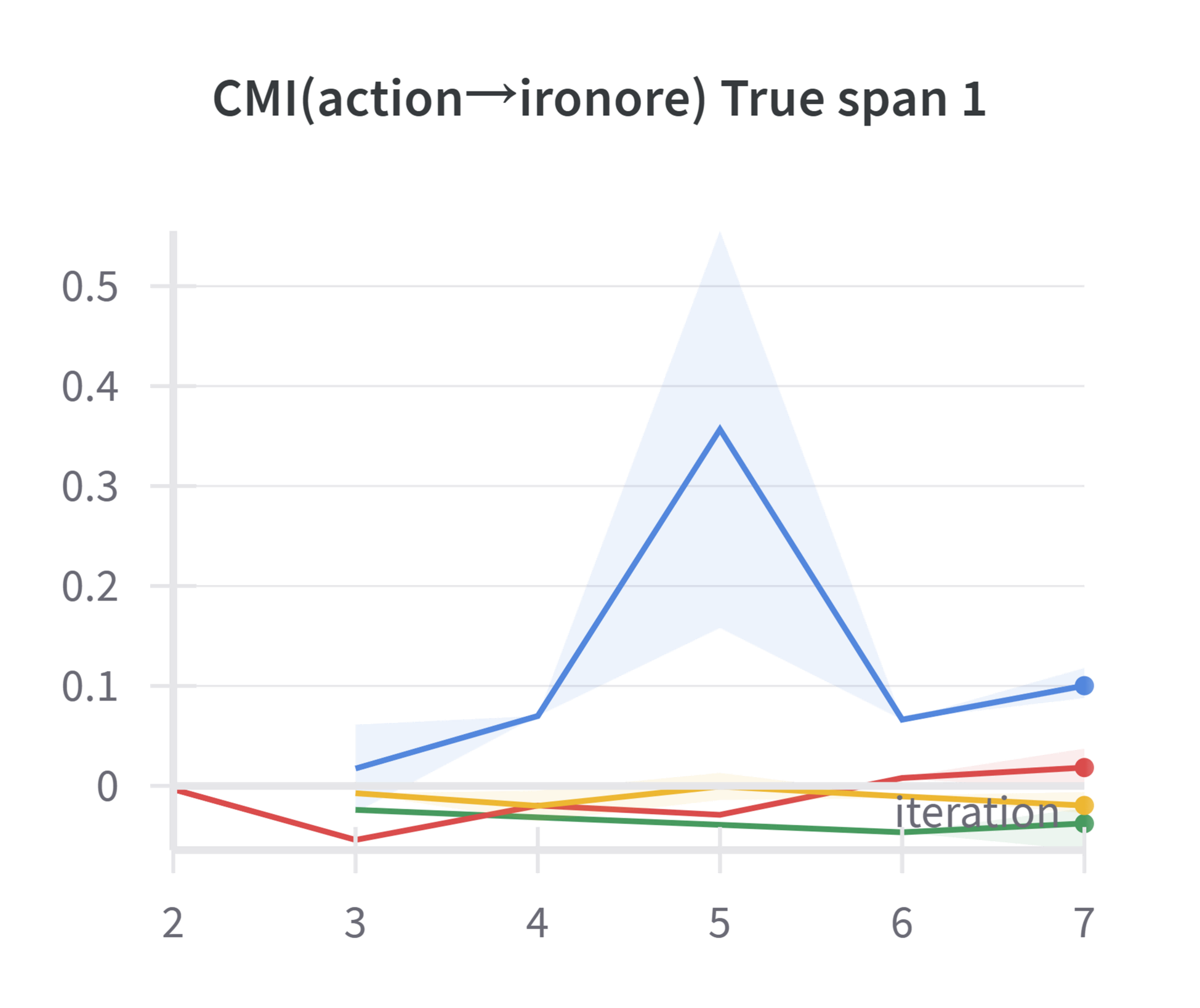} \\
		\includegraphics[width=0.2\textwidth]{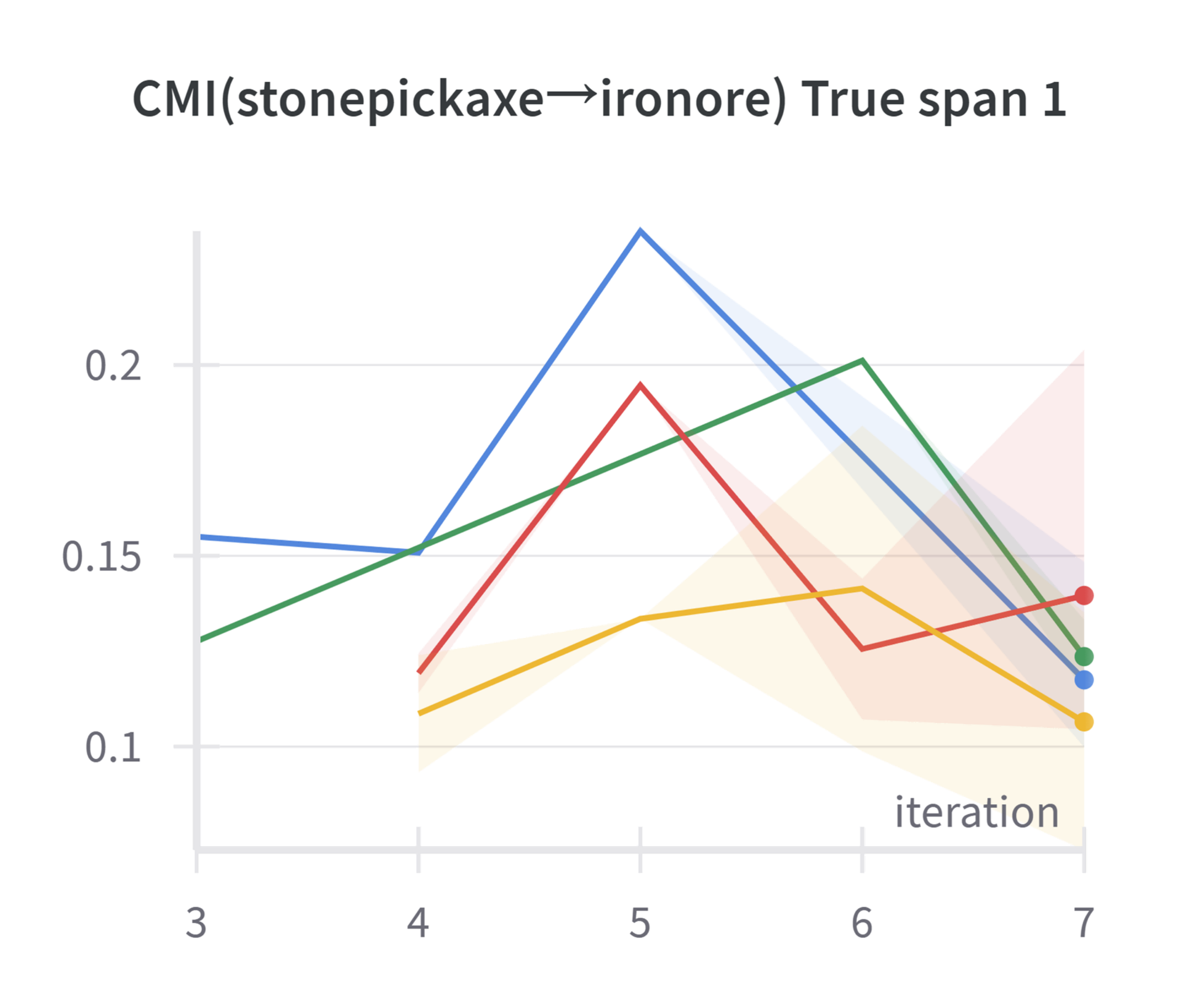} &
		\includegraphics[width=0.2\textwidth]{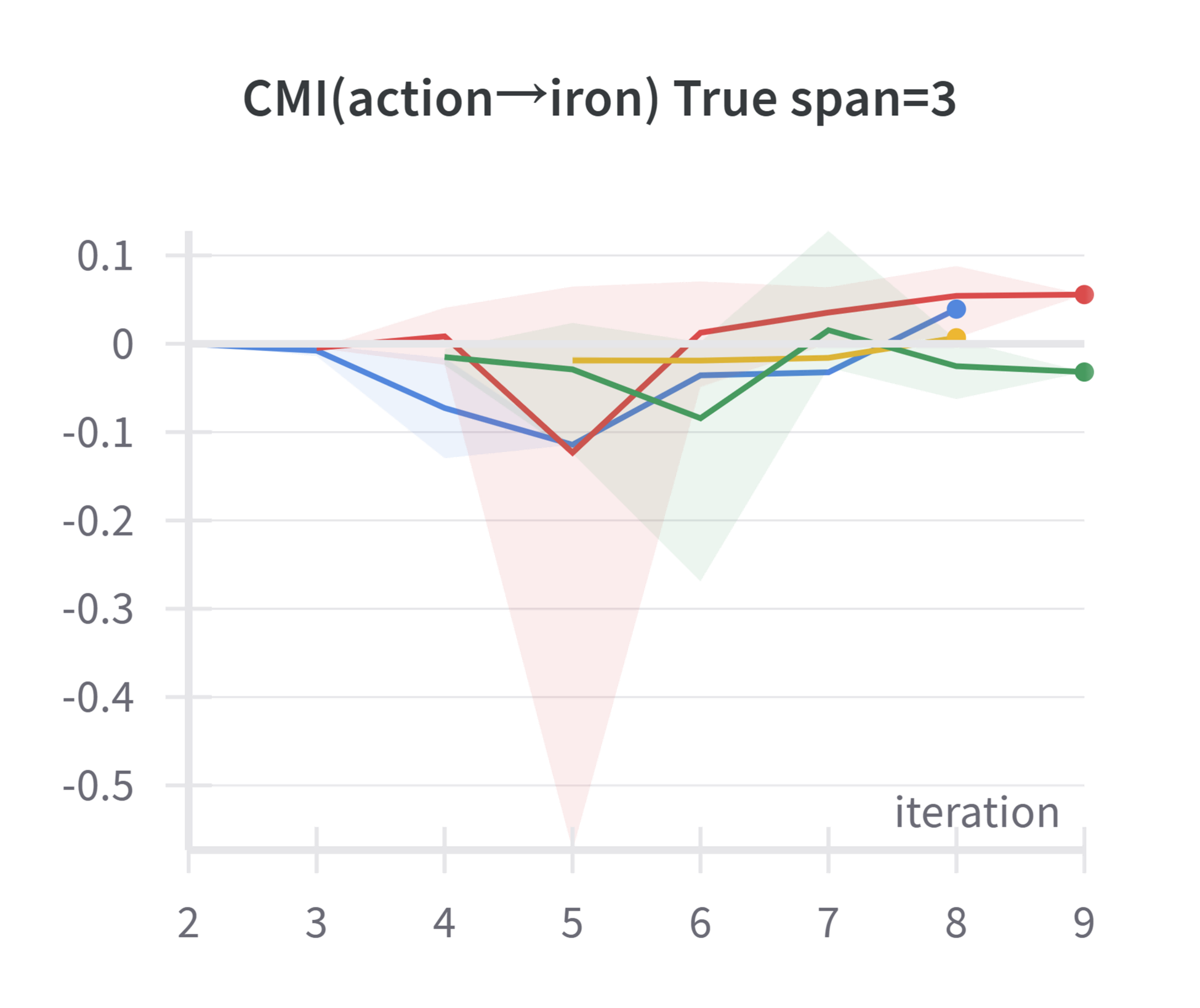} &
		\includegraphics[width=0.2\textwidth]{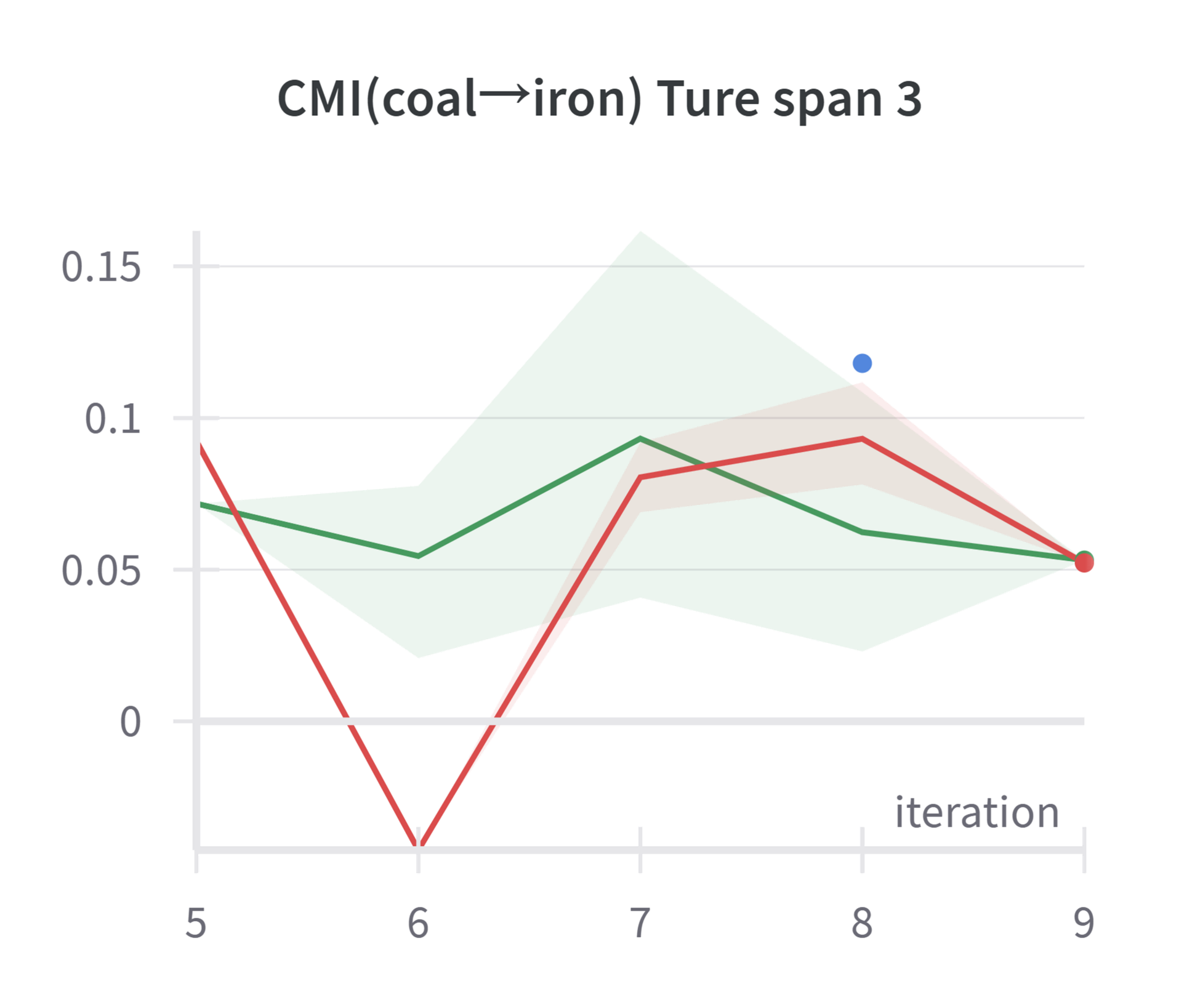} &
		\includegraphics[width=0.2\textwidth]{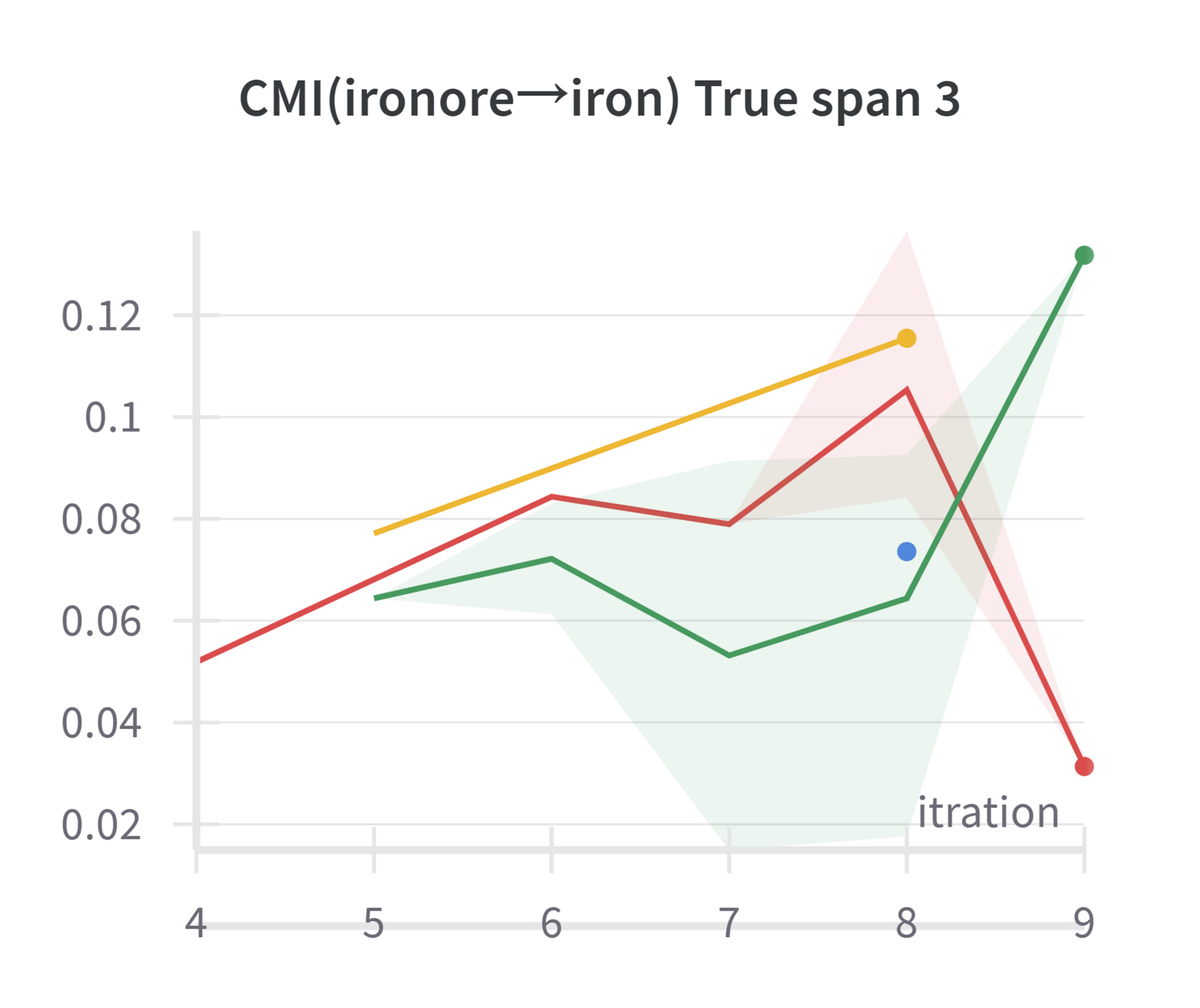} 
	\end{tabular}
	\caption{The CMI of different time spans of \textbf{causal relationships} in GetIron-R0-T1 task with \(\tau_{max}=4\).}
	\label{fig:cmi_}
\end{figure*}

\begin{figure*}[htbp]
	\centering
	\setlength{\tabcolsep}{1pt} % 设置列间距为 2pt
	\begin{tabular}{ccccc}
		\includegraphics[width=0.2\textwidth]{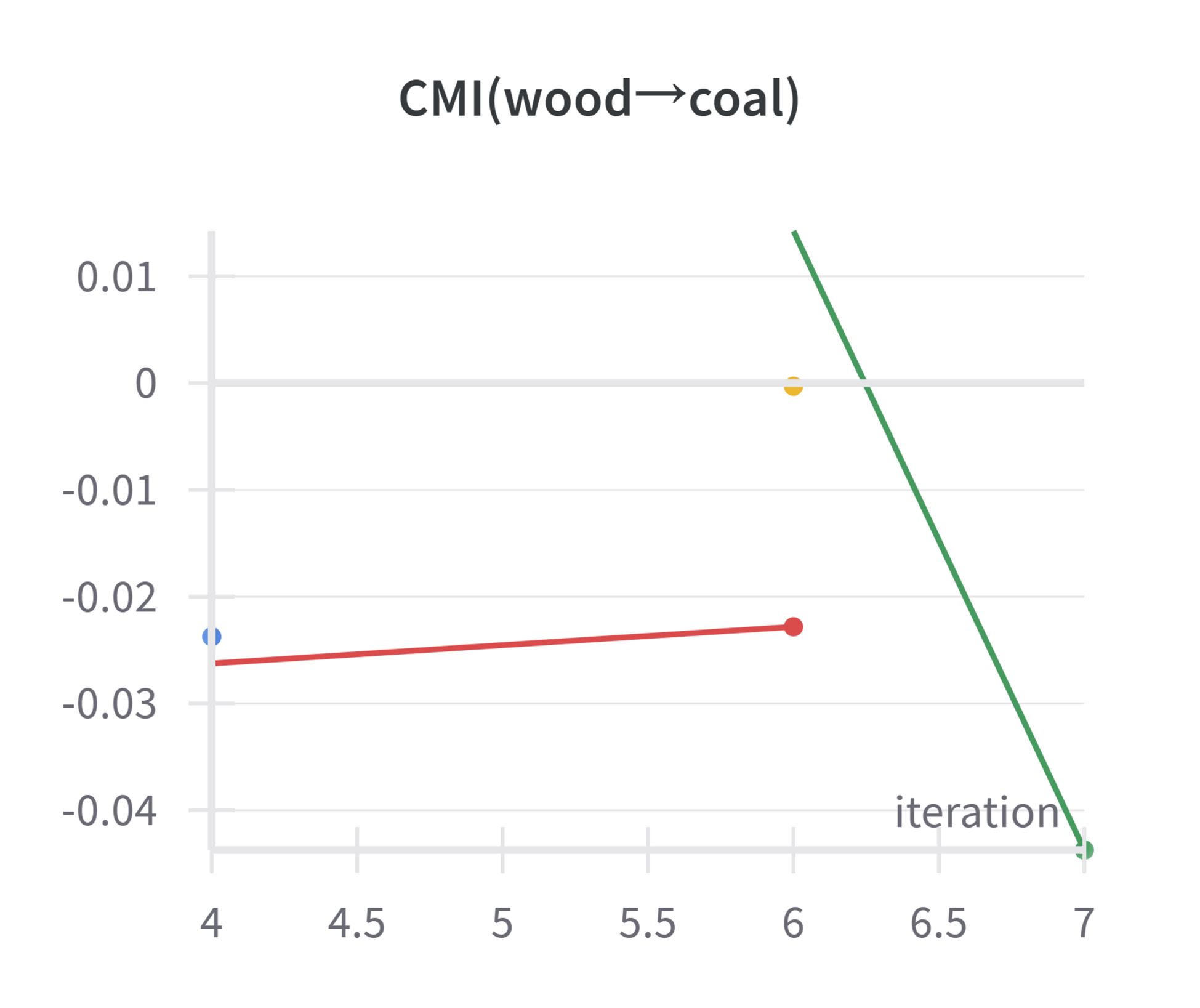} &
		\includegraphics[width=0.2\textwidth]{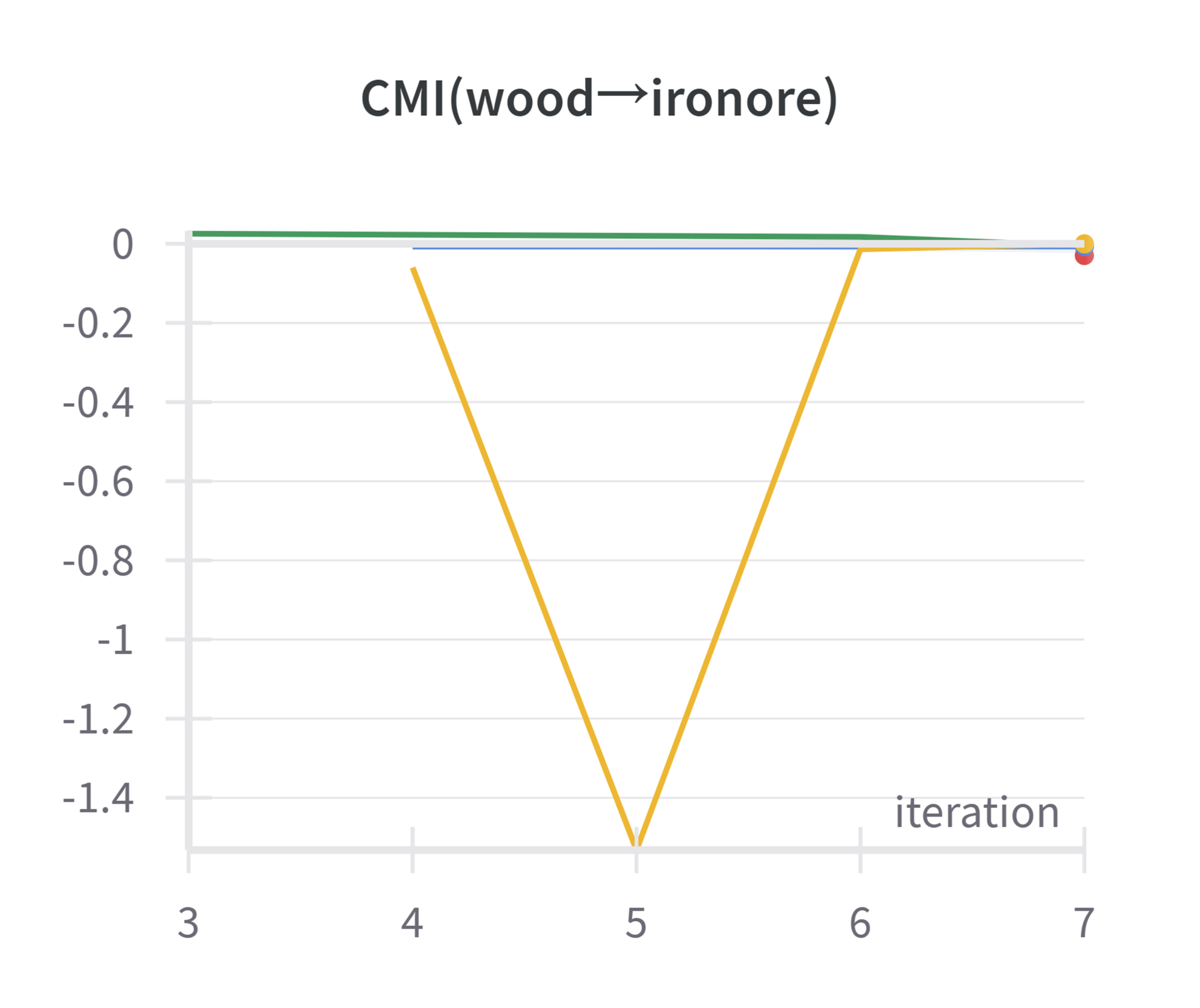} &
		\includegraphics[width=0.2\textwidth]{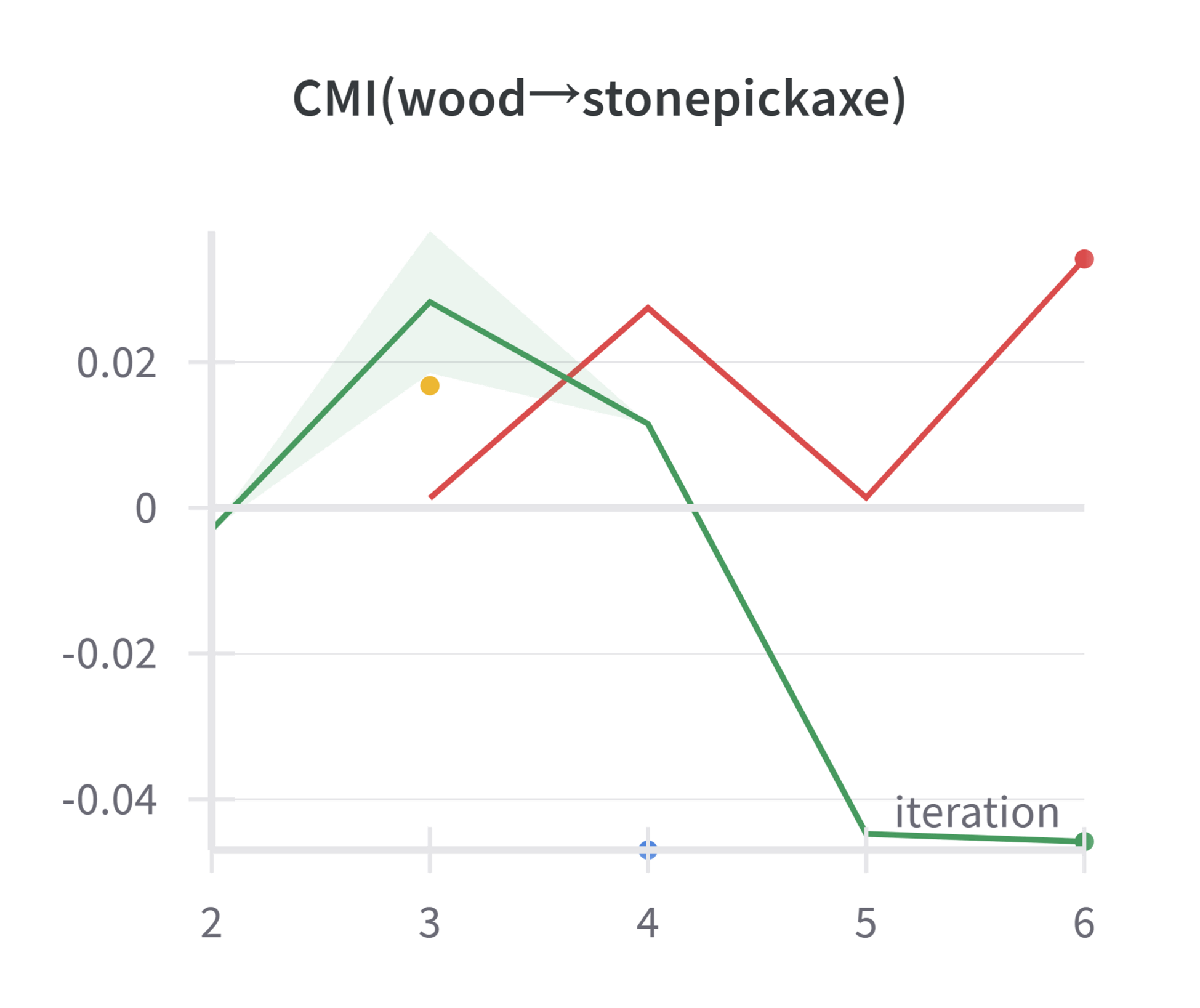} &
		\includegraphics[width=0.2\textwidth]{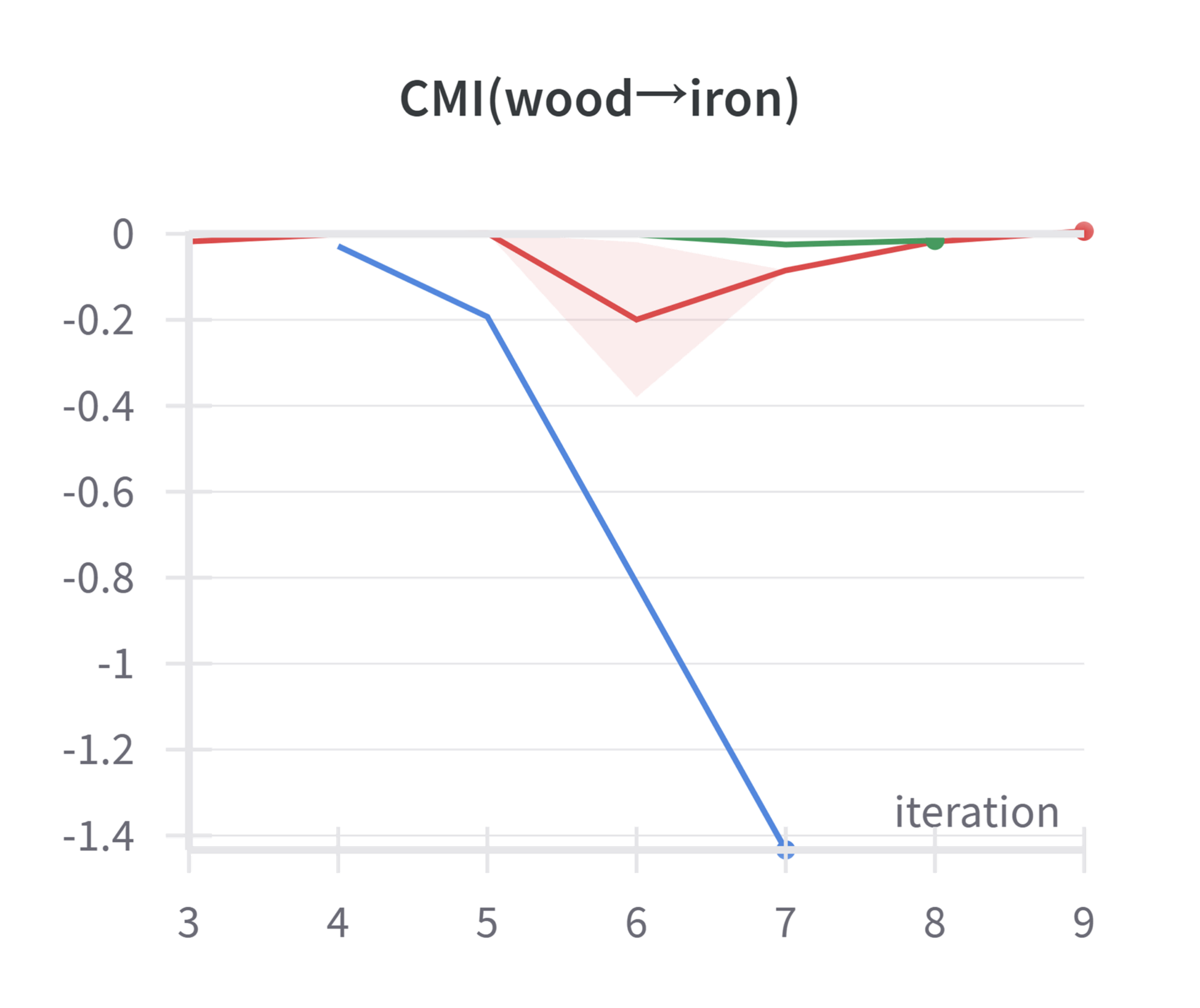} &
		\includegraphics[width=0.2\textwidth]{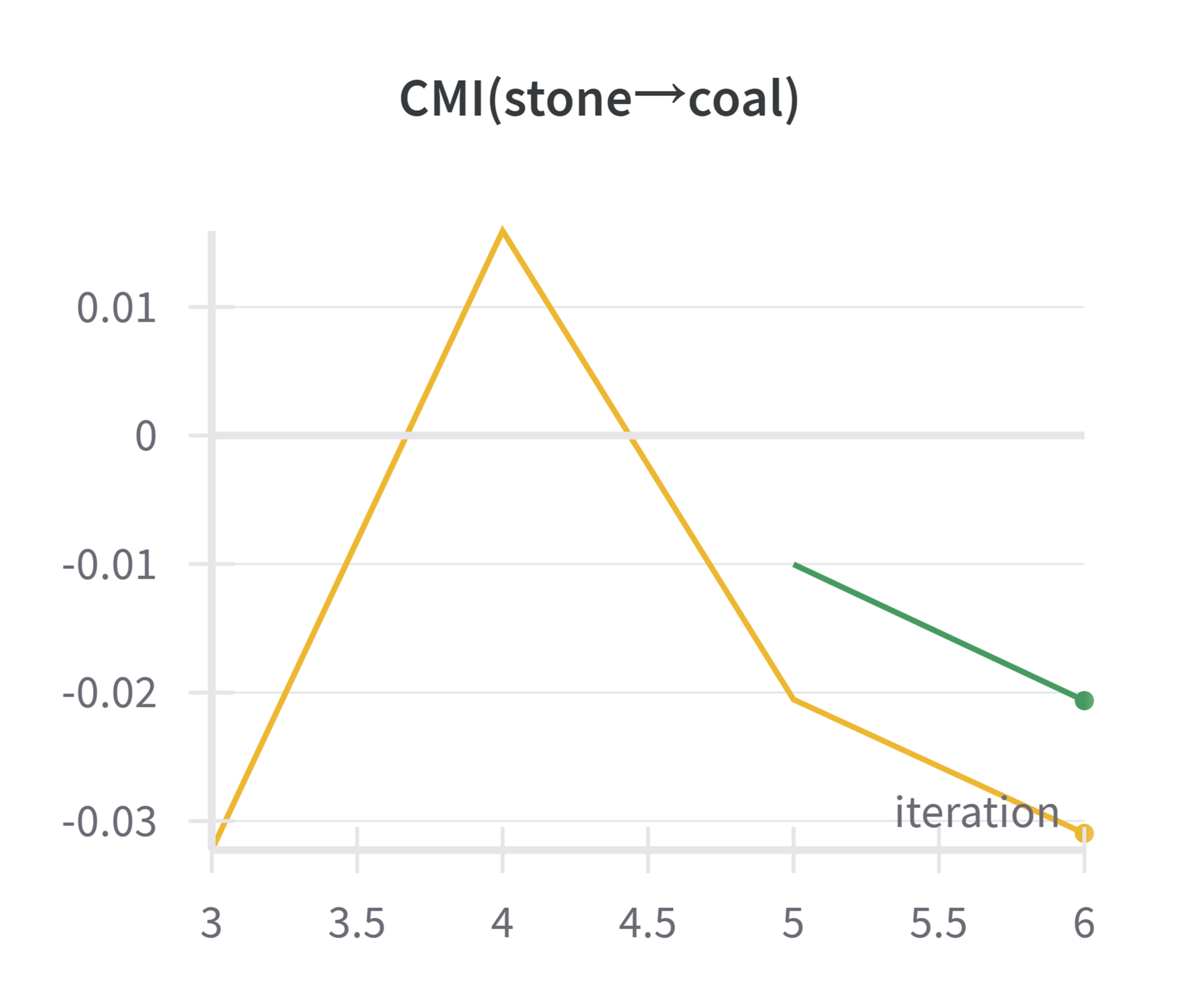} \\
		\includegraphics[width=0.2\textwidth]{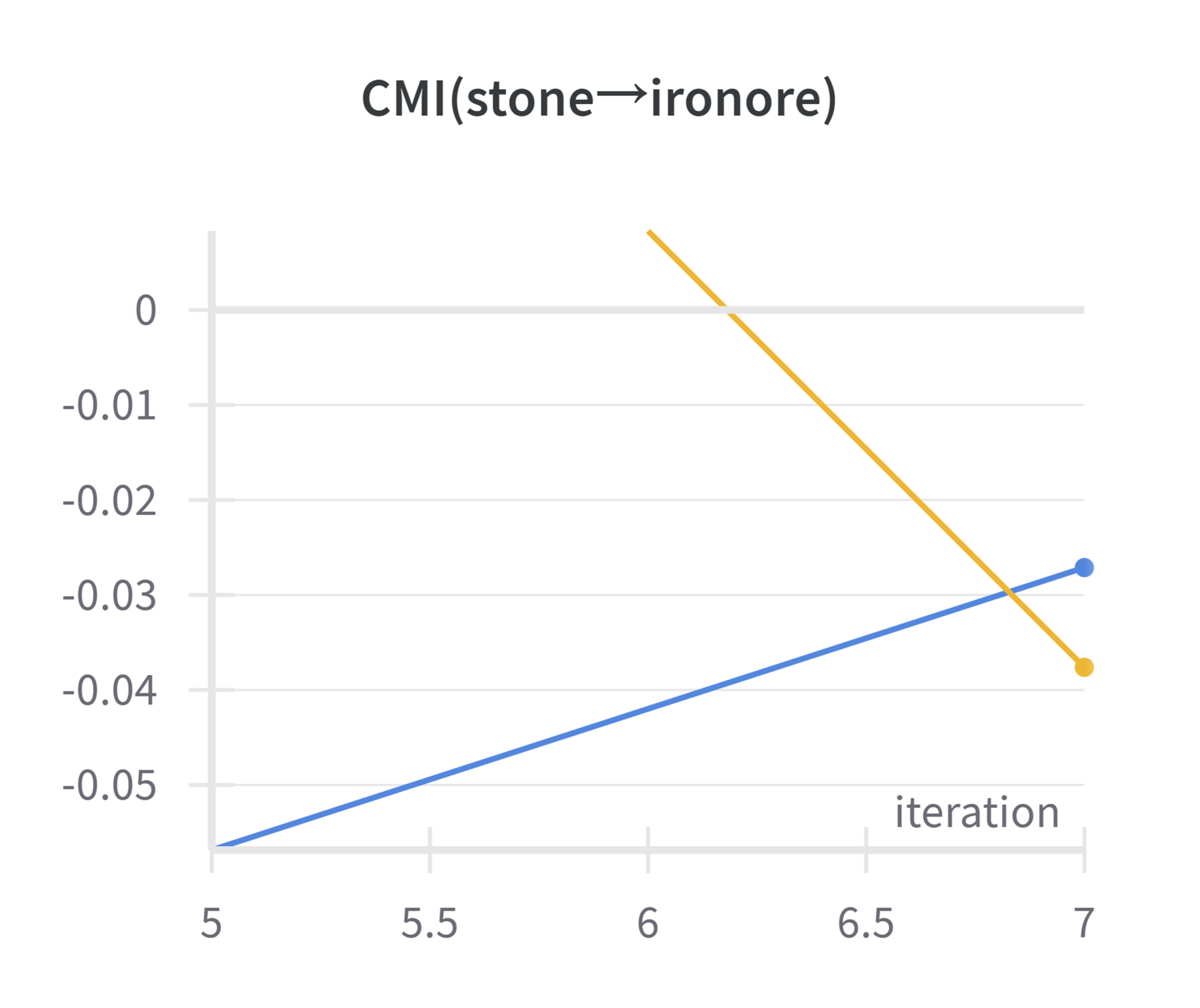} &
		\includegraphics[width=0.2\textwidth]{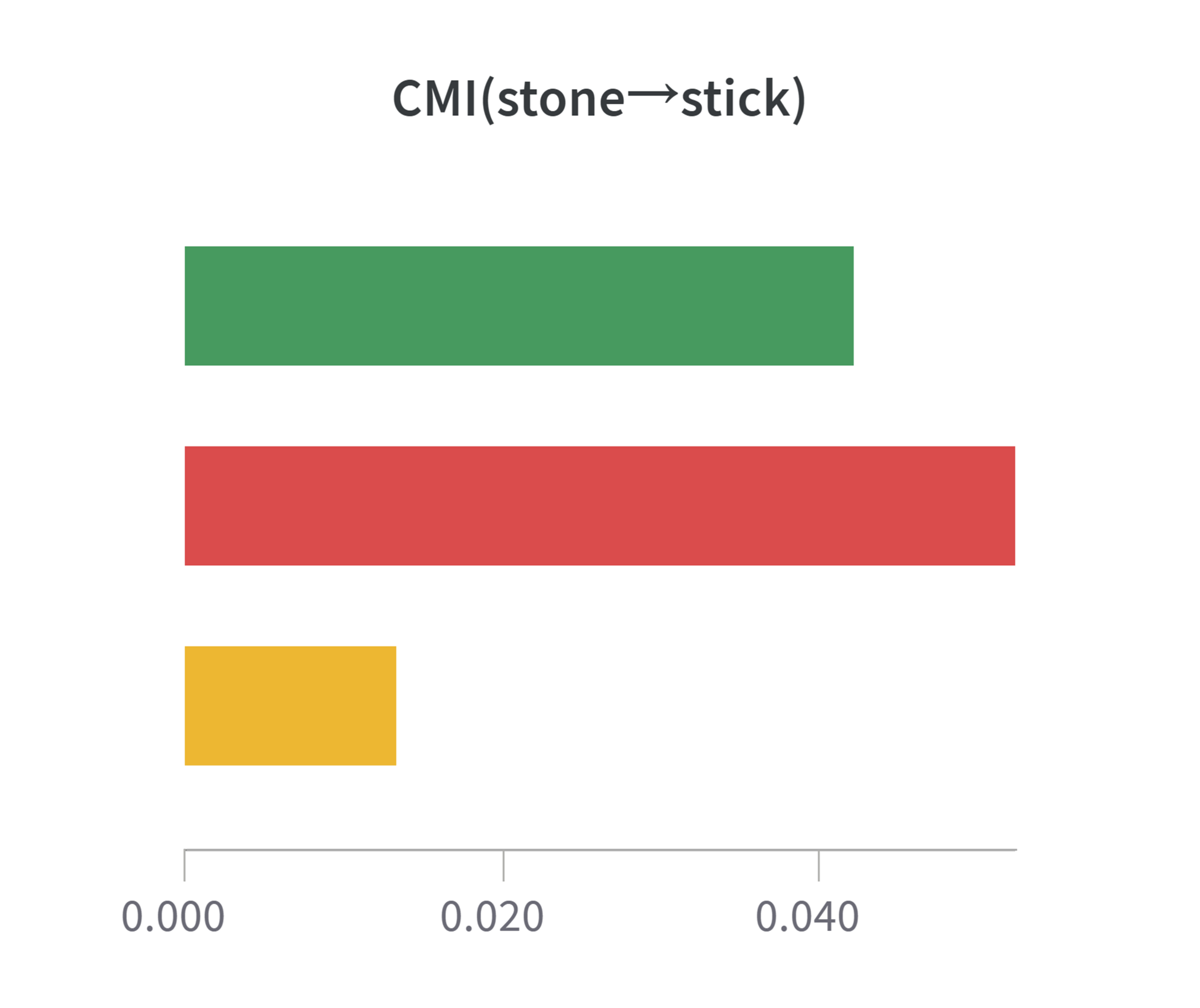} &
		\includegraphics[width=0.2\textwidth]{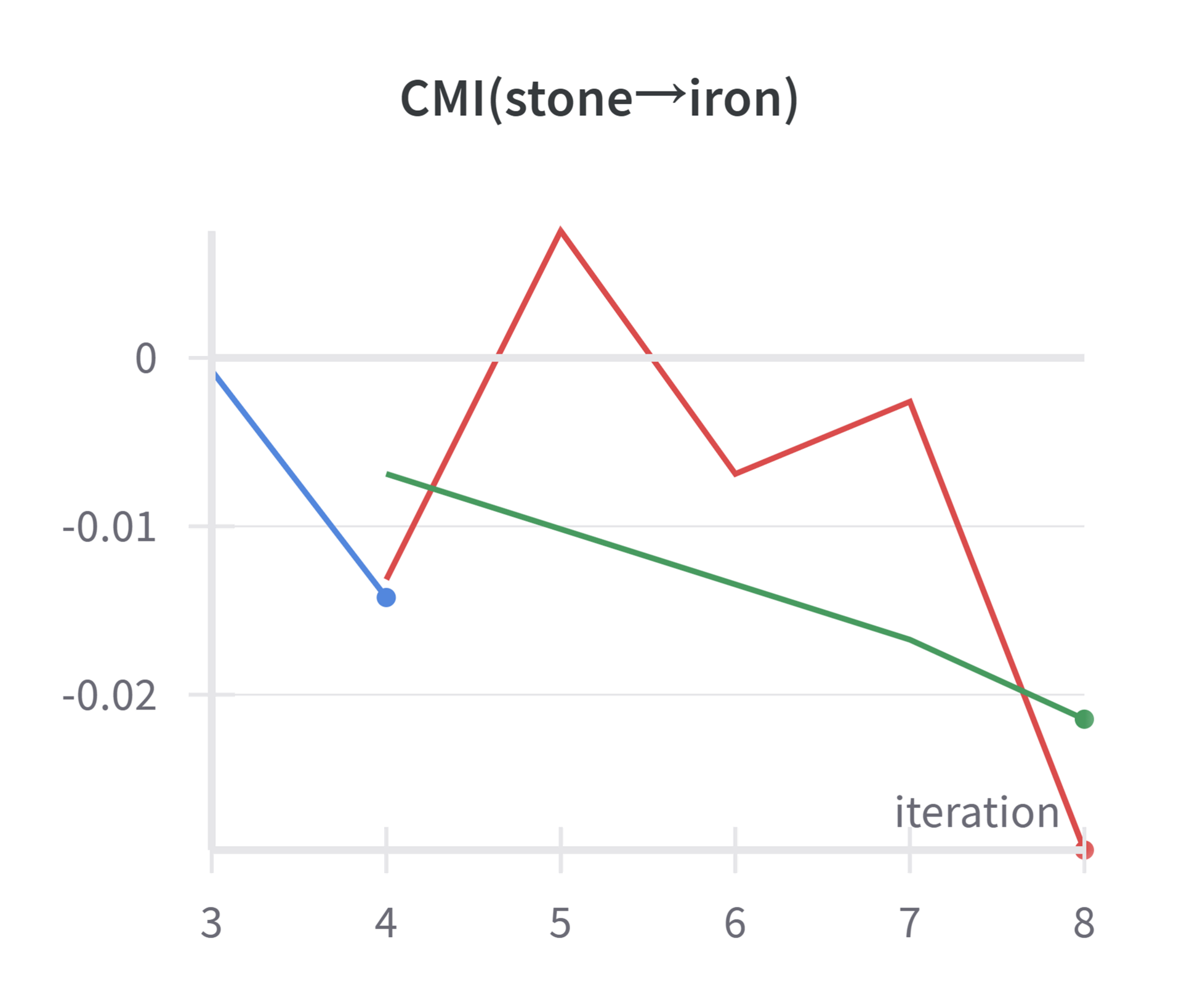} &
		\includegraphics[width=0.2\textwidth]{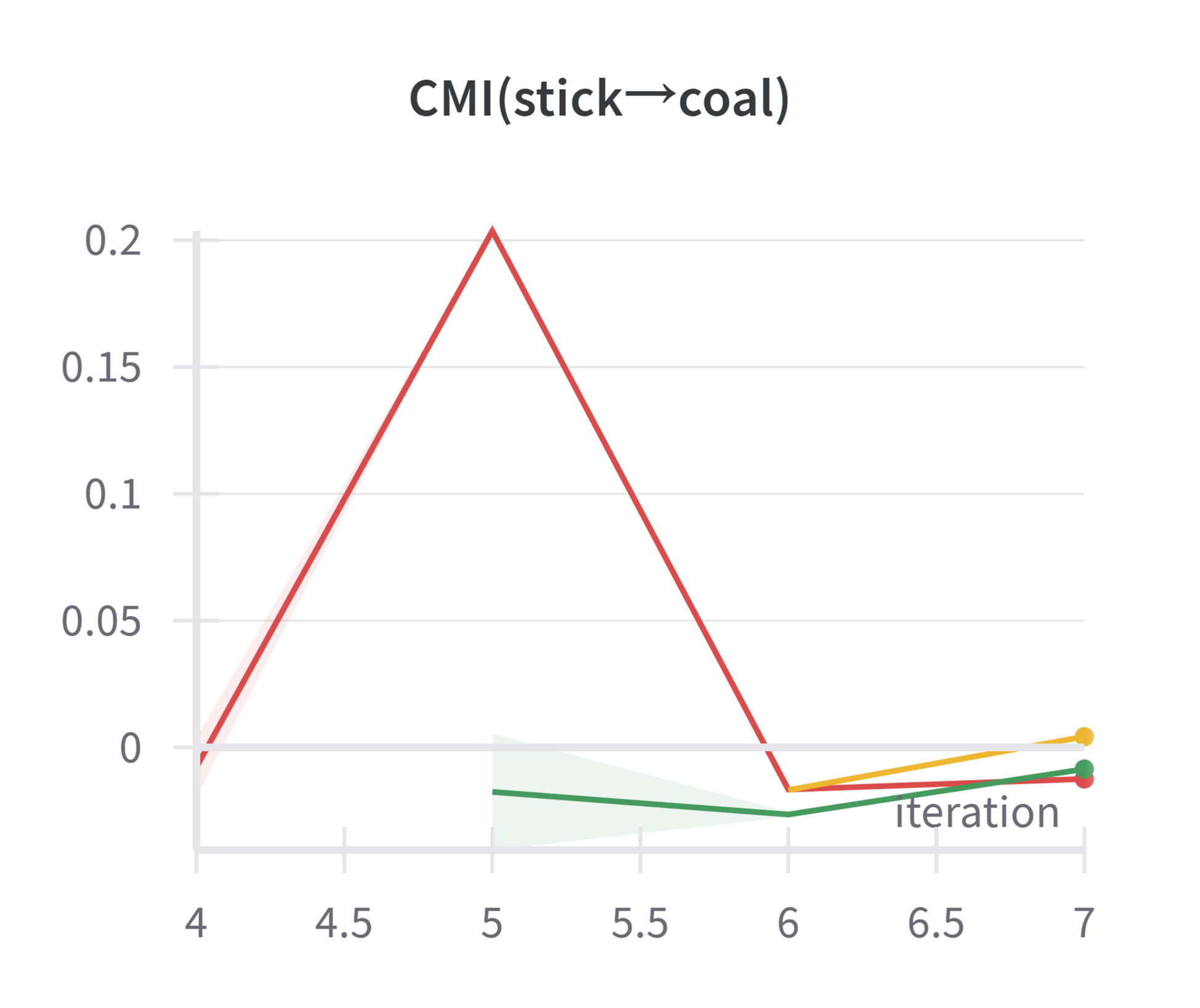} &
		\includegraphics[width=0.2\textwidth]{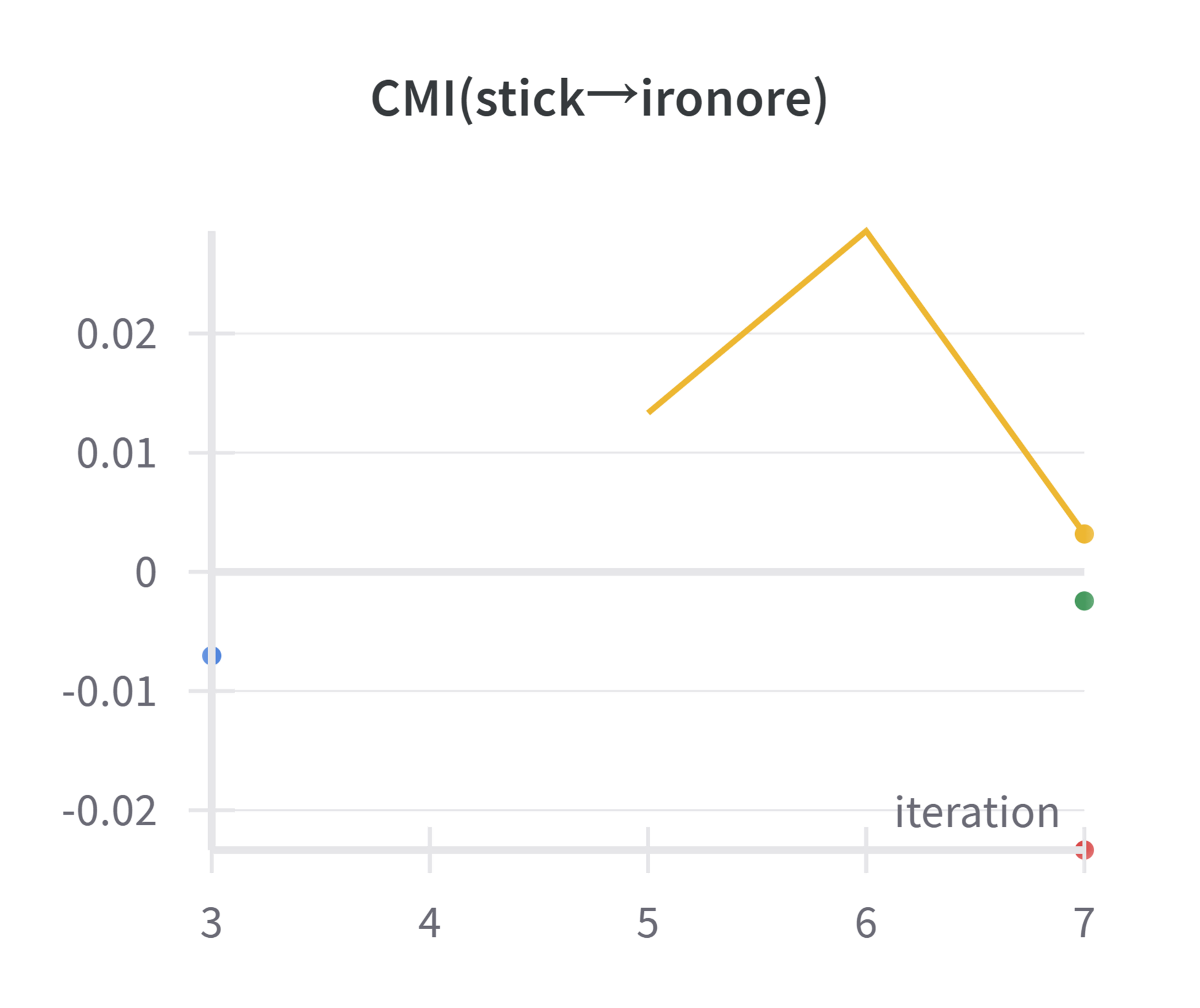} \\
		\includegraphics[width=0.2\textwidth]{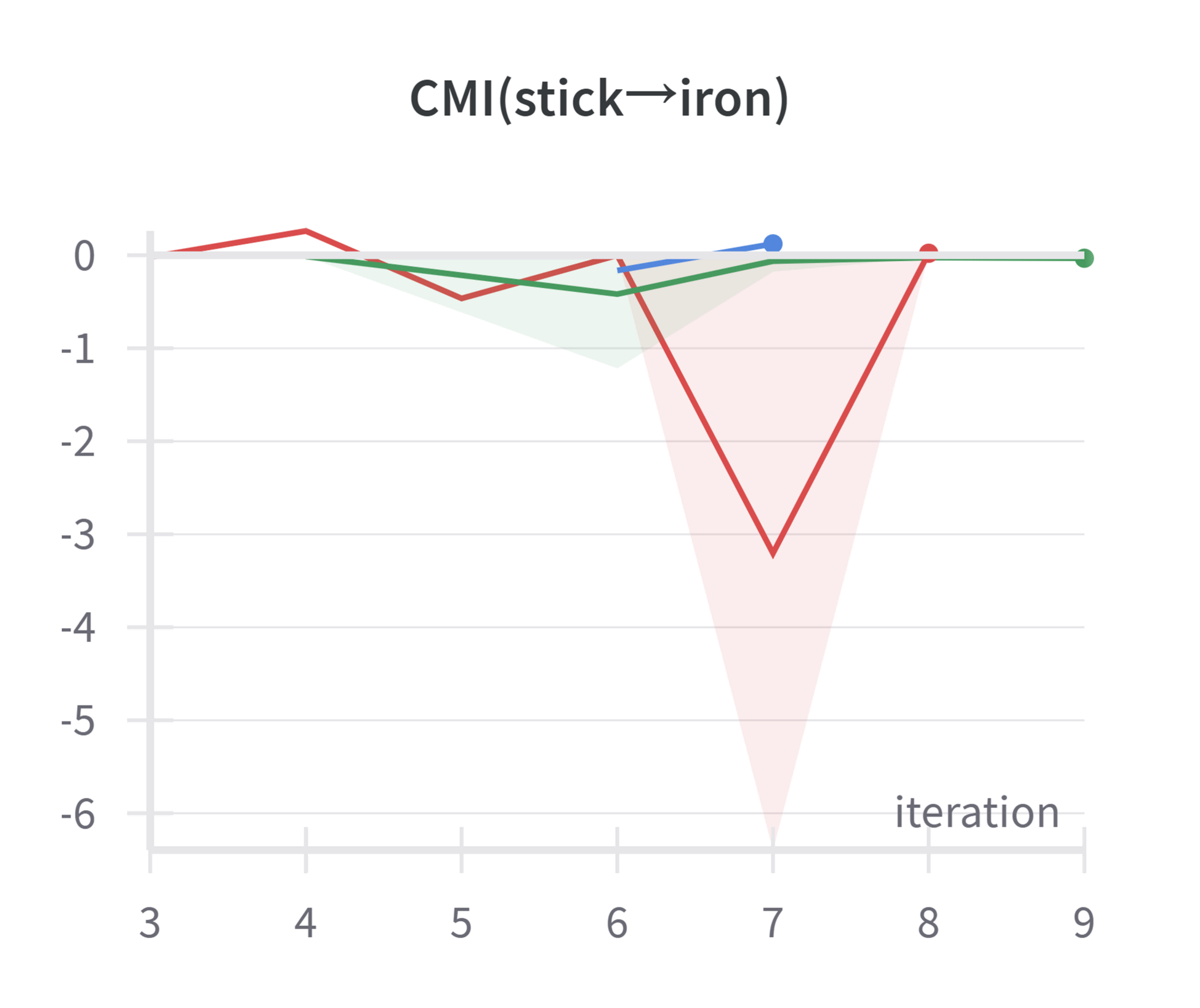} &
		\includegraphics[width=0.2\textwidth]{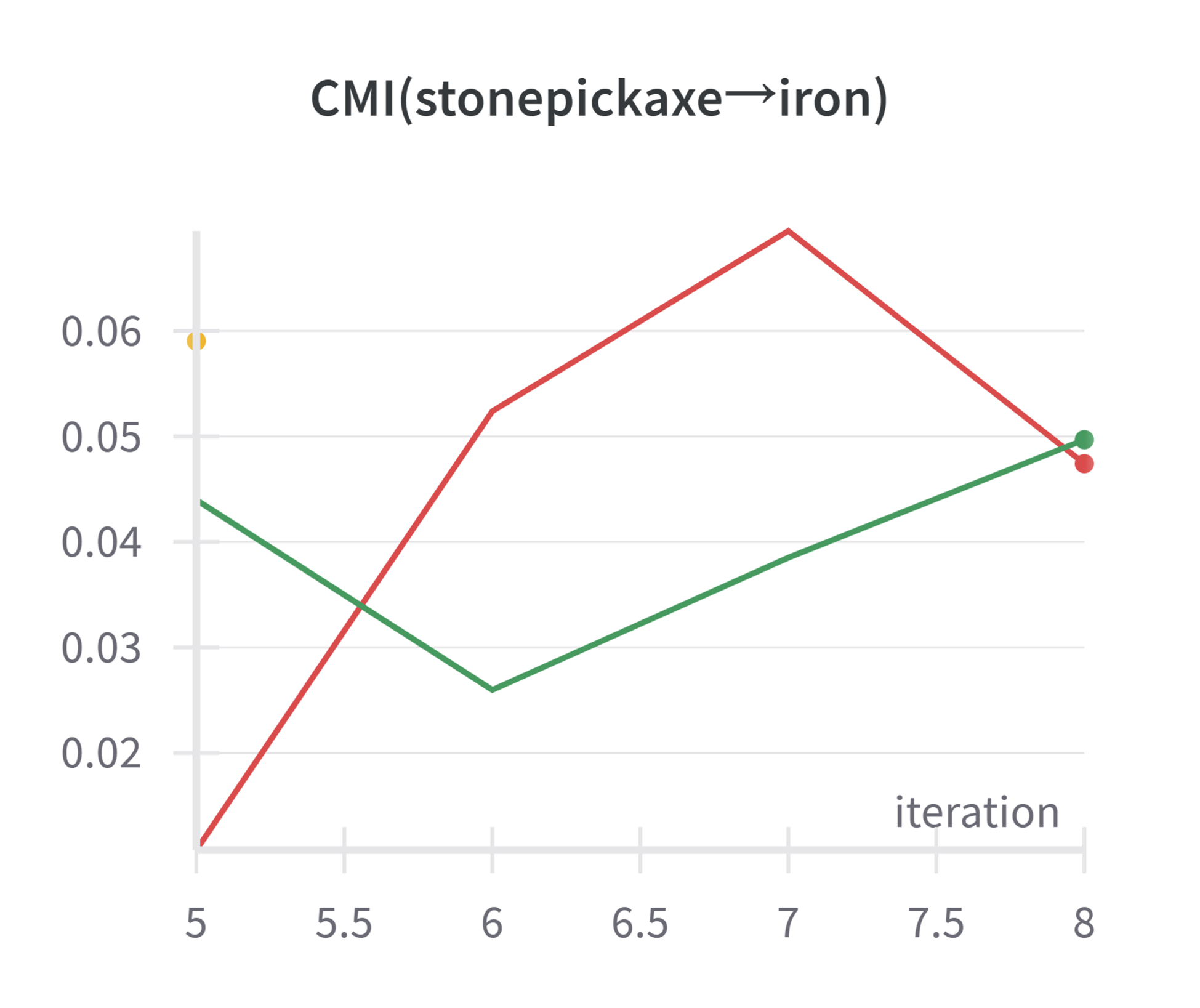} &
		\includegraphics[width=0.2\textwidth]{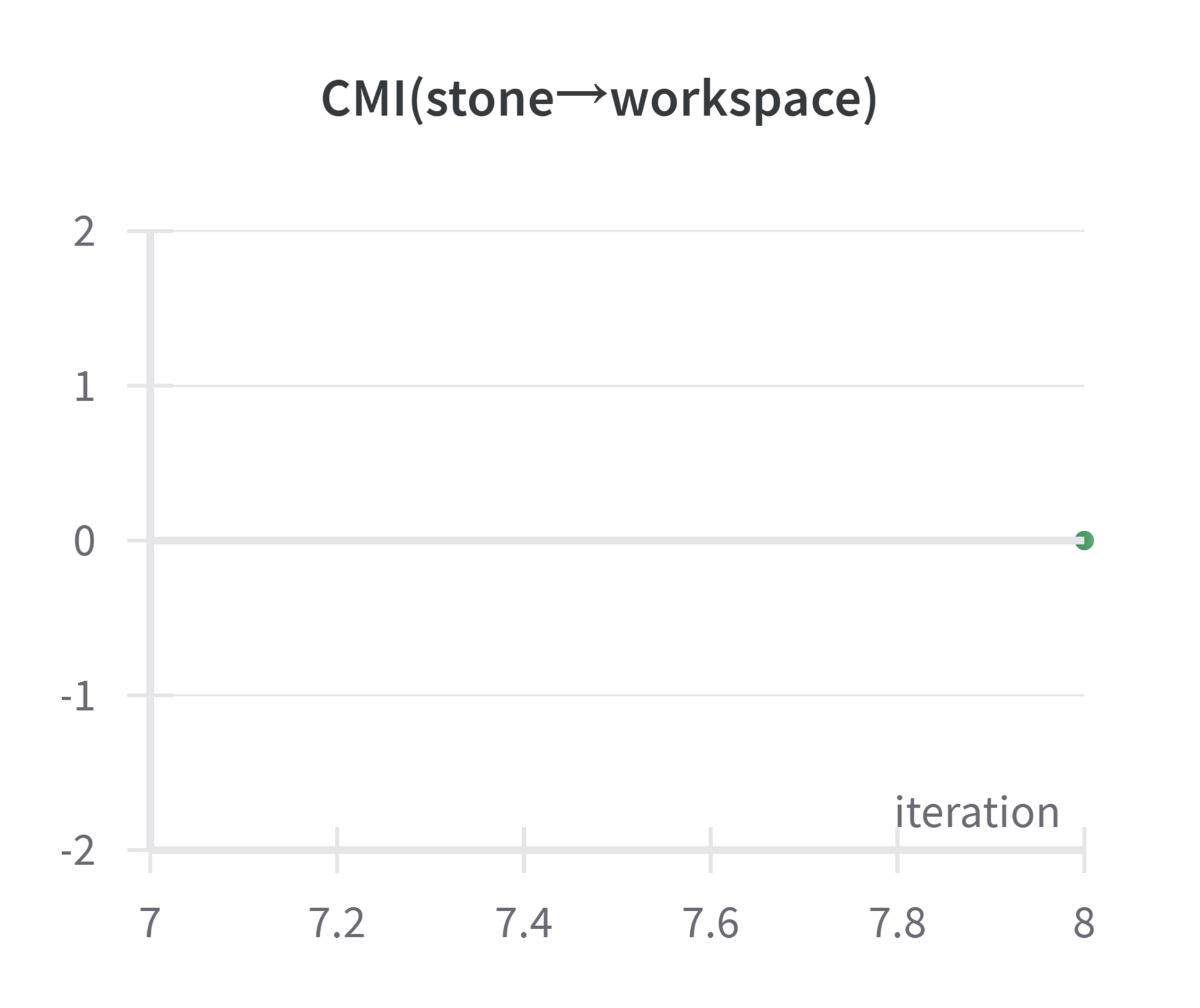} &
		\includegraphics[width=0.2\textwidth]{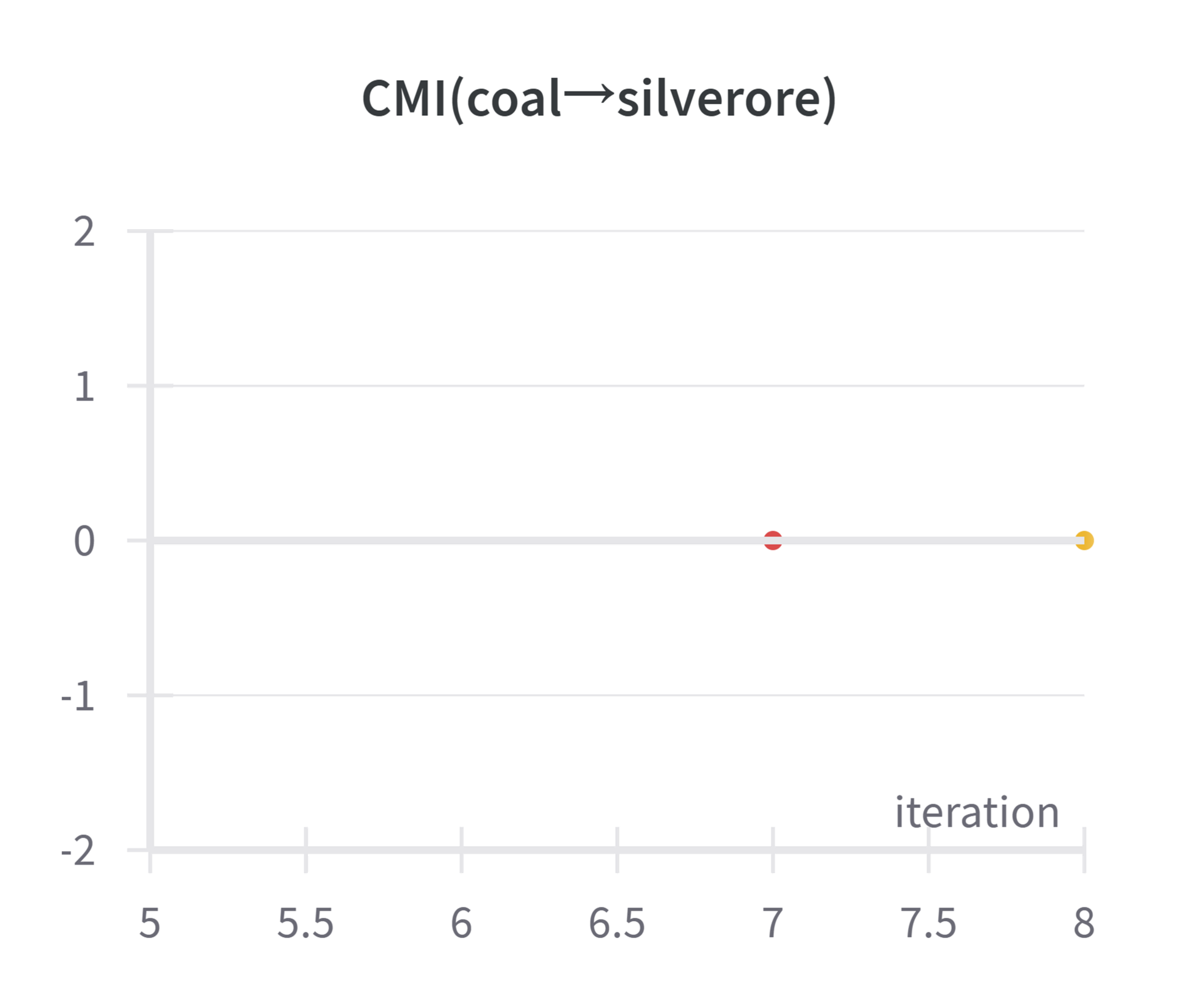}
	\end{tabular}
	\caption{The CMI of different time spans of \textbf{spurious correlations} in GetIron-R0-T1 task with \(\tau_{max}=4\).}
	\label{fig:fcmi_}
\end{figure*}

%\noindent\textbf{3. Does CMI effectively filter out spurious correlations?}
\begin{comment}
	\begin{figure*}[h]
		\centering
		\setlength{\tabcolsep}{1pt} % 设置列间距为 2pt
		\begin{tabular}{ccccccc}
			\includegraphics[width=0.14\textwidth]{figures/single-wood} &
			\includegraphics[width=0.14\textwidth]{figures/single-stone} &
			\includegraphics[width=0.14\textwidth]{figures/single-stick} &
			\includegraphics[width=0.14\textwidth]{figures/single-stonepickaxe} &
			\includegraphics[width=0.14\textwidth]{figures/single-coal} &
			\includegraphics[width=0.14\textwidth]{figures/single-ironore} &
			\includegraphics[width=0.14\textwidth]{figures/single-final4} \\
			\includegraphics[width=0.14\textwidth]{figures/multi-wood} &
			\includegraphics[width=0.14\textwidth]{figures/multi-stone} &
			\includegraphics[width=0.14\textwidth]{figures/multi-stick} &
			\includegraphics[width=0.14\textwidth]{figures/multi-stonepickaxe} &
			\includegraphics[width=0.14\textwidth]{figures/multi-coal} &
			\includegraphics[width=0.14\textwidth]{figures/multi-ironore} &
			\includegraphics[width=0.14\textwidth]{figures/multi-final3} 
		\end{tabular}
		\caption{Comparison of sub-goal training efficiency in the GetIron task with \(\tau_{max}=1\) (Up) and \(\tau_{max}=4\) (Down).}% \textbf{\textcolor{green}{\(\tau_{max}=1\)}} \textcolor{purple}{GetIron} (Up) and \textbf{\textcolor{red}{\(\tau_{max}=4\)}} \textcolor{blue}{GetIron} (Down).
		\label{fig:subgoal_training_}
	\end{figure*}
\end{comment}
\paragraph{Experimental Design}
%We record the CMI values of correlations identified by D3HRL during the SCM training process in the GetIron task
We record the CMI values of partial correlations evaluated by the spurious correlation detection module in GetIron-R0-T1, as shown in Figures \ref{fig:cmi_} and \ref{fig:fcmi_}. Figure \ref{fig:cmi_} shows CMI values for true causal relationships. Figure \ref{fig:fcmi_} shows CMI values for spurious correlations, the first 12 figures are for indirect causal relationships, while the last 2 figures are for spurious correlations. The x-axis denotes the iteration, and the y-axis denotes the CMI value. With \(\tau_{max}=4\), each figure contains up to 4 curves, representing the CMI values of the correlations as indicated in the figure titles at various time spans \(h \in \{\textcolor{b}{1},\textcolor{y}{2},\textcolor{r}{3},\textcolor{g}{4}\}\). 
%The true time span of the causal relationship is also noted in the titles. 

\paragraph{Experimental Results}
It is observed that spurious correlations have CMI values of 0, and most indirect causal relationships have CMI values much less than 0.05. In contrast, most direct causal relationships at their true time spans have CMI values significantly greater than 0.05. Therefore, the CIT through CMI effectively distinguishes spurious correlations from causal relationships with \(\epsilon_{cmi}=0.05\).
For example, the second subplot in Fig 4 shows the CMI values for the relationship $ \textit{A} \rightarrow \textit{stone} $ across 4 various time spans. This relationship (with time spans \textcolor{r}{3} and \textcolor{g}{4}) was preliminarily identified as a true causality by SCM in the 1st and 2nd iterations and underwent spurious correlation detection (SCD). In the 2nd iteration, it passed SCD and was mastered by the hierarchical policy (thus no further SCD was operated), with the CMI value at time span \textcolor{r}{3} exceeding $\epsilon_{cmi}=0.05$, but not at \textcolor{g}{4}. Thus, the true time span was determined to be \textcolor{r}{3}, consistent with the truth. For the first and third subplots of Fig 4, these relationships passed SCD and were mastered in one iteration, thus only one iteration of CMI values is shown as bar charts.
It is important to note that the selection of \(\epsilon_{cmi}\) was determined through a separate statistical experiment conducted prior to our main experiments. Statistical analysis revealed that choosing \(\epsilon_{cmi}=0.05\) is suitable for our tasks.
%setting a CMI threshold of 0.05 effectively distinguishes spurious correlations from true causal relationships. 
%, with their true time spans noted in the titles
%particularly 

\subsubsection{How does D3HRL compare to CDHRL in terms of the quality of the hierarchical framework?}

\begin{figure*}[htbp]
	\centering
	\setlength{\tabcolsep}{1pt} % 设置列间距为 2pt
	\begin{tabular}{cccccc}
		\includegraphics[width=0.16\textwidth]{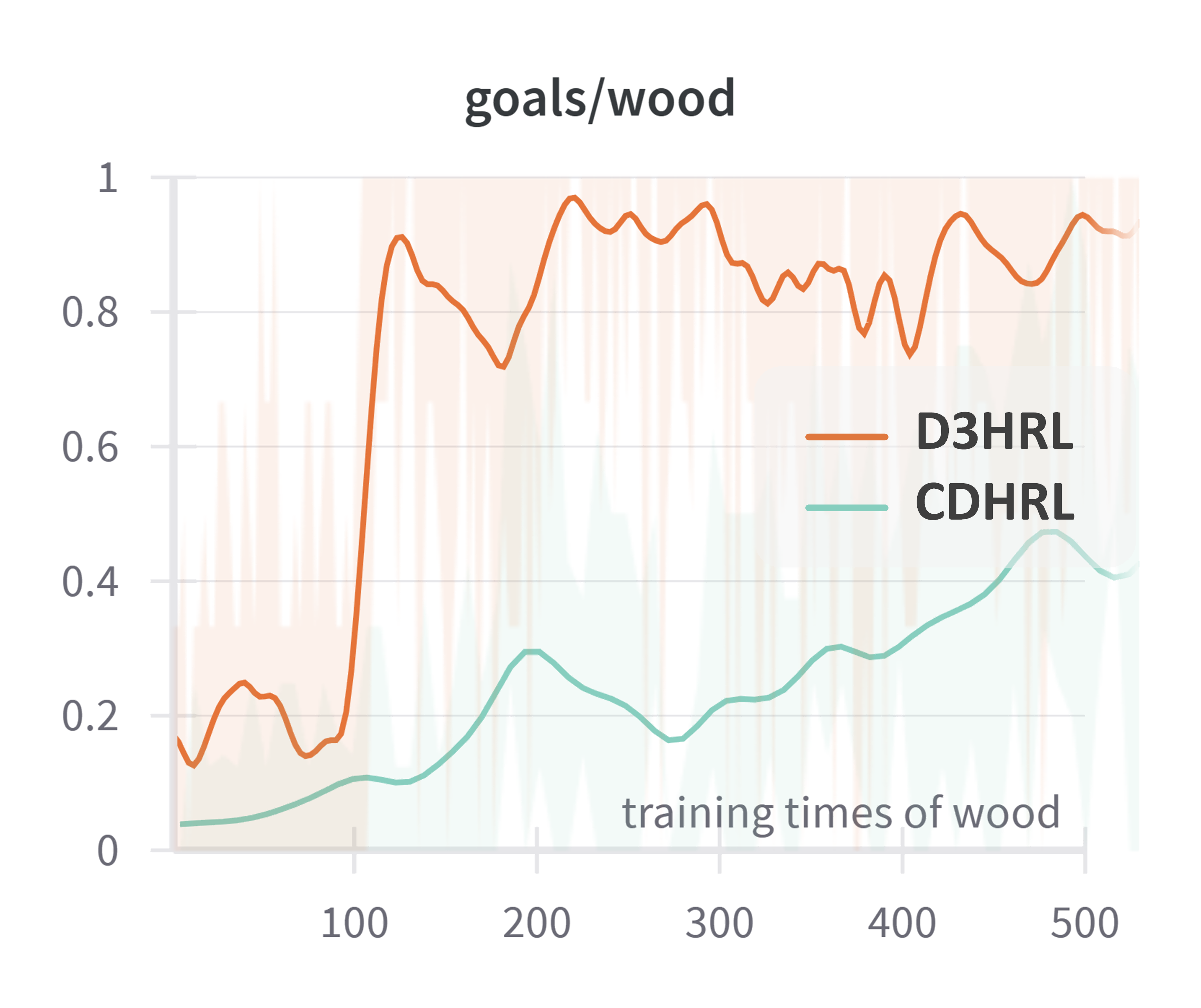} &
		\includegraphics[width=0.16\textwidth]{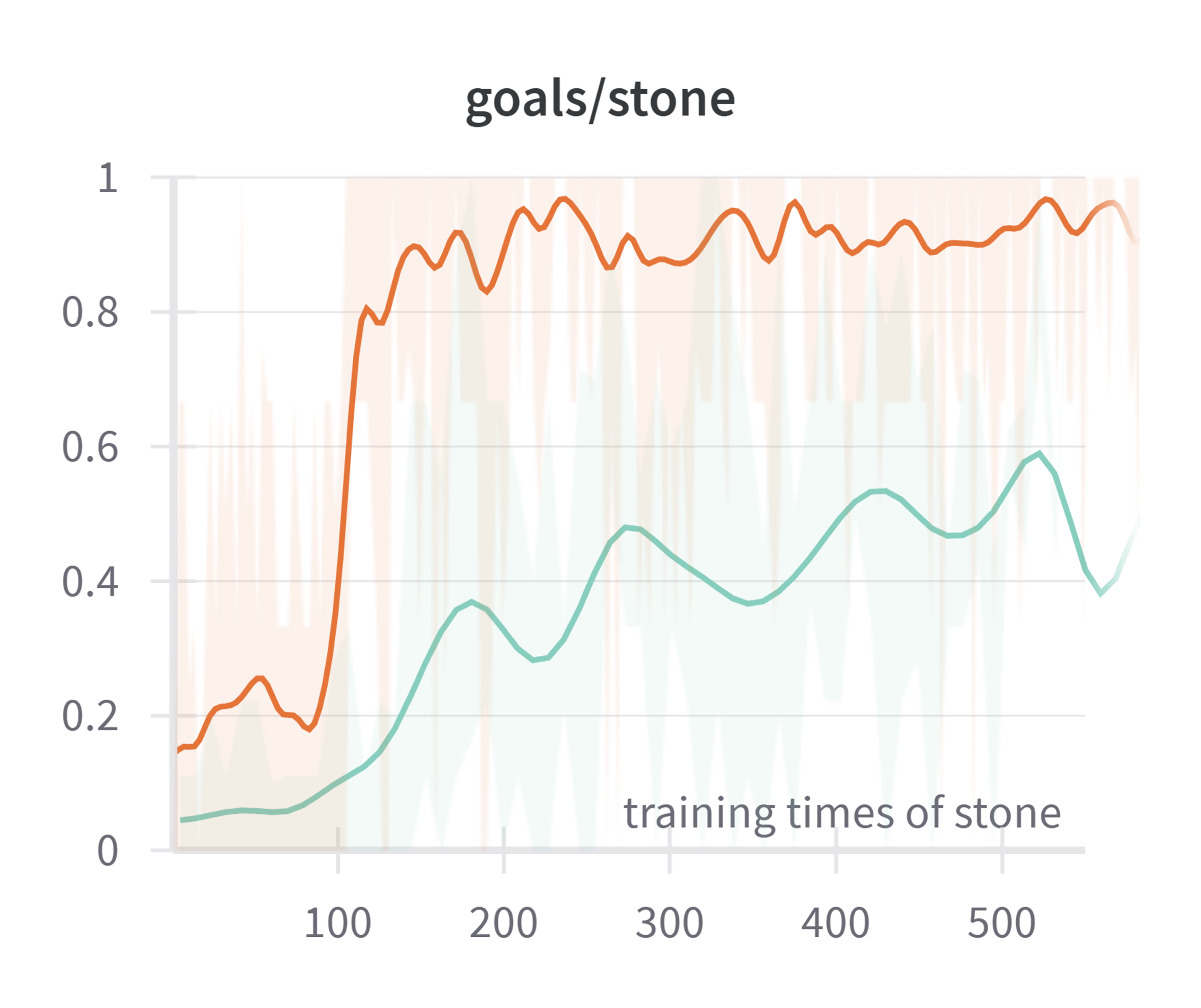} &
		\includegraphics[width=0.16\textwidth]{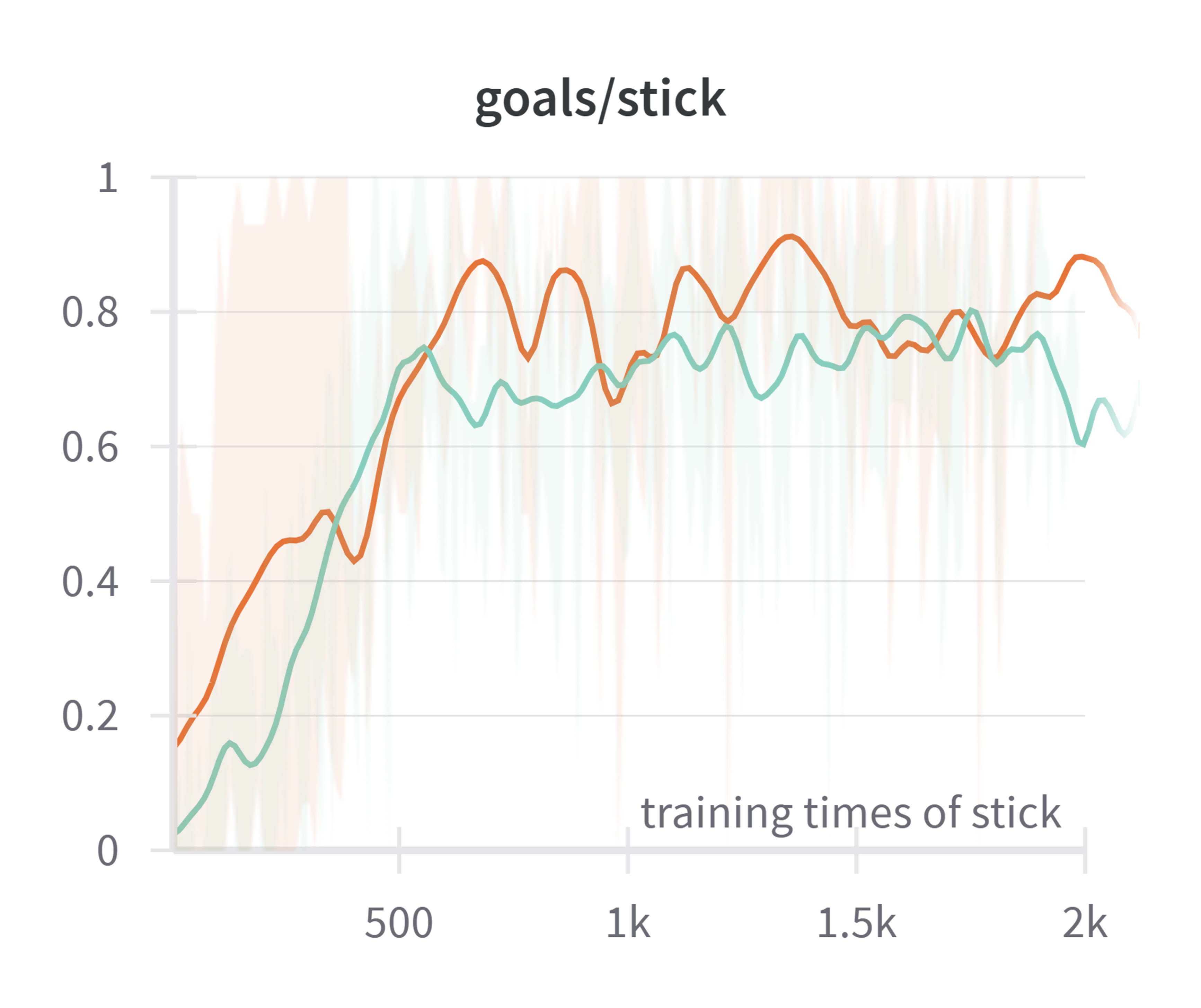} &
		\includegraphics[width=0.16\textwidth]{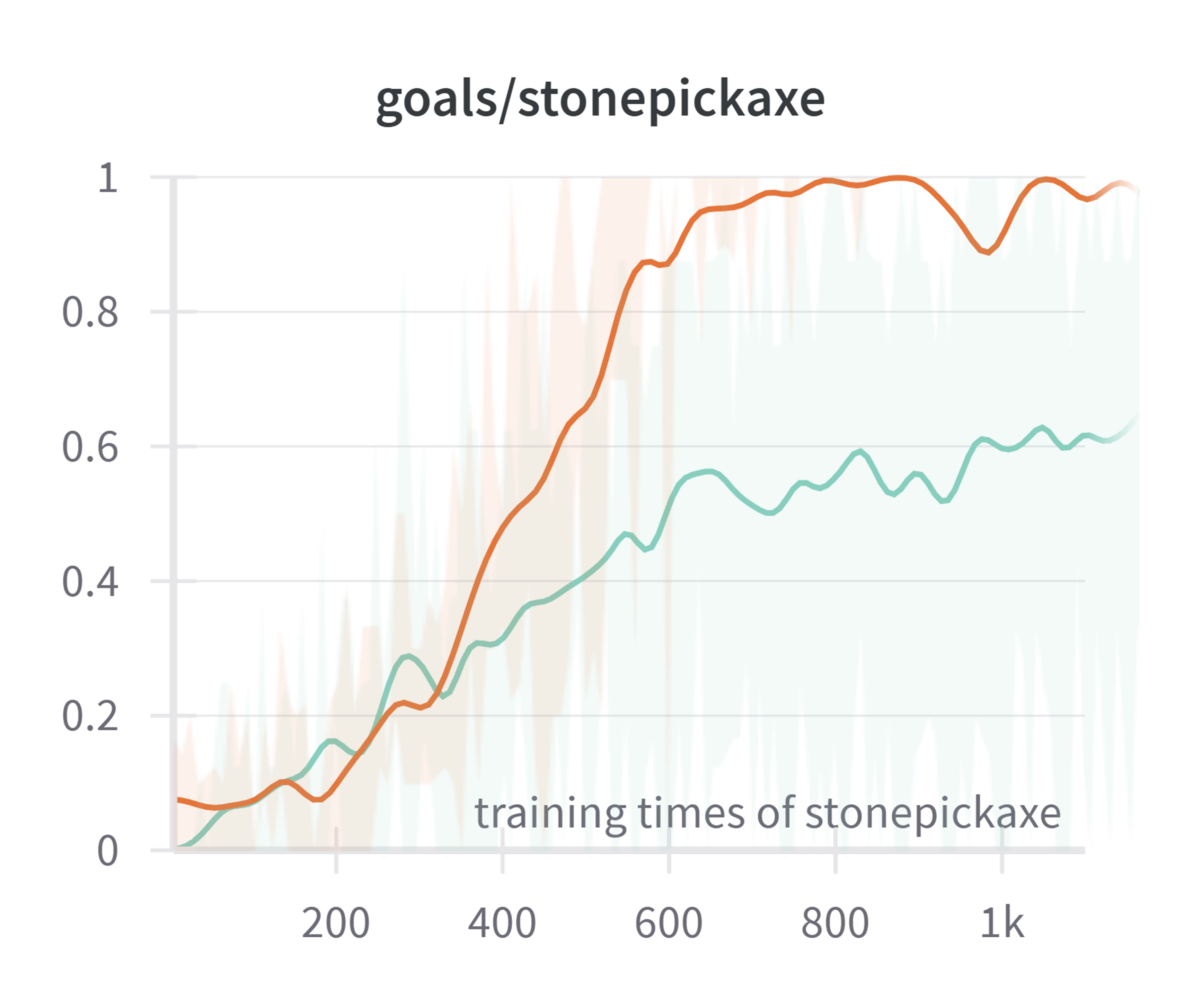} &
		\includegraphics[width=0.16\textwidth]{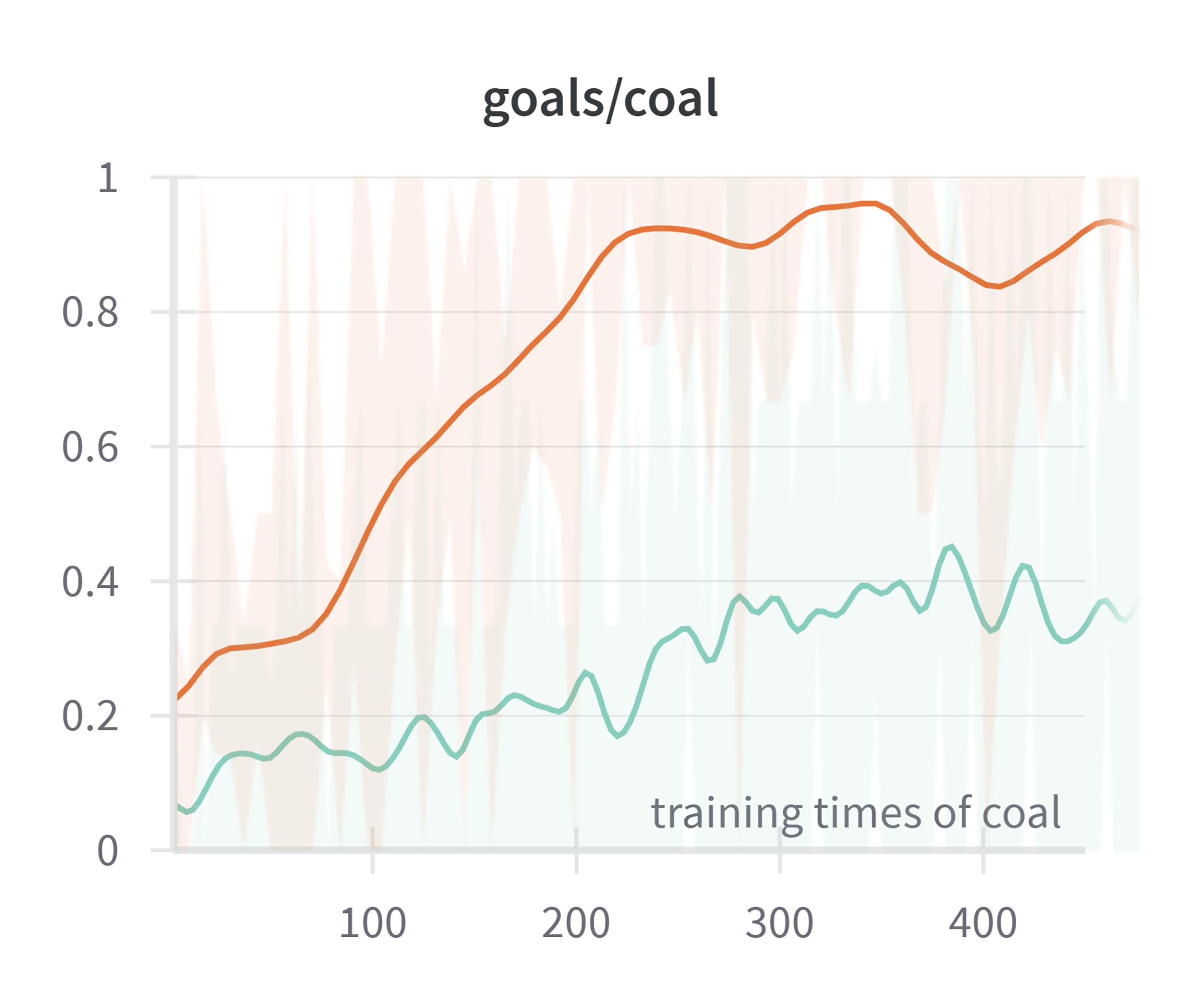} &
		\includegraphics[width=0.16\textwidth]{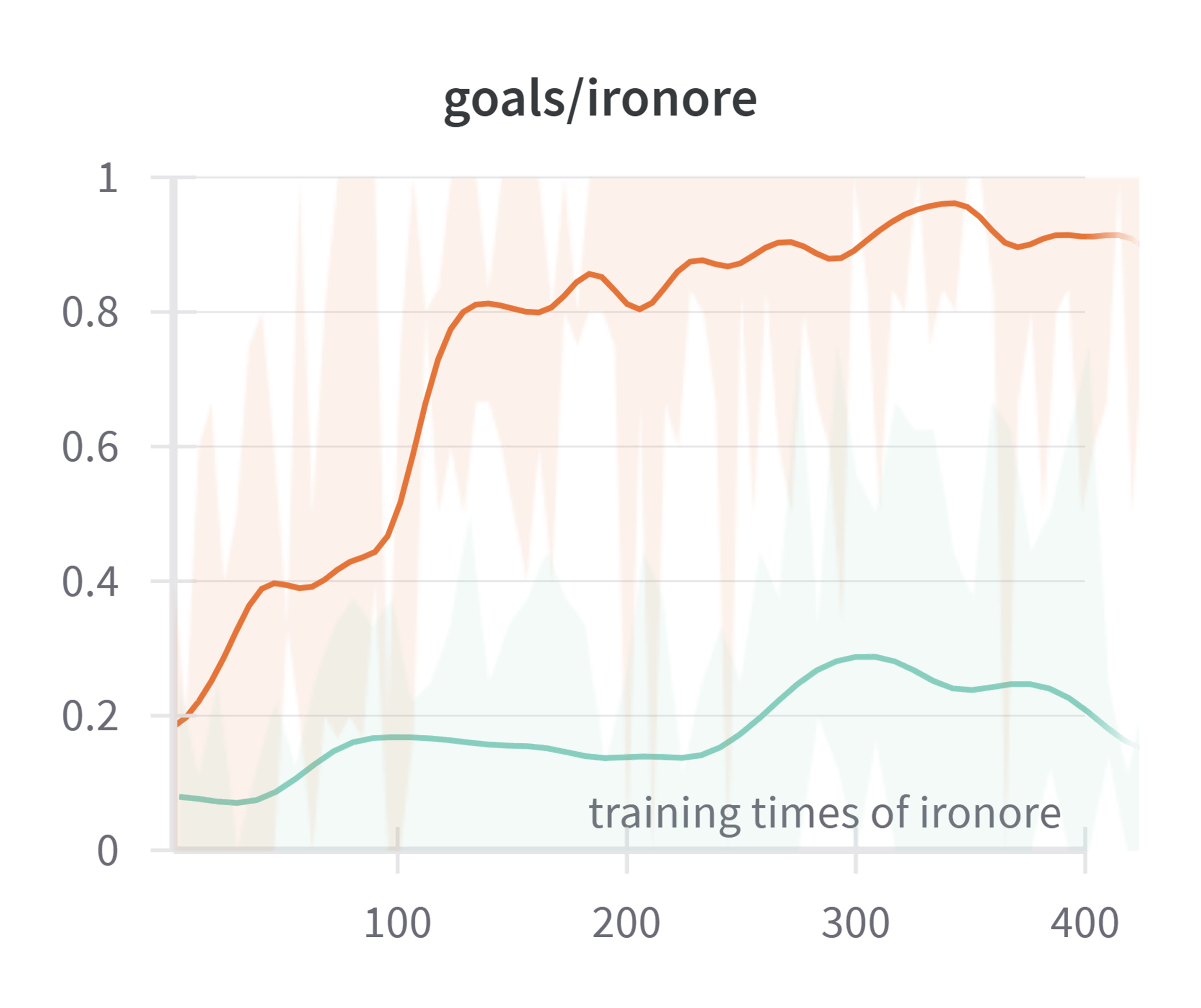} \\
		\includegraphics[width=0.16\textwidth]{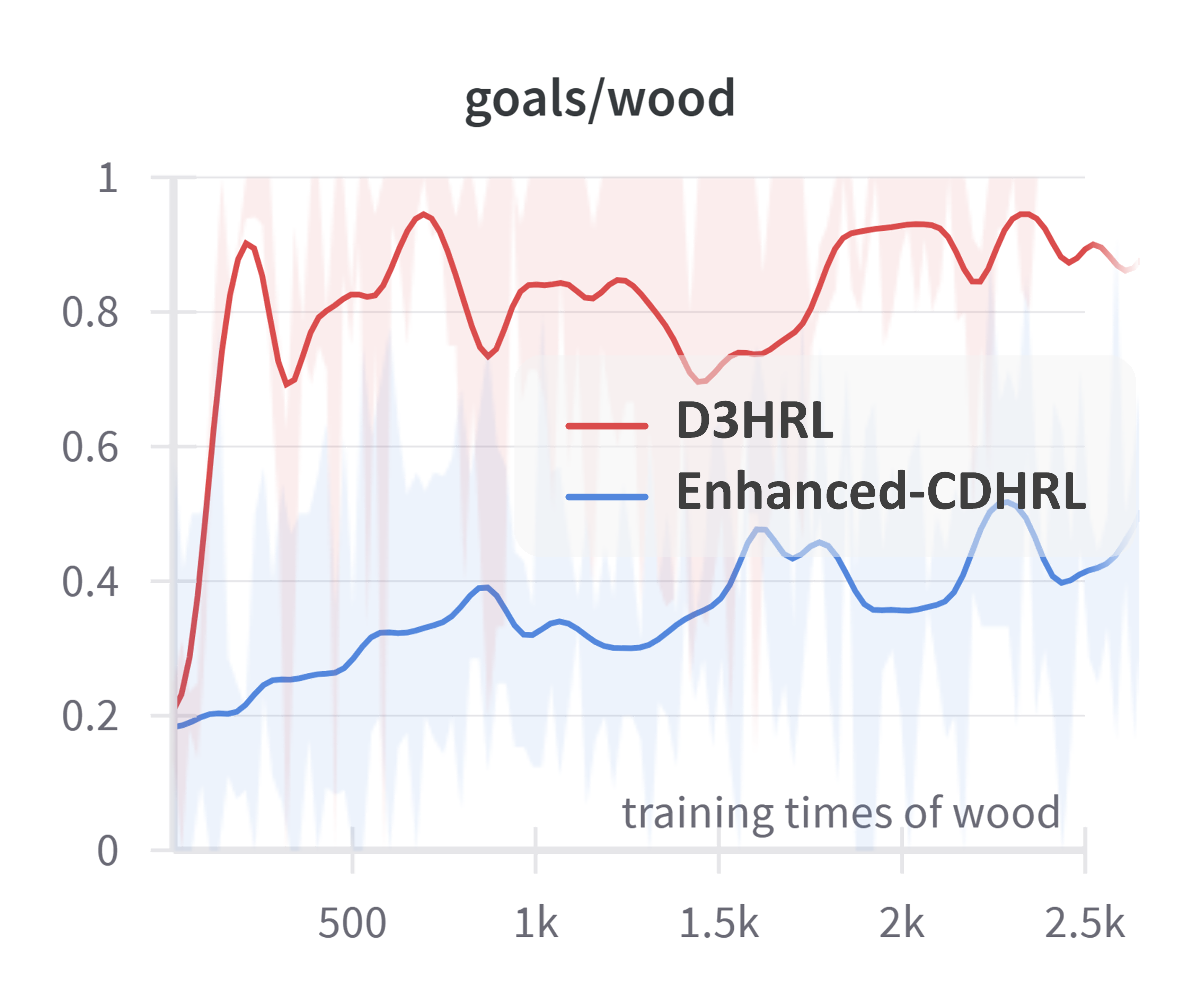} &
		\includegraphics[width=0.16\textwidth]{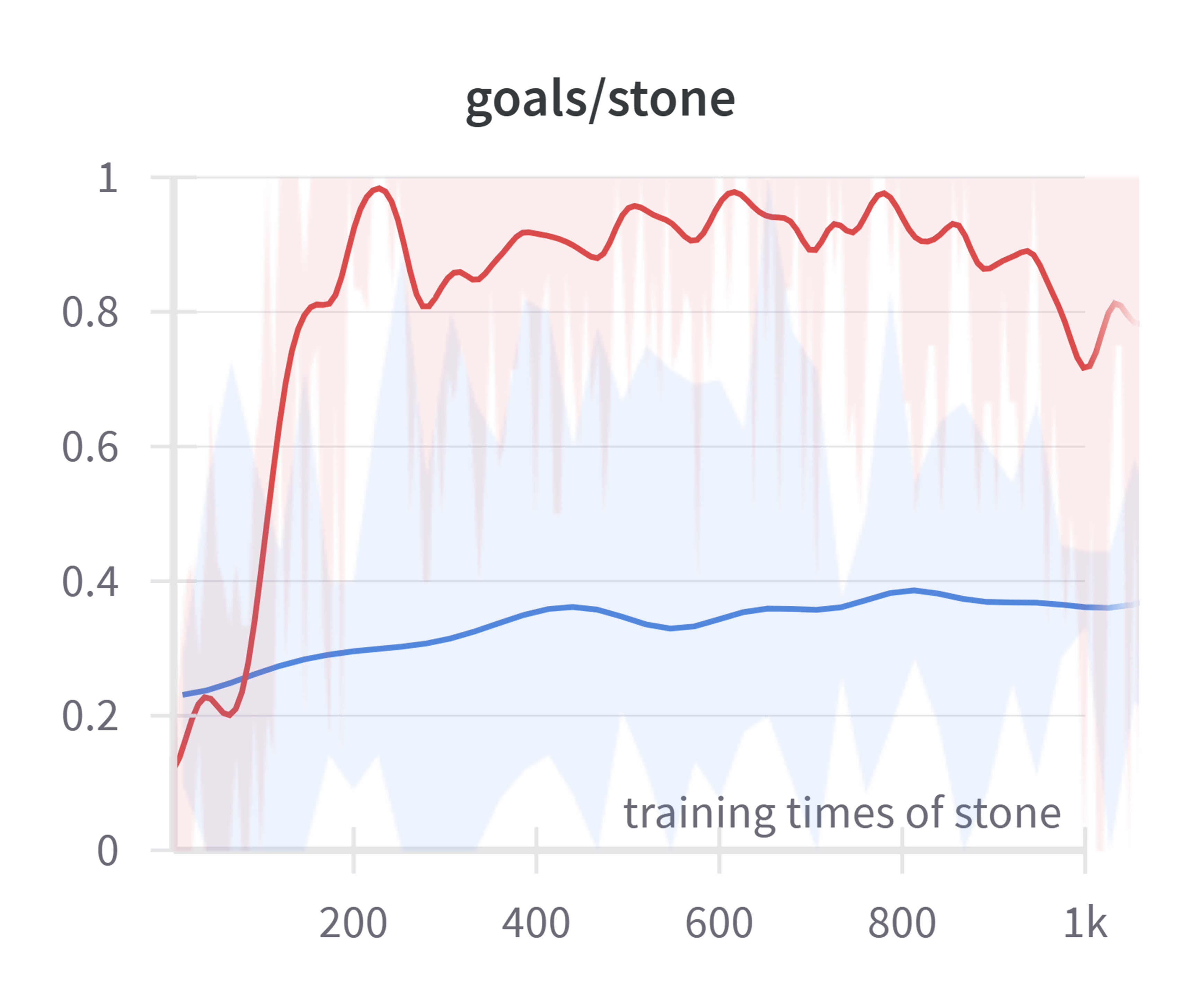} &
		\includegraphics[width=0.16\textwidth]{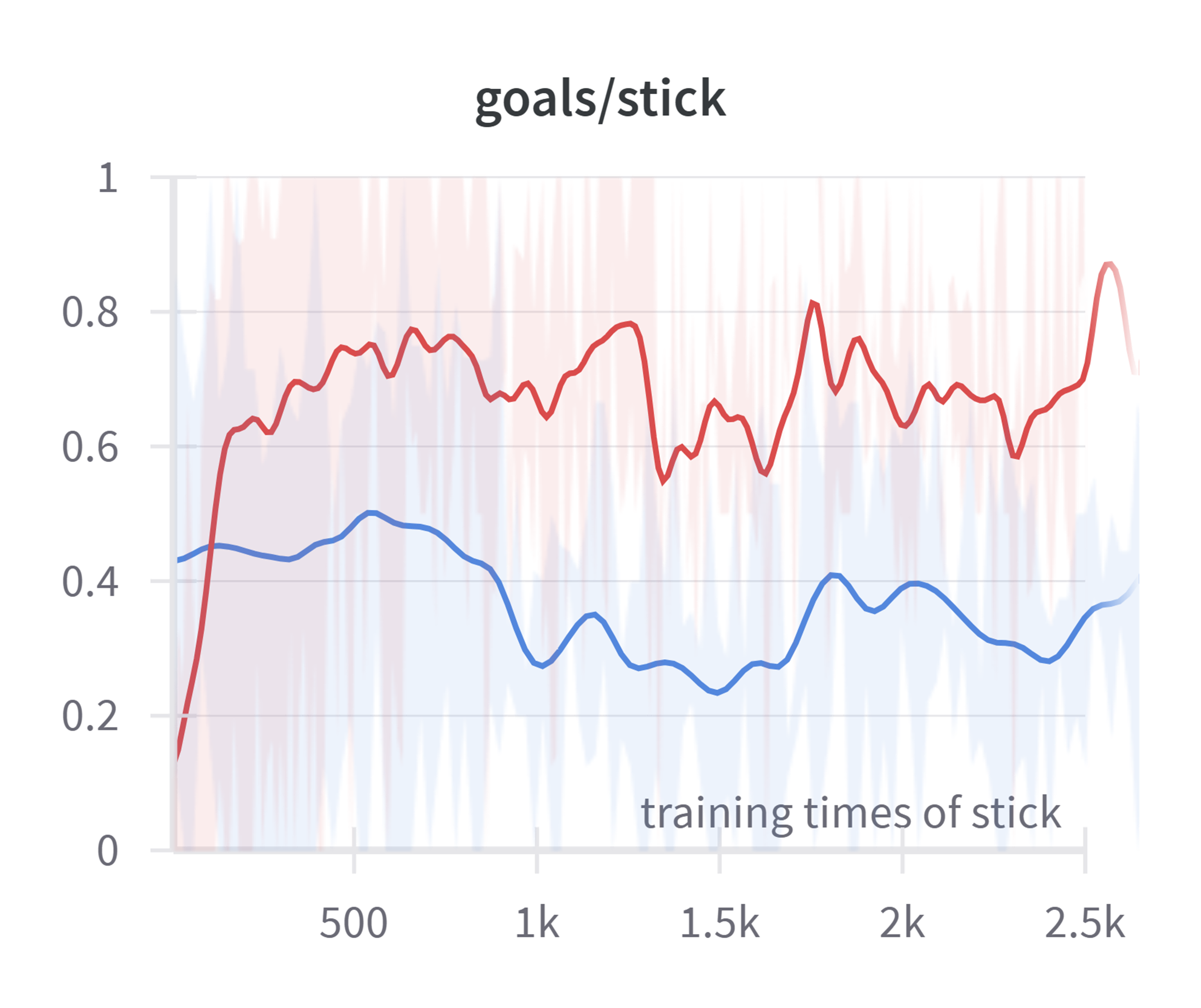} &
		\includegraphics[width=0.16\textwidth]{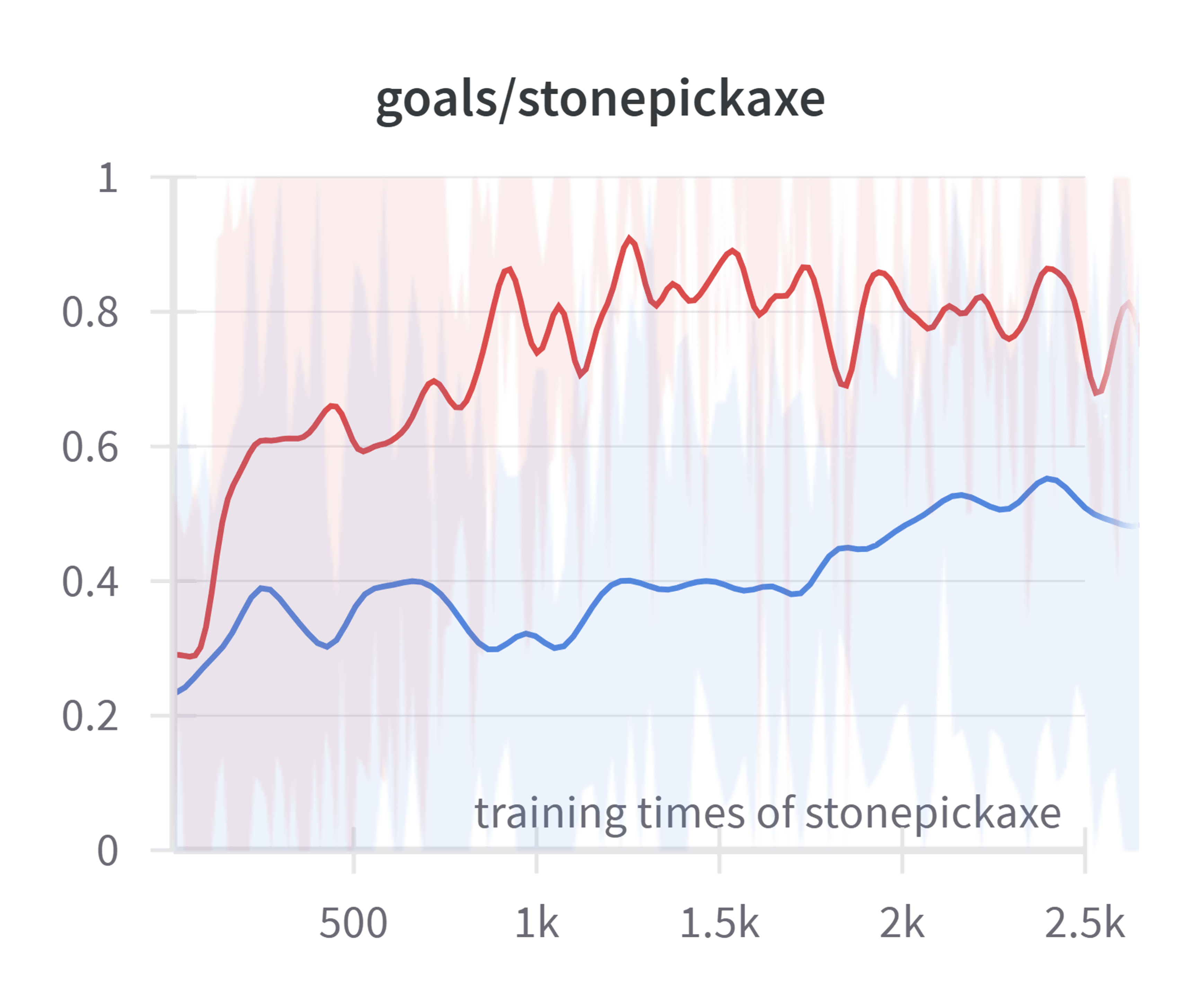} &
		\includegraphics[width=0.16\textwidth]{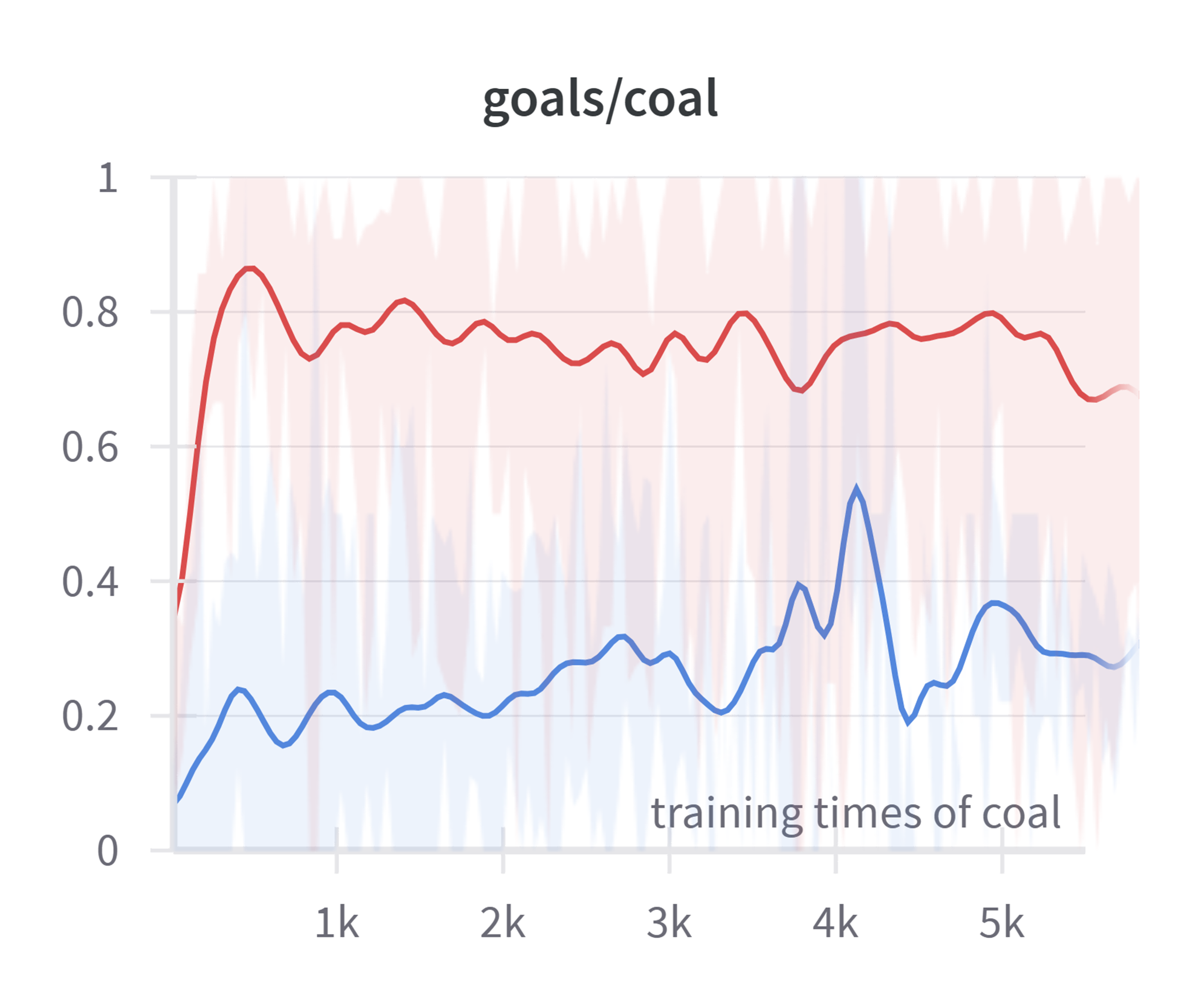} &
		\includegraphics[width=0.16\textwidth]{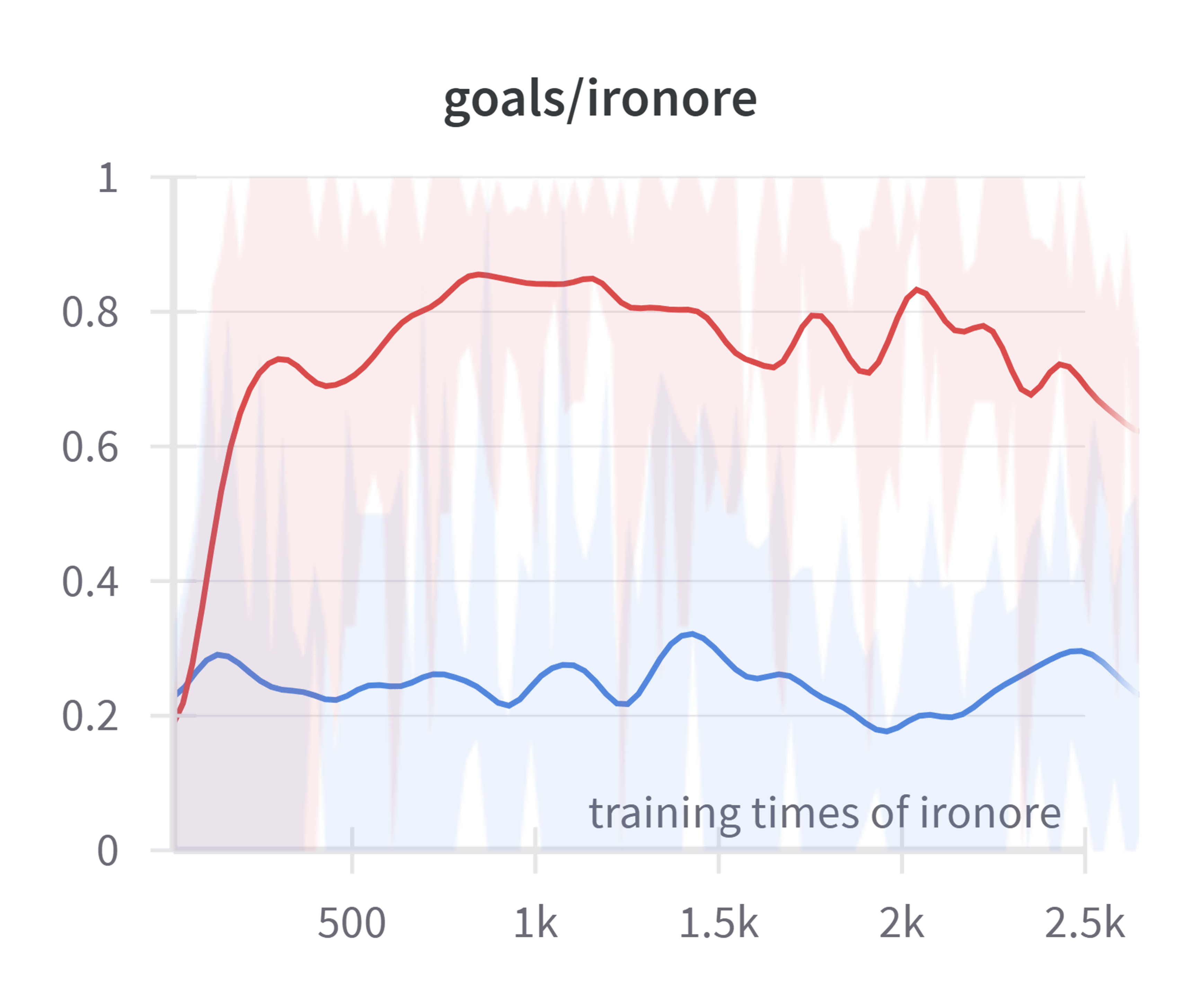}
	\end{tabular}
	\caption{Comparison of sub-goal training efficiency in GetIron-R0-T0 with \(\tau_{max}=1\) (Up) and GetIron-R0-T1 with \(\tau_{max}=4\) (Down).}% \textbf{\textcolor{green}{\(\tau_{max}=1\)}} \textcolor{purple}{GetIron} (Up) and \textbf{\textcolor{red}{\(\tau_{max}=4\)}} \textcolor{blue}{GetIron} (Down).
	\label{fig:subgoal_training_}
\end{figure*}

\paragraph{Experimental Design} 
To ensure a fair comparison, we compared the sub-goals training efficiency of D3HRL and CDHRL in GetIron-R0-T0 (\(\tau_{max}=1\)) and GetIron-R0-T1 (\(\tau_{max}=4\)) tasks, as shown in Figure~\ref{fig:subgoal_training_}. %The sub-goals required to complete GetIron task are shown in Figure~\ref{fig:get_iron} in \ref{app:environment}.
%For a fair comparison, we compared the training efficiency of each sub-goal to complete the GetIron-R1-T0 (\(\tau_{max}=1\)) task between CDHRL and D3HRL. To emphasize the importance of accurately determining time spans of causal relationships, we also compared them in GetIron-R1-T1 (\(\tau_{max}=4\)). In GetIron, the agent completes the task once it obtains iron, the intermediate sub-goals required to achieve this are shown in Figure~\ref{fig:get_iron} in \ref{app:environment}. We compared the training times of D3HRL and CDHRL for each sub-goal, as shown in Figures~\ref{fig:subgoal_training_}.
% as well as the final task
%Since CDHRL cannot percive variable-length state transitions, we set the span for learning causal relationships and collecting transition data to $\tau_{max}$ in its implementation.
%intermediate 

\paragraph{Experimental Results}
The results indicate that D3HRL demonstrates higher training efficiency for most sub-goals compared to CDHRL, achieving better sub-goal completion in shorter training times in both \(\tau_{max}=1\) and \(\tau_{max}=4\) tasks. Overall, the hierarchical framework employed by D3HRL proves to be more efficient in training sub-goals.

\subsubsection{Scalability: How does D3HRL perform across different time spans?}
\begin{figure*}[htbp]
	\centering
	\includegraphics[width=0.5\linewidth]{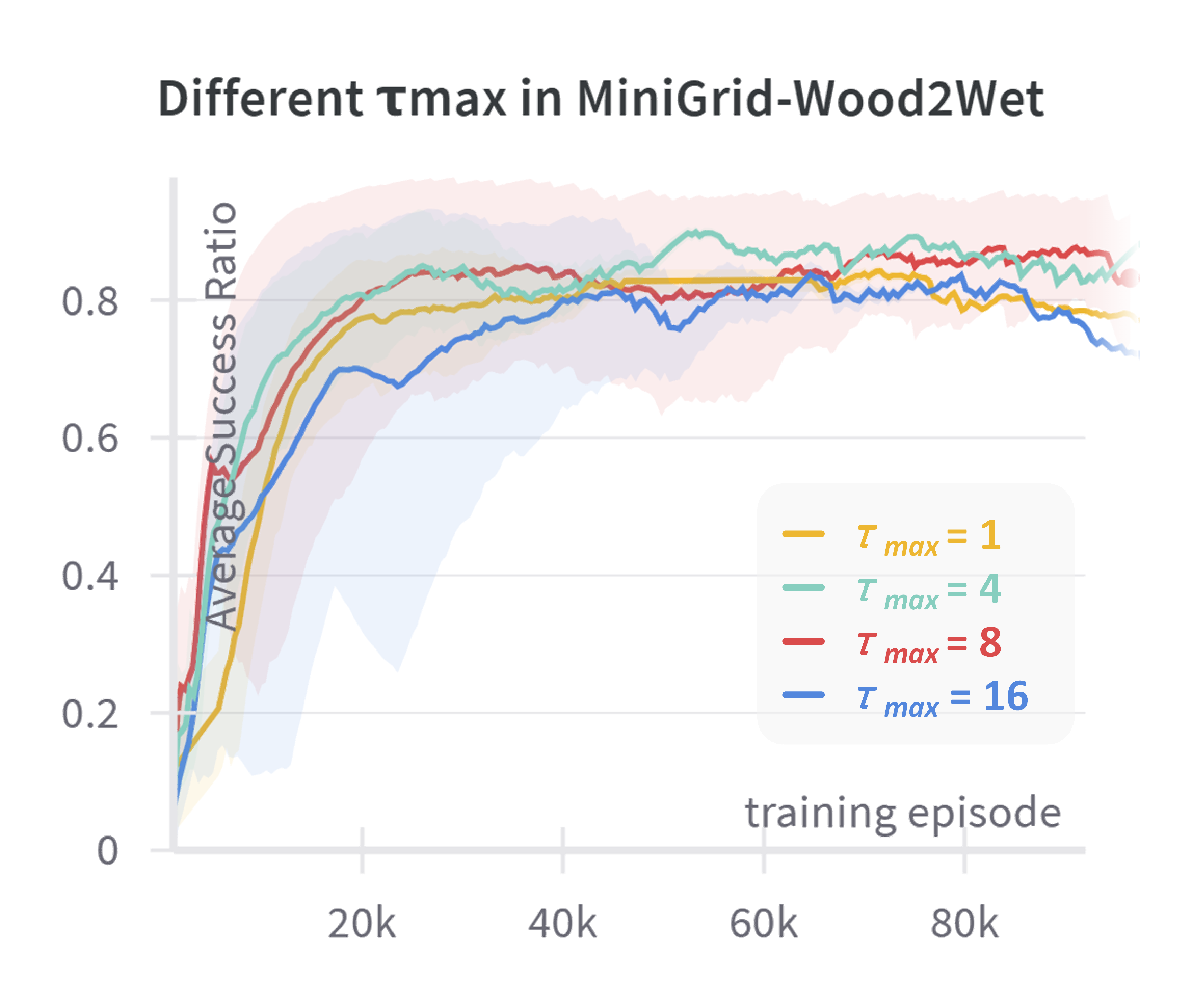}
	\caption{D3HRL's ASR in MiniGrid-Wood2Wet task with varying \(\tau_{max}\).}\label{fig:different_tau}
\end{figure*}

%\noindent\textbf{4. How does D3HRL perform across different time spans?}
\paragraph{Experimental Design}
To validate the ASR of D3HRL under different time spans and the SHD of its causal graph identification, we conducted experiments in Wood2Wet task, varying the values of \(\tau_{max}\) to 1, 4, 8, and 16. The results are shown in Figure~\ref{fig:different_tau} and Table~\ref{tab:shd}. %These experiments help evaluate the performance of D3HRL in tasks with varying complexity of state transitions, particularly in its ability to identify long-horizon causal relationships. 

\paragraph{Experimental Results}
% Figure~\ref{fig:different_tau} and 
As depicted in Figure~\ref{fig:different_tau} and Table~\ref{tab:shd}, D3HRL accurately captures causal relationships in tasks with long state transition spans and performs consistently well across tasks with varying maximum transition spans. In summary, the final average success rates are similar, indicating that the maximum time span (number of parallel processes) has little impact on the learning of causal relationships. However, increasing the number of parallel processes can lead to longer experimental times, especially with limited hardware resources. %The discussion regarding the time and space complexity of learning different time spans of causal relationships in parallel processes is in \ref{app:time_space_complexity}. 

\section{Conclusion}
%Existing HRL methods still face two challenges in long-horizon tasks: variable-length state transitions and spurious correlations. D3HRL addresses these by modeling variable-length state transitions using distributed causal discovery and filtering out spurious correlations through conditional independence testing. This improves the efficiency of hierarchical policy learning, leading to better performance on long-horizon tasks with variable-length state transitions. However, D3HRL struggles with tasks where time spans are randomly distributed. Additionally, it was tested in environments with decouplable state variables, limiting its use in environments where semantic extraction is not feasible. Future work will focus on extending D3HRL to handle various modalities, such as images, and to manage tasks with randomly distributed state transition spans.
%However, while D3HRL can accurately identify causal relationships and their time spans, it struggles with tasks where time spans are randomly distributed.
Existing HRL methods face two key challenges in long-horizon tasks: delay effects (variable-length state transitions) and spurious correlations. D3HRL addresses them by modeling variable-length state transitions with distributed causal discovery and filtering out spurious correlations using conditional independence testing.  However, D3HRL struggles with tasks that have randomly distributed time spans and is limited to tasks with decouplable state variables. Future work will extend D3HRL to handle diverse modalities, such as images, and manage tasks with randomly distributed state transitions.
%This enhances hierarchical policy learning efficiency, improving performance on long-horizon tasks. 
%\section*{Acknowledgments}
%This improves the efficiency of hierarchical policy learning in long-horizon tasks with variable-length state transitions.

\begin{comment}
\section*{Declaration of Generative AI and AI-assisted technologies in the writing process}
During the preparation of this work the authors used Qwen~\cite{qwen} to check grammatical errors. After using them, the authors reviewed and edited the content as needed and take full responsibility for the content of the publication.
\end{comment}

%%%%%%%%%%%%%%%%%%%%%%%%%%%%%%%%%%%%%%%%%%%%%%%%%%%%%%%%%%%%

%% If you have bib database file and want bibtex to generate the
%% bibitems, please use
%%
%%  \bibliographystyle{elsarticle-harv} 
%%  \bibliography{<your bibdatabase>}
\newpage
%% The file named.bst is a bibliography style file for BibTeX 0.99c
%\bibliographystyle{named}

\end{document}